\documentclass{article}

\PassOptionsToPackage{numbers, compress}{natbib}
\usepackage[preprint]{neurips_2026}

\usepackage[utf8]{inputenc} 
\usepackage[T1]{fontenc}    
\usepackage{url}            
\usepackage{booktabs}       
\usepackage{amsfonts}       
\usepackage{nicefrac}       
\usepackage{microtype}      
\usepackage{xcolor}         

\usepackage{amsmath}
\usepackage{colortbl}
\definecolor{iccvblue}{rgb}{0.21,0.49,0.74}
\definecolor{best}{RGB}{255, 179, 179} 
\definecolor{second}{RGB}{255, 217, 179} 
\definecolor{third}{RGB}{255, 255, 179} 
\usepackage[pagebackref=true,breaklinks=true,letterpaper=true,colorlinks,bookmarks=false,citecolor=iccvblue]{hyperref}

\usepackage{subcaption}
\usepackage{textcomp}
\usepackage{booktabs}
\usepackage{multirow}
\usepackage{adjustbox}
\usepackage{afterpage}
\usepackage{bbding}
\usepackage[flushleft]{threeparttable}
\setcitestyle{square}

\usepackage{algorithm}
\usepackage{algpseudocode}
\usepackage{listings}
\usepackage{amssymb}
\usepackage{orcidlink}

\definecolor{defgreen}{RGB}{34, 139, 34}
\definecolor{docbrown}{RGB}{178, 34, 34}
\definecolor{comteal}{RGB}{64, 128, 128}

\usepackage{xr}
\makeatletter
\newcommand*{\addFileDependency}[1]{
  \typeout{(#1)}
  \@addtofilelist{#1}
  \IfFileExists{#1}{}{\typeout{No file #1.}}
}
\makeatother

\newcommand*{\myexternaldocument}[1]{
    \externaldocument{#1}
    \addFileDependency{#1.tex}
    \addFileDependency{#1.aux}
}
\myexternaldocument{supplementary}

\title{At FullTilt: Real-Time Open-Set 3D Macromolecule Detection Directly from Tilted 2D Projections}

\author{%
  Ming-Yang Ho \orcidlink{0000-0002-1767-7494}\\
  Duke University\\
  \texttt{ming-yang.ho@duke.edu} \\
  \And
  Alberto Bartesaghi \orcidlink{0000-0002-7360-1523}\\
  Duke University\\
  \texttt{alberto.bartesaghi@duke.edu} \\
}

\begin{document}

\maketitle

\begin{abstract}
Open-set 3D macromolecule detection in cryogenic electron tomography eliminates the need for target-specific model retraining. However, strict VRAM constraints prohibit processing an entire 3D tomogram, forcing current methods to rely on slow sliding-window inference over extracted subvolumes. To overcome this, we propose FullTilt, an end-to-end framework that redefines 3D detection by operating directly on aligned 2D tilt-series. Because a tilt-series contains significantly fewer images than slices in a reconstructed tomogram, FullTilt eliminates redundant volumetric computation, accelerating inference by orders of magnitude. To process the entire tilt-series simultaneously, we introduce a tilt-series encoder to efficiently fuse cross-view information. We further propose a multiclass visual prompt encoder for flexible prompting, a tilt-aware query initializer to effectively anchor 3D queries, and an auxiliary geometric primitives module to enhance the model's understanding of multi-view geometry while improving robustness to adverse imaging artifacts. Extensive evaluations on three real-world datasets demonstrate that FullTilt achieves state-of-the-art zero-shot performance while drastically reducing runtime and VRAM requirements, paving the way for rapid, large-scale visual proteomics analysis. Code and data are available at \url{https://github.com/Kaminyou/FullTilt}.
\end{abstract}

\section{Introduction}
Determining the 3D structures of macromolecules (hereafter referred to as \textit{particles}) is critical for understanding biological mechanisms and advancing drug development~\cite{blundell1996structure, duran2013structural}. Cryogenic electron tomography (cryo-ET) is an imaging technique that enables \textit{in situ} 3D structural visualization~\cite{turk2020promise, watson2024advances}. Using this method, cellular samples are rotated and imaged by capturing a series of 2D projections at different angles. The resulting \textit{tilt-series} is aligned to a common vertical tilt axis ($x=0.5$ in normalized image coordinates) and subsequently reconstructed into a 3D tomogram. Off-the-shelf computational methods are applied to detect target particles~\cite{wagner2025cryo}, followed by subtomogram averaging and refinement to produce high-resolution structures~\cite{watson2024advances}.

While conventional closed-set detection methods~\cite{wagner2019sphire, moebel2021deep, liu2024deepetpicker} require laborious data annotation and retraining for every new target, recent work has explored open-set detection paradigms~\cite{rice2023tomotwin, zhao2024cryosam, wiedemann2026propicker, uddin2024tomopicker}. By supplying a \textit{visual prompt} (e.g., a reference subvolume), these frameworks localize arbitrary targets by comparing prompt embeddings against 3D subvolumes extracted from the tomogram. This enables \textit{zero-shot} detection without target-specific retraining.

\begin{figure}[t]
\begin{center}
\includegraphics[width=1.0\linewidth]{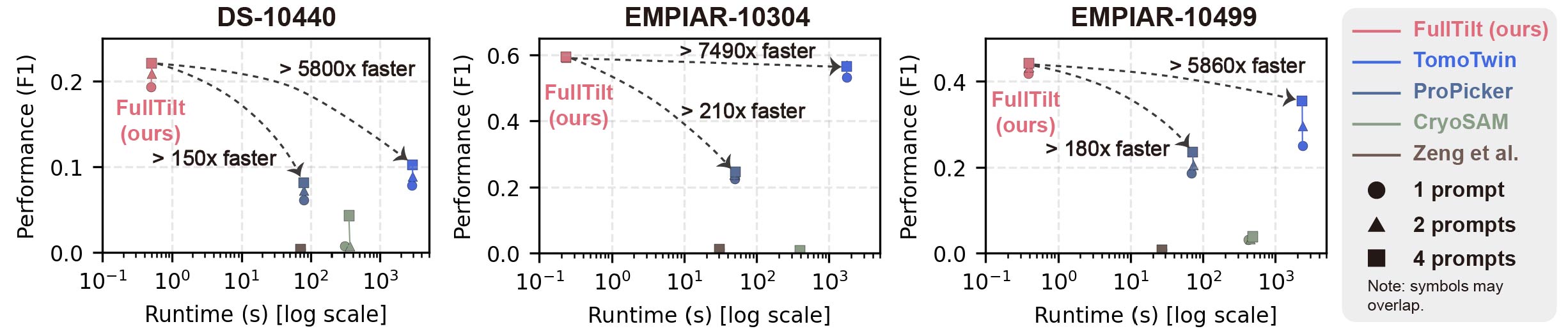}
\end{center}
\caption{\textbf{Runtime and performance comparison.} FullTilt achieves inference speeds hundreds to thousands of times faster than current state-of-the-art open-set macromolecule detection methods, while delivering comparable or superior zero-shot performance across three real-world cryo-ET datasets. The exact acceleration multiplier varies based on the tilt-series and tomogram dimensions.}
\label{fig:teaser}
\end{figure}

\begin{figure}[t]
\begin{center}
\includegraphics[width=1.0\linewidth]{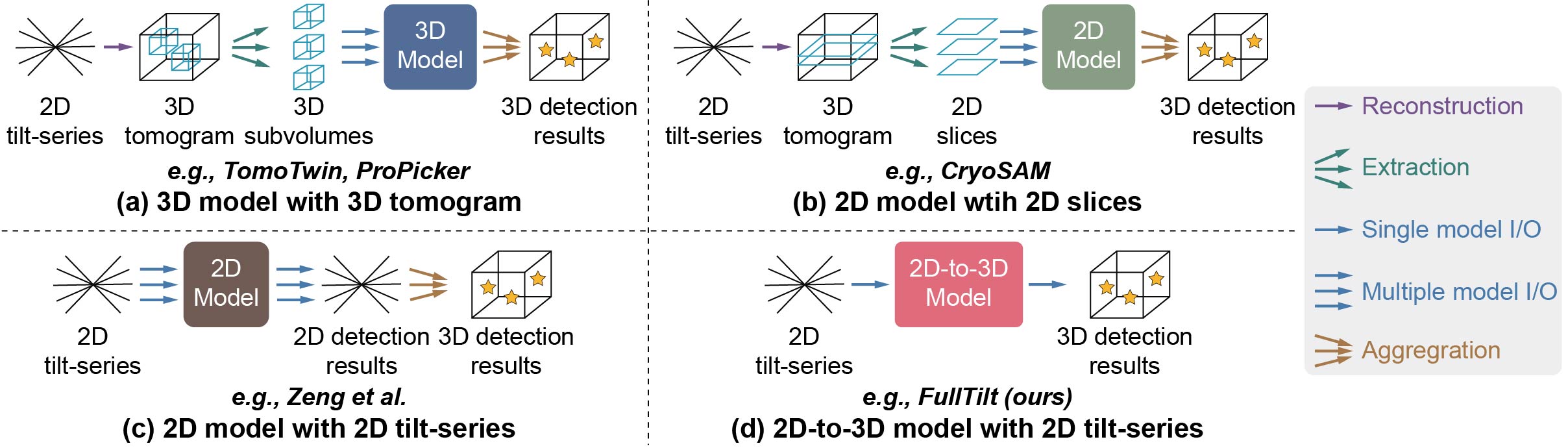}
\end{center}
\caption{\textbf{Comparison of detection mechanisms.} We show existing architectures and inputs for open-set 3D macromolecule detection in cryo-ET (prompting modules omitted for simplicity). (a) TomoTwin~\cite{rice2023tomotwin} and ProPicker~\cite{wiedemann2026propicker} process 3D tomograms using 3D models. Due to strict VRAM limits, they rely on sliding-window subvolume extraction, which requires extensive model I/O and bottlenecks inference speed. (b) CryoSAM~\cite{zhao2024cryosam} applies a 2D SAM model to orthogonal 2D slices of the 3D tomogram, necessitating repeated, costly model I/O operations. (c) Zeng et al.~\cite{zeng2022structure} apply a 2D detector to individual 2D tilt images and back-project the results into 3D space. This approach suffers from low single-image visibility and severe back-projection errors. (d) FullTilt is the first framework to process the 2D tilt-series directly into 3D detection results in an end-to-end manner.}
\label{fig:compare}
\end{figure}

However, strict VRAM constraints prohibit loading an entire 3D tomogram into memory. As a result, current methods rely on sliding-window inference approaches~\cite{rice2023tomotwin, wiedemann2026propicker}, resulting in long inference runtimes (Figs.~\ref{fig:teaser} and ~\ref{fig:compare}a). Others apply SAM~\cite{kirillov2023segment} to 2D slices along each orthogonal axis of the 3D tomogram~\cite{zhao2024cryosam}, but they still exhibit high computational complexity that scales with the tomogram's volumetric dimensions (Figs.~\ref{fig:teaser} and ~\ref{fig:compare}b).

Considering that a 3D tomogram is reconstructed from the aligned 2D tilt-series~\cite{levoy1992volume, winkler2006accurate}, the tilt-series intrinsically contains at least as much information as the tomogram. Moreover, the number of images in a tilt-series (e.g., 41) is usually smaller than the depth dimension of the reconstructed tomogram (e.g., 256). This indicates that tomograms contain redundant information, which wastes computational resources and slows down processing, raising the critical question: \textit{Can we conduct open-set 3D macromolecule  detection directly on the aligned 2D tilt-series to expedite runtime?}

To answer this question, we propose \textit{FullTilt}, an open-set 3D macromolecule detection framework that processes the 2D tilt-series in an end-to-end manner (Fig.~\ref{fig:compare}d). By dramatically reducing the input dimensionality, our framework operates \textit{at full tilt}---running hundreds to thousands of times faster than current state-of-the-art methods (Fig.~\ref{fig:teaser}).

\begin{figure}[h]
\begin{center}
\includegraphics[width=1.0\linewidth]{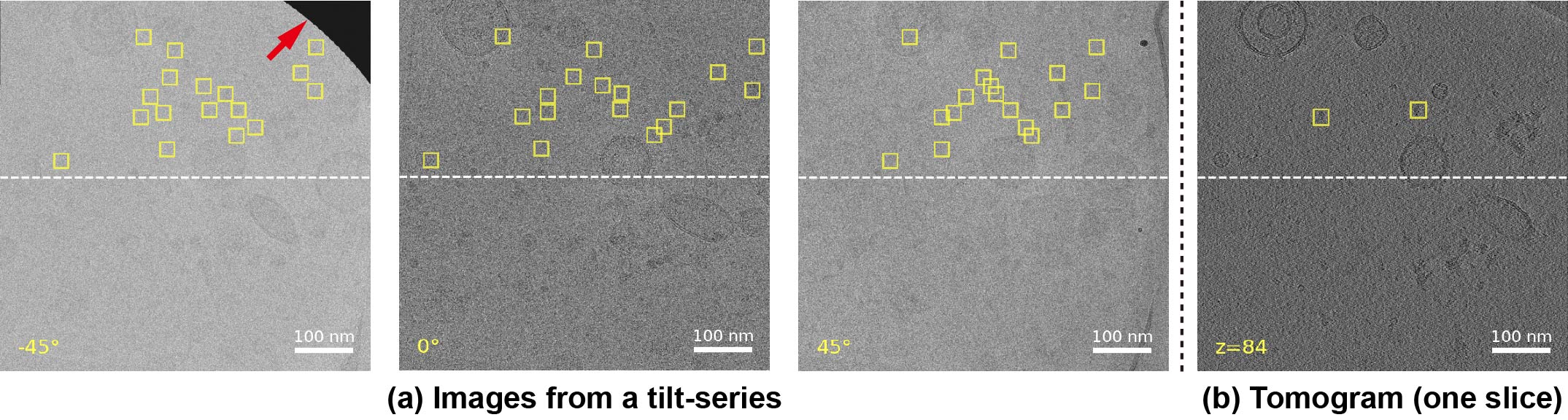}
\end{center}
\caption{\textbf{2D tilt-series vs. 3D tomogram.} We show images at varying tilt angles alongside a tomogram slice reconstructed from the same tilt-series (DS-10440 dataset, \texttt{TS\_6\_6}). Yellow boxes denote thyroglobulin particles (top half). (a) High noise makes recognizing individual particles in a single 2D tilt image difficult without viewing consecutive frames. Furthermore, particles often overlap at certain angles because each image projects through the full sample depth, and raw images often exhibit severe physical occlusion (red arrow). (b) Conversely, within a tomogram Z-slice, particles are separated along the depth axis with no spatial overlap.}
\label{fig:examples}
\end{figure}

FullTilt adopts a DETR-like architecture~\cite{carion2020end}. We first utilize a 2D backbone to extract initial 2D features~\cite{liu2021swin}. Our tilt-series encoder ($\mathcal{E}_{\text{Tilt}}$) then merges these feature maps into 3D-aware features with an efficient global row attention, followed by a multiclass visual prompt encoder ($\mathcal{E}_{\text{Prompt}}$) that extracts prototypes for the user's requested classes. We then effectively initialize 3D anchors with a tilt-aware query initializer ($\mathcal{I}_{\text{Tilt}}$) and leverage a 3D decoder to obtain the 3D coordinates and labels of the targeted particles. To enhance the model's understanding of multi-view geometry and improve its robustness to adverse imaging artifacts such as occlusion and non-uniform illumination (Fig.~\ref{fig:examples}), we introduce an auxiliary geometric primitives module ($\mathcal{G}$) that generates geometric data \textit{on-the-fly}.

Although 3D macromolecule detection from 2D tilt-series shares similarities with multi-view 3D object detection in autonomous driving~\cite{singh2023surround}, directly applying existing frameworks (e.g., DETR3D~\cite{wang2022detr3d} or PETR~\cite{liu2022petr}) yields poor results, rendering such adaptations non-trivial. This is due to the extremely low signal-to-noise ratio (SNR) characteristic of cryo-ET data~\cite{yang2024self}. Unlike natural images where objects are clearly visible in a single frame, macromolecules often only become distinguishable when observing consecutive frames across the tilt-series (Fig. \ref{fig:examples}). For the same reason, naive implementations, such as applying a 2D detector to each tilt image independently and back-projecting the results into 3D space~\cite{zeng2022structure}, also fail (Figs.~\ref{fig:teaser} and~\ref{fig:compare}c). FullTilt successfully addresses this challenge by introducing a dedicated tilt-series encoder, enabling the model to holistically contextualize all tilts at once.

Evaluated on three public, real-world cryo-ET datasets as well as simulated datasets, FullTilt achieves comparable or superior zero-shot performance against existing open-set baselines, while dramatically reducing runtime and VRAM requirements. In summary, our contributions are as follows:
\begin{itemize}
    \item To the best of our knowledge, this is the first study to conduct 3D macromolecule detection directly from 2D tilt-series in an end-to-end manner, achieving state-of-the-art zero-shot performance on real-world cryo-ET data while drastically reducing computational overhead.
    \item We introduce a tilt-series encoder to efficiently merge information across images, paired with a tilt-aware query initializer to effectively establish 3D anchors. Additionally, an auxiliary geometric primitives module is introduced to enhance the model's understanding of multi-view geometry and improve its robustness to adverse imaging artifacts. 
    \item We propose a multiclass visual prompt encoder designed to flexibly handle diverse prompting scenarios (accommodating varying prompt counts and multiple classes per tilt image), which can be seamlessly integrated into any DETR-like framework.
\end{itemize}

\section{Related Work}
\noindent \textbf{3D Macromolecule Detection for Cryo-ET.} Deep learning has largely superseded template matching in cryo-ET~\cite{moebel2021deep}. Methods like crYOLO~\cite{wagner2019sphire} and PickYOLO~\cite{genthe2023pickyolo} merge 2D slice detections into 3D coordinates. Zeng's method~\cite{zeng2022structure} back-projects 2D tilt-series detections into 3D, but suffers from high projection errors and lacks cross-tilt fusion. Volumetric frameworks (e.g., DeepFinder~\cite{moebel2021deep}, DeePiCt~\cite{de2023convolutional}, VP-Detector~\cite{hao2021vp}) utilize 3D CNNs, requiring sliding-window inference and segmentation clustering. To reduce annotation costs, weakly supervised learning (WSL) methods (Huang's method~\cite{huang2022accurate}, TomoPicker~\cite{uddin2024tomopicker}) employ positive-unlabeled learning~\cite{kiryo2017positive}. Crucially, all these approaches operate in a \textit{closed-set} paradigm, requiring target-specific model retraining for novel particle classes. 

\noindent \textbf{Open-Set 3D Macromolecule Detection.} To bypass retraining, open-set detection paradigms have recently emerged~\cite{rice2023tomotwin, wiedemann2026propicker, zhao2024cryosam}. TomoTwin~\cite{rice2023tomotwin} compares a target subvolume's embedding against all 3D tomogram subvolumes. ProPicker~\cite{wiedemann2026propicker} frames this as conditional segmentation to improve speed, while CryoSAM~\cite{zhao2024cryosam} clusters masks generated by applying 2D SAM~\cite{kirillov2023segment} to orthogonal tomogram slices. However, all these methods rely on exhaustive 2D or 3D tomogram patching, which severely bottlenecks inference via repetitive model I/O. Instead, FullTilt directly processes the 2D tilt-series end-to-end, completely bypassing 3D tomogram inference, thus drastically accelerating detection.

\noindent \textbf{Multi-View 3D Object Detection.} End-to-end multi-view 3D object detection is prominent in autonomous driving~\cite{alzahrani2024deep}. DETR3D~\cite{wang2022detr3d} extends DETR~\cite{carion2020end} for multi-view data, while PETR~\cite{liu2022petr} replaces DETR3D's 3D-to-2D projection with a 3D positional encoder. Subsequent models like PETRv2~\cite{liu2023petrv2} incorporate temporal modeling. FullTilt combines DETR3D's training efficiency (via 3D-to-2D projection) with PETR's idea of 3D-aware positional encoding (detailed in~\ref{sec:compare_detr3d_petr}). Crucially, to handle high-noise cryo-ET data and enable open-set detection, FullTilt further introduces four novel components: a tilt-series encoder ($\mathcal{E}_{\text{Tilt}}$), a multiclass visual prompt encoder ($\mathcal{E}_{\text{Prompt}}$), a tilt-aware query initializer ($\mathcal{I}_{\text{Tilt}}$), and an auxiliary geometric primitives module ($\mathcal{G})$.

\begin{figure}[t]
\begin{center}
\includegraphics[width=1.0\linewidth]{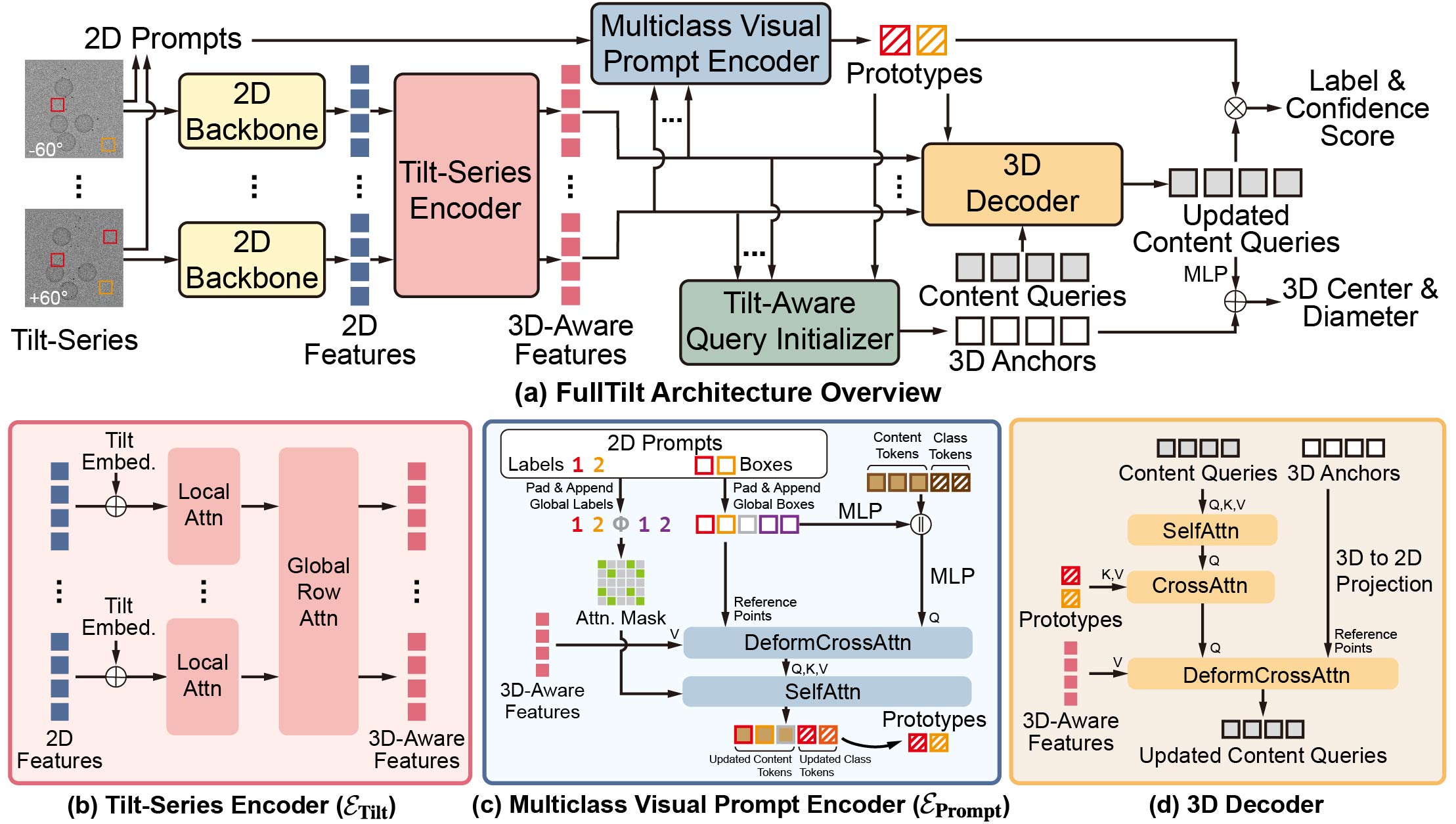}
\end{center}
\caption{\textbf{FullTilt architecture.} (a) Overview of the FullTilt architecture. (b) The tilt-series encoder ($\mathcal{E}_{\text{Tilt}}$) utilizes alternating local and global row attention to fuse information from the 2D features into 3D-aware features. (c) The multiclass visual prompt encoder ($\mathcal{E}_{\text{Prompt}}$) accommodates flexible prompting scenarios by leveraging masked attention to generate prototypes for all requested classes. (d) The 3D decoder updates the content queries via self-attention, cross-attention with the prototypes, and deformable cross-attention with the 3D-aware features, using 2D coordinates projected from the 3D anchors.}
\label{fig:framework}
\end{figure}

\section{FullTilt}
\subsection{Problem Definition}
Given an aligned 2D tilt-series of $N$ images $\{\mathbf{X}_i\}_{i=1}^N$ with $\mathbf{X}_i \in \mathbb{R}^{H \times W}$ and corresponding tilt angles $\{\theta_i\}_{i=1}^N$, each image $\mathbf{X}_i$ is accompanied by $K_i$ visual prompts $\{(\mathbf{P}_{i}^{k}, Y_i^k)\}_{k=1}^{K_i}$. Each prompt comprises a class label $Y_i^k \in \mathbb{Z}^{+}$ and a bounding box $\mathbf{P}_{i}^{k} = (x, y, d, d) \in \mathbb{R}^{4}$ defining the particle's 2D center coordinates $(x, y)$ and diameter $d$.

FullTilt processes these inputs end-to-end to simultaneously estimate a set of $M$ 3D particles $\{(\hat{\mathbf{P}}^m, \hat{Y}^m, s^m)\}_{m=1}^{M}$. Here, $\hat{\mathbf{P}}^m = (\hat{x}, \hat{y}, \hat{z}, \hat{d}) \in \mathbb{R}^{4}$ denotes the estimated 3D center and diameter, $\hat{Y}^m \in \mathbb{Z}^{+}$ is the estimated class label, and $s^m \in [0, 1]$ is the confidence score.

\subsection{Architecture Overview}
FullTilt follows the DETR design principles~\cite{carion2020end}, using a transformer encoder-decoder to estimate a set of objects from learnable queries, with bipartite matching for training. It consists of five sequential modules (Fig.~\ref{fig:framework}a). First, a \textbf{2D backbone} independently extracts multi-level 2D feature maps from each tilt image. A \textbf{tilt-series encoder ($\mathcal{E}_{\text{Tilt}}$)} then collectively processes these features, merging cross-tilt information into 3D-aware features. Subsequently, a \textbf{multiclass visual prompt encoder ($\mathcal{E}_{\text{Prompt}}$)} extracts prototypes for the requested classes based on the 2D prompts. Using these prototypes and the 3D-aware features, a \textbf{tilt-aware query initializer ($\mathcal{I}_{\text{Tilt}}$)} establishes 3D anchors and initializes learnable content queries. Finally, a \textbf{3D decoder} refines these queries to estimate target 3D coordinates, diameters, and labels.

\subsection{Architecture Details}
\noindent \textbf{2D Backbone.} Following the standard DETR design~\cite{DBLP:conf/iclr/ZhuSLLWD21}, a 2D backbone (e.g., Swin Transformer) extracts multi-level feature maps from each 2D tilt image $\mathbf{X}_i$, denoted as $\mathbf{F}_{i,l}\in \mathbb{R}^{H_l \times W_l \times C}$, where $l$ indicates the feature level and $C$ is the hidden channel dimension used throughout the architecture.

\noindent \textbf{Tilt-Series Encoder ($\mathcal{E}_{\text{Tilt}}$).} To lift these 2D features into 3D-aware representations, we aggregate information across all tilt images (Fig.~\ref{fig:framework}b). For a given tilt angle $\theta$, we generate a tilt embedding $\mathbf{e}_\theta \in \mathbb{R}^{C}$ via $\mathbf{e}_\theta = \text{MLP} \Big( \big[ \sin(\theta \omega_0), \dots, \sin(\theta \omega_{\nicefrac{C}{2} - 1}), \cos(\theta \omega_0), \dots, \cos(\theta \omega_{\nicefrac{C}{2} - 1}) \big] \Big)$, where $\omega_{k}=2^{\frac{8k}{\nicefrac{C}{2} - 1}}$. These tilt embeddings, alongside 2D positional and level embeddings, are added to the 2D features. We then adapt an alternating-attention (AA) strategy~\cite{wang2025vggt}, which alternates local (single-image) and global (cross-image) attention to fuse information across images. Specifically, we perform \textit{local attention} within each tilt image's features followed by \textit{global row attention} across all tilts. Since the tilt-series is aligned to a common vertical tilt axis, a target particle consistently occupies the same row across all images. Exploiting this geometric constraint drastically reduces the computational overhead of standard AA---vital given the large number of views $N$ in cryo-ET---while remaining straightforward to implement (Appendix Alg.~\ref{alg:attention_pseudo}). After $L_{\text{AA}}$ layers, we obtain the final 3D-aware features $\{\mathbf{Z}_i\}_{i=1}^{N}$, where $\mathbf{Z}_i \in \mathbb{R}^{\tilde{H}\times \tilde{W}\times C}$ (multi-level notation omitted for simplicity).

\noindent \textbf{Multiclass Visual Prompt Encoder ($\mathcal{E}_{\text{Prompt}}$).} Given the 2D prompts $\{\{(\mathbf{P}_{i}^{k}, Y_i^k)\}_{k=1}^{K_i} \}_{i=1}^{N}$, let $N_{\text{p}}:=\max_i K_i$ be the maximum number of prompts across all images and $N_{\text{c}}$ be the total number of unique classes. For each $i \in \{1, \dots, N\}$, we pad the local prompt labels $\{Y_i^k\}_{k=1}^{K_i}$ to size $N_{\text{p}}$ using an empty token $\phi$, and append the $N_{\text{c}}$ unique global labels $(1, \dots, N_c)$ to form the label vector $\mathbf{V}_i$ of size $N_{\text{p}}+N_{\text{c}}$ (while we assume ordered labels here for simplicity, our framework supports arbitrary $\mathbb{Z}^{+}$ labels; details in~\ref{sec:external_internal_mapping}). To govern information flow, we generate an attention mask $\mathbf{A}_i \in \{0, 1\}^{(N_{\text{p}}+N_{\text{c}})\times (N_{\text{p}}+N_{\text{c}})}$ (Fig.~\ref{fig:framework}c), where the entry at row $r$ and column $c$ permits attention only between valid, identical labels (i.e., $\mathbf{V}_i[r] = \mathbf{V}_i[c] \neq \phi$).

Similarly, we pad the local bounding boxes $\{\mathbf{P}_i^k\}_{k=1}^{K_i}$ to size $N_{\text{p}}$ using dummy coordinates $(0, 0, 0, 0)$. Extending the global bounding box approach introduced in T-Rex2~\cite{jiang2024t}, we append $N_{\text{c}}$ such boxes $(0.5, 0.5, 1.0, 1.0)$ and encode the entire spatial set via an MLP. Next, we initialize $N_{\text{p}}$ learnable \textit{content tokens} and $N_{\text{c}}$ learnable \textit{class tokens}. These are concatenated along the channel dimension with their corresponding encoded boxes and fused via an MLP to form a unified token sequence $\mathbf{E}_i \in \mathbb{R}^{(N_{\text{p}}+N_{\text{c}})\times C}$. This sequence is iteratively refined through $L_{\text{VP}}$ layers. Each layer applies deformable cross-attention with the 3D-aware features, followed by masked self-attention utilizing $\mathbf{A}_i$, which enables the class tokens to aggregate information exclusively from relevant content tokens. After each iteration, the updated class tokens are averaged across all tilt images containing that specific class. The final extracted class tokens serve as our class \textit{prototypes}, denoted as $\mathbf{t}\in \mathbb{R}^{N_\text{c}\times C}$.

\noindent \textbf{Tilt-Aware Query Initializer ($\mathcal{I}_{\text{Tilt}}$).} Because a particle's $x$ and $y$ coordinates align perfectly with its projection at a $0^\circ$ tilt, we strategically initialize the \textit{3D anchors} $\{\mathbf{q}_{\text{Anchor}}^{m}\}_{m=1}^{M}$ using the near-zero tilt image. We first extract the features $\mathbf{Z}_{i^{*}}$ from the tilt closest to $0^\circ$, where $i^{*} = \arg\min_{i} |\theta_i|$. Following previous work~\cite{jiang2024t, liu2024grounding}, we select the initial 2D anchors $(x, y, d, d)$ from the bounding boxes corresponding to the top $M$ entries of $\mathbf{S} \in \mathbb{R}^{\tilde{H}\times\tilde{W}}$, where $S_{h,w} = \max_{1 \leq j \leq N_c}(\mathbf{Z}_{i^{*}}\mathbf{t}^{\top})_{h,w,j}$ takes the maximum over the $N_c$ class prototypes at each spatial location. Assuming a zero-mean depth bias within the tomogram, we initialize the $z$-coordinate to 0.5 (the middle plane), yielding 3D anchors formatted as $\mathbf{q}_{\text{Anchor}}^m=(x, y, z, d)\in \mathbb{R}^{4}$. As is standard practice, the corresponding \textit{content queries} $\{\mathbf{q}_{\text{Content}}^{m}\}_{m=1}^{M}$, where each $\mathbf{q}_{\text{Content}}^m \in \mathbb{R}^{C}$, are initialized as learnable parameters.

\begin{table*}[t]
\centering
\caption{\textbf{Zero-shot intra-instance detection results on the DS-10440 dataset.} The best-performing cryo-ET method for each metric is highlighted in bold. FullTilt achieves the best performance across all evaluated metrics, with the shortest runtime and lowest VRAM usage among cryo-ET methods.}
\label{tab:intra_sample_czii}
\resizebox{0.7\textwidth}{!}{%
\begin{threeparttable}
\begin{tabular}{@{}ll|c|rrrrr@{}}
\toprule
\textbf{Category} &
\textbf{Method} &
\textbf{\# Prompts} & 
\textbf{mAP@0.5r $\uparrow$} & \textbf{mAP@1r $\uparrow$} &
\textbf{F1 $\uparrow$} &
\textbf{T. $\downarrow$} &
\textbf{M. $\downarrow$}
\\

\midrule
\multirow{2}{*}{General MV} & 
DETR3D~\cite{wang2022detr3d}$+\mathcal{E}_{\text{Prompt}}$ & \multirow{7}{*}{1} & 0.048 ± 0.004 & 0.091 ± 0.005 & 0.146 ± 0.006 & 0.214 & 3430 \\

& PETR~\cite{liu2022petr}$+\mathcal{E}_{\text{Prompt}}$ & & 0.000 ± 0.000 & 0.001 ± 0.000 & 0.010 ± 0.000 & 0.571 & 13800 \\

\cmidrule{1-2} \cmidrule{4-8}
\multirow{5}{*}{Cryo-ET} & 
Zeng et al.~\cite{zeng2022structure}$+\mathcal{E}_{\text{Prompt}}$ & &
n.a. & n.a. & 0.004 ± 0.001 & \cellcolor{second}71.3 & \cellcolor{second}4050 \\ 

& TomoTwin~\cite{rice2023tomotwin} & &
\cellcolor{second}0.089 ± 0.021 & \cellcolor{second}0.132 ± 0.024 & \cellcolor{second}0.078 ± 0.012 & 2960 & 20800 \\ 

& CryoSAM~\cite{zhao2024cryosam} & &
n.a. & n.a. & 0.008 ± 0.003 & 309 & 30600 \\ 

& ProPicker~\cite{wiedemann2026propicker} & &
n.a. & n.a. & \cellcolor{third}0.061 ± 0.008 & \cellcolor{third}79.5 & \cellcolor{third}12700 \\ 

 & \textbf{FullTilt (ours)} & &
\cellcolor{best}\textbf{0.118 ± 0.013} & \cellcolor{best}\textbf{0.148 ± 0.013} & \cellcolor{best}\textbf{0.193 ± 0.008} & \cellcolor{best}\textbf{0.499} & \cellcolor{best}\textbf{3430} \\

\midrule
\midrule
\multirow{2}{*}{General MV} & 
DETR3D~\cite{wang2022detr3d}$+\mathcal{E}_{\text{Prompt}}$ & \multirow{7}{*}{2} & 0.056 ± 0.005 & 0.104 ± 0.005 & 0.161 ± 0.006 & 0.214 & 3430 \\

& PETR~\cite{liu2022petr}$+\mathcal{E}_{\text{Prompt}}$ & & 0.000 ± 0.000 & 0.001 ± 0.000 & 0.010 ± 0.000 & 0.572 & 13800 \\

\cmidrule{1-2} \cmidrule{4-8}
\multirow{5}{*}{Cryo-ET} & Zeng et al.~\cite{zeng2022structure}$+\mathcal{E}_{\text{Prompt}}$ & &
n.a. & n.a. & 0.004 ± 0.001 & \cellcolor{second}71.3 & \cellcolor{second}4050\\ 

& TomoTwin~\cite{rice2023tomotwin} & &
\cellcolor{second}0.066 ± 0.012 & \cellcolor{second}0.098 ± 0.015 & \cellcolor{second}0.089 ± 0.010 & 2960 & 20800 \\ 

& CryoSAM~\cite{zhao2024cryosam} & &
n.a. & n.a. & 0.008 ± 0.003 & 365 & 30600 \\ 

& ProPicker~\cite{wiedemann2026propicker} & &
n.a. & n.a. & \cellcolor{third}0.072 ± 0.004 & \cellcolor{third}79.5 & \cellcolor{third}12700 \\ 

& \textbf{FullTilt (ours)}  & &
\cellcolor{best}\textbf{0.129 ± 0.010} & \cellcolor{best}\textbf{0.161 ± 0.011} & \cellcolor{best}\textbf{0.209 ± 0.013} & \cellcolor{best}\textbf{0.503} & \cellcolor{best}\textbf{3430} \\

\midrule
\midrule
\multirow{2}{*}{General MV} & 
DETR3D~\cite{wang2022detr3d}$+\mathcal{E}_{\text{Prompt}}$ & \multirow{7}{*}{4} & 0.062 ± 0.003 & 0.111 ± 0.003 & 0.168 ± 0.005 & 0.215 & 3430 \\

& PETR~\cite{liu2022petr}$+\mathcal{E}_{\text{Prompt}}$ & & 0.000 ± 0.000 & 0.001 ± 0.000 & 0.010 ± 0.000 & 0.572 & 13800 \\

\cmidrule{1-2} \cmidrule{4-8}
\multirow{5}{*}{Cryo-ET} & Zeng et al.~\cite{zeng2022structure}$+\mathcal{E}_{\text{Prompt}}$ & &
n.a. & n.a. & 0.004 ± 0.001 & \cellcolor{second}71.3 & \cellcolor{second}4050 \\ 

& TomoTwin~\cite{rice2023tomotwin} & &
\cellcolor{second}0.067 ± 0.010 & \cellcolor{second}0.105 ± 0.015 & \cellcolor{second}0.102 ± 0.021 & 2960 & 20800 \\ 

& CryoSAM~\cite{zhao2024cryosam} & &
n.a. & n.a. & 0.043 ± 0.003 & 352 & 30600 \\ 

& ProPicker~\cite{wiedemann2026propicker} & &
n.a. & n.a. & \cellcolor{third}0.081 ± 0.004 & \cellcolor{third}79.2 & \cellcolor{third}12700 \\ 

& \textbf{FullTilt (ours)} & &
\cellcolor{best}\textbf{0.133 ± 0.007} & \cellcolor{best}\textbf{0.180 ± 0.007} & \cellcolor{best}\textbf{0.221 ± 0.008} & \cellcolor{best}\textbf{0.510} & \cellcolor{best}\textbf{3430} \\
\bottomrule
\end{tabular}%

\begin{tablenotes}[flushleft]\footnotesize
\item $+\mathcal{E}_{\text{Prompt}}$: with our multiclass visual prompt encoder module; MV: multi-view; n.a.: not applicable because the metric cannot be measured; T.: runtime (s); M.: VRAM (MB); $\downarrow$: lower is better; $\uparrow$: higher is better
\end{tablenotes}
\end{threeparttable}
}
\end{table*}

\noindent \textbf{3D Decoder.} The 3D decoder iteratively updates the content queries and 3D anchors (Fig.~\ref{fig:framework}d). Inspired by GroundingDINO~\cite{liu2024grounding}, we employ a 3-stage attention block repeated $L_{\text{D}}$ times. First, self-attention is applied to the content queries, followed by cross-attention between these queries and the class prototypes. To perform the third stage---deformable cross-attention with the 3D-aware features---we duplicate the content queries for each tilt image and project their 3D anchors $(x, y, z)$ into 2D coordinates $(\tilde{x}, \tilde{y})$ given the tilt angle $\theta$: $\tilde{x} = (x - \frac{W}{2})\cos\theta + (z - \frac{D}{2})\sin\theta + \frac{W}{2}$ and $\tilde{y} = y$, where $D$ is the tomogram depth. After computing deformable cross-attention using these projected coordinates, the output content queries are averaged across all tilt images. The 3D anchors are refined iteratively via a ``look forward twice'' approach~\cite{zhang2022dino}, which propagates gradients from the next decoder layer through the current layer's anchor refinement to stabilize training and accelerate convergence. Ultimately, for each $m\in \{1, \dots, M\}$, we take the final refined 3D anchor as the spatial estimation $\hat{\mathbf{P}}^m = (\hat{x}, \hat{y}, \hat{z}, \hat{d})$. The estimated class label $\hat{Y}^m$ and confidence score $s^m$ are derived from the inner product similarity between the final content query and the prototypes, computed as $\mathbf{q}_{\text{Content}}^{m}\mathbf{t}^{\top}\in \mathbb{R}^{N_c}$. More architecture details are in~\ref{sec:fulltilt_arch_details}.

\subsection{Training Objective} 
Following standard DETR architectures~\cite{jiang2024t, liu2024grounding}, we compute an L1 loss for bounding box regression and a focal loss for classification, applied after bipartite matching between model outputs and ground truth. To accelerate convergence, we incorporate auxiliary losses at each decoder layer~\cite{liu2024grounding} and employ a contrastive denoising strategy~\cite{zhang2022dino}, which additionally suppresses duplicate detections.

\subsection{Auxiliary Geometric Primitives ($\mathcal{G}$)}
Learning multi-view geometry directly from 2D tilt-series is exceptionally challenging, especially given adverse imaging artifacts (Fig.~\ref{fig:examples}). As physics-based cryo-ET simulations are computationally prohibitive and scale poorly, we introduce an auxiliary geometric primitives module ($\mathcal{G}$). This module generates synthetic 2D projections (e.g., circles of varying sizes) alongside their exact 3D ground truth coordinates \textit{on-the-fly}, eliminating I/O storage bottlenecks. $\mathcal{G}$ simulates diverse object distributions (e.g., dense clusters in the central $z$-plane) and mimics adverse imaging artifacts such as physical occlusion and non-uniform illumination. During training, we alternate between this dynamically generated data and the simulated cryo-ET data at every epoch. Details are provided in~\ref{sec:agp}.

\subsection{Inference and Interaction}
FullTilt's flexible design supports multiple interactive workflows. In a standard setup, a user selects a few target particles (e.g., 1–4 per class) from a reconstructed 3D tomogram to serve as prompts. For high-throughput analysis, users can extract class prototypes from one tomogram and reuse them across an entire dataset, enabling rapid zero-shot inference without manual re-prompting. In both cases, FullTilt automatically projects the 3D coordinates onto the 2D tilt images and seamlessly handles varying prompt quantities across views. In addition, users can successfully conduct full tilt-series inference by directly providing 2D prompts solely on the $0^\circ$ tilt image.

\begin{table*}[t]
\centering
\caption{\textbf{Zero-shot intra-instance detection results on the EMPIAR-10304 dataset.} The best-performing cryo-ET method for each metric is highlighted in bold. FullTilt achieves the best detection performance except for mAP@0.5r in the 2- and 4-prompt settings, with the shortest runtime and lowest VRAM usage among cryo-ET methods.}
\label{tab:intra_sample_10304}
\resizebox{0.7\textwidth}{!}{%
\begin{threeparttable}
\begin{tabular}{@{}ll|c|rrrrr@{}}
\toprule
\textbf{Category} &
\textbf{Method} &
\textbf{\# Prompts} & 
\textbf{mAP@0.5r $\uparrow$} & \textbf{mAP@1r $\uparrow$} &
\textbf{F1 $\uparrow$} &
\textbf{T. $\downarrow$} &
\textbf{M. $\downarrow$}
\\

\midrule
\multirow{2}{*}{General MV} & 
DETR3D~\cite{wang2022detr3d}$+\mathcal{E}_{\text{Prompt}}$ & \multirow{7}{*}{1} & 0.011 ± 0.001 & 0.091 ± 0.002 & 0.222 ± 0.003 & 0.136 & 2290 \\

& PETR~\cite{liu2022petr}$+\mathcal{E}_{\text{Prompt}}$ & & 0.000 ± 0.000 & 0.002 ± 0.000 & 0.032 ± 0.001 & 0.233 & 6020 \\

\cmidrule{1-2} \cmidrule{4-8}
\multirow{5}{*}{Cryo-ET} & 
Zeng et al.~\cite{zeng2022structure}$+\mathcal{E}_{\text{Prompt}}$ & &
n.a. & n.a. & 0.012 ± 0.001 & \cellcolor{second}30.2 & \cellcolor{second}2960 \\ 

 & TomoTwin~\cite{rice2023tomotwin} & &
\cellcolor{second}0.189 ± 0.057 & \cellcolor{second}0.339 ± 0.063 & \cellcolor{second}0.532 ± 0.061 & 1730 & 20800\\ 

 & CryoSAM~\cite{zhao2024cryosam} & &
n.a. & n.a. & 0.002 ± 0.002 & 388 & 28400 \\ 

 & ProPicker~\cite{wiedemann2026propicker} & &
n.a. & n.a. & \cellcolor{third}0.224 ± 0.014 & \cellcolor{third}49.2 & \cellcolor{third}12500 \\ 

 & \textbf{FullTilt (ours)} & &
\cellcolor{best}\textbf{0.199 ± 0.004} & \cellcolor{best}\textbf{0.437 ± 0.002} & \cellcolor{best}\textbf{0.593 ± 0.002} & \cellcolor{best}\textbf{0.229} & \cellcolor{best}\textbf{2410} \\

\midrule
\midrule
\multirow{2}{*}{General MV} & 
DETR3D~\cite{wang2022detr3d}$+\mathcal{E}_{\text{Prompt}}$ & \multirow{7}{*}{2} & 0.012 ± 0.001 & 0.096 ± 0.003 & 0.227 ± 0.002 & 0.134 & 2290 \\

& PETR~\cite{liu2022petr}$+\mathcal{E}_{\text{Prompt}}$ & & 0.000 ± 0.000 & 0.002 ± 0.000 & 0.032 ± 0.001 & 0.232 & 6020 \\

\cmidrule{1-2} \cmidrule{4-8}
\multirow{5}{*}{Cryo-ET} & Zeng et al.~\cite{zeng2022structure}$+\mathcal{E}_{\text{Prompt}}$ & &
n.a. & n.a. & 0.013 ± 0.001 & \cellcolor{second}30.2 & \cellcolor{second}2960 \\ 

& TomoTwin~\cite{rice2023tomotwin} & &
\cellcolor{best}\textbf{0.240 ± 0.046} & \cellcolor{second}0.373 ± 0.050 & \cellcolor{second}0.569 ± 0.048 & 1720 & 20800 \\ 

& CryoSAM~\cite{zhao2024cryosam} & &
n.a. & n.a. & 0.004 ± 0.003 & 394 & 28400 \\ 

& ProPicker~\cite{wiedemann2026propicker} & &
n.a. & n.a. & \cellcolor{third}0.239 ± 0.021 & \cellcolor{third}49.1 & \cellcolor{third}12500 \\ 

& \textbf{FullTilt (ours)}  & &
\cellcolor{second}0.200 ± 0.004 & \cellcolor{best}\textbf{0.437 ± 0.003} & \cellcolor{best}\textbf{0.592 ± 0.002} & \cellcolor{best}\textbf{0.230} & \cellcolor{best}\textbf{2410} \\

\midrule
\midrule
\multirow{2}{*}{General MV} & 
DETR3D~\cite{wang2022detr3d}$+\mathcal{E}_{\text{Prompt}}$ & \multirow{7}{*}{4} & 0.012 ± 0.001 & 0.100 ± 0.003 & 0.230 ± 0.001 & 0.134 & 2290 \\

& PETR~\cite{liu2022petr}$+\mathcal{E}_{\text{Prompt}}$ & & 0.000 ± 0.000 & 0.002 ± 0.000 & 0.032 ± 0.001 & 0.232 & 6020 \\

\cmidrule{1-2} \cmidrule{4-8}
\multirow{5}{*}{Cryo-ET} & Zeng et al.~\cite{zeng2022structure}$+\mathcal{E}_{\text{Prompt}}$ & &
n.a. & n.a. & 0.012 ± 0.001 & \cellcolor{second}30.2 & \cellcolor{second}2960 \\ 

& TomoTwin~\cite{rice2023tomotwin} & &
\cellcolor{best}\textbf{0.241 ± 0.022} & \cellcolor{second}0.367 ± 0.026 & \cellcolor{second}0.565 ± 0.030 & 1720 & 20800 \\ 

& CryoSAM~\cite{zhao2024cryosam} & &
n.a. & n.a. & 0.008 ± 0.002 & 396 & 28400 \\ 

& ProPicker~\cite{wiedemann2026propicker} & &
n.a. & n.a. & \cellcolor{third}0.247 ± 0.016	 & \cellcolor{third}50.2 & \cellcolor{third}12500 \\ 

& \textbf{FullTilt (ours)} & &
\cellcolor{second}0.201 ± 0.004 & \cellcolor{best}\textbf{0.439 ± 0.001} & \cellcolor{best}\textbf{0.593 ± 0.002} & \cellcolor{best}\textbf{0.230} & \cellcolor{best}\textbf{2410} \\
\bottomrule
\end{tabular}%

\begin{tablenotes}[flushleft]\footnotesize
\item $+\mathcal{E}_{\text{Prompt}}$: with our multiclass visual prompt encoder module; MV: multi-view; n.a.: not applicable because the metric cannot be measured; T.: runtime (s); M.: VRAM (MB); $\downarrow$: lower is better; $\uparrow$: higher is better
\end{tablenotes}
\end{threeparttable}
}
\end{table*}

\begin{table*}[t]
\centering
\caption{\textbf{Zero-shot intra-instance detection results on the EMPIAR-10499 dataset.} The best-performing cryo-ET method for each metric is highlighted in bold. FullTilt achieves the best performance across all evaluated metrics, with the shortest runtime and lowest VRAM usage among cryo-ET methods.}
\label{tab:intra_sample_10499}
\resizebox{0.7\textwidth}{!}{%
\begin{threeparttable}
\begin{tabular}{@{}ll|c|rrrrr@{}}
\toprule
\textbf{Category} &
\textbf{Method} &
\textbf{\# Prompts} & 
\textbf{mAP@0.5r $\uparrow$} & \textbf{mAP@1r $\uparrow$} &
\textbf{F1 $\uparrow$} &
\textbf{T. $\downarrow$} &
\textbf{M. $\downarrow$}
\\

\midrule
\multirow{2}{*}{General MV} & 
DETR3D~\cite{wang2022detr3d}$+\mathcal{E}_{\text{Prompt}}$ & \multirow{7}{*}{1} & 0.039 ± 0.002 & 0.137 ± 0.004 & 0.271 ± 0.006 & 0.198 & 3230 \\

& PETR~\cite{liu2022petr}$+\mathcal{E}_{\text{Prompt}}$ & & 0.000 ± 0.000 & 0.000 ± 0.000 & 0.017 ± 0.000 & 0.421 & 10600 \\ 

\cmidrule{1-2} \cmidrule{4-8}
\multirow{5}{*}{Cryo-ET} & 
Zeng et al.~\cite{zeng2022structure}$+\mathcal{E}_{\text{Prompt}}$ & &
n.a. & n.a. & 0.008 ± 0.001 & \cellcolor{second}26.9 & \cellcolor{second}4710 \\ 

 & TomoTwin~\cite{rice2023tomotwin} & &
\cellcolor{second}0.035 ± 0.008 & \cellcolor{second}0.113 ± 0.020 & \cellcolor{second}0.249 ± 0.027 & 2330 & 20800 \\ 

 & CryoSAM~\cite{zhao2024cryosam} & &
n.a. & n.a. & 0.032 ± 0.002 & 421 & 29200 \\ 

 & ProPicker~\cite{wiedemann2026propicker} & &
n.a. & n.a. & \cellcolor{third}0.187 ± 0.011 & \cellcolor{third}69.1 & \cellcolor{third}12700 \\ 

 & \textbf{FullTilt (ours)} & &
\cellcolor{best}\textbf{0.159 ± 0.004} & \cellcolor{best}\textbf{0.288 ± 0.004} & \cellcolor{best}\textbf{0.418 ± 0.005} & \cellcolor{best}\textbf{0.389} & \cellcolor{best}\textbf{3410} \\

\midrule
\midrule
\multirow{2}{*}{General MV} & 
DETR3D~\cite{wang2022detr3d}$+\mathcal{E}_{\text{Prompt}}$ & \multirow{7}{*}{2} & 0.049 ± 0.002 & 0.160 ± 0.004 & 0.299 ± 0.005 & 0.202 & 3230 \\

& PETR~\cite{liu2022petr}$+\mathcal{E}_{\text{Prompt}}$ & & 0.000 ± 0.000 & 0.000 ± 0.000 & 0.017 ± 0.000 & 0.423 & 10600 \\

\cmidrule{1-2} \cmidrule{4-8}
\multirow{5}{*}{Cryo-ET} & Zeng et al.~\cite{zeng2022structure}$+\mathcal{E}_{\text{Prompt}}$ & &
n.a. & n.a. & 0.008 ± 0.001 & \cellcolor{second}26.8 & \cellcolor{second}4710 \\ 

& TomoTwin~\cite{rice2023tomotwin} & &
\cellcolor{second}0.057 ± 0.011 & \cellcolor{second}0.155 ± 0.023 & \cellcolor{second}0.296 ± 0.032 & 2320 & 20800 \\ 

& CryoSAM~\cite{zhao2024cryosam} & &
n.a. & n.a. & 0.034 ± 0.001 & 444 & 29200 \\ 

& ProPicker~\cite{wiedemann2026propicker} & &
n.a. & n.a. & \cellcolor{third}0.206 ± 0.012 & \cellcolor{third}72.8 & \cellcolor{third}12700 \\ 

& \textbf{FullTilt (ours)}  & &
\cellcolor{best}\textbf{0.169 ± 0.006} & \cellcolor{best}\textbf{0.302 ± 0.009} & \cellcolor{best}\textbf{0.432 ± 0.010} & \cellcolor{best}\textbf{0.394} & \cellcolor{best}\textbf{3410} \\

\midrule
\midrule
\multirow{2}{*}{General MV} & 
DETR3D~\cite{wang2022detr3d}$+\mathcal{E}_{\text{Prompt}}$ & \multirow{7}{*}{4} & 0.056 ± 0.001 & 0.177 ± 0.002 & 0.317 ± 0.001 & 0.201 & 3230 \\

& PETR~\cite{liu2022petr}$+\mathcal{E}_{\text{Prompt}}$ & & 0.000 ± 0.000 & 0.000 ± 0.000 & 0.016 ± 0.000 & 0.423 & 10600 \\ 

\cmidrule{1-2} \cmidrule{4-8}
\multirow{5}{*}{Cryo-ET} & Zeng et al.~\cite{zeng2022structure}$+\mathcal{E}_{\text{Prompt}}$ & &
n.a. & n.a. & 0.008 ± 0.001 & \cellcolor{second}26.9 & \cellcolor{second}4710 \\ 

& TomoTwin~\cite{rice2023tomotwin} & &
\cellcolor{second}0.078 ± 0.008 & \cellcolor{second}0.198 ± 0.016 & \cellcolor{second}0.354 ± 0.018 & 2310 & 20800 \\ 

& CryoSAM~\cite{zhao2024cryosam} & &
n.a. & n.a. & 0.038 ± 0.001 & 486 & 29200 \\ 

& ProPicker~\cite{wiedemann2026propicker} & &
n.a. & n.a. & \cellcolor{third}0.235 ± 0.012 & \cellcolor{third}71.8 & \cellcolor{third}12700 \\ 

& \textbf{FullTilt (ours)} & &
\cellcolor{best}\textbf{0.173 ± 0.002} & \cellcolor{best}\textbf{0.309 ± 0.003} & \cellcolor{best}\textbf{0.441 ± 0.005} & \cellcolor{best}\textbf{0.394} & \cellcolor{best}\textbf{3410} \\

\bottomrule
\end{tabular}%

\begin{tablenotes}[flushleft]\footnotesize
\item $+\mathcal{E}_{\text{Prompt}}$: with our multiclass visual prompt encoder module; MV: multi-view; n.a.: not applicable because the metric cannot be measured; T.: runtime (s); M.: VRAM (MB); $\downarrow$: lower is better; $\uparrow$: higher is better
\end{tablenotes}
\end{threeparttable}
}
\end{table*}

\begin{table*}[h]
\centering
\caption{\textbf{Zero-shot cross-instance detection results on real-world cryo-ET datasets.} The best-performing open-set method for each metric is highlighted in bold. FullTilt outperforms all baselines, excepting mAP@0.5r under 2- and 4-prompt settings on EMPIAR-10304.}
\label{tab:cross_sample_real}
\resizebox{1.0\textwidth}{!}{%
\begin{threeparttable}
\begin{tabular}{@{}ll|c|rrr|rrr|rrr@{}}
\toprule
& & & \multicolumn{3}{c}{\textbf{DS-10440}} & \multicolumn{3}{c}{\textbf{EMPIAR-10304}} & \multicolumn{3}{c}{\textbf{EMPIAR-10499}}\\ 
\cmidrule(lr){4-6}\cmidrule(lr){7-9}\cmidrule(lr){10-12}
\textbf{Cat.} &
\textbf{Method} &
\textbf{\# P.} & 
\textbf{mAP@0.5r $\uparrow$} & \textbf{mAP@1r $\uparrow$} &
\textbf{F1 $\uparrow$} &
\textbf{mAP@0.5r $\uparrow$} & \textbf{mAP@1r $\uparrow$} &
\textbf{F1 $\uparrow$} &
\textbf{mAP@0.5r $\uparrow$} & \textbf{mAP@1r $\uparrow$} &
\textbf{F1 $\uparrow$}
\\

\midrule
\multirow{2}{*}{WSL} & 
Huang et al.~\cite{huang2022accurate} & \multirow{5}{*}{1} & - & - & - & - & - & - & - & - & - \\ 
& TomoPicker~\cite{uddin2024tomopicker} & &
0.002 ± 0.001 & 0.035 ± 0.006 & 0.061 ± 0.010 &
0.027 ± 0.030 & 0.056 ± 0.038 & 0.205 ± 0.108 &
0.002 ± 0.001 & 0.017 ± 0.012 & 0.092 ± 0.055 \\  

\cmidrule{1-2} \cmidrule{4-12}
\multirow{3}{*}{OS} & TomoTwin~\cite{rice2023tomotwin} & &
\cellcolor{second}0.077 ± 0.026 & \cellcolor{second}0.121 ± 0.029 & \cellcolor{second}0.095 ± 0.026 &
\cellcolor{second}0.203 ± 0.137 & \cellcolor{second}0.348 ± 0.189 & \cellcolor{second}0.508 ± 0.214 &
\cellcolor{second}0.049 ± 0.051 & \cellcolor{second}0.164 ± 0.136 & \cellcolor{second}0.265 ± 0.196 \\ 

 & ProPicker~\cite{wiedemann2026propicker} & &
n.a. & n.a. & \cellcolor{third}0.076 ± 0.013 &
n.a. & n.a. & \cellcolor{third}0.221 ± 0.071 &
n.a. & n.a. & \cellcolor{third}0.196 ± 0.108 \\ 

 & \textbf{FullTilt (ours)} & &
\cellcolor{best}\textbf{0.112 ± 0.033} & \cellcolor{best}\textbf{0.137 ± 0.036} & \cellcolor{best}\textbf{0.167 ± 0.037} &
\cellcolor{best}\textbf{0.204 ± 0.006} & \cellcolor{best}\textbf{0.439 ± 0.005} & \cellcolor{best}\textbf{0.593 ± 0.004} &
\cellcolor{best}\textbf{0.149 ± 0.051} & \cellcolor{best}\textbf{0.271 ± 0.083} & \cellcolor{best}\textbf{0.394 ± 0.134} \\

\midrule
\midrule

\multirow{2}{*}{WSL} & 
Huang et al.~\cite{huang2022accurate} & \multirow{5}{*}{2} &
0.000 ± 0.000 & 0.000 ± 0.000 & 0.000 ± 0.000 &
0.000 ± 0.000 & 0.000 ± 0.000 & 0.000 ± 0.000 &
0.000 ± 0.000 & 0.000 ± 0.000 & 0.000 ± 0.000\\ 
& TomoPicker~\cite{uddin2024tomopicker} & &
0.002 ± 0.001 & 0.033 ± 0.009 & 0.055 ± 0.013 &
0.100 ± 0.131 & 0.149 ± 0.141 & 0.321 ± 0.188 &
0.003 ± 0.001 & 0.039 ± 0.019 & 0.159 ± 0.044 \\ 

\cmidrule{1-2} \cmidrule{4-12}
\multirow{3}{*}{OS} & TomoTwin~\cite{rice2023tomotwin}  & &
\cellcolor{second}0.091 ± 0.044 & \cellcolor{second}0.140 ± 0.057 & \cellcolor{second}0.078 ± 0.042 &
\cellcolor{best}\textbf{0.216 ± 0.152} & \cellcolor{second}0.336 ± 0.177 & \cellcolor{second}0.503 ± 0.206 &
\cellcolor{second}0.055 ± 0.045 & \cellcolor{second}0.143 ± 0.105 & \cellcolor{second}0.265 ± 0.158 \\ 

& ProPicker~\cite{wiedemann2026propicker} & &
n.a. & n.a. & \cellcolor{third}0.068 ± 0.018 &
n.a. & n.a. & \cellcolor{third}0.218 ± 0.050 &
n.a. & n.a. & \cellcolor{third}0.220 ± 0.109 \\ 

& \textbf{FullTilt (ours)} & &
\cellcolor{best}\textbf{0.124 ± 0.020} & \cellcolor{best}\textbf{0.154 ± 0.022} & \cellcolor{best}\textbf{0.186 ± 0.025} &
\cellcolor{second}0.199 ± 0.003 & \cellcolor{best}\textbf{0.437 ± 0.003} & \cellcolor{best}\textbf{0.591 ± 0.002} &
\cellcolor{best}\textbf{0.147 ± 0.039} & \cellcolor{best}\textbf{0.273 ± 0.062} & \cellcolor{best}\textbf{0.401 ± 0.091} \\

\midrule
\midrule

\multirow{2}{*}{WSL} & 
Huang et al.~\cite{huang2022accurate} & \multirow{5}{*}{4} & 0.000 ± 0.000 & 0.000 ± 0.000 & 0.004 ± 0.002 &
0.000 ± 0.000 & 0.000 ± 0.000 & 0.000 ± 0.000 &
0.000 ± 0.000 & 0.000 ± 0.000 & 0.000 ± 0.000 \\ 

& TomoPicker~\cite{uddin2024tomopicker} & &
0.006 ± 0.001 & 0.042 ± 0.004 & 0.080 ± 0.011 &
0.074 ± 0.054 & 0.215 ± 0.117 & 0.421 ± 0.127 &
0.018 ± 0.007 & 0.107 ± 0.020 & 0.256 ± 0.012 \\

\cmidrule{1-2} \cmidrule{4-12}
\multirow{3}{*}{OS} & TomoTwin~\cite{rice2023tomotwin}  & &
\cellcolor{second}0.106 ± 0.026 & \cellcolor{second}0.160 ± 0.038 & \cellcolor{second}0.111 ± 0.061 &
 \cellcolor{best}\textbf{0.238 ± 0.167} & \cellcolor{second}0.357 ± 0.211 & \cellcolor{second}0.506 ± 0.262 &
 \cellcolor{second}0.064 ± 0.066 & \cellcolor{second}0.163 ± 0.123 & \cellcolor{second}0.272 ± 0.186 \\ 

 & ProPicker~\cite{wiedemann2026propicker} & &
n.a. & n.a. & \cellcolor{third}0.083 ± 0.011 &
n.a. & n.a. & \cellcolor{third}0.249 ± 0.035 &
n.a. & n.a. & \cellcolor{third}0.191 ± 0.079 \\ 

 & \textbf{FullTilt (ours)} & &
\cellcolor{best}\textbf{0.136 ± 0.009} & \cellcolor{best}\textbf{0.169 ± 0.009} & \cellcolor{best}\textbf{0.200 ± 0.013} &
\cellcolor{second}0.200 ± 0.008 & \cellcolor{best}\textbf{0.437 ± 0.006} & \cellcolor{best}\textbf{0.591 ± 0.004} &
\cellcolor{best}\textbf{0.177 ± 0.019} & \cellcolor{best}\textbf{0.316 ± 0.023} & \cellcolor{best}\textbf{0.456 ± 0.022} \\

\bottomrule
\end{tabular}%

\begin{tablenotes}[flushleft]\footnotesize
\item Cat.: category; \# P.: \# prompts; WSL: weakly-supervised learning; OS: open-set; -: Huang et al. needs at least 2 prompts; n.a.: not applicable because the metric cannot be measured; $\downarrow$: lower is better; $\uparrow$: higher is better
\end{tablenotes}
\end{threeparttable}
}
\end{table*}

\section{Experiments}
\subsection{Datasets and Metrics}
\noindent \textbf{Training Dataset.} To ensure fair comparisons with open-set methods~\cite{rice2023tomotwin, wiedemann2026propicker}, we generate 1,138 simulated tomograms and corresponding aligned tilt-series (noisy and clean) using the same 119 proteins and established TEM simulation protocol from prior work~\cite{rice2023tomotwin, rullgaard2011simulation}.

\noindent \textbf{Evaluation Dataset.} We evaluate FullTilt's zero-shot performance on three public real-world cryo-ET datasets: CZII CryoET Object Identification Challenge (DS-10440)~\cite{peck2025realistic}, EMPIAR-10304~\cite{eisenstein2019improved}, and EMPIAR-10499~\cite{tegunov2021multi}. 
We additionally evaluate on a simulated dataset containing 10 unseen proteins alongside a training subset. Further dataset preparation details are provided in~\ref{sec:data_details}.

\noindent \textbf{Metrics.} We report mean Average Precision (mAP) at distance thresholds of 0.5 and 1.0 times the target radius (mAP@0.5r, mAP@1.0r)~\cite{huang2022accurate}. As some baselines lack confidence scores, we supplement this with an F1 score (1.0 radius threshold). In all cases, inference runtime (T.) and peak VRAM (M.) are measured on a single NVIDIA RTX 6000 Ada GPU and 48-core CPU. Further details regarding the benchmark settings are provided in~\ref{sec:supp_benchmark_details}.

\subsection{Implementation Details}
Following previous work~\cite{zhang2022dino, liu2024grounding, jiang2024t}, we use the Swin Transformer~\cite{liu2021swin} as our 2D backbone, setting $L_{\text{AA}}$, $L_{\text{VP}}$, and $L_{\text{D}}$ to 6, $C$ to 256, and $M$ to 900. FullTilt is implemented in PyTorch 2.5.1 and trained for 200 epochs via AdamW~\cite{loshchilovdecoupled} (learning rate $1 \times 10^{-4}$). Training is distributed across four NVIDIA RTX 6000 Ada GPUs with a per-GPU batch size of 1. Further details regarding training and baseline implementations are provided in~\ref{sec:training_details} and \ref{sec:baseline_details}, respectively.

\subsection{Intra-Instance Detection}
For intra-instance detection (i.e., prompts and targets reside in the same instance), we evaluate FullTilt against zero-shot open-set methods (TomoTwin~\cite{rice2023tomotwin}, CryoSAM~\cite{zhao2024cryosam}, ProPicker~\cite{wiedemann2026propicker}). To benchmark against 2D back-projection (Zeng's method~\cite{zeng2022structure}) and general multi-view frameworks (DETR3D~\cite{wang2022detr3d}, PETR~\cite{liu2022petr}), we integrated our prompt module ($\mathcal{E}_{\text{Prompt}}$) into their architectures. Fully supervised methods are omitted due to our focus on zero-shot detection and the absence of separate training sets.

We evaluate 1, 2, and 4 prompts over 10 independent trials (Tables~\ref{tab:intra_sample_czii}--\ref{tab:intra_sample_10499}; extended details and qualitative results in~\ref{sec:supp_eval_intra_sample} and~\ref{sec:supp_result_visual}). Expectedly, providing more prompts generally enhances performance (except EMPIAR-10304). FullTilt achieves comparable or superior real-world performance across all metrics, reducing inference to under one second with the lowest VRAM requirements. Notably, FullTilt's margin over TomoTwin grows on challenging native cellular (EMPIAR-10499) versus purified (EMPIAR-10304) ribosomes. Among the baselines, TomoTwin outperforms ProPicker but incurs significantly longer runtimes, whereas CryoSAM struggles because its natural-image pre-training translates poorly to cryo-ET~\cite{wiedemann2026propicker}. Despite being trained on identical data, general multi-view models (DETR3D, PETR) fail to effectively fuse tilt-series information. PETR's particularly poor performance indicates that attention across all tilts in the decoder cannot capture critical features.

\subsection{Cross-Instance Detection}
For cross-instance detection, a prompt from one instance is used for dataset-wide inference. We compare FullTilt against the top open-set baselines (TomoTwin~\cite{rice2023tomotwin}, ProPicker~\cite{wiedemann2026propicker}) and WSL methods (Huang's method~\cite{huang2022accurate}, TomoPicker~\cite{uddin2024tomopicker}). We randomly select one instance per dataset to extract prototype embeddings or train WSL models, using 1, 2, or 4 prompts. Experiments span 10 trials (limited to 3 for WSL models due to extensive training times). As shown in Table~\ref{tab:cross_sample_real}, FullTilt outperforms baselines across all real-world datasets, except for a marginally lower mAP@0.5r than TomoTwin on EMPIAR-10304. As expected, WSL methods fail to generalize effectively with so few prompts, though TomoPicker surpasses Huang's method (details provided in~\ref{sec:supp_eval_croo_sample}).


\subsection{Multiclass Prompting Detection}
Finally, we evaluate $\mathcal{E}_{\text{Prompt}}$ in multiclass scenarios for both intra- and cross-instance settings using the 6-class DS-10440 dataset. We test all possible class combinations over 10 independent trials (Table~\ref{tab:multiclass_intra} and~\ref{sec:additional_multiclass_prompting}). Although querying more classes marginally compromises detection performance, FullTilt processes all requested classes simultaneously with negligible runtime and VRAM overhead.

\subsection{Ablation Studies}

\noindent \textbf{Number of Tilt Images vs. Projections per Prompt.} We evaluate FullTilt under two distinct data reduction scenarios (Table~\ref{tab:ablation_sparsity_prompts}). First, retaining fewer tilt images (by progressively doubling the angle increment) proportionally degrades detection performance, though it correspondingly improves runtime and VRAM usage. Conversely, supplying visual prompts in only a sparse subset of images, down to an extreme ``single-image prompt'' setting, only slightly compromises performance. More ablation experiments are provided in~\ref{sec:supp_abl}.


\begin{table*}[h]
\centering
\caption{\textbf{Multi-class prompting performance comparison (intra-instance).} FullTilt maintains stable computational efficiency regardless of the number of target classes.}
\label{tab:multiclass_intra}
\resizebox{\textwidth}{!}{%
\begin{threeparttable}
\begin{tabular}{@{}l|rrrr|rrrr|rrrr@{}}
\toprule
 & \multicolumn{4}{c}{\textbf{1 Prompt}} & \multicolumn{4}{c}{\textbf{2 Prompts}} & \multicolumn{4}{c}{\textbf{4 Prompts}}\\ 
\cmidrule(lr){2-5}\cmidrule(lr){6-9}\cmidrule(lr){10-13}
\textbf{\# Classes} & 
\textbf{mAP@0.5r $\uparrow$} & \textbf{mAP@1r $\uparrow$} &
\textbf{T. $\downarrow$} & \textbf{M. $\downarrow$} &
\textbf{mAP@0.5r $\uparrow$} & \textbf{mAP@1r $\uparrow$} &
\textbf{T. $\downarrow$} & \textbf{M. $\downarrow$} &
\textbf{mAP@0.5r $\uparrow$} & \textbf{mAP@1r $\uparrow$} &
\textbf{T. $\downarrow$} & \textbf{M. $\downarrow$}
\\

\midrule
1 &
0.118 ± 0.013 & 0.148 ± 0.013 & 0.499 & 3430 &
0.129 ± 0.010 & 0.161 ± 0.011 & 0.503 & 3430 &
0.133 ± 0.007 & 0.180 ± 0.007 & 0.510 & 3430 \\ 
2 &
0.112 ± 0.005 & 0.142 ± 0.004 & 0.518 & 3430 &
0.123 ± 0.003 & 0.155 ± 0.003 & 0.522 & 3430 &
0.127 ± 0.003 & 0.161 ± 0.003 & 0.523 & 3430 \\
3 &
0.111 ± 0.004 & 0.142 ± 0.004 & 0.521 & 3440 &
0.119 ± 0.002 & 0.151 ± 0.002 & 0.524 & 3440 &
0.123 ± 0.001 & 0.156 ± 0.001 & 0.525 & 3440 \\
4 &
0.111 ± 0.002 & 0.142 ± 0.003 & 0.505 & 3440 &
0.118 ± 0.002 & 0.151 ± 0.001 & 0.507 & 3440 &
0.120 ± 0.001 & 0.153 ± 0.001 & 0.510 & 3440 \\
5 &
0.108 ± 0.003 & 0.139 ± 0.004 & 0.514 & 3440 &
0.118 ± 0.002 & 0.150 ± 0.002 & 0.521 & 3440 &
0.119 ± 0.001 & 0.150 ± 0.002 & 0.524 & 3440 \\
6 & 
0.113 ± 0.008 & 0.144 ± 0.007 & 0.508 & 3440 &
0.115 ± 0.006 & 0.147 ± 0.006 & 0.507 & 3440 &
0.117 ± 0.003 & 0.149 ± 0.003 & 0.513 & 3440 \\
\bottomrule
\end{tabular}%

\begin{tablenotes}[flushleft]\footnotesize
\item T.: runtime (s); M.: VRAM (MB); $\downarrow$: lower is better; $\uparrow$: higher is better
\end{tablenotes}
\end{threeparttable}
}
\end{table*}

\begin{table*}[h]
\centering
\caption{\textbf{Ablation study on the number of tilt images and the number of projections per prompt.} Reducing the number of tilt images decreases runtime and memory at the cost of overall detection performance; reducing the number of projections per prompt only slightly compromises performance.}
\label{tab:ablation_sparsity_prompts}
\resizebox{\textwidth}{!}{%
\begin{threeparttable}
\begin{tabular}{@{}l|rrrr|rrrr|rrrr@{}}
\toprule
 & \multicolumn{4}{c}{\textbf{DS-10440}} & \multicolumn{4}{c}{\textbf{EMPIAR-10304}} & \multicolumn{4}{c}{\textbf{EMPIAR-10499}}\\ 
\cmidrule(lr){2-5}\cmidrule(lr){6-9}\cmidrule(lr){10-13}
\textbf{Configuration} & 
\textbf{mAP@0.5r $\uparrow$} & \textbf{mAP@1r $\uparrow$} &
\textbf{T. $\downarrow$} & \textbf{M. $\downarrow$} &
\textbf{mAP@0.5r $\uparrow$} & \textbf{mAP@1r $\uparrow$} &
\textbf{T. $\downarrow$} & \textbf{M. $\downarrow$} &
\textbf{mAP@0.5r $\uparrow$} & \textbf{mAP@1r $\uparrow$} &
\textbf{T. $\downarrow$} & \textbf{M. $\downarrow$}
\\

\midrule
Baseline (100\%) & 
0.118 ± 0.013 & 0.148 ± 0.013 & 0.499 & 3430 &
0.199 ± 0.004 & 0.437 ± 0.002 & 0.229 & 2410 &
0.159 ± 0.004 & 0.288 ± 0.004 & 0.389 & 3410 \\

\midrule
\multicolumn{13}{l}{\textit{Ablation: Reduced Tilt Images}} \\
\midrule
50\% & 
0.100 ± 0.013 & 0.133 ± 0.011 & 0.246 & 1860 & 
0.184 ± 0.002 & 0.427 ± 0.002 & 0.118 & 1380 &
0.127 ± 0.006 & 0.260 ± 0.007 & 0.202 & 1950 \\

25\% & 
0.085 ± 0.010 & 0.120 ± 0.011 & 0.153 & 1280 &
0.157 ± 0.003 & 0.391 ± 0.003 & 0.079 & 860 &
0.060 ± 0.005 & 0.165 ± 0.008 & 0.116 & 1220 \\

12.5\% & 
0.038 ± 0.007 & 0.073 ± 0.014 & 0.087 & 700 & 
0.084 ± 0.003 & 0.268 ± 0.003 & 0.069 & 600 &
0.010 ± 0.001 & 0.048 ± 0.003 & 0.083 & 790 \\

\midrule
\multicolumn{13}{l}{\textit{Ablation: Reduced Projections per Prompt}} \\
\midrule
50\% & 
0.111 ± 0.008 & 0.142 ± 0.009 & 0.514 & 3430 &
0.199 ± 0.003 & 0.437 ± 0.002 & 0.223 & 2420 &
0.162 ± 0.005 & 0.290 ± 0.005 & 0.392 & 3410 \\

25\% & 
0.107 ± 0.013 & 0.136 ± 0.014 & 0.513 & 3430 & 
0.199 ± 0.004 & 0.437 ± 0.003 & 0.223 & 2420 &
0.158 ± 0.006 & 0.285 ± 0.007 & 0.391 & 3410 \\

12.5\% & 
0.108 ± 0.009 & 0.137 ± 0.010 & 0.512 & 3430 &
0.197 ± 0.005 & 0.436 ± 0.003 & 0.224 & 2420 &
0.158 ± 0.005 & 0.285 ± 0.006 & 0.391 & 3410 \\

Only One & 
0.106 ± 0.014 & 0.134 ± 0.016 & 0.513 & 3430 &
0.190 ± 0.004 & 0.430 ± 0.004 & 0.224 & 2420 &
0.155 ± 0.009 & 0.278 ± 0.011 & 0.390 & 3410 \\

\bottomrule
\end{tabular}%

\begin{tablenotes}[flushleft]\footnotesize
\item T.: runtime (s); M.: VRAM (MB); $\downarrow$: lower is better; $\uparrow$: higher is better
\end{tablenotes}
\end{threeparttable}
}
\end{table*}

\begin{table*}[h!]
\centering
\caption{\textbf{Ablation study on FullTilt components.} We ablate the tilt-series encoder ($\mathcal{E}_{\text{Tilt}}$), tilt-aware query initializer ($\mathcal{I}_{\text{Tilt}}$), and auxiliary geometric primitives ($\mathcal{G}$) on detection performance across three real-world datasets. The integration of all three modules yields the optimal performance.}
\label{tab:ablation_module_real}
\resizebox{\textwidth}{!}{%
\begin{threeparttable}
\begin{tabular}{@{}ccc|rrrr|rrrr|rrrr@{}}
\toprule
 & & & \multicolumn{4}{c}{\textbf{DS-10440}} & \multicolumn{4}{c}{\textbf{EMPIAR-10304}} & \multicolumn{4}{c}{\textbf{EMPIAR-10499}}\\ 
\cmidrule(lr){4-7}\cmidrule(lr){8-11}\cmidrule(lr){12-15}
$\mathcal{E}_{\text{Tilt}}$ & $\mathcal{I}_{\text{Tilt}}$ & $\mathcal{G}$ & 
\textbf{mAP@0.5r $\uparrow$} & \textbf{mAP@1r $\uparrow$} &
\textbf{T. $\downarrow$} & \textbf{M. $\downarrow$} &
\textbf{mAP@0.5r $\uparrow$} & \textbf{mAP@1r $\uparrow$} &
\textbf{T. $\downarrow$} & \textbf{M. $\downarrow$} &
\textbf{mAP@0.5r $\uparrow$} & \textbf{mAP@1r $\uparrow$} &
\textbf{T. $\downarrow$} & \textbf{M. $\downarrow$}
\\

\midrule

& & & 
0.001 ± 0.001 & 0.003 ± 0.001 & 0.215 & 3430 &
0.000 ± 0.000 & 0.001 ± 0.000 & 0.133 & 2280 &
0.000 ± 0.000 & 0.000 ± 0.000 & 0.200 & 3170 \\

\midrule

\checkmark & & & 
0.002 ± 0.001 & 0.008 ± 0.001 & 0.519 & 3430 &
0.000 ± 0.000 & 0.005 ± 0.001 & 0.229 & 2300 &
0.000 ± 0.000 & 0.002 ± 0.000 & 0.380 & 3190 \\

& \checkmark & & 
0.040 ± 0.002 & 0.099 ± 0.003 & 0.223 & 3430 &
0.004 ± 0.000 & 0.106 ± 0.003 & 0.141 & 2400 &
0.009 ± 0.001 & 0.082 ± 0.003 & 0.207 & 3380 \\

& & \checkmark & 
0.025 ± 0.002 & 0.063 ± 0.003 & 0.211 & 3430 &
0.000 ± 0.000 & 0.016 ± 0.001 & 0.139 & 2280 &
0.001 ± 0.000 & 0.011 ± 0.001 & 0.198 & 3170 \\

\midrule

\checkmark & \checkmark & & 
0.099 ± 0.007 & 0.131 ± 0.008 & 0.517 & 3430 &
0.008 ± 0.001 & 0.136 ± 0.009 & 0.228 & 2420 &
0.049 ± 0.003 & 0.168 ± 0.006 & 0.392 & 3390 \\

& \checkmark & \checkmark & 
0.094 ± 0.004 & 0.153 ± 0.005 & 0.222 & 3430 &
0.030 ± 0.001 & 0.181 ± 0.003 & 0.142 & 2400 &
0.033 ± 0.001 & 0.127 ± 0.003 & 0.207 & 3380 \\

\checkmark & & \checkmark & 
0.002 ± 0.001 & 0.008 ± 0.002 & 0.520 & 3430 &
0.001 ± 0.000 & 0.017 ± 0.001 & 0.229 & 2300 &
0.000 ± 0.000 & 0.004 ± 0.000 & 0.379 & 3190 \\

\midrule

\checkmark & \checkmark & \checkmark & 
0.118 ± 0.013 & 0.148 ± 0.013 & 0.499 & 3430 &
0.199 ± 0.004 & 0.437 ± 0.002 & 0.229 & 2410 &
0.159 ± 0.004 & 0.288 ± 0.004 & 0.389 & 3410 \\

\bottomrule
\end{tabular}%

\begin{tablenotes}[flushleft]\footnotesize
\item \checkmark: has this feature; $\mathcal{E}_{\text{Tilt}}$: tilt-series encoder; $\mathcal{I}_{\text{Tilt}}$: tilt-aware query initializer; $\mathcal{G}$: auxiliary geometric primitives; T.: runtime (s); M.: VRAM (MB); $\downarrow$: lower is better; $\uparrow$: higher is better
\end{tablenotes}
\end{threeparttable}
}
\end{table*}


\noindent \textbf{Component Ablation.} We ablate our proposed modules: the tilt-series encoder ($\mathcal{E}_{\text{Tilt}}$), tilt-aware query initializer ($\mathcal{I}_{\text{Tilt}}$), and auxiliary geometric primitives module ($\mathcal{G}$) (Table~\ref{tab:ablation_module_real}; granular design ablations in~\ref{sec:abl_tilt_encoder} and~\ref{sec:abl_tilt_initializer}). Without these modules, FullTilt fails. Among single-module configurations, only $\mathcal{I}_{\text{Tilt}}$ yields viable detections. Integrating $\mathcal{E}_{\text{Tilt}}$ subsequently drives a dramatic performance surge, underscoring the absolute necessity of cross-tilt fusion for cryo-ET data. Finally, adding $\mathcal{G}$ further boosts performance by enhancing the model's understanding of multi-view geometry and improving its robustness to adverse imaging artifacts. Overall, integrating all three modules yields the best performance across the evaluated metrics, except for mAP@1r on DS-10440.

\section{Conclusion}
In this work, we introduced FullTilt, an end-to-end framework that redefines open-set 3D macromolecule detection by operating directly on aligned 2D cryo-ET tilt-series. By bypassing computationally expensive 3D sliding-window inference, FullTilt eliminates redundant volumetric computation, reducing runtime by orders of magnitude while strictly limiting VRAM usage. Extensive evaluations demonstrate that FullTilt achieves state-of-the-art zero-shot detection performance on real-world datasets. We anticipate that demonstrating the efficiency and effectiveness of this direct 2D-to-3D detection paradigm will facilitate rapid, large-scale object detection in structural biology and beyond.


\noindent \textbf{Limitations and Future Work.} FullTilt currently struggles with particularly small macromolecules, with near-zero AP for all evaluated protein types below 100~kDa (see~\ref{sec:g_analysis}), consistent with the small-object detection challenges of DETR-like architectures~\cite{carion2020end}. Its relatively weaker performance on simulated data may arise because FullTilt and competing methods overfit to different simulation-specific cues, such as particle-placement patterns and noise statistics. Our preliminary experiments suggest that broader training-data coverage can improve FullTilt's performance on simulated data (see~\ref{sec:broader}); future work will further investigate how reliance on these cues affects generalization. Moreover, zero-shot performance remains below that of fully supervised methods, potentially due to the substantial domain gap between simulated and real-world data. This motivates real-world fine-tuning and further large-scale training, for which FullTilt can serve as a foundation. Future work will also integrate our macromolecule detection framework into downstream subtomogram averaging~\cite{watson2024advances}.

\section*{Acknowledgements}
Research reported in this publication was supported by the National Institute Of General Medical Sciences of the National Institutes of Health under Award Number R35GM163737. The content is solely the responsibility of the authors and does not necessarily represent the official views of the National Institutes of Health. M.-Y.H. also acknowledges support from the Hung Taiwan-Duke University Fellowship. This study utilized the computational resources offered by Duke Research Computing (http://rc.duke.edu). We thank Tracy Futhey, Charley Kneifel, Katie Kilroy, Mike Newton, Victor Orlikowski, Tom Milledge, and David Lane from the Duke Office of Information Technology and Research Computing for providing assistance with the computing environment.


{\small
\bibliographystyle{ieeetr}
\bibliography{main}

@inproceedings{wang2022detr3d,
  title={Detr3d: 3d object detection from multi-view images via 3d-to-2d queries},
  author={Wang, Yue and Guizilini, Vitor Campagnolo and Zhang, Tianyuan and Wang, Yilun and Zhao, Hang and Solomon, Justin},
  booktitle={Conference on robot learning},
  pages={180--191},
  year={2022},
  organization={PMLR}
}

@inproceedings{liu2022petr,
  title={Petr: Position embedding transformation for multi-view 3d object detection},
  author={Liu, Yingfei and Wang, Tiancai and Zhang, Xiangyu and Sun, Jian},
  booktitle={European conference on computer vision},
  pages={531--548},
  year={2022},
  organization={Springer}
}

@inproceedings{huang2022accurate,
  title={Accurate detection of proteins in cryo-electron tomograms from sparse labels},
  author={Huang, Qinwen and Zhou, Ye and Liu, Hsuan-Fu and Bartesaghi, Alberto},
  booktitle={European Conference on Computer Vision},
  pages={644--660},
  year={2022},
  organization={Springer}
}

@article{uddin2024tomopicker,
  title={TomoPicker: annotation-efficient particle picking in cryo-electron tomograms},
  author={Uddin, Mostofa Rafid and Ahmed, Ajmain Yasar and Tahmid, Md Toki and Alam, Md Zarif Ul and Freyberg, Zachary and Xu, Min},
  journal={bioRxiv},
  year={2024}
}

@article{zeng2022structure,
  title={Structure detection in three-dimensional cellular cryoelectron tomograms by reconstructing two-dimensional annotated tilt series},
  author={Zeng, Xiangrui and Lin, Ziqian and Uddin, Mostofa Rafid and Zhou, Bo and Cheng, Chao and Zhang, Jing and Freyberg, Zachary and Xu, Min},
  journal={Journal of Computational Biology},
  volume={29},
  number={8},
  pages={932--941},
  year={2022},
  publisher={SAGE Publications Sage CA: Los Angeles, CA}
}

@inproceedings{zhao2024cryosam,
  title={Cryosam: Training-free cryoet tomogram segmentation with foundation models},
  author={Zhao, Yizhou and Bian, Hengwei and Mu, Michael and Uddin, Mostofa R and Li, Zhenyang and Li, Xiang and Wang, Tianyang and Xu, Min},
  booktitle={International Conference on Medical Image Computing and Computer-Assisted Intervention},
  pages={124--134},
  year={2024},
  organization={Springer}
}

@article{rice2023tomotwin,
  title={TomoTwin: generalized 3D localization of macromolecules in cryo-electron tomograms with structural data mining},
  author={Rice, Gavin and Wagner, Thorsten and Stabrin, Markus and Sitsel, Oleg and Prumbaum, Daniel and Raunser, Stefan},
  journal={Nature methods},
  volume={20},
  number={6},
  pages={871--880},
  year={2023},
  publisher={Nature Publishing Group US New York}
}

@article{wiedemann2026propicker,
  title={ProPicker: promptable segmentation for particle picking in cryogenic electron tomography},
  author={Wiedemann, Simon and Fabian, Zalan and Soltanolkotabi, Mahdi and Heckel, Reinhard},
  journal={Journal of Structural Biology},
  pages={108298},
  year={2026},
  publisher={Elsevier}
}

@inproceedings{carion2020end,
  title={End-to-end object detection with transformers},
  author={Carion, Nicolas and Massa, Francisco and Synnaeve, Gabriel and Usunier, Nicolas and Kirillov, Alexander and Zagoruyko, Sergey},
  booktitle={European conference on computer vision},
  pages={213--229},
  year={2020},
  organization={Springer}
}

@inproceedings{DBLP:conf/iclr/ZhuSLLWD21,
  author       = {Xizhou Zhu and
                  Weijie Su and
                  Lewei Lu and
                  Bin Li and
                  Xiaogang Wang and
                  Jifeng Dai},
  title        = {Deformable {DETR:} Deformable Transformers for End-to-End Object Detection},
  booktitle    = {9th International Conference on Learning Representations, {ICLR} 2021,
                  Virtual Event, Austria, May 3-7, 2021},
  publisher    = {OpenReview.net},
  year         = {2021},
  url          = {https://openreview.net/forum?id=gZ9hCDWe6ke},
  bibsource    = {dblp computer science bibliography, https://dblp.org}
}

@inproceedings{wang2025vggt,
  title={Vggt: Visual geometry grounded transformer},
  author={Wang, Jianyuan and Chen, Minghao and Karaev, Nikita and Vedaldi, Andrea and Rupprecht, Christian and Novotny, David},
  booktitle={Proceedings of the Computer Vision and Pattern Recognition Conference},
  pages={5294--5306},
  year={2025}
}

@inproceedings{jiang2024t,
  title={T-rex2: Towards generic object detection via text-visual prompt synergy},
  author={Jiang, Qing and Li, Feng and Zeng, Zhaoyang and Ren, Tianhe and Liu, Shilong and Zhang, Lei},
  booktitle={European Conference on Computer Vision},
  pages={38--57},
  year={2024},
  organization={Springer}
}

@inproceedings{liu2024grounding,
  title={Grounding dino: Marrying dino with grounded pre-training for open-set object detection},
  author={Liu, Shilong and Zeng, Zhaoyang and Ren, Tianhe and Li, Feng and Zhang, Hao and Yang, Jie and Jiang, Qing and Li, Chunyuan and Yang, Jianwei and Su, Hang and others},
  booktitle={European conference on computer vision},
  pages={38--55},
  year={2024},
  organization={Springer}
}

@misc{zhang2022dino,
      title={DINO: DETR with Improved DeNoising Anchor Boxes for End-to-End Object Detection}, 
      author={Hao Zhang and Feng Li and Shilong Liu and Lei Zhang and Hang Su and Jun Zhu and Lionel M. Ni and Heung-Yeung Shum},
      year={2022},
      eprint={2203.03605},
      archivePrefix={arXiv},
      primaryClass={cs.CV}
}

@article{duran2013structural,
  title={Structural systems pharmacology: the role of 3D structures in next-generation drug development},
  author={Duran-Frigola, Miquel and Mosca, Roberto and Aloy, Patrick},
  journal={Chemistry \& biology},
  volume={20},
  number={5},
  pages={674--684},
  year={2013},
  publisher={Elsevier}
}

@article{blundell1996structure,
  title={Structure-based drug design},
  author={Blundell, Tom L},
  journal={Nature},
  volume={384},
  number={6604},
  pages={23},
  year={1996},
  publisher={[London: Macmillan Journals], 1869-}
}

@article{turk2020promise,
  title={The promise and the challenges of cryo-electron tomography},
  author={Turk, Martin and Baumeister, Wolfgang},
  journal={FEBS letters},
  volume={594},
  number={20},
  pages={3243--3261},
  year={2020},
  publisher={Wiley Online Library}
}

@article{watson2024advances,
  title={Advances in cryo-ET data processing: meeting the demands of visual proteomics},
  author={Watson, Abigail JI and Bartesaghi, Alberto},
  journal={Current Opinion in Structural Biology},
  volume={87},
  pages={102861},
  year={2024},
  publisher={Elsevier}
}

@article{wagner2025cryo,
  title={Cryo-electron tomography: Challenges and computational strategies for particle picking},
  author={Wagner, Thorsten and Raunser, Stefan},
  journal={Current Opinion in Structural Biology},
  volume={93},
  pages={103113},
  year={2025},
  publisher={Elsevier}
}

@article{wagner2019sphire,
  title={SPHIRE-crYOLO is a fast and accurate fully automated particle picker for cryo-EM},
  author={Wagner, Thorsten and Merino, Felipe and Stabrin, Markus and Moriya, Toshio and Antoni, Claudia and Apelbaum, Amir and Hagel, Philine and Sitsel, Oleg and Raisch, Tobias and Prumbaum, Daniel and others},
  journal={Communications biology},
  volume={2},
  number={1},
  pages={218},
  year={2019},
  publisher={Nature Publishing Group UK London}
}

@article{moebel2021deep,
  title={Deep learning improves macromolecule identification in 3D cellular cryo-electron tomograms},
  author={Moebel, Emmanuel and Martinez-Sanchez, Antonio and Lamm, Lorenz and Righetto, Ricardo D and Wietrzynski, Wojciech and Albert, Sahradha and Larivi{\`e}re, Damien and Fourmentin, Eric and Pfeffer, Stefan and Ortiz, Julio and others},
  journal={Nature methods},
  volume={18},
  number={11},
  pages={1386--1394},
  year={2021},
  publisher={Nature Publishing Group US New York}
}

@article{liu2024deepetpicker,
  title={DeepETPicker: Fast and accurate 3D particle picking for cryo-electron tomography using weakly supervised deep learning},
  author={Liu, Guole and Niu, Tongxin and Qiu, Mengxuan and Zhu, Yun and Sun, Fei and Yang, Ge},
  journal={Nature Communications},
  volume={15},
  number={1},
  pages={2090},
  year={2024},
  publisher={Nature Publishing Group UK London}
}

@inproceedings{kirillov2023segment,
  title={Segment anything},
  author={Kirillov, Alexander and Mintun, Eric and Ravi, Nikhila and Mao, Hanzi and Rolland, Chloe and Gustafson, Laura and Xiao, Tete and Whitehead, Spencer and Berg, Alexander C and Lo, Wan-Yen and others},
  booktitle={Proceedings of the IEEE/CVF international conference on computer vision},
  pages={4015--4026},
  year={2023}
}

@book{levoy1992volume,
  title={Volume rendering using the Fourier projection-slice theorem},
  author={Levoy, Marc},
  year={1992},
  publisher={Computer Systems Laboratory, Stanford University}
}

@article{winkler2006accurate,
  title={Accurate marker-free alignment with simultaneous geometry determination and reconstruction of tilt series in electron tomography},
  author={Winkler, Hanspeter and Taylor, Kenneth A},
  journal={Ultramicroscopy},
  volume={106},
  number={3},
  pages={240--254},
  year={2006},
  publisher={Elsevier}
}

@inproceedings{liu2021swin,
  title={Swin transformer: Hierarchical vision transformer using shifted windows},
  author={Liu, Ze and Lin, Yutong and Cao, Yue and Hu, Han and Wei, Yixuan and Zhang, Zheng and Lin, Stephen and Guo, Baining},
  booktitle={Proceedings of the IEEE/CVF international conference on computer vision},
  pages={10012--10022},
  year={2021}
}

@inproceedings{singh2023surround,
  title={Surround-view vision-based 3d detection for autonomous driving: A survey},
  author={Singh, Apoorv},
  booktitle={Proceedings of the IEEE/CVF International Conference on Computer Vision},
  pages={3243--3252},
  year={2023}
}

@article{yang2024self,
  title={Self-supervised noise modeling and sparsity guided electron tomography volumetric image denoising},
  author={Yang, Zhidong and Zang, Dawei and Li, Hongjia and Zhang, Zhao and Zhang, Fa and Han, Renmin},
  journal={Ultramicroscopy},
  volume={255},
  pages={113860},
  year={2024},
  publisher={Elsevier}
}

@article{genthe2023pickyolo,
  title={PickYOLO: Fast deep learning particle detector for annotation of cryo electron tomograms},
  author={Genthe, Erik and Miletic, Sean and Tekkali, Indira and James, Rory Hennell and Marlovits, Thomas C and Heuser, Philipp},
  journal={Journal of structural biology},
  volume={215},
  number={3},
  pages={107990},
  year={2023},
  publisher={Elsevier}
}

@article{de2023convolutional,
  title={Convolutional networks for supervised mining of molecular patterns within cellular context},
  author={de Teresa-Trueba, Irene and Goetz, Sara K and Mattausch, Alexander and Stojanovska, Frosina and Zimmerli, Christian E and Toro-Nahuelpan, Mauricio and Cheng, Dorothy WC and Tollervey, Fergus and Pape, Constantin and Beck, Martin and others},
  journal={Nature Methods},
  volume={20},
  number={2},
  pages={284--294},
  year={2023},
  publisher={Nature Publishing Group US New York}
}

@article{hao2021vp,
  title={VP-Detector: A 3D convolutional neural network for automated macromolecule localization and classification in cryo-electron tomograms},
  author={Hao, Yu and Zhang, Biao and Wan, Xiaohua and Yan, Rui and Liu, Zhiyong and Li, Jintao and Zhang, Shihua and Cui, Xuefeng and Zhang, Fa},
  journal={bioRxiv},
  pages={2021--05},
  year={2021},
  publisher={Cold Spring Harbor Laboratory}
}

@article{kiryo2017positive,
  title={Positive-unlabeled learning with non-negative risk estimator},
  author={Kiryo, Ryuichi and Niu, Gang and Du Plessis, Marthinus C and Sugiyama, Masashi},
  journal={Advances in neural information processing systems},
  volume={30},
  year={2017}
}

@inproceedings{liu2023petrv2,
  title={Petrv2: A unified framework for 3d perception from multi-camera images},
  author={Liu, Yingfei and Yan, Junjie and Jia, Fan and Li, Shuailin and Gao, Aqi and Wang, Tiancai and Zhang, Xiangyu},
  booktitle={Proceedings of the IEEE/CVF international conference on computer vision},
  pages={3262--3272},
  year={2023}
}

@article{alzahrani2024deep,
  title={Deep models for multi-view 3D object recognition: a review},
  author={Alzahrani, Mona and Usman, Muhammad and Jarraya, Salma Kammoun and Anwar, Saeed and Helmy, Tarek},
  journal={Artificial Intelligence Review},
  volume={57},
  number={12},
  pages={323},
  year={2024},
  publisher={Springer}
}

@article{peck2025realistic,
  title={A realistic phantom dataset for benchmarking cryo-ET data annotation},
  author={Peck, Ariana and Yu, Yue and Schwartz, Jonathan and Cheng, Anchi and Ermel, Utz Heinrich and Hutchings, Joshua and Kandel, Saugat and Kimanius, Dari and Montabana, Elizabeth A and Serwas, Daniel and others},
  journal={Nature Methods},
  volume={22},
  number={9},
  pages={1819--1823},
  year={2025},
  publisher={Nature Publishing Group US New York}
}

@article{eisenstein2019improved,
  title={Improved applicability and robustness of fast cryo-electron tomography data acquisition},
  author={Eisenstein, Fabian and Danev, Radostin and Pilhofer, Martin},
  journal={Journal of structural biology},
  volume={208},
  number={2},
  pages={107--114},
  year={2019},
  publisher={Elsevier}
}

@article{tegunov2021multi,
  title={Multi-particle cryo-EM refinement with M visualizes ribosome-antibiotic complex at 3.5 {\AA} in cells},
  author={Tegunov, Dimitry and Xue, Liang and Dienemann, Christian and Cramer, Patrick and Mahamid, Julia},
  journal={Nature methods},
  volume={18},
  number={2},
  pages={186--193},
  year={2021},
  publisher={Nature Publishing Group US New York}
}

@article{rullgaard2011simulation,
  title={Simulation of transmission electron microscope images of biological specimens},
  author={Rullg{\aa}rd, Hans and {\"O}fverstedt, L-G and Masich, Sergey and Daneholt, Bertil and {\"O}ktem, Ozan},
  journal={Journal of microscopy},
  volume={243},
  number={3},
  pages={234--256},
  year={2011},
  publisher={Wiley Online Library}
}

@inproceedings{loshchilovdecoupled,
  title={Decoupled Weight Decay Regularization},
  author={Loshchilov, Ilya and Hutter, Frank},
  booktitle={International Conference on Learning Representations},
  year={2019}
}

@inproceedings{wang2025yoloe,
  title={Yoloe: Real-time seeing anything},
  author={Wang, Ao and Liu, Lihao and Chen, Hui and Lin, Zijia and Han, Jungong and Ding, Guiguang},
  booktitle={Proceedings of the IEEE/CVF International Conference on Computer Vision},
  pages={24591--24602},
  year={2025}
}

@article{jain2024knowledge,
  title={Knowledge transfer from macro-world to micro-world: enhancing 3D Cryo-ET classification through fine-tuning video-based deep models},
  author={Jain, Sabhay and Li, Xingjian and Xu, Min},
  journal={Bioinformatics},
  volume={40},
  number={7},
  pages={btae368},
  year={2024},
  publisher={Oxford University Press}
}

@article{ren2016faster,
  title={Faster R-CNN: Towards real-time object detection with region proposal networks},
  author={Ren, Shaoqing and He, Kaiming and Girshick, Ross and Sun, Jian},
  journal={IEEE transactions on pattern analysis and machine intelligence},
  volume={39},
  number={6},
  pages={1137--1149},
  year={2016},
  publisher={IEEE}
}

@article{fischler1981random,
  title={Random sample consensus: a paradigm for model fitting with applications to image analysis and automated cartography},
  author={Fischler, Martin A and Bolles, Robert C},
  journal={Communications of the ACM},
  volume={24},
  number={6},
  pages={381--395},
  year={1981},
  publisher={ACM New York, NY, USA}
}

@article{dao2022flashattention,
  title={Flashattention: Fast and memory-efficient exact attention with io-awareness},
  author={Dao, Tri and Fu, Dan and Ermon, Stefano and Rudra, Atri and R{\'e}, Christopher},
  journal={Advances in neural information processing systems},
  volume={35},
  pages={16344--16359},
  year={2022}
}

@article{liu2023nextpyp,
  title={nextPYP: a comprehensive and scalable platform for characterizing protein variability in situ using single-particle cryo-electron tomography},
  author={Liu, Hsuan-Fu and Zhou, Ye and Huang, Qinwen and Piland, Jonathan and Jin, Weisheng and Mandel, Justin and Du, Xiaochen and Martin, Jeffrey and Bartesaghi, Alberto},
  journal={Nature Methods},
  volume={20},
  number={12},
  pages={1909--1919},
  year={2023},
  publisher={Nature Publishing Group US New York}
}

@article{bartesaghi2014structure,
  title={Structure of $\beta$-galactosidase at 3.2-{\AA} resolution obtained by cryo-electron microscopy},
  author={Bartesaghi, Alberto and Matthies, Doreen and Banerjee, Soojay and Merk, Alan and Subramaniam, Sriram},
  journal={Proceedings of the National Academy of Sciences},
  volume={111},
  number={32},
  pages={11709--11714},
  year={2014},
  publisher={National Academy of Sciences}
}
}

\clearpage
\appendix
\renewcommand{\thesection}{Appendix \Alph{section}}
\renewcommand{\thefigure}{A\arabic{figure}}
\renewcommand{\thetable}{A\arabic{table}}
\renewcommand{\thealgorithm}{A\arabic{algorithm}}
\setcounter{figure}{0}
\setcounter{table}{0}
{\LARGE \bfseries Appendix\par}
\vspace{1em}
\section{Comparison with DETR3D and PETR, and FullTilt Design Rationale}
\label{sec:compare_detr3d_petr}
DETR3D~\cite{wang2022detr3d} extends the DETR~\cite{carion2020end} framework for multi-view 3D object detection. In each decoder layer, it projects 3D anchors onto each 2D image to incorporate local features into the object queries. Conversely, PETR eliminates the explicit 3D-to-2D projection step by directly infusing 2D features with 3D spatial information via 3D positional encoding. In the PETR decoder layer, the 3D object queries perform multi-head attention over all flattened features to update their representations.

Both frameworks present distinct advantages and limitations. For DETR3D, the 3D-to-2D projection operation ensures training efficiency and stability by avoiding attention over all features across all images. However, this projection requires careful geometric engineering. PETR addresses the projection bottleneck by enabling features to inherently possess 3D positional awareness; however, performing attention across all features from all images is computationally expensive and often leads to unstable training.

This instability is significantly exacerbated in cryo-ET data. Unlike autonomous driving scenarios where a small number of large, distinguishable objects (e.g., cars) are present, cryo-ET data suffers from extremely low signal-to-noise ratios and contains high concentration of small particles that are difficult to distinguish in any single tilt image. In such densely populated, low-contrast environments, PETR's global attention mechanism (i.e., attention across all features from all images) struggles to accurately localize the exact same target across different tilts, even with 3D-aware features. Furthermore, while particles become identifiable when observing consecutive frames, neither DETR3D nor PETR incorporates a fusion module to integrate information across the entire tilt-series.

Our FullTilt architecture addresses these challenges by synthesizing the strengths of both frameworks while introducing critical, cryo-ET-tailored modules. In our tilt-series encoder, we introduce a tilt embedding that grants the 2D features 3D spatial awareness, analogous to PETR's approach. Moreover, we incorporate a local attention and global row attention to fuse information across all tilt images.

A naive implementation of this fusion would flatten all features from all tilt images and apply self-attention~\cite{wang2025vggt}. However, a standard tilt-series contains over 30 images, each potentially yielding more than 4,000 features from multi-level feature maps, rendering global self-attention computationally prohibitive. To overcome this, we exploit the geometric prior of cryo-ET data: because the tilt-series is aligned along the y-axis, a given particle will consistently appear at the same y-axis coordinate across all tilt images. By restricting our module to a global row attention mechanism, combined with local attention, we efficiently fuse multi-tilt information into the corresponding features.

Finally, to support open-set detection scenarios, FullTilt incorporates a multiclass visual prompt encoder. A tilt-aware query initializer is introduced to provide 3D anchor priors for the decoder to refine. Within the decoder, we adopt DETR3D's 3D-to-2D projection strategy to strictly attend to localized target particles, ensuring that our training remains both highly stable and computationally efficient.

\section{FullTilt Architecture Details}
\label{sec:fulltilt_arch_details}
\subsection{External and Internal Label Mapping}
\label{sec:external_internal_mapping}
To accommodate arbitrary user-defined class identifiers ($Y \in \mathbb{Z}^+$)—including sparse, excessively large, or non-contiguous label sets—we implement a dynamic mapping mechanism. In the beginning, all unique \textit{external} labels provided by the user are hashed and mapped to a contiguous, zero-indexed \textit{internal} label space ($0, 1, \dots, N_c-1$, where $N_c$ is the number of unique classes). This internal representation facilitates efficient tensor indexing and matrix gathering operations within the network pipeline. Following inference, the estimated internal class indices are deterministically mapped back to the user's original external labels.

\subsection{Normalized Coordinate Space}
Following standard DETR design principles~\cite{carion2020end, DBLP:conf/iclr/ZhuSLLWD21, liu2024grounding}, all 3D spatial coordinates within the pipeline are strictly normalized to the $[0, 1]$ range. Because the intermediate transformer layer outputs exist in unconstrained real space ($\mathbb{R}$), we utilize standard sigmoid ($\sigma$) and inverse sigmoid ($\sigma^{-1}$) transformations for stable coordinate refinement. For instance, to update a normalized 3D anchor using the decoder's content queries, we first apply $\sigma^{-1}$ to the anchor, add the unconstrained output from the model, and then apply $\sigma$ to project the updated coordinate back into the normalized $[0, 1]$ range. This normalization strategy fundamentally impacts the 3D-to-2D projection geometry, necessitating precise anisotropic scaling factors to account for the varying absolute dimensions of the volumetric space. Additionally, the voxel or pixel size becomes unequal across dimensions within this normalized space. To address this spatial distortion in practice, the decoder explicitly regresses the diameter $d$ as independent height ($h$) and width ($w$) variables, deriving the final scalar diameter by averaging these two values. In the final step, all estimated coordinates are scaled back from the normalized space to the absolute spatial dimensions of the original 3D tomogram.

\section{Auxiliary Geometric Primitives ($\mathcal{G}$) Implementation Details} \label{sec:agp}
\begin{figure}[h]
\begin{center}
\includegraphics[width=0.8\linewidth]{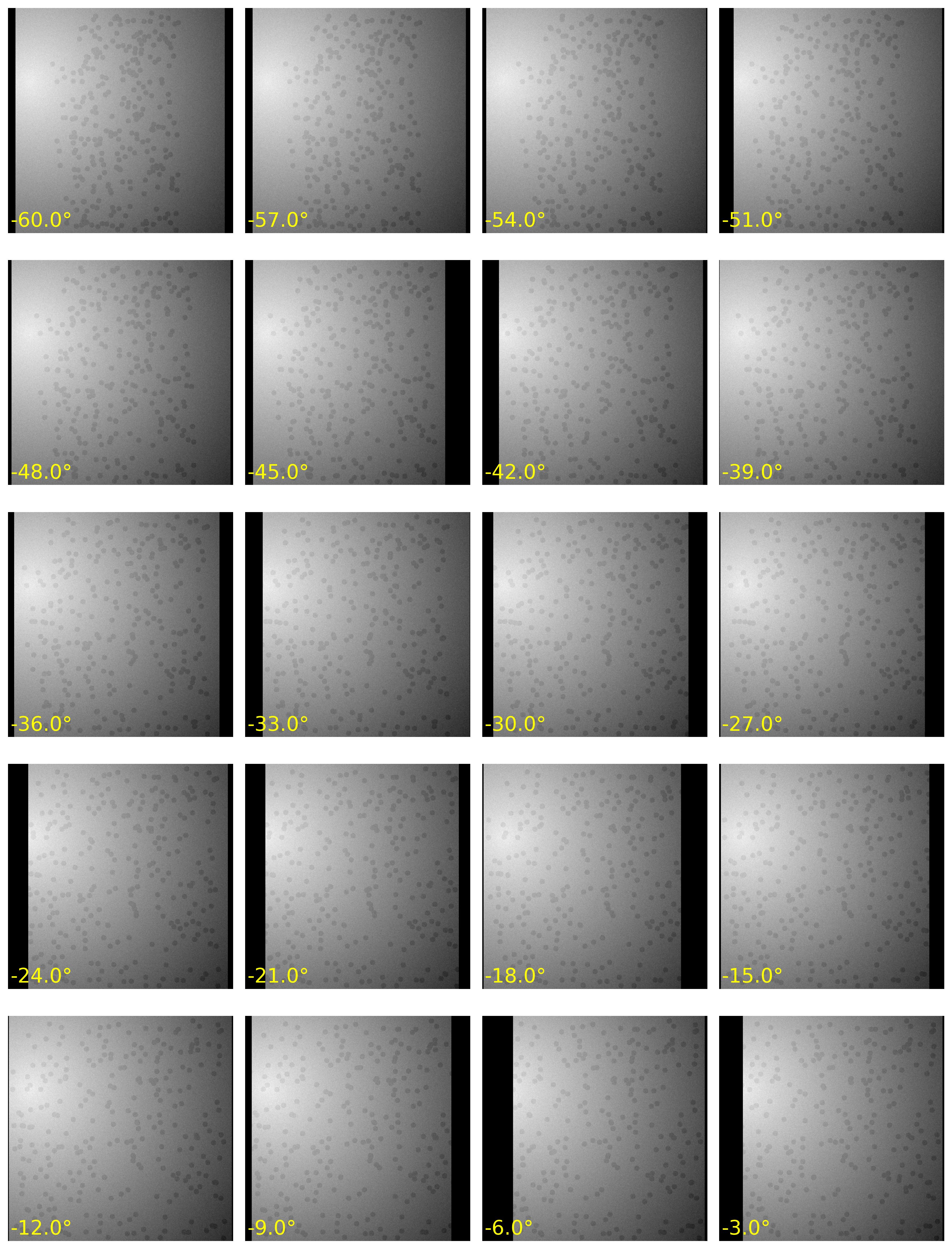}
\end{center}
\caption{\textbf{A tilt-series generated by our auxiliary geometric primitives module.}}
\label{fig:geo_examples_A_1}
\end{figure}

\begin{figure}[h]
\begin{center}
\includegraphics[width=0.8\linewidth]{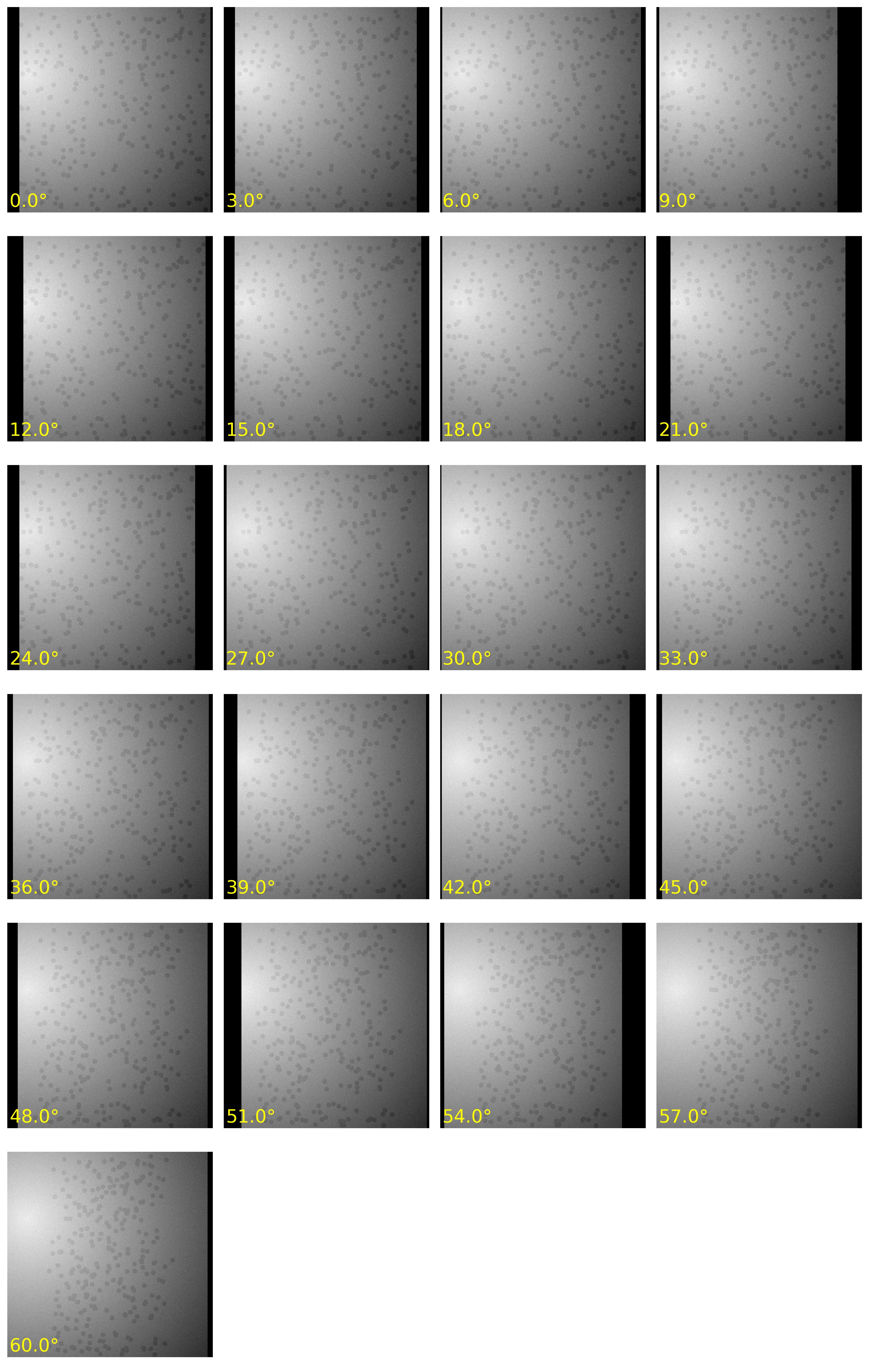}
\end{center}
\caption{\textbf{A tilt-series generated by our auxiliary geometric primitives module (cont.).}}
\label{fig:geo_examples_A_2}
\end{figure}

The auxiliary geometric primitives module ($\mathcal{G}$) has two primary objectives: (1) to enable FullTilt to learn multi-view geometry without relying exclusively on computationally expensive simulated cryo-ET data, which cannot be generated \textit{on-the-fly} and incurs significant storage overhead (typically over 2 GB per simulation); and (2) to improve the model's robustness to adverse imaging artifacts, such as severe physical occlusion and non-uniform illumination. One example of the resulting synthetic tilt-series is provided in Figs.~\ref{fig:geo_examples_A_1} and~\ref{fig:geo_examples_A_2}. The generation pipeline, implemented entirely in \texttt{NumPy}, synthesizes a complete tilt-series of varying dimensions and angular increments in approximately 0.5 to 1.0 seconds.

\noindent \textbf{Parameter Sampling and Coordinate Generation.}
The process begins by \textit{empirically} sampling the spatial dimensions from discrete ($\mathcal{U}_{\text{Int}}$) or continuous ($\mathcal{U}$) uniform distributions: image height $H \sim \mathcal{U}_{\text{Int}}(300, 512)$, width $W \sim \mathcal{U}_{\text{Int}}(300, 512)$, and tomogram depth $D \sim \mathcal{U}_{\text{Int}}(120, 300)$. To control particle density along the $z$-axis, we define a central bounding slab of depth $D' \sim \mathcal{U}_{\text{Int}}(80, 120)$, ensuring $D' \leq D$. We then uniformly sample $M \sim \mathcal{U}_{\text{Int}}(100, 600)$ 3D coordinates within this $D'$ slab. The particle diameter $d \sim \mathcal{U}_{\text{Int}}(6, 16)$ and opacity $c \sim \mathcal{U}(0.03, 0.10)$ are also randomly assigned. Additional parameters governing the illumination gradients, Gaussian noise, and boundary occlusions are drawn from empirical distributions designed to cover the variance observed in real-world cryo-ET datasets.

\noindent \textbf{Projection and Visibility Masking.}
After selecting a random angular range and tilt increment, the 3D coordinates are projected onto each 2D plane. To enforce strict cross-view geometric consistency, any particle that is not fully visible across \textit{all} generated tilt images is discarded.

\noindent \textbf{Rendering and Adverse Condition Simulation.}
For each tilt angle, the valid projected coordinates are rendered as solid 2D disk kernels (circles). We then apply the sampled illumination gradient and Gaussian noise. Finally, we inject random left and right physical occlusions (simulating field-of-view truncation). These synthetic 2D images and their corresponding 3D ground truth coordinates are subsequently passed through our training augmentation pipeline (e.g., random resizing, padding, and flipping).

\noindent \textbf{Limitations.}
While $\mathcal{G}$ effectively models multi-view geometry and core adverse imaging artifacts, incorporating a wider variety of non-spherical geometric primitives and complex physical conditions could further enrich the synthetic diversity; we leave this to future work. Additionally, due to the extensive computational time required for full model training, conducting an exhaustive hyperparameter search or an ablation study for $\mathcal{G}$ remains computationally prohibitive.

\clearpage

\section{Data Preparation Details}
\label{sec:data_details}
\subsection{Training Dataset}
Following the protocol established by TomoTwin~\cite{rice2023tomotwin}, we utilize the TEM Simulator~\cite{rullgaard2011simulation} to generate our simulated cryo-ET training dataset. Each generated instance comprises a 2D tilt-series, the ground truth 3D spatial coordinates of the particles, and a fully reconstructed 3D tomogram. Notably, our framework strictly utilizes the tilt-series and 3D coordinates, entirely discarding the reconstructed tomogram during training. To ensure a fair and direct comparison with state-of-the-art open-set 3D macromolecule detection methods (i.e., TomoTwin~\cite{rice2023tomotwin} and ProPicker~\cite{wiedemann2026propicker}), we construct our simulations using the same core set of proteins. The complete list of the 119 proteins utilized in our study is provided in Table~\ref{tab:training_pdbs}. Note that while the original TomoTwin protocol included 120 proteins, we omitted PDB entry \texttt{5MRC.pdb} due to download unavailability, resulting in our final count of 119.

The primary deviation from the TomoTwin simulation protocol lies in the distribution of particle classes and particle numbers per instance. We generated a diverse dataset consisting of 338 single-class, 300 two-class, 300 four-class, and 300 eight-class instances. For the single-class instances, we systematically cycled through the 119 proteins to guarantee each was represented at least twice, whereas the proteins for the multi-class instances were selected randomly. To control spatial crowding, the number of particles per protein was sampled randomly based on the total class count: $120$--$320$ for single-class, $60$--$160$ for two-class, $30$--$80$ for four-class, and $15$--$40$ for eight-class instances.

This specific compositional balance contrasts with TomoTwin's highly dense simulation settings (e.g., 150 particles distributed across 10 classes). Because subvolume-matching methods like TomoTwin extract small, localized 3D patches (e.g., $37 \times 37 \times 37$ voxels), generating densely populated tomograms is highly efficient, allowing them to harvest numerous positive subvolumes from a single simulation. In contrast, FullTilt solves a fundamentally different and highly complex multi-view geometry problem directly from multi-view projections. If the 3D volume is excessively dense, the corresponding 2D projections suffer from severe projection overlap and ambiguity. Therefore, our adjusted density distribution is specifically designed to mitigate extreme projection overlap, facilitating robust geometric learning and accurate 3D localization. We acknowledge this controlled density as a limitation of our current training paradigm and plan to address it in future work.

\begin{table*}[h]
\caption{\textbf{Training PDBs.}}
\label{tab:training_pdbs}
\centering
\small
\begin{tabular}{llllllllll}
\texttt{1BXN} & \texttt{1G3I} & \texttt{1N9G} & \texttt{1QVR} & \texttt{1S3X} & \texttt{1SS8} & \texttt{1U6G} & \texttt{1UL1} & \texttt{2CG9} & \texttt{2DFS} \\
\texttt{2R9R} & \texttt{2RHS} & \texttt{2VZ9} & \texttt{2WW2} & \texttt{2XNX} & \texttt{3CF3} & \texttt{3D2F} & \texttt{3GL1} & \texttt{3H84} & \texttt{3LUE} \\
\texttt{3MKQ} & \texttt{3QM1} & \texttt{3ULV} & \texttt{4CR2} & \texttt{4UIC} & \texttt{4WRM} & \texttt{4XK8} & \texttt{5A20} & \texttt{5CSA} & \texttt{5G04} \\
\texttt{5H0S} & \texttt{5JH9} & \texttt{5LJO} & \texttt{5O32} & \texttt{5OOL} & \texttt{5VKQ} & \texttt{6AHU} & \texttt{6BQ1} & \texttt{6CES} & \texttt{6CNJ} \\
\texttt{6DUZ} & \texttt{6EMK} & \texttt{6F8L} & \texttt{6GY6} & \texttt{6GYM} & \texttt{6I0D} & \texttt{6ID1} & \texttt{6IGC} & \texttt{6JY0} & \texttt{6KLH} \\
\texttt{6KRK} & \texttt{6KSP} & \texttt{6LMT} & \texttt{6LX3} & \texttt{6LXK} & \texttt{6LXV} & \texttt{6M04} & \texttt{6MRC} & \texttt{6PIF} & \texttt{6SCJ} \\
\texttt{6TA5} & \texttt{6TAV} & \texttt{6TGC} & \texttt{6TPS} & \texttt{6U8Q} & \texttt{6UP6} & \texttt{6VGR} & \texttt{6VN1} & \texttt{6VZ8} & \texttt{6W6M} \\
\texttt{6WZT} & \texttt{6X5Z} & \texttt{6X9Q} & \texttt{6XF8} & \texttt{6YT5} & \texttt{6Z3A} & \texttt{6Z6O} & \texttt{6Z80} & \texttt{6ZIU} & \texttt{6ZQJ} \\
\texttt{7AMV} & \texttt{7B5S} & \texttt{7B7U} & \texttt{7BKC} & \texttt{7BLQ} & \texttt{7BLR} & \texttt{7DD9} & \texttt{7E1Y} & \texttt{7E6G} & \texttt{7E8H} \\
\texttt{7EEP} & \texttt{7EGD} & \texttt{7EGE} & \texttt{7EGQ} & \texttt{7ETM} & \texttt{7EY7} & \texttt{7JSN} & \texttt{7K5X} & \texttt{7KDV} & \texttt{7KFE} \\
\texttt{7KJ2} & \texttt{7LSY} & \texttt{7MEI} & \texttt{7NHS} & \texttt{7NIU} & \texttt{7NYZ} & \texttt{7O01} & \texttt{7Q21} & \texttt{7QJ0} & \texttt{7R04} \\
\texttt{7S7K} & \texttt{7SFW} & \texttt{7SGM} & \texttt{7SHK} & \texttt{7SN7} & \texttt{7T3U} & \texttt{7VTQ} & \texttt{7WBT} & \texttt{7WOO} & \\
\end{tabular}
\end{table*}

\subsection{Evaluation Datasets}
We evaluate FullTilt across a comprehensive suite of real-world and simulated cryo-ET datasets to rigorously assess its detection capabilities and generalization. We first benchmark our framework on the CZII CryoET Object Identification Challenge dataset (CryoET Data Portal Dataset ID: DS-10440)~\cite{peck2025realistic}, as it serves as a premier, multi-class real-world benchmark with meticulously curated ground truth annotations. To further validate our method against established standards, we select EMPIAR-10304~\cite{eisenstein2019improved} and EMPIAR-10499~\cite{tegunov2021multi} due to their widespread adoption in the literature~\cite{huang2022accurate, zhao2024cryosam}. These datasets represent single-class scenarios, evaluating performance in both a purified (EMPIAR-10304) and a native cellular environment (EMPIAR-10499). Crucially, all real-world evaluations are conducted in a strictly zero-shot setting. However, acknowledging the inherent challenge of obtaining exhaustive, perfect ground truth for empirical EMPIAR data, we supplement our real-world benchmarks with fully annotated, multiclass simulation datasets. These simulations provide exact 3D ground truth coordinates, enabling precise quantification of model performance in both ``unseen'' (zero-shot) and ``seen'' (non-zero-shot) scenarios. A statistical summary of all evaluation datasets is provided in Table~\ref{tab:supp_dataset_details}.

\begin{table*}[h]
\centering
\caption{\textbf{Dataset Details.} Characteristics of the real-world and simulated Cryo-ET datasets used to evaluate FullTilt. Only the ``seen'' simulated cryo-ET dataset is not evaluated in a zero-shot setting.}
\resizebox{\textwidth}{!}{%
\begin{threeparttable}
\begin{tabular}{@{}l|rrr|rr@{}}
\toprule
 & \multicolumn{3}{c}{\textbf{Real-World Cryo-ET Dataset}} & \multicolumn{2}{c}{\textbf{Simulated Cryo-ET Dataset}} \\ 
\cmidrule(lr){2-4}\cmidrule(lr){5-6}
& 
\textbf{DS-10440}~\cite{peck2025realistic} & \textbf{EMPIAR-10304}~\cite{eisenstein2019improved} &
\textbf{EMPIAR-10499}~\cite{tegunov2021multi} & \textbf{Seen} &
\textbf{Unseen} \\

\midrule
\# of Instances & 7 & 12 & 65 & 20 & 20 \\
\# of Classes & 6 & 1 & 1 & 10 & 10 \\
\midrule
\# of Particles (per instance) & 181.3 $\pm$ 35.2\ [140, 225] & 979.2 $\pm$ 72.6\ [819, 1111] & 382.1 $\pm$ 49.1\ [257, 490] & 100.0 $\pm$ 0.0\ [100, 100] & 100.0 $\pm$ 0.0\ [100, 100]\\
\# of Particles (per instance and per class) & 30.2 $\pm$ 20.0\ [9, 95] & 979.2 $\pm$ 72.6\ [819, 1111] & 382.1 $\pm$ 49.1\ [257, 490] & 50.0 $\pm$ 30.6\ [25, 100] & 50.0 $\pm$ 30.6\ [25, 100]\\
\midrule
XY Pixel space (\AA) & 10 & 25.2 & 13.6 & 10.2 & 10.2 \\
Tilt Range \& Increment & $-45^{\circ}$ to $+45^{\circ}$ ($3^{\circ}$ step) & $-60^{\circ}$ to $+60^{\circ}$ ($3^{\circ}$ step) & $-60^{\circ}$ to $+60^{\circ}$ ($3^{\circ}$ step) & $-60^{\circ}$ to $+60^{\circ}$ ($2^{\circ}$ step) & $-60^{\circ}$ to $+60^{\circ}$ ($2^{\circ}$ step) \\
\# of Projections & 31 & 41 & 41 & 61 & 61 \\
Projection Size (H$\times$W) & 630$\times$630 & 480$\times$340 & 464$\times$480 & 512$\times$512 & 512$\times$512 \\ 
\midrule
Rec. Tomogram Size (D$\times$H$\times$W) & 184$\times$630$\times$630 & 256$\times$480$\times$340 & 256$\times$464$\times$480 & 
200$\times$512$\times$512 & 200$\times$512$\times$512 \\ 

\bottomrule
\end{tabular}
\begin{tablenotes}[flushleft]\footnotesize
\item Rec.: reconstructed; \# of particles is reported in mean $\pm$ std\ [min, max] 
\end{tablenotes}
\end{threeparttable}
}
\label{tab:supp_dataset_details}
\end{table*}

\subsubsection{Zero-Shot Setting}
\noindent \textbf{DS-10440~\cite{peck2025realistic}.} We downloaded all seven raw tilt-series from the CryoET Data Portal (Deposition ID: CZCDP-10310; Dataset ID:
DS-10440) and performed alignment according to the official protocol using the provided alignment files. Each tilt-series contains six distinct protein classes: apoferritin, $\beta$-amylase, $\beta$-galactosidase, ribosome, thyroglobulin, and virus-like particles. The corresponding 3D ground truth coordinates were obtained from the official Kaggle competition dataset.

\noindent \textbf{EMPIAR-10304~\cite{eisenstein2019improved}.} We downloaded all 12 raw tilt-series from the EMPIAR archive and aligned them using IMOD. This dataset features a single target class: the purified \textit{Escherichia coli} 70S ribosome. Due to the prohibitive difficulty of exhaustive manual annotation, 3D ground truth coordinates were first generated using an established size-based particle detection pipeline~\cite{liu2023nextpyp, bartesaghi2014structure} with optimized parameters, followed by meticulous manual examination to ensure high fidelity. While a small fraction of particles may remain unlabeled, these annotations are qualitatively robust for empirical benchmarking (see Fig.~\ref{fig:detection_result_tomo_10304_tilt1_gt}). Notably, our computational efficiency benchmarks (inference runtime and VRAM consumption) are entirely independent of label completeness. During inference, we omit the first and last five tilt images of each series due to severe adverse imaging artifacts (e.g., total physical occlusion). 

\noindent \textbf{EMPIAR-10499~\cite{tegunov2021multi}.} We downloaded all 65 raw tilt-series from the EMPIAR archive, which were similarly aligned using IMOD. This dataset contains a single target class: the ribosome from native \textit{Mycoplasma pneumoniae} cells treated with chloramphenicol. Similar to EMPIAR-10304, due to the prohibitive difficulty of exhaustive manual annotation, 3D ground truth coordinates were generated using an established size-based particle detection pipeline~\cite{liu2023nextpyp, bartesaghi2014structure} with optimized parameters, followed by meticulous manual examination to ensure high fidelity. While a small fraction of particles may remain unlabeled, these annotations are qualitatively robust for empirical benchmarking (see Fig.~\ref{fig:detection_result_tomo_10499_TS_01_gt}). Notably, our computational efficiency benchmarks (inference runtime and VRAM consumption) are entirely independent of label completeness.

\noindent \textbf{Simulation (Unseen).} We utilized 10 external proteins (i.e., strictly disjoint from the training set) sourced from prior studies~\cite{rice2023tomotwin, jain2024knowledge}  to generate 20 instances: \texttt{1E9R}, \texttt{1F1B}, \texttt{1FPY}, \texttt{1JZ8}, \texttt{1OAO}, \texttt{2DF7}, \texttt{2H12}, \texttt{3HHB}, \texttt{4D4R}, and \texttt{6T3E}. Following the same generation protocol used for the training data, we controlled the particle count to maintain consistent density: 100 particles for single-class volumes (10 instances, systematically cycling through all 10 proteins), 50 particles per class for two-class volumes (5 instances, randomly selected proteins), and 25 particles per class for four-class volumes (5 instances, randomly selected proteins).

\subsubsection{Non-Zero-Shot Setting}
\noindent \textbf{Simulation (Seen).} Due to the large computational overhead required for exhaustive benchmarking and our primary architectural focus on the zero-shot setting, we evaluated non-zero-shot performance on a random subset of PDBs from the training set. Specifically, we randomly sampled 10 proteins from the 119-protein training set to generate 20 instances: \texttt{1BXN}, \texttt{1S3X}, \texttt{3CF3}, \texttt{3GL1}, \texttt{5JH9}, \texttt{5OOL}, \texttt{6IGC}, \texttt{6LXV}, \texttt{7DD9}, and \texttt{7ETM}. The generation protocol and particle density distributions precisely mirror those of the ``unseen'' simulation dataset (10 single-class, 5 two-class, and 5 four-class instances).

\section{Benchmark Setting Details}
\label{sec:supp_benchmark_details}
\subsection{Prompt Quantities and Evaluation Trials}
We evaluated all open-set methods under 1-, 2-, and 4-prompt settings. We capped the maximum number of visual prompts at four for two primary reasons. First, this represents a realistic empirical constraint: because the sparsest target class across our evaluation datasets contains only nine ground truth particles (Table~\ref{tab:supp_dataset_details}), supplying four prompts would already consume nearly half of the available targets in the volume. Second, manually identifying four reference particles requires negligible human effort (taking only seconds for an annotator), strictly adhering to the fundamental objective of open-set detection paradigms to minimize annotation burden.

To ensure statistical robustness, all open-set experiments were averaged across 10 independent trials. For each trial, the reference prompts were randomly re-sampled from the available ground truth pool. In contrast, due to the prohibitive computational cost of repeatedly training models from scratch, evaluations for the WSL baselines were limited to three independent trials.

\subsection{Auxiliary F1 Score Metric}
Because several baseline methods do not output confidence scores, calculating the standard mean Average Precision (mAP) across all evaluated frameworks is not feasible. To ensure a fair and comprehensive comparison, we report the F1 score as an auxiliary evaluation metric. For methods that do yield confidence scores, we determine the optimal F1 score by performing a discrete grid search over the confidence threshold space. Specifically, we sweep the threshold from $0.0$ to $1.0$ at intervals of $0.05$, systematically filtering out low-confidence outputs at each step to report the maximum achieved F1 score for that method.

\subsection{Runtime and Memory Benchmarking Protocol}
To effectively isolate the model's intrinsic computational efficiency from highly variable I/O bottlenecks (e.g., disk read latency, CPU page faults, and OS-level scheduling), we strictly measure the core inference execution time. Specifically, runtime is recorded using Python's high-resolution \texttt{time.perf\_counter()} from the precise moment the data is fully loaded into memory until the final 3D coordinates and associated metadata (e.g., confidence scores) are generated.

Peak VRAM consumption is exclusively profiled within this exact execution window using \texttt{torch.cuda.max\_memory\_allocated()}. To guarantee independent and accurate memory statistics across sequential trials, we actively clear the profiler state via \texttt{torch.cuda.reset\_peak\_memory\_stats()} prior to each measurement.

\section{Training Details}
\label{sec:training_details}
\subsection{Augmentation}
After loading the paired tilt-series images and their 3D ground truth coordinates, we apply consistent 3D-aware augmentations across the series, followed by 2D-only augmentations on individual images. The 3D-aware transformations include random resizing (scale factors $0.7$ to $1.0$), padding (scale factors $1.0$ to $1.1$), and spatial flipping. To maintain geometric consistency, a single set of transformation parameters is sampled and applied uniformly to both the 2D tilt images and the corresponding 3D ground truth. To accommodate the 48 GB VRAM limit of our NVIDIA RTX 6000 Ada GPUs, padding is discarded if the combined resizing and padding operations would result in an image larger than the original resolution. Subsequently, 2D-only augmentations are applied independently to each image, which include the addition of random Gaussian noise ($\sigma=0.1$) and a random grayscale rectangular occlusion (up to 150 pixels in size) to simulate physical occlusion and other adverse imaging artifacts.

Next, we randomly sample the set of input tilts and visual prompts. Specifically, we first determine the number of target classes (ranging from 1 to the total number of unique classes present in the instance). For each selected class, we uniformly sample between 1 and 5 visual prompts. Subsequently, the order of the tilt images and their corresponding prompts are randomly shuffled. This prevents the model from overfitting to a specific sequential order, forcing it to learn the underlying multi-view geometry. Finally, we randomly sample a subset of 21 to 31 tilt images from the full sequence (also to strictly adhere to the aforementioned VRAM constraints).

During each training epoch, we iterate over the entire simulated cryo-ET dataset, followed by iterations over dynamically generated data from our auxiliary geometric primitives module ($\mathcal{G}$) equivalent to half the size of the simulated dataset. We empirically set these augmentation hyperparameters following established practices in standard visual prompting architectures~\cite{wang2025yoloe, jiang2024t}. Due to the extensive computational cost of training, an exhaustive ablation study of these augmentation hyperparameters was not feasible.

\subsection{Implementation and Training Time}
FullTilt is implemented in PyTorch 2.5.1 and trained for 200 epochs using the AdamW optimizer~\cite{loshchilovdecoupled} with a learning rate of $1 \times 10^{-4}$. Distributed training is conducted across four NVIDIA RTX 6000 Ada GPUs with a per-GPU batch size of 1, requiring approximately two weeks to converge. To better understand the computational bottleneck, we conducted empirical profiling of the training loop. We observed that the core network operations execute efficiently: the forward pass takes approximately 400 ms, the backward pass requires roughly 700 ms, and distributed gradient synchronization (e.g., AllReduce) takes less than 200 ms. In stark contrast, the bipartite matching required for the loss calculation dominates the computational overhead, consuming nearly 2 seconds per iteration.

\section{Baseline Implementation Details}
\label{sec:baseline_details}

\subsection{Open-Set Methods for Cryo-ET Data}
\noindent \textbf{TomoTwin~\cite{rice2023tomotwin}.} We utilized the official codebase and pretrained weights provided by the authors. Inference was conducted using all officially recommended hyperparameters, with the sole exception of increasing the batch size from 256 to 512. This modification accelerated the embedding process by leveraging the larger VRAM capacity of our NVIDIA RTX 6000 Ada GPU.

\begin{figure}[h]
    \centering
    \begin{subfigure}[b]{0.45\textwidth}
        \includegraphics[width=\textwidth]{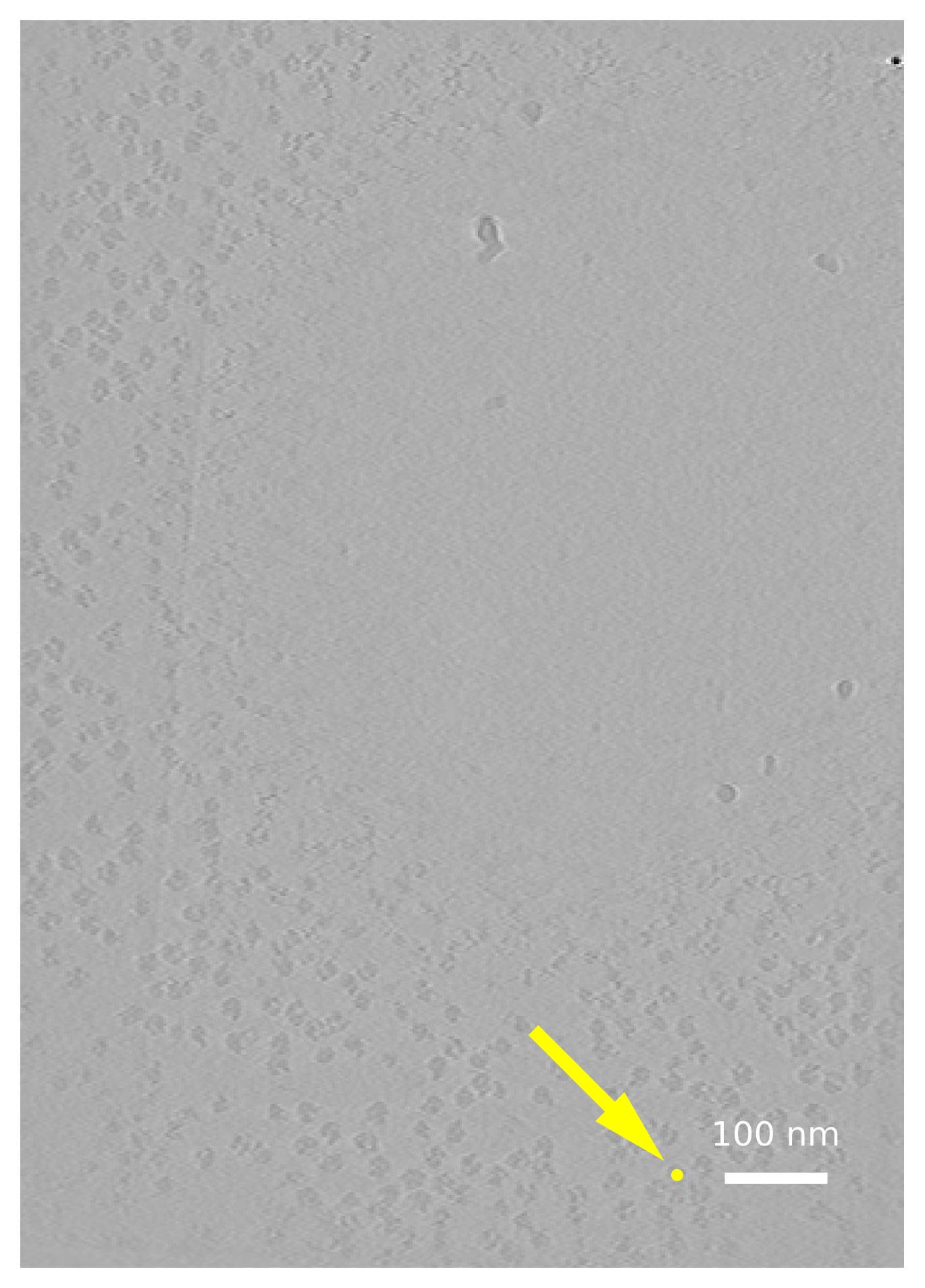}
        \caption{Prompt}
    \end{subfigure}
    \hfill
    \begin{subfigure}[b]{0.45\textwidth}
        \includegraphics[width=\textwidth]{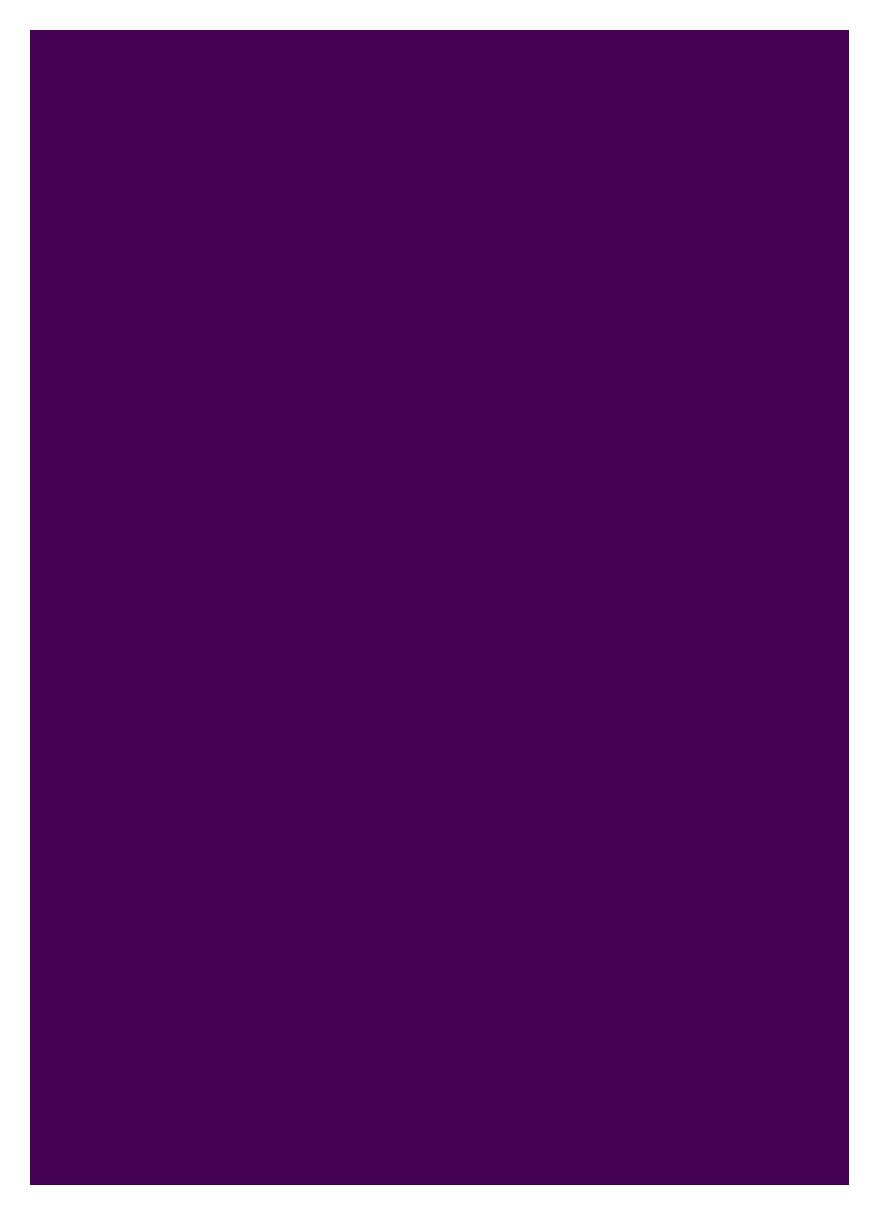} 
        \caption{Mask}
    \end{subfigure}
    \caption{\textbf{Example of CryoSAM failure on EMPIAR-10304.} Given the prompt (a small yellow dot indicated by the yellow arrow), CryoSAM fails to estimate any masked region (i.e., yields an empty mask).}
    \label{fig:cryosam_fail2}
\end{figure}

\begin{figure}[h]
    \centering
    \begin{subfigure}[b]{0.45\textwidth}
        \includegraphics[width=\textwidth]{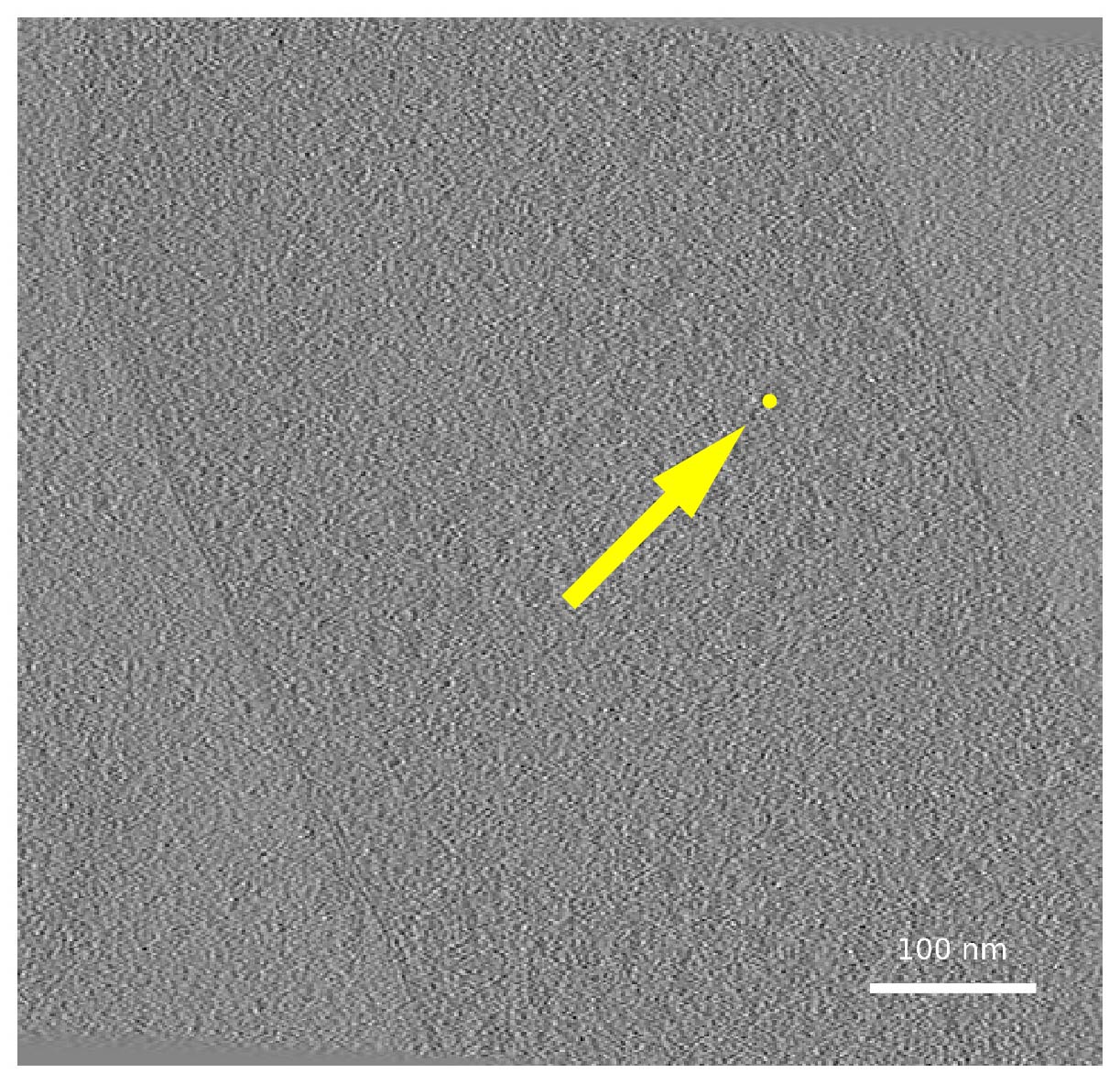}
        \caption{Prompt}
    \end{subfigure}
    \hfill
    \begin{subfigure}[b]{0.45\textwidth}
        \includegraphics[width=\textwidth]{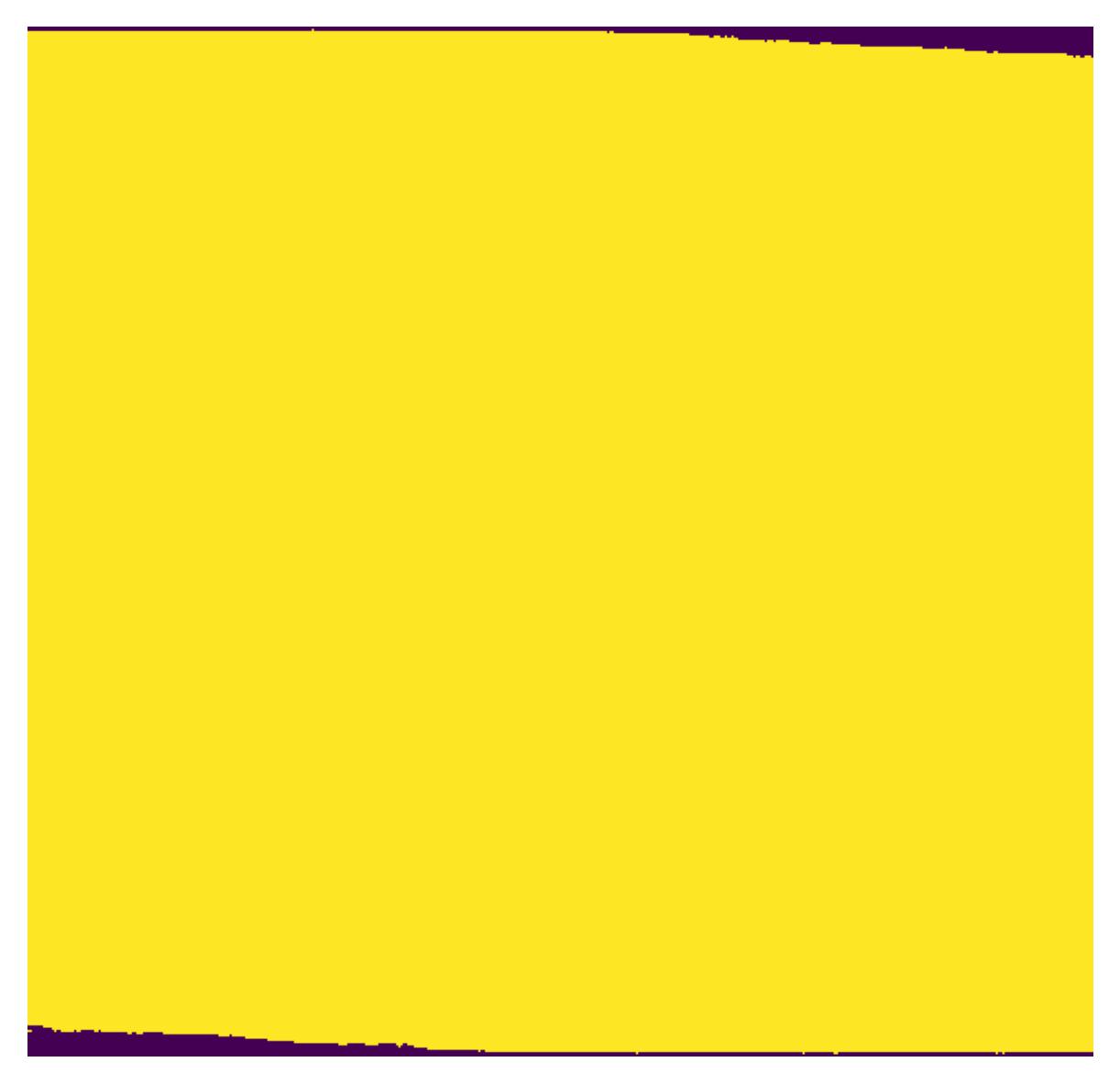} 
        \caption{Mask}
    \end{subfigure}
    \caption{\textbf{Example of CryoSAM failure on EMPIAR-10499.} Given the prompt (a small yellow dot indicated by the yellow arrow), CryoSAM incorrectly estimates a mask covering the entire image.}
    \label{fig:cryosam_fail1}
\end{figure}

\noindent \textbf{CryoSAM~\cite{zhao2024cryosam}.} We set up the official environment and loaded the designated pretrained weights for SAM~\cite{kirillov2023segment}. Despite strictly adhering to the official inference protocols and experimenting with various prompts, we were unable to reproduce the results reported in the original paper. We attribute this failure to a severe zero-shot domain gap: SAM, pretrained on natural images, struggled to generalize to the inherently low-contrast, noisy slices of cryo-ET tomograms, consistently yielding empty or nonsensical segmentation masks (Figs.~\ref{fig:cryosam_fail2} and~\ref{fig:cryosam_fail1}). This irreproducibility has also been documented in recent literature~\cite{wiedemann2026propicker}.

\noindent \textbf{ProPicker~\cite{wiedemann2026propicker}.} We utilized the official implementation with the provided pretrained weights (ProPicker and TomoTwin~\cite{rice2023tomotwin}). Inference was performed exactly according to the officially suggested parameter configurations.

\subsection{Weakly Supervised Learning Methods for Cryo-ET Data}
\noindent \textbf{Huang et al.~\cite{huang2022accurate}.} We configured the environment using the official repository. The model was trained and evaluated strictly following the authors' recommended hyperparameters.

\noindent \textbf{TomoPicker~\cite{uddin2024tomopicker}.} Using the official codebase, we trained the model with the suggested \texttt{GE-KL} objective, which is documented as the optimal configuration. Inference was subsequently conducted using all default hyperparameters.

\subsection{Other Methods for Cryo-ET Data}
\noindent \textbf{Zeng et al.~\cite{zeng2022structure}.} As the authors do not provide source code or specific implementation details for their 2D-to-3D backprojection mechanics, we rigorously re-implemented their approach to ensure a fair comparison. While the original paper utilized Faster R-CNN~\cite{ren2016faster} as the 2D detector, we upgraded the architecture to a DETR-based detector to align with our framework's baseline capabilities. Specifically, we integrated DETR with our multiclass visual prompt encoder and replaced the standard decoder with a 3-stage formulation~\cite{liu2024grounding}. To handle the inherently high false-positive rate of independent 2D detections during 3D backprojection, we implemented a RANSAC-based~\cite{fischler1981random} algorithm. The model was trained using the exact same dataset and hyperparameters as FullTilt. Training was conducted using PyTorch 2.5.1 across four NVIDIA RTX 6000 Ada GPUs.

\subsection{Multi-View 3D Object Detection Methods}
\noindent \textbf{DETR3D~\cite{wang2022detr3d}.} We adapted the DETR3D architecture by integrating our multiclass visual prompt encoder and swapping its default decoder for a 3-stage decoder~\cite{liu2024grounding}. The model was trained using the exact same dataset as FullTilt, plus data generated by our auxiliary geometric primitives ($\mathcal{G}$) module. Training was executed with PyTorch 2.5.1 on four NVIDIA RTX 6000 Ada GPUs.

\noindent \textbf{PETR~\cite{liu2022petr}.} Similar to DETR3D, we integrated our multiclass visual prompt encoder into the PETR architecture and utilized a 3-stage decoder~\cite{liu2024grounding}. The model was trained on the identical FullTilt dataset (plus $\mathcal{G}$). Due to the extensively higher VRAM requirements of PETR's global attention mechanism, training was conducted on four NVIDIA H200 GPUs. However, to ensure a fair latency comparison against our method, all inference benchmarking for PETR was performed on a single NVIDIA RTX 6000 Ada GPU.

\section{Comprehensive Evaluation}
\label{sec:supp_eval}
\subsection{Additional Intra-Instance Detection Results}
\label{sec:supp_eval_intra_sample}
The evaluation results on the two simulation datasets are provided in Tables~\ref{tab:intra_sample_sim_same} and~\ref{tab:intra_sample_sim_diff}. While FullTilt demonstrates superior performance on real-world datasets, it slightly underperforms compared to TomoTwin and ProPicker on some metrics within the simulated environments. Nevertheless, FullTilt remains the fastest and most memory-efficient framework among all evaluated open-set cryo-ET detection methods. This performance discrepancy on synthetic data might suggest that TomoTwin and ProPicker (which relies on TomoTwin's pretrained embeddings) overfit to the idealized imaging conditions of the simulation distribution.

Furthermore, we provide detailed class-level performance breakdowns across the DS-10440, ``seen''simulation, and ``unseen'' simulation datasets. These comprehensive metrics are tabulated individually for TomoTwin (Tables~\ref{tab:intra_sample_detail_czii_tomotwin}, \ref{tab:intra_sample_detail_sim_same_tomotwin}, and \ref{tab:intra_sample_detail_sim_diff_tomotwin}), ProPicker (Tables~\ref{tab:intra_sample_detail_czii_propicker}, \ref{tab:intra_sample_detail_sim_same_propicker}, and \ref{tab:intra_sample_detail_sim_diff_propicker}), and FullTilt (Tables~\ref{tab:intra_sample_detail_czii_fulltilt}, \ref{tab:intra_sample_detail_sim_same_fulltilt}, and \ref{tab:intra_sample_detail_sim_diff_fulltilt}).

\begin{table*}[h]
\centering
\caption{\textbf{Intra-instance detection results on the simulation dataset (seen).} The best-performing cryo-ET method for each metric is highlighted in bold.}
\label{tab:intra_sample_sim_same}
\resizebox{\textwidth}{!}{%
\begin{threeparttable}
\begin{tabular}{@{}ll|c|rrrrr@{}}
\toprule
\textbf{Category} &
\textbf{Method} &
\textbf{\# Prompts} & 
\textbf{mAP@0.5r $\uparrow$} & \textbf{mAP@1r $\uparrow$} &
\textbf{F1 $\uparrow$} &
\textbf{T. $\downarrow$} &
\textbf{M. $\downarrow$}
\\

\midrule
\multirow{2}{*}{General MV} & 
DETR3D~\cite{wang2022detr3d}$+\mathcal{E}_{\text{Prompt}}$ & \multirow{7}{*}{1} & 0.317 ± 0.013 & 0.520 ± 0.020 & 0.519 ± 0.007 & 0.322 & 4800 \\

& PETR~\cite{liu2022petr}$+\mathcal{E}_{\text{Prompt}}$ & & 0.000 ± 0.000 & 0.001 ± 0.000 & 0.009 ± 0.000 & 0.849 & 17700 \\

\cmidrule{1-2} \cmidrule{4-8}
\multirow{5}{*}{Cryo-ET} & 
Zeng et al.~\cite{zeng2022structure}$+\mathcal{E}_{\text{Prompt}}$ & &
n.a. & n.a. & 0.003 ± 0.000 & \cellcolor{second}16.0 & \cellcolor{second}6320 \\ 

 & TomoTwin~\cite{rice2023tomotwin} & &
\cellcolor{second}0.578 ± 0.018 & \cellcolor{best}\textbf{0.708 ± 0.021} & \cellcolor{best}\textbf{0.732 ± 0.019} & 2160 & 20800 \\ 

 & CryoSAM~\cite{zhao2024cryosam} & &
n.a. & n.a. & 0.003 ± 0.001 & 296 & 29100 \\ 

 & ProPicker~\cite{wiedemann2026propicker} & &
n.a. & n.a. & \cellcolor{second}0.683 ± 0.021 & \cellcolor{third}65.6 & \cellcolor{third}12600 \\ 

 & \textbf{FullTilt (ours)} & &
\cellcolor{best}\textbf{0.632 ± 0.006} & \cellcolor{second}0.670 ± 0.007 & \cellcolor{third}0.637 ± 0.008 & \cellcolor{best}\textbf{0.682} & \cellcolor{best}\textbf{5160} \\

\midrule
\midrule
\multirow{2}{*}{General MV} & 
DETR3D~\cite{wang2022detr3d}$+\mathcal{E}_{\text{Prompt}}$ & \multirow{7}{*}{2} & 0.350 ± 0.007 & 0.541 ± 0.007 & 0.533 ± 0.003 & 0.321 & 4800 \\

& PETR~\cite{liu2022petr}$+\mathcal{E}_{\text{Prompt}}$ & & 0.000 ± 0.000 & 0.001 ± 0.000 & 0.009 ± 0.000 & 0.849 & 17700 \\

\cmidrule{1-2} \cmidrule{4-8}
\multirow{5}{*}{Cryo-ET} & Zeng et al.~\cite{zeng2022structure}$+\mathcal{E}_{\text{Prompt}}$ & &
n.a. & n.a. & 0.002 ± 0.000 & \cellcolor{second}16.0 & \cellcolor{second}6320 \\ 

& TomoTwin~\cite{rice2023tomotwin} & &
\cellcolor{second}0.597 ± 0.014 & \cellcolor{best}\textbf{0.726 ± 0.020} & \cellcolor{best}\textbf{0.757 ± 0.021} & 2160 & 20800 \\ 

& CryoSAM~\cite{zhao2024cryosam} & &
n.a. & n.a. & 0.003 ± 0.001 & 295 & 29100 \\ 

& ProPicker~\cite{wiedemann2026propicker} & &
n.a. & n.a. & \cellcolor{second}0.696 ± 0.010 & \cellcolor{third}64.3 & \cellcolor{third}12600 \\ 

& \textbf{FullTilt (ours)}  & &
\cellcolor{best}\textbf{0.631 ± 0.008} & \cellcolor{second}0.670 ± 0.007 & \cellcolor{third}0.645 ± 0.009 & \cellcolor{best}\textbf{0.685} & \cellcolor{best}\textbf{5160} \\

\midrule
\midrule
\multirow{2}{*}{General MV} & 
DETR3D~\cite{wang2022detr3d}$+\mathcal{E}_{\text{Prompt}}$ & \multirow{7}{*}{4} & 0.352 ± 0.005 & 0.533 ± 0.005 & 0.527 ± 0.002 & 0.325 & 4800 \\

& PETR~\cite{liu2022petr}$+\mathcal{E}_{\text{Prompt}}$ & & 0.000 ± 0.000 & 0.001 ± 0.000 & 0.009 ± 0.000 & 0.847 & 17700 \\

\cmidrule{1-2} \cmidrule{4-8}
\multirow{5}{*}{Cryo-ET} & Zeng et al.~\cite{zeng2022structure}$+\mathcal{E}_{\text{Prompt}}$ & &
n.a. & n.a. & 0.002 ± 0.000 & \cellcolor{second}16.0 & \cellcolor{second}6320 \\ 

& TomoTwin~\cite{rice2023tomotwin} & &
\cellcolor{second}0.604 ± 0.013 & \cellcolor{best}\textbf{0.739 ± 0.026} & \cellcolor{best}\textbf{0.772 ± 0.030} & 2160 & 20800 \\ 

& CryoSAM~\cite{zhao2024cryosam} & &
n.a. & n.a. & 0.004 ± 0.001 & 297 & 29200 \\ 

& ProPicker~\cite{wiedemann2026propicker} & &
n.a. & n.a. & \cellcolor{second}0.700 ± 0.009 & \cellcolor{third}64.3 & \cellcolor{third}12600 \\ 

& \textbf{FullTilt (ours)} & &
\cellcolor{best}\textbf{0.627 ± 0.008} & \cellcolor{second}0.662 ± 0.007 & \cellcolor{third}0.648 ± 0.005 & \cellcolor{best}\textbf{0.687} & \cellcolor{best}\textbf{5160} \\

\bottomrule
\end{tabular}%

\begin{tablenotes}[flushleft]\footnotesize
\item $+\mathcal{E}_{\text{Prompt}}$: with our multiclass visual prompt encoder module; MV: multi-view; n.a.: not applicable because the metric cannot be measured; T.: runtime (s); M.: VRAM (MB); $\downarrow$: lower is better; $\uparrow$: higher is better
\end{tablenotes}
\end{threeparttable}
}
\end{table*}

\begin{table*}[h]
\centering
\caption{\textbf{Zero-shot intra-instance detection results on the simulation dataset (unseen).} The best-performing cryo-ET method for each metric is highlighted in bold.}
\label{tab:intra_sample_sim_diff}
\resizebox{\textwidth}{!}{%
\begin{threeparttable}
\begin{tabular}{@{}ll|c|rrrrr@{}}
\toprule
\textbf{Category} &
\textbf{Method} &
\textbf{\# Prompts} & 
\textbf{mAP@0.5r $\uparrow$} & \textbf{mAP@1r $\uparrow$} &
\textbf{F1 $\uparrow$} &
\textbf{T. $\downarrow$} &
\textbf{M. $\downarrow$}
\\

\midrule
\multirow{2}{*}{General MV} & 
DETR3D~\cite{wang2022detr3d}$+\mathcal{E}_{\text{Prompt}}$ & \multirow{7}{*}{1} & 0.208 ± 0.009 & 0.392 ± 0.009 & 0.426 ± 0.006 & 0.321 & 4800 \\

& PETR~\cite{liu2022petr}$+\mathcal{E}_{\text{Prompt}}$ & & 0.000 ± 0.000 & 0.000 ± 0.000 & 0.004 ± 0.000 & 0.848 & 17700 \\

\cmidrule{1-2} \cmidrule{4-8}
\multirow{5}{*}{Cryo-ET} & 
Zeng et al.~\cite{zeng2022structure}$+\mathcal{E}_{\text{Prompt}}$ & &
n.a. & n.a. & 0.001 ± 0.000 & \cellcolor{second}17.5 & \cellcolor{second}6310 \\ 

 & TomoTwin~\cite{rice2023tomotwin} & &
\cellcolor{best}\textbf{0.531 ± 0.010} & \cellcolor{best}\textbf{0.678 ± 0.021} & \cellcolor{best}\textbf{0.697 ± 0.025} & 2160 & 20800 \\ 

 & CryoSAM~\cite{zhao2024cryosam} & &
n.a. & n.a. & 0.002 ± 0.001 & 293 & 29100 \\ 

 & ProPicker~\cite{wiedemann2026propicker} & &
n.a. & n.a. & \cellcolor{second}0.663 ± 0.017 & \cellcolor{third}62.9 & \cellcolor{third}12600 \\ 

 & \textbf{FullTilt (ours)} & &
\cellcolor{second}0.502 ± 0.012 & \cellcolor{second}0.542 ± 0.013 & \cellcolor{third}0.521 ± 0.012 & \cellcolor{best}\textbf{0.694} & \cellcolor{best}\textbf{5160} \\

\midrule
\midrule
\multirow{2}{*}{General MV} & 
DETR3D~\cite{wang2022detr3d}$+\mathcal{E}_{\text{Prompt}}$ & \multirow{7}{*}{2} & 0.248 ± 0.009 & 0.418 ± 0.008 & 0.435 ± 0.003 & 0.322 & 4800 \\

& PETR~\cite{liu2022petr}$+\mathcal{E}_{\text{Prompt}}$ & & 0.000 ± 0.000 & 0.000 ± 0.000 & 0.004 ± 0.000 & 0.849 & 17700 \\

\cmidrule{1-2} \cmidrule{4-8}
\multirow{5}{*}{Cryo-ET} & Zeng et al.~\cite{zeng2022structure}$+\mathcal{E}_{\text{Prompt}}$ & &
n.a. & n.a. & 0.001 ± 0.000 & \cellcolor{second}17.5 & \cellcolor{second}6310 \\ 

& TomoTwin~\cite{rice2023tomotwin} & &
\cellcolor{best}\textbf{0.549 ± 0.011} & \cellcolor{best}\textbf{0.697 ± 0.017} & \cellcolor{best}\textbf{0.700 ± 0.019} & 2160 & 20800 \\ 

& CryoSAM~\cite{zhao2024cryosam} & &
n.a. & n.a. & 0.003 ± 0.001 & 294 & 29100 \\ 

& ProPicker~\cite{wiedemann2026propicker} & &
n.a. & n.a. & \cellcolor{second}0.683 ± 0.013 & \cellcolor{third}66.1 & \cellcolor{third}12600 \\ 

& \textbf{FullTilt (ours)}  & &
\cellcolor{second}0.513 ± 0.010  & \cellcolor{second}0.552 ± 0.011 & \cellcolor{third}0.536 ± 0.015 & \cellcolor{best}\textbf{0.698} & \cellcolor{best}\textbf{5160} \\

\midrule
\midrule
\multirow{2}{*}{General MV} & 
DETR3D~\cite{wang2022detr3d}$+\mathcal{E}_{\text{Prompt}}$ & \multirow{7}{*}{4} & 0.256 ± 0.006 & 0.415 ± 0.005 & 0.436 ± 0.003 & 0.320 & 4800 \\

& PETR~\cite{liu2022petr}$+\mathcal{E}_{\text{Prompt}}$ & & 0.000 ± 0.000 & 0.000 ± 0.000 & 0.004 ± 0.000 & 0.848 & 17700 \\

\cmidrule{1-2} \cmidrule{4-8}
\multirow{5}{*}{Cryo-ET} & Zeng et al.~\cite{zeng2022structure}$+\mathcal{E}_{\text{Prompt}}$ & &
n.a. & n.a. & 0.001 ± 0.000 & \cellcolor{second}17.5 & \cellcolor{second}6310 \\ 

& TomoTwin~\cite{rice2023tomotwin} & &
\cellcolor{best}\textbf{0.563 ± 0.007} & \cellcolor{best}\textbf{0.706 ± 0.012} & \cellcolor{best}\textbf{0.717 ± 0.015} & 2160 & 20800 \\ 

& CryoSAM~\cite{zhao2024cryosam} & &
n.a. & n.a. & 0.002 ± 0.000 & 296 & 29100 \\ 

& ProPicker~\cite{wiedemann2026propicker} & &
n.a. & n.a. & \cellcolor{second}0.689 ± 0.011 & \cellcolor{third}61.5 & \cellcolor{third}12600 \\ 

& \textbf{FullTilt (ours)} & &
\cellcolor{second}0.513 ± 0.004 & \cellcolor{second}0.553 ± 0.005 & \cellcolor{third}0.545 ± 0.005 & \cellcolor{best}\textbf{0.699} & \cellcolor{best}\textbf{5160} \\

\bottomrule
\end{tabular}%

\begin{tablenotes}[flushleft]\footnotesize
\item $+\mathcal{E}_{\text{Prompt}}$: with our multiclass visual prompt encoder module; MV: multi-view; n.a.: not applicable because the metric cannot be measured; T.: runtime (s); M.: VRAM (MB); $\downarrow$: lower is better; $\uparrow$: higher is better
\end{tablenotes}
\end{threeparttable}
}
\end{table*}

\begin{table*}[h]
\centering
\caption{\textbf{Detailed zero-shot intra-instance detection results on the DS-10440 dataset (TomoTwin).}}
\label{tab:intra_sample_detail_czii_tomotwin}
\resizebox{\textwidth}{!}{%
\begin{threeparttable}
\begin{tabular}{@{}l|rrr|rrr|rrr@{}}
\toprule
& \multicolumn{3}{c}{\textbf{1 Prompt}} & \multicolumn{3}{c}{\textbf{2 Prompts}} &
\multicolumn{3}{c}{\textbf{4 Prompts}}\\ 
\cmidrule(lr){2-4}\cmidrule(lr){5-7}\cmidrule(lr){8-10}
\textbf{Class} &
\textbf{AP@0.5r $\uparrow$} & \textbf{AP@1r $\uparrow$} &
\textbf{F1 $\uparrow$} &
\textbf{AP@0.5r $\uparrow$} & \textbf{AP@1r $\uparrow$} &
\textbf{F1 $\uparrow$} &
\textbf{AP@0.5r $\uparrow$} & \textbf{AP@1r $\uparrow$} &
\textbf{F1 $\uparrow$}
\\

\midrule
apoferritin &
0.039 ± 0.030 & 0.073 ± 0.043 & 0.102 ± 0.051 &
0.047 ± 0.021 & 0.082 ± 0.031 & 0.146 ± 0.046 &
0.093 ± 0.057 & 0.158 ± 0.081 & 0.210 ± 0.088 \\

$\beta$-amylase &
0.040 ± 0.012 & 0.052 ± 0.018 & 0.029 ± 0.022 &
0.021 ± 0.014 & 0.025 ± 0.016 & 0.032 ± 0.033 &
0.025 ± 0.010 & 0.030 ± 0.010 & 0.043 ± 0.028 \\

$\beta$-galactosidase &
0.065 ± 0.016 & 0.095 ± 0.027 & 0.089 ± 0.023 &
0.047 ± 0.025 & 0.069 ± 0.038 & 0.099 ± 0.057 &
0.042 ± 0.022 & 0.063 ± 0.027 & 0.087 ± 0.043 \\

ribosome &
0.079 ± 0.011 & 0.211 ± 0.023 & 0.079 ± 0.018 &
0.061 ± 0.014 & 0.159 ± 0.044 & 0.079 ± 0.026 &
0.063 ± 0.016 & 0.177 ± 0.044 & 0.081 ± 0.012 \\

thyroglobulin &
0.089 ± 0.012 & 0.117 ± 0.012 & 0.121 ± 0.014 &
0.058 ± 0.011 & 0.078 ± 0.013 & 0.103 ± 0.043 &
0.074 ± 0.016 & 0.091 ± 0.018 & 0.118 ± 0.034 \\

virus-like particles &
0.222 ± 0.102 & 0.241 ± 0.108 & 0.050 ± 0.009 &
0.165 ± 0.068 & 0.178 ± 0.069 & 0.077 ± 0.019 &
0.107 ± 0.052 & 0.113 ± 0.055 & 0.072 ± 0.029 \\

\midrule
mean &
0.089 ± 0.021 & 0.132 ± 0.024 & 0.078 ± 0.012 &
0.066 ± 0.012 & 0.098 ± 0.015 & 0.089 ± 0.010 & 
0.067 ± 0.010 & 0.105 ± 0.015 & 0.102 ± 0.021 \\

\bottomrule
\end{tabular}%

\begin{tablenotes}[flushleft]\footnotesize
\item $\downarrow$: lower is better; $\uparrow$: higher is better
\end{tablenotes}
\end{threeparttable}
}
\end{table*}

\begin{table*}[h]
\centering
\caption{\textbf{Detailed intra-instance detection results on the simulation dataset (seen; TomoTwin).}}
\label{tab:intra_sample_detail_sim_same_tomotwin}
\resizebox{\textwidth}{!}{%
\begin{threeparttable}
\begin{tabular}{@{}l|rrr|rrr|rrr@{}}
\toprule
& \multicolumn{3}{c}{\textbf{1 Prompt}} & \multicolumn{3}{c}{\textbf{2 Prompts}} &
\multicolumn{3}{c}{\textbf{4 Prompts}}\\ 
\cmidrule(lr){2-4}\cmidrule(lr){5-7}\cmidrule(lr){8-10}
\textbf{Class} &
\textbf{AP@0.5r $\uparrow$} & \textbf{AP@1r $\uparrow$} &
\textbf{F1 $\uparrow$} &
\textbf{AP@0.5r $\uparrow$} & \textbf{AP@1r $\uparrow$} &
\textbf{F1 $\uparrow$} &
\textbf{AP@0.5r $\uparrow$} & \textbf{AP@1r $\uparrow$} &
\textbf{F1 $\uparrow$}
\\

\midrule
1BXN &
0.812 ± 0.044 & 0.898 ± 0.035 & 0.897 ± 0.028 &
0.854 ± 0.041 & 0.922 ± 0.030 & 0.933 ± 0.020 &
0.878 ± 0.013 & 0.926 ± 0.015 & 0.920 ± 0.034 \\

1S3X &
0.010 ± 0.009 & 0.170 ± 0.072 & 0.234 ± 0.106 &
0.012 ± 0.009 & 0.141 ± 0.084 & 0.192 ± 0.120 &
0.018 ± 0.014 & 0.228 ± 0.128 & 0.291 ± 0.169 \\

3CF3 &
0.321 ± 0.159 & 0.527 ± 0.255 & 0.544 ± 0.236 &
0.484 ± 0.085 & 0.664 ± 0.108 & 0.733 ± 0.087 &
0.505 ± 0.108 & 0.687 ± 0.134 & 0.772 ± 0.109 \\

3GL1 &
0.010 ± 0.007 & 0.101 ± 0.050 & 0.170 ± 0.059 &
0.013 ± 0.012 & 0.137 ± 0.073 & 0.207 ± 0.090 &
0.018 ± 0.011 & 0.159 ± 0.080 & 0.183 ± 0.112 \\

5JH9 &
0.890 ± 0.012 & 0.900 ± 0.012 & 0.911 ± 0.008 &
0.892 ± 0.004 & 0.901 ± 0.004 & 0.926 ± 0.017 & 
0.893 ± 0.002 & 0.902 ± 0.002 & 0.934 ± 0.002 \\

5OOL &
0.900 ± 0.010 & 0.903 ± 0.006 & 0.927 ± 0.019 & 
0.904 ± 0.008 & 0.905 ± 0.009 & 0.946 ± 0.005 &
0.901 ± 0.004 & 0.901 ± 0.004 & 0.945 ± 0.003 \\

6IGC &
0.912 ± 0.005 & 0.944 ± 0.006 & 0.971 ± 0.003 & 
0.910 ± 0.007 & 0.943 ± 0.005 & 0.970 ± 0.004 & 
0.914 ± 0.005 & 0.944 ± 0.007 & 0.971 ± 0.004 \\

6LXV &
0.926 ± 0.005 & 0.926 ± 0.005 & 0.958 ± 0.003 & 0.916 ± 0.016 & 0.916 ± 0.016 & 0.950 ± 0.014 & 
0.922 ± 0.007 & 0.922 ± 0.007 & 0.956 ± 0.004 \\

7DD9 &
0.842 ± 0.005 & 0.842 ± 0.005 & 0.881 ± 0.005 &
0.829 ± 0.033 & 0.833 ± 0.025 & 0.853 ± 0.007 & 
0.833 ± 0.014 & 0.833 ± 0.014 & 0.866 ± 0.005 \\

7ETM &
0.151 ± 0.033 & 0.864 ± 0.053 & 0.829 ± 0.013 & 
0.156 ± 0.024 & 0.897 ± 0.029 & 0.864 ± 0.018 & 0.158 ± 0.031 & 0.888 ± 0.036 & 0.877 ± 0.017 \\

\midrule
mean & 
0.578 ± 0.018 & 0.708 ± 0.021 & 0.732 ± 0.019 & 0.597 ± 0.014 & 0.726 ± 0.020 & 0.757 ± 0.021 & 0.604 ± 0.013 & 0.739 ± 0.026 & 0.772 ± 0.030 \\

\bottomrule
\end{tabular}%

\begin{tablenotes}[flushleft]\footnotesize
\item $\downarrow$: lower is better; $\uparrow$: higher is better
\end{tablenotes}
\end{threeparttable}
}
\end{table*}

\begin{table*}[h]
\centering
\caption{\textbf{Detailed zero-shot intra-instance detection results on the simulation dataset (unseen; TomoTwin).}}
\label{tab:intra_sample_detail_sim_diff_tomotwin}
\resizebox{\textwidth}{!}{%
\begin{threeparttable}
\begin{tabular}{@{}l|rrr|rrr|rrr@{}}
\toprule
& \multicolumn{3}{c}{\textbf{1 Prompt}} & \multicolumn{3}{c}{\textbf{2 Prompts}} &
\multicolumn{3}{c}{\textbf{4 Prompts}}\\ 
\cmidrule(lr){2-4}\cmidrule(lr){5-7}\cmidrule(lr){8-10}
\textbf{Class} &
\textbf{AP@0.5r $\uparrow$} & \textbf{AP@1r $\uparrow$} &
\textbf{F1 $\uparrow$} &
\textbf{AP@0.5r $\uparrow$} & \textbf{AP@1r $\uparrow$} &
\textbf{F1 $\uparrow$} &
\textbf{AP@0.5r $\uparrow$} & \textbf{AP@1r $\uparrow$} &
\textbf{F1 $\uparrow$}
\\

\midrule
1E9R &
0.885 ± 0.006 & 0.888 ± 0.007 & 0.928 ± 0.004 &
0.880 ± 0.016 & 0.891 ± 0.012 & 0.900 ± 0.005 &
0.883 ± 0.011 & 0.888 ± 0.009 & 0.930 ± 0.007 \\

1F1B &
0.214 ± 0.053 & 0.664 ± 0.085 & 0.670 ± 0.073 & 
0.278 ± 0.050 & 0.794 ± 0.057 & 0.782 ± 0.051 & 
0.357 ± 0.050 & 0.794 ± 0.068 & 0.801 ± 0.046 \\

1FPY & 
0.896 ± 0.046 & 0.929 ± 0.032 & 0.940 ± 0.031 & 
0.924 ± 0.013 & 0.935 ± 0.011 & 0.947 ± 0.016 & 
0.935 ± 0.013 & 0.942 ± 0.010 & 0.952 ± 0.009 \\

1JZ8 &
0.778 ± 0.059 & 0.830 ± 0.044 & 0.836 ± 0.044 &
0.833 ± 0.023 & 0.844 ± 0.026 & 0.851 ± 0.014 & 
0.850 ± 0.024 & 0.864 ± 0.023 & 0.865 ± 0.022 \\

1OAO &
0.691 ± 0.091 & 0.803 ± 0.071 & 0.816 ± 0.058 & 
0.723 ± 0.037 & 0.822 ± 0.046 & 0.806 ± 0.073 & 
0.718 ± 0.025 & 0.813 ± 0.036 & 0.795 ± 0.061 \\

2DF7 & 
0.961 ± 0.007 & 0.961 ± 0.007 & 0.978 ± 0.003 &
0.963 ± 0.005 & 0.963 ± 0.005 & 0.981 ± 0.003 &
0.960 ± 0.005 & 0.960 ± 0.005 & 0.980 ± 0.003 \\
2H12 &
0.728 ± 0.071 & 0.903 ± 0.030 & 0.893 ± 0.036 & 
0.706 ± 0.079 & 0.873 ± 0.085 & 0.883 ± 0.079 &
0.750 ± 0.035 & 0.905 ± 0.015 & 0.890 ± 0.007 \\

3HHB & 
0.076 ± 0.029 & 0.531 ± 0.124 & 0.589 ± 0.112 & 
0.088 ± 0.037 & 0.542 ± 0.118 & 0.528 ± 0.211 & 
0.077 ± 0.019 & 0.581 ± 0.065 & 0.619 ± 0.056 \\
4D4R &
0.003 ± 0.003 & 0.023 ± 0.022 & 0.051 ± 0.032 & 
0.005 ± 0.006 & 0.026 ± 0.028 & 0.050 ± 0.050 & 0.001 ± 0.001 & 0.012 ± 0.003 & 0.034 ± 0.018 \\

6T3E &
0.079 ± 0.045 & 0.250 ± 0.085 & 0.273 ± 0.107 & 
0.093 ± 0.034 & 0.276 ± 0.097 & 0.270 ± 0.105 & 
0.100 ± 0.028 & 0.304 ± 0.081 & 0.305 ± 0.083 \\

\midrule
mean & 
0.531 ± 0.010 & 0.678 ± 0.021 & 0.697 ± 0.025 &
0.549 ± 0.011 & 0.697 ± 0.017 & 0.700 ± 0.019 & 
0.563 ± 0.007 & 0.706 ± 0.012 & 0.717 ± 0.015 \\

\bottomrule
\end{tabular}%

\begin{tablenotes}[flushleft]\footnotesize
\item $\downarrow$: lower is better; $\uparrow$: higher is better
\end{tablenotes}
\end{threeparttable}
}
\end{table*}

\begin{table*}[h]
\centering
\caption{\textbf{Detailed zero-shot intra-instance detection results on the DS-10440 dataset (ProPicker).}}
\label{tab:intra_sample_detail_czii_propicker}
\resizebox{\textwidth}{!}{%
\begin{threeparttable}
\begin{tabular}{@{}l|rrr|rrr|rrr@{}}
\toprule
& \multicolumn{3}{c}{\textbf{1 Prompt}} & \multicolumn{3}{c}{\textbf{2 Prompts}} &
\multicolumn{3}{c}{\textbf{4 Prompts}}\\ 
\cmidrule(lr){2-4}\cmidrule(lr){5-7}\cmidrule(lr){8-10}
\textbf{Class} &
\textbf{AP@0.5r $\uparrow$} & \textbf{AP@1r $\uparrow$} &
\textbf{F1 $\uparrow$} &
\textbf{AP@0.5r $\uparrow$} & \textbf{AP@1r $\uparrow$} &
\textbf{F1 $\uparrow$} &
\textbf{AP@0.5r $\uparrow$} & \textbf{AP@1r $\uparrow$} &
\textbf{F1 $\uparrow$}
\\

\midrule
apoferritin &
n.a. & n.a. & 0.131 ± 0.032 &
n.a. & n.a. & 0.159 ± 0.019 &
n.a. & n.a. & 0.175 ± 0.024 \\

$\beta$-amylase &
n.a. & n.a. & 0.013 ± 0.004 &
n.a. & n.a. & 0.016 ± 0.004 &
n.a. & n.a. & 0.018 ± 0.001 \\

$\beta$-galactosidase &
n.a. & n.a. & 0.029 ± 0.004 &
n.a. & n.a. & 0.027 ± 0.003 &
n.a. & n.a. & 0.030 ± 0.003 \\

ribosome &
n.a. & n.a. & 0.085 ± 0.026 &
n.a. & n.a. & 0.119 ± 0.017 &
n.a. & n.a. & 0.145 ± 0.018 \\

thyroglobulin &
n.a. & n.a. & 0.071 ± 0.006 &
n.a. & n.a. & 0.073 ± 0.005 &
n.a. & n.a. & 0.077 ± 0.003 \\

virus-like particles &
n.a. & n.a. & 0.039 ± 0.008 &
n.a. & n.a. & 0.037 ± 0.008 & 
n.a. & n.a. & 0.041 ± 0.006 \\

\midrule
mean &
n.a. & n.a. & 0.061 ± 0.008 &
n.a. & n.a. & 0.072 ± 0.004 &
n.a. & n.a. & 0.081 ± 0.004 \\

\bottomrule
\end{tabular}%

\begin{tablenotes}[flushleft]\footnotesize
\item n.a.: not applicable because the metric cannot be measured; $\downarrow$: lower is better; $\uparrow$: higher is better
\end{tablenotes}
\end{threeparttable}
}
\end{table*}

\begin{table*}[h]
\centering
\caption{\textbf{Detailed intra-instance detection results on the simulation dataset (seen; ProPicker).}}
\label{tab:intra_sample_detail_sim_same_propicker}
\resizebox{\textwidth}{!}{%
\begin{threeparttable}
\begin{tabular}{@{}l|rrr|rrr|rrr@{}}
\toprule
& \multicolumn{3}{c}{\textbf{1 Prompt}} & \multicolumn{3}{c}{\textbf{2 Prompts}} &
\multicolumn{3}{c}{\textbf{4 Prompts}}\\ 
\cmidrule(lr){2-4}\cmidrule(lr){5-7}\cmidrule(lr){8-10}
\textbf{Class} &
\textbf{AP@0.5r $\uparrow$} & \textbf{AP@1r $\uparrow$} &
\textbf{F1 $\uparrow$} &
\textbf{AP@0.5r $\uparrow$} & \textbf{AP@1r $\uparrow$} &
\textbf{F1 $\uparrow$} &
\textbf{AP@0.5r $\uparrow$} & \textbf{AP@1r $\uparrow$} &
\textbf{F1 $\uparrow$}
\\

\midrule
1BXN &
n.a. & n.a. & 0.814 ± 0.018 &
n.a. & n.a. & 0.803 ± 0.019 &
n.a. & n.a. & 0.810 ± 0.011 \\

1S3X &
n.a. & n.a. & 0.247 ± 0.106 &
n.a. & n.a. & 0.202 ± 0.047 &
n.a. & n.a. & 0.248 ± 0.064 \\

3CF3 &
n.a. & n.a. & 0.535 ± 0.156 &
n.a. & n.a. & 0.606 ± 0.109 &
n.a. & n.a. & 0.611 ± 0.071 \\

3GL1 &
n.a. & n.a. & 0.153 ± 0.031 &
n.a. & n.a. & 0.189 ± 0.053 &
n.a. & n.a. & 0.166 ± 0.023 \\

5JH9 &
n.a. & n.a. & 0.828 ± 0.001 &
n.a. & n.a. & 0.827 ± 0.001 &
n.a. & n.a. & 0.829 ± 0.002 \\

5OOL &
n.a. & n.a. & 0.887 ± 0.073 &
n.a. & n.a. & 0.906 ± 0.026 &
n.a. & n.a. & 0.919 ± 0.007 \\

6IGC &
n.a. & n.a. & 0.882 ± 0.002 &
n.a. & n.a. & 0.882 ± 0.003 & 
n.a. & n.a. & 0.878 ± 0.004 \\

6LXV &
n.a. & n.a. & 0.949 ± 0.004 &
n.a. & n.a. & 0.949 ± 0.004 & 
n.a. & n.a. & 0.951 ± 0.008 \\

7DD9 &
n.a. & n.a. & 0.790 ± 0.163 &
n.a. & n.a. & 0.842 ± 0.009 &
n.a. & n.a. & 0.844 ± 0.006 \\

7ETM &
n.a. & n.a. & 0.745 ± 0.018 &
n.a. & n.a. & 0.749 ± 0.010 &
n.a. & n.a. & 0.743 ± 0.006 \\

\midrule
mean & 
n.a. & n.a. & 0.683 ± 0.021 &
n.a. & n.a. & 0.696 ± 0.010 &
n.a. & n.a. & 0.700 ± 0.009 \\

\bottomrule
\end{tabular}%

\begin{tablenotes}[flushleft]\footnotesize
\item n.a.: not applicable because the metric cannot be measured; $\downarrow$: lower is better; $\uparrow$: higher is better
\end{tablenotes}
\end{threeparttable}
}
\end{table*}

\begin{table*}[h]
\centering
\caption{\textbf{Detailed zero-shot intra-instance detection results on the simulation dataset (unseen; ProPicker).}}
\label{tab:intra_sample_detail_sim_diff_propicker}
\resizebox{\textwidth}{!}{%
\begin{threeparttable}
\begin{tabular}{@{}l|rrr|rrr|rrr@{}}
\toprule
& \multicolumn{3}{c}{\textbf{1 Prompt}} & \multicolumn{3}{c}{\textbf{2 Prompts}} &
\multicolumn{3}{c}{\textbf{4 Prompts}}\\ 
\cmidrule(lr){2-4}\cmidrule(lr){5-7}\cmidrule(lr){8-10}
\textbf{Class} &
\textbf{AP@0.5r $\uparrow$} & \textbf{AP@1r $\uparrow$} &
\textbf{F1 $\uparrow$} &
\textbf{AP@0.5r $\uparrow$} & \textbf{AP@1r $\uparrow$} &
\textbf{F1 $\uparrow$} &
\textbf{AP@0.5r $\uparrow$} & \textbf{AP@1r $\uparrow$} &
\textbf{F1 $\uparrow$}
\\

\midrule
1E9R &
n.a. & n.a. & 0.903 ± 0.004 &
n.a. & n.a. & 0.895 ± 0.013 &
n.a. & n.a. & 0.895 ± 0.006 \\

1F1B &
n.a. & n.a. & 0.781 ± 0.066 &
n.a. & n.a. & 0.793 ± 0.036 &
n.a. & n.a. & 0.816 ± 0.005 \\

1FPY & 
n.a. & n.a. & 0.756 ± 0.110 &
n.a. & n.a. & 0.812 ± 0.070 &
n.a. & n.a. & 0.826 ± 0.056 \\

1JZ8 &
n.a. & n.a. & 0.762 ± 0.104 &
n.a. & n.a. & 0.783 ± 0.072 &
n.a. & n.a. & 0.786 ± 0.080 \\

1OAO &
n.a. & n.a. & 0.791 ± 0.096 &
n.a. & n.a. & 0.842 ± 0.028 &
n.a. & n.a. & 0.869 ± 0.022 \\

2DF7 & 
n.a. & n.a. & 0.897 ± 0.024 &
n.a. & n.a. & 0.893 ± 0.038 &
n.a. & n.a. & 0.873 ± 0.029 \\

2H12 &
n.a. & n.a. & 0.797 ± 0.035 &
n.a. & n.a. & 0.791 ± 0.017 &
n.a. & n.a. & 0.789 ± 0.015 \\

3HHB & 
n.a. & n.a. & 0.558 ± 0.112 &
n.a. & n.a. & 0.595 ± 0.074 &
n.a. & n.a. & 0.616 ± 0.032 \\

4D4R &
n.a. & n.a. & 0.105 ± 0.013 &
n.a. & n.a. & 0.120 ± 0.020 &
n.a. & n.a. & 0.117 ± 0.011 \\

6T3E &
n.a. & n.a. & 0.284 ± 0.076 &
n.a. & n.a. & 0.309 ± 0.059 &
n.a. & n.a. & 0.300 ± 0.043 \\

\midrule
mean & 
n.a. & n.a. & 0.663 ± 0.017 &
n.a. & n.a. & 0.683 ± 0.013 &
n.a. & n.a. & 0.689 ± 0.011 \\

\bottomrule
\end{tabular}%

\begin{tablenotes}[flushleft]\footnotesize
\item n.a.: not applicable because the metric cannot be measured; $\downarrow$: lower is better; $\uparrow$: higher is better
\end{tablenotes}
\end{threeparttable}
}
\end{table*}

\begin{table*}[h]
\centering
\caption{\textbf{Detailed zero-shot intra-instance detection results on the DS-10440 dataset (FullTilt).}}
\label{tab:intra_sample_detail_czii_fulltilt}
\resizebox{\textwidth}{!}{%
\begin{threeparttable}
\begin{tabular}{@{}l|rrr|rrr|rrr@{}}
\toprule
& \multicolumn{3}{c}{\textbf{1 Prompt}} & \multicolumn{3}{c}{\textbf{2 Prompts}} &
\multicolumn{3}{c}{\textbf{4 Prompts}}\\ 
\cmidrule(lr){2-4}\cmidrule(lr){5-7}\cmidrule(lr){8-10}
\textbf{Class} &
\textbf{AP@0.5r $\uparrow$} & \textbf{AP@1r $\uparrow$} &
\textbf{F1 $\uparrow$} &
\textbf{AP@0.5r $\uparrow$} & \textbf{AP@1r $\uparrow$} &
\textbf{F1 $\uparrow$} &
\textbf{AP@0.5r $\uparrow$} & \textbf{AP@1r $\uparrow$} &
\textbf{F1 $\uparrow$}
\\

\midrule
apoferritin &
0.002 ± 0.001 & 0.009 ± 0.002 & 0.056 ± 0.006 & 
0.002 ± 0.000 & 0.009 ± 0.001 & 0.055 ± 0.007 &
0.003 ± 0.000 & 0.011 ± 0.001 & 0.060 ± 0.005 \\  

$\beta$-amylase &
0.018 ± 0.007 & 0.027 ± 0.010 & 0.083 ± 0.016 & 
0.021 ± 0.007 & 0.032 ± 0.010 & 0.087 ± 0.012 &
0.024 ± 0.004 & 0.036 ± 0.005 & 0.084 ± 0.015 \\

$\beta$-galactosidase &
0.052 ± 0.015 & 0.073 ± 0.017 & 0.132 ± 0.009 & 
0.057 ± 0.009 & 0.073 ± 0.009 & 0.144 ± 0.016 &
0.057 ± 0.008 & 0.072 ± 0.009 & 0.142 ± 0.010 \\

ribosome &
0.503 ± 0.067 & 0.581 ± 0.071 & 0.559 ± 0.048 & 
0.559 ± 0.045 & 0.641 ± 0.044 & 0.622 ± 0.046 &
0.570 ± 0.029 & 0.654 ± 0.025 & 0.641 ± 0.030 \\

thyroglobulin &
0.087 ± 0.016 & 0.115 ± 0.018 & 0.197 ± 0.019 &
0.087 ± 0.010 & 0.116 ± 0.009 & 0.204 ± 0.023 &
0.101 ± 0.015 & 0.134 ± 0.014 & 0.235 ± 0.021 \\

virus-like particles &
0.045 ± 0.020 & 0.084 ± 0.025 & 0.131 ± 0.038 & 
0.048 ± 0.020 & 0.094 ± 0.030 & 0.141 ± 0.035 & 
0.044 ± 0.013 & 0.098 ± 0.015 & 0.166 ± 0.018 \\

\midrule
mean &
0.118 ± 0.013 & 0.148 ± 0.013 & 0.193 ± 0.008 & 
0.129 ± 0.010 & 0.161 ± 0.011 & 0.209 ± 0.013 & 
0.133 ± 0.007 & 0.180 ± 0.007 & 0.221 ± 0.008 \\

\bottomrule
\end{tabular}%

\begin{tablenotes}[flushleft]\footnotesize
\item $\downarrow$: lower is better; $\uparrow$: higher is better
\end{tablenotes}
\end{threeparttable}
}
\end{table*}

\begin{table*}[h]
\centering
\caption{\textbf{Detailed intra-instance detection results on the simulation dataset (seen; FullTilt).}}
\label{tab:intra_sample_detail_sim_same_fulltilt}
\resizebox{\textwidth}{!}{%
\begin{threeparttable}
\begin{tabular}{@{}l|rrr|rrr|rrr@{}}
\toprule
& \multicolumn{3}{c}{\textbf{1 Prompt}} & \multicolumn{3}{c}{\textbf{2 Prompts}} &
\multicolumn{3}{c}{\textbf{4 Prompts}}\\ 
\cmidrule(lr){2-4}\cmidrule(lr){5-7}\cmidrule(lr){8-10}
\textbf{Class} &
\textbf{AP@0.5r $\uparrow$} & \textbf{AP@1r $\uparrow$} &
\textbf{F1 $\uparrow$} &
\textbf{AP@0.5r $\uparrow$} & \textbf{AP@1r $\uparrow$} &
\textbf{F1 $\uparrow$} &
\textbf{AP@0.5r $\uparrow$} & \textbf{AP@1r $\uparrow$} &
\textbf{F1 $\uparrow$}
\\

\midrule
1BXN &
0.802 ± 0.053 & 0.833 ± 0.054 & 0.770 ± 0.062 &
0.847 ± 0.017 & 0.876 ± 0.018 & 0.809 ± 0.028 &
0.844 ± 0.007 & 0.870 ± 0.007 & 0.818 ± 0.009 \\

1S3X &
0.000 ± 0.000 & 0.000 ± 0.000 & 0.003 ± 0.002 &
0.000 ± 0.000 & 0.000 ± 0.000 & 0.002 ± 0.003 & 
0.000 ± 0.000 & 0.000 ± 0.000 & 0.001 ± 0.001 \\

3CF3 &
0.583 ± 0.015 & 0.647 ± 0.015 & 0.588 ± 0.017 & 
0.575 ± 0.016 & 0.645 ± 0.015 & 0.586 ± 0.013 & 
0.566 ± 0.012 & 0.634 ± 0.013 & 0.576 ± 0.011 \\

3GL1 &
0.000 ± 0.000 & 0.000 ± 0.000 & 0.003 ± 0.002 & 
0.000 ± 0.000 & 0.000 ± 0.000 & 0.004 ± 0.002 & 
0.000 ± 0.000 & 0.000 ± 0.000 & 0.003 ± 0.002 \\

5JH9 &
0.786 ± 0.056 & 0.808 ± 0.055 & 0.768 ± 0.019 & 
0.779 ± 0.029 & 0.809 ± 0.030 & 0.775 ± 0.022 & 
0.783 ± 0.045 & 0.808 ± 0.042 & 0.791 ± 0.003 \\

5OOL &
0.806 ± 0.035 & 0.839 ± 0.035 & 0.828 ± 0.007 & 
0.790 ± 0.017 & 0.824 ± 0.017 & 0.840 ± 0.008 & 
0.783 ± 0.006 & 0.814 ± 0.008 & 0.843 ± 0.005 \\

6IGC &
0.985 ± 0.004 & 0.991 ± 0.002 & 0.987 ± 0.003 & 
0.984 ± 0.002 & 0.992 ± 0.002 & 0.987 ± 0.002 & 
0.984 ± 0.003 & 0.991 ± 0.003 & 0.986 ± 0.003 \\

6LXV &
0.949 ± 0.018 & 0.951 ± 0.017 & 0.862 ± 0.024 & 
0.957 ± 0.008 & 0.959 ± 0.006 & 0.870 ± 0.021 &
0.961 ± 0.004 & 0.962 ± 0.004 & 0.891 ± 0.009 \\

7DD9 &
0.809 ± 0.065 & 0.827 ± 0.068 & 0.809 ± 0.013 & 
0.779 ± 0.070 & 0.792 ± 0.066 & 0.806 ± 0.010 & 
0.725 ± 0.034 & 0.738 ± 0.034 & 0.802 ± 0.008 \\

7ETM &
0.602 ± 0.027 & 0.799 ± 0.020 & 0.758 ± 0.048 & 
0.601 ± 0.026 & 0.800 ± 0.017 & 0.770 ± 0.047 &
0.625 ± 0.017 & 0.805 ± 0.016 & 0.770 ± 0.038 \\

\midrule
mean & 
0.632 ± 0.006 & 0.670 ± 0.007 & 0.637 ± 0.008 & 
0.631 ± 0.008 & 0.670 ± 0.007 & 0.645 ± 0.009 & 
0.627 ± 0.008 & 0.662 ± 0.007 & 0.648 ± 0.005 \\

\bottomrule
\end{tabular}%

\begin{tablenotes}[flushleft]\footnotesize
\item $\downarrow$: lower is better; $\uparrow$: higher is better
\end{tablenotes}
\end{threeparttable}
}
\end{table*}

\begin{table*}[h]
\centering
\caption{\textbf{Detailed zero-shot intra-instance detection results on the simulation dataset (unseen; FullTilt).}}
\label{tab:intra_sample_detail_sim_diff_fulltilt}
\resizebox{\textwidth}{!}{%
\begin{threeparttable}
\begin{tabular}{@{}l|rrr|rrr|rrr@{}}
\toprule
& \multicolumn{3}{c}{\textbf{1 Prompt}} & \multicolumn{3}{c}{\textbf{2 Prompts}} &
\multicolumn{3}{c}{\textbf{4 Prompts}}\\ 
\cmidrule(lr){2-4}\cmidrule(lr){5-7}\cmidrule(lr){8-10}
\textbf{Class} &
\textbf{AP@0.5r $\uparrow$} & \textbf{AP@1r $\uparrow$} &
\textbf{F1 $\uparrow$} &
\textbf{AP@0.5r $\uparrow$} & \textbf{AP@1r $\uparrow$} &
\textbf{F1 $\uparrow$} &
\textbf{AP@0.5r $\uparrow$} & \textbf{AP@1r $\uparrow$} &
\textbf{F1 $\uparrow$}
\\

\midrule
1E9R &
0.654 ± 0.055 & 0.711 ± 0.051 & 0.681 ± 0.013 &
0.662 ± 0.047 & 0.720 ± 0.045 & 0.683 ± 0.013 &
0.661 ± 0.015 & 0.715 ± 0.020 & 0.666 ± 0.031 \\

1F1B &
0.281 ± 0.044 & 0.486 ± 0.060 & 0.596 ± 0.030 & 
0.278 ± 0.025 & 0.485 ± 0.040 & 0.612 ± 0.016 & 
0.281 ± 0.008 & 0.491 ± 0.016 & 0.618 ± 0.010 \\

1FPY & 
0.860 ± 0.039 & 0.864 ± 0.038 & 0.789 ± 0.011 & 
0.895 ± 0.013 & 0.899 ± 0.013 & 0.805 ± 0.006 & 
0.910 ± 0.008 & 0.915 ± 0.007 & 0.819 ± 0.007 \\

1JZ8 &
0.850 ± 0.064 & 0.857 ± 0.065 & 0.774 ± 0.014 & 
0.893 ± 0.020 & 0.898 ± 0.021 & 0.813 ± 0.037 & 
0.902 ± 0.009 & 0.908 ± 0.008 & 0.831 ± 0.004 \\

1OAO &
0.612 ± 0.044 & 0.660 ± 0.046 & 0.685 ± 0.016 & 
0.618 ± 0.044 & 0.664 ± 0.048 & 0.674 ± 0.012 & 
0.586 ± 0.039 & 0.631 ± 0.042 & 0.680 ± 0.012 \\

2DF7 & 
0.927 ± 0.042 & 0.968 ± 0.042 & 0.891 ± 0.123 & 
0.941 ± 0.070 & 0.975 ± 0.070 & 0.944 ± 0.150 & 
0.965 ± 0.003 & 0.999 ± 0.001 & 0.996 ± 0.001 \\

2H12 &
0.832 ± 0.015 & 0.863 ± 0.013 & 0.766 ± 0.055 & 
0.840 ± 0.008 & 0.874 ± 0.008 & 0.805 ± 0.022 & 
0.826 ± 0.010 & 0.866 ± 0.011 & 0.808 ± 0.017\\

3HHB & 
0.000 ± 0.000 & 0.003 ± 0.001 & 0.014 ± 0.005 & 
0.000 ± 0.000 & 0.003 ± 0.001 & 0.014 ± 0.007 & 
0.000 ± 0.000 & 0.001 ± 0.000 & 0.009 ± 0.004 \\

4D4R &
0.000 ± 0.000 & 0.000 ± 0.000 & 0.001 ± 0.000 & 
0.000 ± 0.000 & 0.000 ± 0.000 & 0.001 ± 0.001 &
0.000 ± 0.000 & 0.000 ± 0.000 & 0.001 ± 0.001 \\

6T3E &
0.000 ± 0.001 & 0.005 ± 0.003 & 0.016 ± 0.008 & 
0.000 ± 0.000 & 0.002 ± 0.003 & 0.013 ± 0.010 & 
0.000 ± 0.000 & 0.002 ± 0.001 & 0.019 ± 0.008 \\

\midrule
mean & 
0.502 ± 0.012 & 0.542 ± 0.013 & 0.521 ± 0.012 &  
0.513 ± 0.010 & 0.552 ± 0.011 & 0.536 ± 0.015 & 
0.513 ± 0.004 & 0.553 ± 0.005 & 0.545 ± 0.005 \\

\bottomrule
\end{tabular}%

\begin{tablenotes}[flushleft]\footnotesize
\item $\downarrow$: lower is better; $\uparrow$: higher is better
\end{tablenotes}
\end{threeparttable}
}
\end{table*}

\clearpage

\subsection{Additional Cross-Instance Detection Results}
\label{sec:supp_eval_croo_sample}
We present cross-instance detection results on the simulated cryo-ET datasets in Table~\ref{tab:cross_sample_sim}, evaluating both the non-zero-shot setting (seen) and the zero-shot setting (unseen). FullTilt achieves the highest mAP@0.5r on the ``seen'' dataset, while consistently securing the second-best performance across the majority of the remaining metrics and dataset splits.

Furthermore, we provide detailed class-level performance breakdowns across the DS-10440, ``seen'' simulation, and ``unseen'' simulation datasets. These comprehensive metrics are tabulated individually for TomoTwin (Tables~\ref{tab:cross_sample_detail_czii_tomotwin}, \ref{tab:cross_sample_detail_sim_same_tomotwin}, and \ref{tab:cross_sample_detail_sim_diff_tomotwin}), ProPicker (Tables~\ref{tab:cross_sample_detail_czii_propicker}, \ref{tab:cross_sample_detail_sim_same_propicker}, and \ref{tab:cross_sample_detail_sim_diff_propicker}), and FullTilt (Tables~\ref{tab:cross_sample_detail_czii_fulltilt}, \ref{tab:cross_sample_detail_sim_same_fulltilt}, and \ref{tab:cross_sample_detail_sim_diff_fulltilt}).

\begin{table*}[h]
\centering
\caption{\textbf{Cross-instance detection results on simulated cryo-ET datasets.} The best-performing open-set method for each metric is highlighted in bold.}
\label{tab:cross_sample_sim}
\resizebox{\textwidth}{!}{%
\begin{threeparttable}
\begin{tabular}{@{}ll|c|rrr|rrr@{}}
\toprule
& & & \multicolumn{3}{c}{\textbf{Simulation (Seen)}} & \multicolumn{3}{c}{\textbf{Simulation (Unseen)}} \\ 
\cmidrule(lr){4-6}\cmidrule(lr){7-9}
\textbf{Cat.} &
\textbf{Method} &
\textbf{\# P.} & 
\textbf{mAP@0.5r $\uparrow$} & \textbf{mAP@1r $\uparrow$} &
\textbf{F1 $\uparrow$} &
\textbf{mAP@0.5r $\uparrow$} & \textbf{mAP@1r $\uparrow$} &
\textbf{F1 $\uparrow$}
\\

\midrule
\multirow{2}{*}{WSL} & 
Huang et al.~\cite{huang2022accurate} & \multirow{5}{*}{1} & n.a. & n.a. & n.a. & n.a. & n.a. & n.a. \\ 
 & TomoPicker~\cite{uddin2024tomopicker} & & 
0.028 ± 0.018 & 0.108 ± 0.037 & 0.138 ± 0.054 &
0.010 ± 0.003 & 0.072 ± 0.010 & 0.088 ± 0.010 \\  

\cmidrule{1-2} \cmidrule{4-9}
\multirow{3}{*}{OS} & TomoTwin~\cite{rice2023tomotwin} & &
\cellcolor{second}0.582 ± 0.023 & \cellcolor{best}\textbf{0.730 ± 0.033} & \cellcolor{best}\textbf{0.761 ± 0.036} &
\cellcolor{best}\textbf{0.543 ± 0.029} & \cellcolor{best}\textbf{0.677 ± 0.049} & \cellcolor{best}\textbf{0.695 ± 0.050} \\  

& ProPicker~\cite{wiedemann2026propicker} & &
n.a. & n.a. & \cellcolor{second}0.707 ± 0.029 &
n.a. & n.a. & \cellcolor{second}0.674 ± 0.063 \\

& \textbf{FullTilt (ours)} & &
\cellcolor{best}\textbf{0.633 ± 0.006} & \cellcolor{second}0.672 ± 0.007 & \cellcolor{third}0.632 ± 0.007 &
\cellcolor{second}0.500 ± 0.022 & \cellcolor{second}0.538 ± 0.021 & \cellcolor{third}0.514 ± 0.028 \\

\midrule
\midrule

\multirow{2}{*}{WSL} & 
Huang et al.~\cite{huang2022accurate} & \multirow{5}{*}{2} &
0.000 ± 0.000 & 0.000 ± 0.000 & 0.000 ± 0.000 &
0.000 ± 0.000 & 0.000 ± 0.000 & 0.000 ± 0.000 \\  

 & TomoPicker~\cite{uddin2024tomopicker} & &
0.062 ± 0.006 & 0.167 ± 0.013 & 0.189 ± 0.029 &
0.019 ± 0.008 & 0.128 ± 0.013 & 0.125 ± 0.028 \\

\cmidrule{1-2} \cmidrule{4-9}
\multirow{3}{*}{OS} & TomoTwin~\cite{rice2023tomotwin}  & &
\cellcolor{second}0.584 ± 0.034 & \cellcolor{best}\textbf{0.749 ± 0.027} & \cellcolor{best}\textbf{0.781 ± 0.027} &
\cellcolor{best}\textbf{0.564 ± 0.022} & \cellcolor{best}\textbf{0.728 ± 0.025} & \cellcolor{best}\textbf{0.710 ± 0.039} \\

 & ProPicker~\cite{wiedemann2026propicker} & &
n.a. & n.a. & \cellcolor{second}0.714 ± 0.035 &
n.a. & n.a. & \cellcolor{second}0.708 ± 0.022 \\

 & \textbf{FullTilt (ours)} & &
\cellcolor{best}\textbf{0.636 ± 0.008} & \cellcolor{second}0.671 ± 0.007 & \cellcolor{third}0.636 ± 0.008 &
\cellcolor{second}0.510 ± 0.005 & \cellcolor{third}0.546 ± 0.005 & \cellcolor{third}0.528 ± 0.008 \\  

\midrule
\midrule

\multirow{2}{*}{WSL} & 
Huang et al.~\cite{huang2022accurate} & \multirow{5}{*}{4} &
0.000 ± 0.000 & 0.001 ± 0.000 & 0.003 ± 0.003 &
0.004 ± 0.004 & 0.012 ± 0.014 & 0.015 ± 0.017 \\  

 & TomoPicker~\cite{uddin2024tomopicker} & &
0.068 ± 0.018 & 0.209 ± 0.023 & 0.197 ± 0.028 &
0.029 ± 0.002 & 0.132 ± 0.032 & 0.110 ± 0.002 \\  

\cmidrule{1-2} \cmidrule{4-9}
\multirow{3}{*}{OS} & TomoTwin~\cite{rice2023tomotwin} & &
\cellcolor{second}0.608 ± 0.010 & \cellcolor{best}\textbf{0.760 ± 0.017} & \cellcolor{best}\textbf{0.773 ± 0.023} &
\cellcolor{best}\textbf{0.573 ± 0.017} & \cellcolor{best}\textbf{0.741 ± 0.016} & \cellcolor{best}\textbf{0.715 ± 0.029} \\

 & ProPicker~\cite{wiedemann2026propicker} & &
n.a. & n.a. & \cellcolor{second}0.714 ± 0.015 &
n.a. & n.a. & \cellcolor{second}0.694 ± 0.027 \\

 & \textbf{FullTilt (ours)} & &
\cellcolor{best}\textbf{0.625 ± 0.006} & \cellcolor{second}0.660 ± 0.006 & \cellcolor{third}0.635 ± 0.006 &
\cellcolor{second}0.508 ± 0.005 & \cellcolor{third}0.545 ± 0.005 & \cellcolor{third}0.532 ± 0.008 \\

\bottomrule
\end{tabular}%

\begin{tablenotes}[flushleft]\footnotesize
\item Cat.: category; \# P.: \# prompts; WSL: weakly-supervised learning; OS: open-set; -: Huang et al. needs at least 2 prompts; n.a.: not applicable because the metric cannot be measured; $\downarrow$: lower is better; $\uparrow$: higher is better
\end{tablenotes}
\end{threeparttable}
}
\end{table*}

\begin{table*}[h]
\centering
\caption{\textbf{Detailed zero-shot cross-instance detection results on the DS-10440 dataset (TomoTwin).}}
\label{tab:cross_sample_detail_czii_tomotwin}
\resizebox{\textwidth}{!}{%
\begin{threeparttable}
\begin{tabular}{@{}l|rrr|rrr|rrr@{}}
\toprule
& \multicolumn{3}{c}{\textbf{1 Prompt}} & \multicolumn{3}{c}{\textbf{2 Prompts}} &
\multicolumn{3}{c}{\textbf{4 Prompts}}\\ 
\cmidrule(lr){2-4}\cmidrule(lr){5-7}\cmidrule(lr){8-10}
\textbf{Class} &
\textbf{AP@0.5r $\uparrow$} & \textbf{AP@1r $\uparrow$} &
\textbf{F1 $\uparrow$} &
\textbf{AP@0.5r $\uparrow$} & \textbf{AP@1r $\uparrow$} &
\textbf{F1 $\uparrow$} &
\textbf{AP@0.5r $\uparrow$} & \textbf{AP@1r $\uparrow$} &
\textbf{F1 $\uparrow$}
\\

\midrule
apoferritin &
0.037 ± 0.032 & 0.079 ± 0.072 & 0.125 ± 0.135 & 
0.116 ± 0.070 & 0.202 ± 0.108 & 0.161 ± 0.147 &
0.123 ± 0.072 & 0.214 ± 0.111 & 0.191 ± 0.149 \\

$\beta$-amylase &
0.024 ± 0.016 & 0.030 ± 0.022 & 0.047 ± 0.048 & 
0.003 ± 0.003 & 0.006 ± 0.005 & 0.010 ± 0.011 &
0.003 ± 0.003 & 0.007 ± 0.005 & 0.017 ± 0.022 \\

$\beta$-galactosidase &
0.051 ± 0.034 & 0.086 ± 0.045 & 0.108 ± 0.055 & 
0.027 ± 0.016 & 0.047 ± 0.020 & 0.048 ± 0.043 & 
0.033 ± 0.009 & 0.058 ± 0.015 & 0.053 ± 0.044 \\

ribosome &
0.085 ± 0.019 & 0.235 ± 0.043 & 0.089 ± 0.079 & 
0.081 ± 0.016 & 0.233 ± 0.035 & 0.079 ± 0.054 & 
0.087 ± 0.016 & 0.244 ± 0.035 & 0.136 ± 0.123 \\

thyroglobulin &
0.072 ± 0.034 & 0.097 ± 0.038 & 0.105 ± 0.038 & 
0.095 ± 0.030 & 0.119 ± 0.030 & 0.091 ± 0.054 &
0.090 ± 0.021 & 0.117 ± 0.020 & 0.113 ± 0.077 \\

virus-like particles &
0.193 ± 0.142 & 0.200 ± 0.152 & 0.093 ± 0.070 & 
0.223 ± 0.140 & 0.234 ± 0.150 & 0.078 ± 0.094 &
0.301 ± 0.041 & 0.321 ± 0.045 & 0.156 ± 0.138 \\

\midrule
mean &
0.077 ± 0.026 & 0.121 ± 0.029 & 0.095 ± 0.026 & 
0.091 ± 0.044 & 0.140 ± 0.057 & 0.078 ± 0.042 &
0.106 ± 0.026 & 0.160 ± 0.038 & 0.111 ± 0.061 \\

\bottomrule
\end{tabular}%

\begin{tablenotes}[flushleft]\footnotesize
\item $\downarrow$: lower is better; $\uparrow$: higher is better
\end{tablenotes}
\end{threeparttable}
}
\end{table*}

\begin{table*}[h]
\centering
\caption{\textbf{Detailed cross-instance detection results on the simulation dataset (seen; TomoTwin).}}
\label{tab:cross_sample_detail_sim_same_tomotwin}
\resizebox{\textwidth}{!}{%
\begin{threeparttable}
\begin{tabular}{@{}l|rrr|rrr|rrr@{}}
\toprule
& \multicolumn{3}{c}{\textbf{1 Prompt}} & \multicolumn{3}{c}{\textbf{2 Prompts}} &
\multicolumn{3}{c}{\textbf{4 Prompts}}\\ 
\cmidrule(lr){2-4}\cmidrule(lr){5-7}\cmidrule(lr){8-10}
\textbf{Class} &
\textbf{AP@0.5r $\uparrow$} & \textbf{AP@1r $\uparrow$} &
\textbf{F1 $\uparrow$} &
\textbf{AP@0.5r $\uparrow$} & \textbf{AP@1r $\uparrow$} &
\textbf{F1 $\uparrow$} &
\textbf{AP@0.5r $\uparrow$} & \textbf{AP@1r $\uparrow$} &
\textbf{F1 $\uparrow$}
\\

\midrule
1BXN &
0.832 ± 0.088 & 0.929 ± 0.015 & 0.933 ± 0.022 & 
0.761 ± 0.244 & 0.914 ± 0.049 & 0.865 ± 0.049 & 
0.894 ± 0.052 & 0.930 ± 0.009 & 0.909 ± 0.026 \\

1S3X &
0.019 ± 0.018 & 0.295 ± 0.132 & 0.416 ± 0.151 & 
0.034 ± 0.019 & 0.346 ± 0.119 & 0.401 ± 0.132 & 
0.020 ± 0.011 & 0.296 ± 0.141 & 0.364 ± 0.207 \\

3CF3 &
0.327 ± 0.211 & 0.526 ± 0.318 & 0.545 ± 0.304 & 
0.384 ± 0.177 & 0.578 ± 0.222 & 0.678 ± 0.192 & 
0.505 ± 0.085 & 0.689 ± 0.072 & 0.763 ± 0.092 \\

3GL1 &
0.012 ± 0.013 & 0.130 ± 0.114 & 0.157 ± 0.135 & 
0.024 ± 0.019 & 0.237 ± 0.135 & 0.286 ± 0.162 & 
0.026 ± 0.018 & 0.262 ± 0.118 & 0.129 ± 0.127 \\

5JH9 &
0.897 ± 0.008 & 0.907 ± 0.008 & 0.936 ± 0.005 &
0.895 ± 0.006 & 0.904 ± 0.006 & 0.935 ± 0.004 & 
0.892 ± 0.005 & 0.902 ± 0.005 & 0.934 ± 0.003 \\

5OOL &
0.906 ± 0.006 & 0.906 ± 0.006 & 0.947 ± 0.003 & 
0.903 ± 0.007 & 0.903 ± 0.007 & 0.944 ± 0.003 & 
0.901 ± 0.005 & 0.901 ± 0.005 & 0.945 ± 0.002 \\

6IGC &
0.914 ± 0.005 & 0.947 ± 0.005 & 0.972 ± 0.003 & 
0.908 ± 0.010 & 0.939 ± 0.011 & 0.969 ± 0.006 & 
0.916 ± 0.003 & 0.946 ± 0.004 & 0.972 ± 0.002 \\

6LXV &
0.911 ± 0.021 & 0.911 ± 0.021 & 0.944 ± 0.022 & 
0.922 ± 0.009 & 0.922 ± 0.009 & 0.956 ± 0.003 & 
0.926 ± 0.008 & 0.926 ± 0.008 & 0.957 ± 0.004 \\

7DD9 &
0.841 ± 0.005 & 0.841 ± 0.005 & 0.882 ± 0.008 & 
0.844 ± 0.007 & 0.844 ± 0.007 & 0.885 ± 0.006 &
0.842 ± 0.006 & 0.842 ± 0.006 & 0.875 ± 0.007 \\

7ETM &
0.160 ± 0.023 & 0.906 ± 0.017 & 0.881 ± 0.015 & 
0.160 ± 0.028 & 0.902 ± 0.017 & 0.886 ± 0.008 & 
0.154 ± 0.018 & 0.911 ± 0.011 & 0.878 ± 0.022 \\

\midrule
mean & 
0.582 ± 0.023 & 0.730 ± 0.033 & 0.761 ± 0.036 & 
0.584 ± 0.034 & 0.749 ± 0.027 & 0.781 ± 0.027 & 
0.608 ± 0.010 & 0.760 ± 0.017 & 0.773 ± 0.023 \\

\bottomrule
\end{tabular}%

\begin{tablenotes}[flushleft]\footnotesize
\item $\downarrow$: lower is better; $\uparrow$: higher is better
\end{tablenotes}
\end{threeparttable}
}
\end{table*}

\begin{table*}[h]
\centering
\caption{\textbf{Detailed zero-shot cross-instance detection results on the simulation dataset (unseen; TomoTwin).}}
\label{tab:cross_sample_detail_sim_diff_tomotwin}
\resizebox{\textwidth}{!}{%
\begin{threeparttable}
\begin{tabular}{@{}l|rrr|rrr|rrr@{}}
\toprule
& \multicolumn{3}{c}{\textbf{1 Prompt}} & \multicolumn{3}{c}{\textbf{2 Prompts}} &
\multicolumn{3}{c}{\textbf{4 Prompts}}\\ 
\cmidrule(lr){2-4}\cmidrule(lr){5-7}\cmidrule(lr){8-10}
\textbf{Class} &
\textbf{AP@0.5r $\uparrow$} & \textbf{AP@1r $\uparrow$} &
\textbf{F1 $\uparrow$} &
\textbf{AP@0.5r $\uparrow$} & \textbf{AP@1r $\uparrow$} &
\textbf{F1 $\uparrow$} &
\textbf{AP@0.5r $\uparrow$} & \textbf{AP@1r $\uparrow$} &
\textbf{F1 $\uparrow$}
\\

\midrule
1E9R &
0.882 ± 0.007 & 0.890 ± 0.011 & 0.928 ± 0.007 & 
0.883 ± 0.007 & 0.897 ± 0.012 & 0.936 ± 0.010 & 
0.886 ± 0.007 & 0.886 ± 0.007 & 0.930 ± 0.006 \\

1F1B &
0.256 ± 0.101 & 0.716 ± 0.178 & 0.734 ± 0.177 &
0.316 ± 0.098 & 0.814 ± 0.073 & 0.812 ± 0.063 & 
0.330 ± 0.102 & 0.858 ± 0.040 & 0.842 ± 0.045 \\

1FPY & 
0.884 ± 0.076 & 0.917 ± 0.057 & 0.930 ± 0.023 & 
0.928 ± 0.019 & 0.939 ± 0.023 & 0.950 ± 0.015 & 
0.924 ± 0.021 & 0.938 ± 0.015 & 0.939 ± 0.016 \\

1JZ8 &
0.830 ± 0.031 & 0.854 ± 0.033 & 0.854 ± 0.017 & 
0.846 ± 0.027 & 0.857 ± 0.029 & 0.854 ± 0.015 & 
0.847 ± 0.019 & 0.859 ± 0.016 & 0.860 ± 0.013 \\

1OAO &
0.697 ± 0.234 & 0.773 ± 0.255 & 0.775 ± 0.263 & 
0.695 ± 0.151 & 0.798 ± 0.181 & 0.736 ± 0.246 & 
0.762 ± 0.090 & 0.846 ± 0.064 & 0.747 ± 0.054 \\

2DF7 & 
0.961 ± 0.006 & 0.961 ± 0.006 & 0.980 ± 0.003 &
0.960 ± 0.006 & 0.960 ± 0.006 & 0.980 ± 0.003 &
0.960 ± 0.007 & 0.960 ± 0.007 & 0.980 ± 0.003 \\
2H12 &
0.757 ± 0.041 & 0.915 ± 0.009 & 0.917 ± 0.018 & 
0.764 ± 0.055 & 0.913 ± 0.008 & 0.909 ± 0.007 & 
0.778 ± 0.038 & 0.910 ± 0.011 & 0.882 ± 0.015 \\

3HHB & 
0.062 ± 0.052 & 0.426 ± 0.288 & 0.454 ± 0.290 & 
0.086 ± 0.023 & 0.627 ± 0.090 & 0.542 ± 0.278 & 
0.119 ± 0.020 & 0.688 ± 0.049 & 0.577 ± 0.124 \\
4D4R &
0.006 ± 0.006 & 0.062 ± 0.067 & 0.094 ± 0.082 & 
0.006 ± 0.004 & 0.067 ± 0.049 & 0.051 ± 0.068 &
0.007 ± 0.004 & 0.109 ± 0.062 & 0.110 ± 0.120 \\

6T3E &
0.098 ± 0.081 & 0.261 ± 0.158 & 0.282 ± 0.182 & 
0.153 ± 0.061 & 0.404 ± 0.122 & 0.331 ± 0.164 &
0.117 ± 0.067 & 0.355 ± 0.127 & 0.280 ± 0.169 \\

\midrule
mean & 
0.543 ± 0.029 & 0.677 ± 0.049 & 0.695 ± 0.050 & 
0.564 ± 0.022 & 0.728 ± 0.025 & 0.710 ± 0.039 & 
0.573 ± 0.017 & 0.741 ± 0.016 & 0.715 ± 0.029 \\

\bottomrule
\end{tabular}%

\begin{tablenotes}[flushleft]\footnotesize
\item $\downarrow$: lower is better; $\uparrow$: higher is better
\end{tablenotes}
\end{threeparttable}
}
\end{table*}

\begin{table*}[h]
\centering
\caption{\textbf{Detailed zero-shot cross-instance detection results on the DS-10440 dataset (ProPicker).}}
\label{tab:cross_sample_detail_czii_propicker}
\resizebox{\textwidth}{!}{%
\begin{threeparttable}
\begin{tabular}{@{}l|rrr|rrr|rrr@{}}
\toprule
& \multicolumn{3}{c}{\textbf{1 Prompt}} & \multicolumn{3}{c}{\textbf{2 Prompts}} &
\multicolumn{3}{c}{\textbf{4 Prompts}}\\ 
\cmidrule(lr){2-4}\cmidrule(lr){5-7}\cmidrule(lr){8-10}
\textbf{Class} &
\textbf{AP@0.5r $\uparrow$} & \textbf{AP@1r $\uparrow$} &
\textbf{F1 $\uparrow$} &
\textbf{AP@0.5r $\uparrow$} & \textbf{AP@1r $\uparrow$} &
\textbf{F1 $\uparrow$} &
\textbf{AP@0.5r $\uparrow$} & \textbf{AP@1r $\uparrow$} &
\textbf{F1 $\uparrow$}
\\

\midrule
apoferritin &
n.a. & n.a. & 0.175 ± 0.067 &
n.a. & n.a. & 0.141 ± 0.062 &
n.a. & n.a. & 0.188 ± 0.041 \\

$\beta$-amylase &
n.a. & n.a. & 0.017 ± 0.011 &
n.a. & n.a. & 0.015 ± 0.010 &
n.a. & n.a. & 0.015 ± 0.007 \\

$\beta$-galactosidase &
n.a. & n.a. & 0.028 ± 0.013 &
n.a. & n.a. & 0.026 ± 0.008 &
n.a. & n.a. & 0.035 ± 0.011 \\

ribosome &
n.a. & n.a. & 0.119 ± 0.042 &
n.a. & n.a. & 0.118 ± 0.067 &
n.a. & n.a. & 0.141 ± 0.044 \\

thyroglobulin &
n.a. & n.a. & 0.073 ± 0.006 &
n.a. & n.a. & 0.077 ± 0.007 &
n.a. & n.a. & 0.078 ± 0.006 \\

virus-like particles &
n.a. & n.a. & 0.042 ± 0.011 &
n.a. & n.a. & 0.031 ± 0.020 &
n.a. & n.a. & 0.042 ± 0.019 \\

\midrule
mean &
n.a. & n.a. & 0.076 ± 0.013 &
n.a. & n.a. & 0.068 ± 0.018 &
n.a. & n.a. & 0.083 ± 0.011 \\

\bottomrule
\end{tabular}%

\begin{tablenotes}[flushleft]\footnotesize
\item n.a.: not applicable because the metric cannot be measured; $\downarrow$: lower is better; $\uparrow$: higher is better
\end{tablenotes}
\end{threeparttable}
}
\end{table*}

\begin{table*}[h]
\centering
\caption{\textbf{Detailed cross-instance detection results on the simulation dataset (seen; ProPicker).}}
\label{tab:cross_sample_detail_sim_same_propicker}
\resizebox{\textwidth}{!}{%
\begin{threeparttable}
\begin{tabular}{@{}l|rrr|rrr|rrr@{}}
\toprule
& \multicolumn{3}{c}{\textbf{1 Prompt}} & \multicolumn{3}{c}{\textbf{2 Prompts}} &
\multicolumn{3}{c}{\textbf{4 Prompts}}\\ 
\cmidrule(lr){2-4}\cmidrule(lr){5-7}\cmidrule(lr){8-10}
\textbf{Class} &
\textbf{AP@0.5r $\uparrow$} & \textbf{AP@1r $\uparrow$} &
\textbf{F1 $\uparrow$} &
\textbf{AP@0.5r $\uparrow$} & \textbf{AP@1r $\uparrow$} &
\textbf{F1 $\uparrow$} &
\textbf{AP@0.5r $\uparrow$} & \textbf{AP@1r $\uparrow$} &
\textbf{F1 $\uparrow$}
\\

\midrule
1BXN &
n.a. & n.a. & 0.804 ± 0.036 &
n.a. & n.a. & 0.810 ± 0.024 &
n.a. & n.a. & 0.798 ± 0.011 \\

1S3X &
n.a. & n.a. & 0.285 ± 0.160 &
n.a. & n.a. & 0.355 ± 0.141 &
n.a. & n.a. & 0.269 ± 0.099 \\

3CF3 &
n.a. & n.a. & 0.610 ± 0.145 &
n.a. & n.a. & 0.672 ± 0.106 &
n.a. & n.a. & 0.692 ± 0.063 \\

3GL1 &
n.a. & n.a. & 0.277 ± 0.166 &
n.a. & n.a. & 0.203 ± 0.110 &
n.a. & n.a. & 0.200 ± 0.078 \\

5JH9 &
n.a. & n.a. & 0.830 ± 0.002 &
n.a. & n.a. & 0.828 ± 0.002 &
n.a. & n.a. & 0.826 ± 0.006 \\

5OOL &
n.a. & n.a. & 0.916 ± 0.013 &
n.a. & n.a. & 0.910 ± 0.006 &
n.a. & n.a. & 0.921 ± 0.009 \\

6IGC &
n.a. & n.a. & 0.880 ± 0.003 &
n.a. & n.a. & 0.883 ± 0.003 &
n.a. & n.a. & 0.880 ± 0.004 \\

6LXV &
n.a. & n.a. & 0.948 ± 0.002 &
n.a. & n.a. & 0.948 ± 0.002 &
n.a. & n.a. & 0.951 ± 0.006 \\

7DD9 &
n.a. & n.a. & 0.847 ± 0.005 &
n.a. & n.a. & 0.843 ± 0.008 &
n.a. & n.a. & 0.848 ± 0.005 \\

7ETM &
n.a. & n.a. & 0.679 ± 0.204 &
n.a. & n.a. & 0.686 ± 0.207 &
n.a. & n.a. & 0.752 ± 0.007 \\

\midrule
mean & 
n.a. & n.a. & 0.707 ± 0.029 &
n.a. & n.a. & 0.714 ± 0.035 &
n.a. & n.a. & 0.714 ± 0.015 \\

\bottomrule
\end{tabular}%

\begin{tablenotes}[flushleft]\footnotesize
\item n.a.: not applicable because the metric cannot be measured; $\downarrow$: lower is better; $\uparrow$: higher is better
\end{tablenotes}
\end{threeparttable}
}
\end{table*}

\begin{table*}[h]
\centering
\caption{\textbf{Detailed zero-shot cross-instance detection results on the simulation dataset (unseen; ProPicker).}}
\label{tab:cross_sample_detail_sim_diff_propicker}
\resizebox{\textwidth}{!}{%
\begin{threeparttable}
\begin{tabular}{@{}l|rrr|rrr|rrr@{}}
\toprule
& \multicolumn{3}{c}{\textbf{1 Prompt}} & \multicolumn{3}{c}{\textbf{2 Prompts}} &
\multicolumn{3}{c}{\textbf{4 Prompts}}\\ 
\cmidrule(lr){2-4}\cmidrule(lr){5-7}\cmidrule(lr){8-10}
\textbf{Class} &
\textbf{AP@0.5r $\uparrow$} & \textbf{AP@1r $\uparrow$} &
\textbf{F1 $\uparrow$} &
\textbf{AP@0.5r $\uparrow$} & \textbf{AP@1r $\uparrow$} &
\textbf{F1 $\uparrow$} &
\textbf{AP@0.5r $\uparrow$} & \textbf{AP@1r $\uparrow$} &
\textbf{F1 $\uparrow$}
\\

\midrule
1E9R &
n.a. & n.a. & 0.901 ± 0.003 &
n.a. & n.a. & 0.901 ± 0.007 &
n.a. & n.a. & 0.900 ± 0.007 \\

1F1B &
n.a. & n.a. & 0.793 ± 0.075 &
n.a. & n.a. & 0.818 ± 0.005 &
n.a. & n.a. & 0.821 ± 0.002 \\

1FPY & 
n.a. & n.a. & 0.717 ± 0.158 &
n.a. & n.a. & 0.849 ± 0.111 &
n.a. & n.a. & 0.831 ± 0.121 \\

1JZ8 &
n.a. & n.a. & 0.762 ± 0.086 &
n.a. & n.a. & 0.811 ± 0.005 &
n.a. & n.a. & 0.814 ± 0.008 \\

1OAO &
n.a. & n.a. & 0.729 ± 0.187 &
n.a. & n.a. & 0.833 ± 0.133 &
n.a. & n.a. & 0.898 ± 0.032 \\

2DF7 & 
n.a. & n.a. & 0.819 ± 0.225 &
n.a. & n.a. & 0.889 ± 0.045 &
n.a. & n.a. & 0.893 ± 0.030 \\

2H12 &
n.a. & n.a. & 0.792 ± 0.015 &
n.a. & n.a. & 0.793 ± 0.013 &
n.a. & n.a. & 0.796 ± 0.016 \\

3HHB & 
n.a. & n.a. & 0.644 ± 0.055 &
n.a. & n.a. & 0.607 ± 0.099 &
n.a. & n.a. & 0.561 ± 0.128 \\

4D4R &
n.a. & n.a. & 0.117 ± 0.063 &
n.a. & n.a. & 0.172 ± 0.104 &
n.a. & n.a. & 0.123 ± 0.036 \\

6T3E &
n.a. & n.a. & 0.466 ± 0.164 &
n.a. & n.a. & 0.411 ± 0.221 &
n.a. & n.a. & 0.302 ± 0.180 \\

\midrule
mean & 
n.a. & n.a. & 0.674 ± 0.063 &
n.a. & n.a. & 0.708 ± 0.022 &
n.a. & n.a. & 0.694 ± 0.027 \\

\bottomrule
\end{tabular}%

\begin{tablenotes}[flushleft]\footnotesize
\item n.a.: not applicable because the metric cannot be measured; $\downarrow$: lower is better; $\uparrow$: higher is better
\end{tablenotes}
\end{threeparttable}
}
\end{table*}

\begin{table*}[h]
\centering
\caption{\textbf{Detailed zero-shot cross-instance detection results on the DS-10440 dataset (FullTilt).}}
\label{tab:cross_sample_detail_czii_fulltilt}
\resizebox{\textwidth}{!}{%
\begin{threeparttable}
\begin{tabular}{@{}l|rrr|rrr|rrr@{}}
\toprule
& \multicolumn{3}{c}{\textbf{1 Prompt}} & \multicolumn{3}{c}{\textbf{2 Prompts}} &
\multicolumn{3}{c}{\textbf{4 Prompts}}\\ 
\cmidrule(lr){2-4}\cmidrule(lr){5-7}\cmidrule(lr){8-10}
\textbf{Class} &
\textbf{AP@0.5r $\uparrow$} & \textbf{AP@1r $\uparrow$} &
\textbf{F1 $\uparrow$} &
\textbf{AP@0.5r $\uparrow$} & \textbf{AP@1r $\uparrow$} &
\textbf{F1 $\uparrow$} &
\textbf{AP@0.5r $\uparrow$} & \textbf{AP@1r $\uparrow$} &
\textbf{F1 $\uparrow$}
\\

\midrule
apoferritin &
0.002 ± 0.001 & 0.009 ± 0.002 & 0.048 ± 0.015 & 
0.002 ± 0.001 & 0.009 ± 0.002 & 0.049 ± 0.014 &
0.003 ± 0.001 & 0.010 ± 0.002 & 0.053 ± 0.010 \\

$\beta$-amylase &
0.009 ± 0.003 & 0.014 ± 0.004 & 0.053 ± 0.020 & 
0.008 ± 0.003 & 0.014 ± 0.003 & 0.054 ± 0.024 & 
0.011 ± 0.003 & 0.016 ± 0.004 & 0.066 ± 0.018 \\

$\beta$-galactosidase &
0.033 ± 0.012 & 0.050 ± 0.016 & 0.104 ± 0.024 & 
0.041 ± 0.011 & 0.056 ± 0.013 & 0.116 ± 0.037 &
0.041 ± 0.014 & 0.055 ± 0.017 & 0.106 ± 0.047 \\

ribosome &
0.534 ± 0.191 & 0.601 ± 0.211 & 0.544 ± 0.181 & 
0.585 ± 0.104 & 0.659 ± 0.112 & 0.584 ± 0.150 &
0.633 ± 0.049 & 0.715 ± 0.048 & 0.607 ± 0.064 \\

thyroglobulin &
0.064 ± 0.018 & 0.086 ± 0.024 & 0.159 ± 0.039 & 
0.065 ± 0.019 & 0.086 ± 0.024 & 0.160 ± 0.036 &
0.080 ± 0.010 & 0.106 ± 0.012 & 0.183 ± 0.037 \\

virus-like particles &
0.026 ± 0.023 & 0.059 ± 0.037 & 0.096 ± 0.065 & 
0.041 ± 0.027 & 0.096 ± 0.045 & 0.155 ± 0.066 &
0.049 ± 0.014 & 0.111 ± 0.018 & 0.183 ± 0.026 \\

\midrule
mean &
0.112 ± 0.033 & 0.137 ± 0.036 & 0.167 ± 0.037 & 
0.124 ± 0.020 & 0.154 ± 0.022 & 0.186 ± 0.025 &
0.136 ± 0.009 & 0.169 ± 0.009 & 0.200 ± 0.013 \\
\bottomrule
\end{tabular}%

\begin{tablenotes}[flushleft]\footnotesize
\item $\downarrow$: lower is better; $\uparrow$: higher is better
\end{tablenotes}
\end{threeparttable}
}
\end{table*}

\begin{table*}[h]
\centering
\caption{\textbf{Detailed cross-instance detection results on the simulation dataset (seen; FullTilt).}}
\label{tab:cross_sample_detail_sim_same_fulltilt}
\resizebox{\textwidth}{!}{%
\begin{threeparttable}
\begin{tabular}{@{}l|rrr|rrr|rrr@{}}
\toprule
& \multicolumn{3}{c}{\textbf{1 Prompt}} & \multicolumn{3}{c}{\textbf{2 Prompts}} &
\multicolumn{3}{c}{\textbf{4 Prompts}}\\ 
\cmidrule(lr){2-4}\cmidrule(lr){5-7}\cmidrule(lr){8-10}
\textbf{Class} &
\textbf{AP@0.5r $\uparrow$} & \textbf{AP@1r $\uparrow$} &
\textbf{F1 $\uparrow$} &
\textbf{AP@0.5r $\uparrow$} & \textbf{AP@1r $\uparrow$} &
\textbf{F1 $\uparrow$} &
\textbf{AP@0.5r $\uparrow$} & \textbf{AP@1r $\uparrow$} &
\textbf{F1 $\uparrow$}
\\

\midrule
1BXN &
0.833 ± 0.019 & 0.865 ± 0.015 & 0.763 ± 0.010 &
0.842 ± 0.012 & 0.868 ± 0.013 & 0.795 ± 0.026 &
0.838 ± 0.012 & 0.864 ± 0.010 & 0.784 ± 0.018 \\

1S3X &
0.000 ± 0.000 & 0.000 ± 0.000 & 0.002 ± 0.002 & 
0.000 ± 0.000 & 0.000 ± 0.000 & 0.002 ± 0.001 &
0.000 ± 0.000 & 0.000 ± 0.000 &0.002 ± 0.000 \\

3CF3 &
0.528 ± 0.037 & 0.591 ± 0.043 & 0.559 ± 0.014 &
0.537 ± 0.031 & 0.604 ± 0.037 & 0.568 ± 0.020 & 
0.511 ± 0.045 & 0.570 ± 0.051 & 0.554 ± 0.018 \\

3GL1 &
0.000 ± 0.000 & 0.000 ± 0.000 & 0.004 ± 0.002 &
0.000 ± 0.000 & 0.000 ± 0.000 & 0.004 ± 0.002 &
0.000 ± 0.000 & 0.000 ± 0.000 & 0.002 ± 0.001 \\

5JH9 &
0.843 ± 0.014 & 0.872 ± 0.015 & 0.778 ± 0.009 &
0.825 ± 0.034 & 0.848 ± 0.038 & 0.786 ± 0.008 &
0.799 ± 0.020 & 0.828 ± 0.020 & 0.793 ± 0.005 \\

5OOL &
0.783 ± 0.024 & 0.816 ± 0.024 & 0.828 ± 0.013 &
0.775 ± 0.009 & 0.806 ± 0.007 & 0.837 ± 0.007 &
0.781 ± 0.007 & 0.808 ± 0.007 & 0.839 ± 0.005 \\

6IGC &
0.984 ± 0.003 & 0.991 ± 0.003 & 0.985 ± 0.003 &
0.985 ± 0.005 & 0.992 ± 0.003 & 0.987 ± 0.003 & 
0.984 ± 0.003 & 0.991 ± 0.003 & 0.986 ± 0.004 \\

6LXV &
0.943 ± 0.005 & 0.943 ± 0.005 & 0.844 ± 0.018 &
0.951 ± 0.010 & 0.952 ± 0.009 & 0.849 ± 0.012 &
0.950 ± 0.017 & 0.951 ± 0.017 & 0.854 ± 0.006 \\

7DD9 &
0.818 ± 0.043 & 0.844 ± 0.039 & 0.809 ± 0.008 &
0.830 ± 0.033 & 0.850 ± 0.027 & 0.804 ± 0.008 &
0.768 ± 0.042 & 0.787 ± 0.036 & 0.799 ± 0.008 \\

7ETM &
0.602 ± 0.017 & 0.795 ± 0.011 & 0.747 ± 0.044 &
0.610 ± 0.027 & 0.793 ± 0.012 & 0.732 ± 0.052 &
0.623 ± 0.016 & 0.802 ± 0.015 & 0.739 ± 0.054 \\

\midrule
mean & 
0.633 ± 0.006 & 0.672 ± 0.007 & 0.632 ± 0.007 &
0.636 ± 0.008 & 0.671 ± 0.007 & 0.636 ± 0.008 &
0.625 ± 0.006 & 0.660 ± 0.006 & 0.635 ± 0.006 \\

\bottomrule
\end{tabular}%

\begin{tablenotes}[flushleft]\footnotesize
\item $\downarrow$: lower is better; $\uparrow$: higher is better
\end{tablenotes}
\end{threeparttable}
}
\end{table*}

\begin{table*}[h]
\centering
\caption{\textbf{Detailed zero-shot cross-instance detection results on the simulation dataset (unseen; FullTilt).}}
\label{tab:cross_sample_detail_sim_diff_fulltilt}
\resizebox{\textwidth}{!}{%
\begin{threeparttable}
\begin{tabular}{@{}l|rrr|rrr|rrr@{}}
\toprule
& \multicolumn{3}{c}{\textbf{1 Prompt}} & \multicolumn{3}{c}{\textbf{2 Prompts}} &
\multicolumn{3}{c}{\textbf{4 Prompts}}\\ 
\cmidrule(lr){2-4}\cmidrule(lr){5-7}\cmidrule(lr){8-10}
\textbf{Class} &
\textbf{AP@0.5r $\uparrow$} & \textbf{AP@1r $\uparrow$} &
\textbf{F1 $\uparrow$} &
\textbf{AP@0.5r $\uparrow$} & \textbf{AP@1r $\uparrow$} &
\textbf{F1 $\uparrow$} &
\textbf{AP@0.5r $\uparrow$} & \textbf{AP@1r $\uparrow$} &
\textbf{F1 $\uparrow$}
\\

\midrule
1E9R &
0.652 ± 0.026 & 0.702 ± 0.021 & 0.652 ± 0.028 &
0.640 ± 0.016 & 0.701 ± 0.014 & 0.652 ± 0.018 &
0.644 ± 0.020 & 0.694 ± 0.020 & 0.645 ± 0.023 \\

1F1B &
0.250 ± 0.018 & 0.458 ± 0.023 & 0.589 ± 0.042 &
0.259 ± 0.009 & 0.466 ± 0.009 & 0.609 ± 0.043 &
0.267 ± 0.007 & 0.470 ± 0.009 & 0.616 ± 0.016 \\

1FPY & 
0.876 ± 0.022 & 0.880 ± 0.022 & 0.791 ± 0.014 &
0.895 ± 0.010 & 0.899 ± 0.012 & 0.813 ± 0.008 &
0.891 ± 0.017 & 0.898 ± 0.018 & 0.815 ± 0.016 \\

1JZ8 &
0.876 ± 0.024 & 0.881 ± 0.024 & 0.776 ± 0.048 &
0.869 ± 0.029 & 0.874 ± 0.030 & 0.768 ± 0.034 & 
0.896 ± 0.014 & 0.903 ± 0.014 & 0.820 ± 0.016 \\

1OAO &
0.625 ± 0.051 & 0.671 ± 0.052 & 0.678 ± 0.010 &
0.620 ± 0.031 & 0.667 ± 0.034 & 0.679 ± 0.009 &
0.588 ± 0.039 & 0.633 ± 0.045 & 0.675 ± 0.008 \\

2DF7 & 
0.892 ± 0.204 & 0.915 ± 0.187 & 0.823 ± 0.271 &
0.992 ± 0.018 & 0.997 ± 0.004 & 0.979 ± 0.029 &
0.976 ± 0.035 & 0.992 ± 0.014 & 0.959 ± 0.057 \\

2H12 &
0.831 ± 0.009 & 0.864 ± 0.008 & 0.787 ± 0.031 &
0.825 ± 0.019 & 0.857 ± 0.014 & 0.764 ± 0.037 & 
0.819 ± 0.011 & 0.856 ± 0.012 & 0.778 ± 0.030 \\

3HHB & 
0.000 ± 0.000 & 0.004 ± 0.002 & 0.017 ± 0.005 & 
0.000 ± 0.000 & 0.001 ± 0.001 & 0.003 ± 0.002 & 
0.000 ± 0.000 & 0.000 ± 0.000 & 0.004 ± 0.002 \\

4D4R &
0.000 ± 0.000 & 0.000 ± 0.000 & 0.001 ± 0.001 &
0.000 ± 0.000 & 0.000 ± 0.000 & 0.001 ± 0.001 &
0.000 ± 0.000 & 0.000 ± 0.000 & 0.001 ± 0.001 \\

6T3E &
0.001 ± 0.002 & 0.005 ± 0.003 & 0.020 ± 0.010 &
0.000 ± 0.000 & 0.002 ± 0.002 & 0.016 ± 0.014 &
0.000 ± 0.000 & 0.001 ± 0.001 & 0.011 ± 0.005 \\

\midrule
mean & 
0.500 ± 0.022 & 0.538 ± 0.021 & 0.514 ± 0.028 &
0.510 ± 0.005 & 0.546 ± 0.005 & 0.528 ± 0.008 &
0.508 ± 0.005 & 0.545 ± 0.005 &0.532 ± 0.008 \\

\bottomrule
\end{tabular}%

\begin{tablenotes}[flushleft]\footnotesize
\item $\downarrow$: lower is better; $\uparrow$: higher is better
\end{tablenotes}
\end{threeparttable}
}
\end{table*}

\clearpage
\subsection{Additional Multiclass Prompting Detection Results}
\label{sec:additional_multiclass_prompting}
We evaluate our multiclass visual prompt encoder ($\mathcal{E}_{\text{Prompt}}$) under cross-instance scenarios using the 6-class DS-10440 dataset. To ensure a comprehensive evaluation, we systematically test all possible class combinations (e.g., $\binom{6}{3}=15$ unique combinations for a 3-class query) averaged over 10 independent trials (Table~\ref{tab:multiclass_cross}). While simultaneously querying a larger number of classes incurs a minor trade-off in detection performance, FullTilt efficiently processes all requested classes in a single forward pass, introducing negligible runtime and VRAM overhead.

\begin{table*}[h]
\centering
\caption{\textbf{Multi-class prompting performance comparison (cross-instance).} FullTilt maintains stable computational efficiency regardless of the number of target classes.}
\label{tab:multiclass_cross}
\resizebox{\textwidth}{!}{%
\begin{threeparttable}
\begin{tabular}{@{}l|rrrr|rrrr|rrrr@{}}
\toprule
 & \multicolumn{4}{c}{\textbf{1 Prompt}} & \multicolumn{4}{c}{\textbf{2 Prompts}} & \multicolumn{4}{c}{\textbf{4 Prompts}}\\ 
\cmidrule(lr){2-5}\cmidrule(lr){6-9}\cmidrule(lr){10-13}
\textbf{\# Classes} & 
\textbf{mAP@0.5r $\uparrow$} & \textbf{mAP@1r $\uparrow$} &
\textbf{T. $\downarrow$} & \textbf{M. $\downarrow$} &
\textbf{mAP@0.5r $\uparrow$} & \textbf{mAP@1r $\uparrow$} &
\textbf{T. $\downarrow$} & \textbf{M. $\downarrow$} &
\textbf{mAP@0.5r $\uparrow$} & \textbf{mAP@1r $\uparrow$} &
\textbf{T. $\downarrow$} & \textbf{M. $\downarrow$}
\\

\midrule
1 &
0.112 ± 0.033 & 0.137 ± 0.036 & 0.494 & 3460 &
0.124 ± 0.020 & 0.154 ± 0.022 & 0.498 & 3460 &
0.136 ± 0.009 & 0.169 ± 0.009 & 0.498 & 3460 \\

2 &
0.111 ± 0.011 & 0.139 ± 0.013 & 0.508 & 3460 &
0.123 ± 0.006 & 0.154 ± 0.008 & 0.499 & 3460 &
0.128 ± 0.004 & 0.158 ± 0.005 & 0.498 & 3460 \\

3 &
0.108 ± 0.005 & 0.138 ± 0.006 & 0.500 & 3470 &
0.123 ± 0.003 & 0.154 ± 0.004 & 0.502 & 3470 &
0.124 ± 0.003 & 0.154 ± 0.004 & 0.502 & 3470 \\

4 &
0.115 ± 0.005 & 0.146 ± 0.006 & 0.505 & 3470 &
0.122 ± 0.003 & 0.154 ± 0.003 & 0.507 & 3470 &
0.124 ± 0.003 & 0.155 ± 0.003 & 0.506 & 3480 \\

5 &
0.109 ± 0.011 & 0.138 ± 0.012 & 0.494 & 3470 &
0.119 ± 0.005 & 0.151 ± 0.005 & 0.497 & 3470 &
0.122 ± 0.004 & 0.153 ± 0.005 & 0.518 & 3470 \\

6 & 
0.109 ± 0.022 & 0.137 ± 0.026 & 0.481 & 3470 &
0.121 ± 0.003 & 0.152 ± 0.004 & 0.487 & 3470 &
0.117 ± 0.009 & 0.148 ± 0.011 & 0.489 & 3470 \\

\bottomrule
\end{tabular}%

\begin{tablenotes}[flushleft]\footnotesize
\item T.: runtime (s); M.: VRAM (MB); $\downarrow$: lower is better; $\uparrow$: higher is better
\end{tablenotes}
\end{threeparttable}
}
\end{table*}

\subsection{Extended Hardware and Compilation Benchmarks}
To further evaluate the computational flexibility and efficiency of FullTilt, we benchmarked its inference runtime and memory consumption under two additional execution paradigms: (1) utilizing an optimized static computational graph via PyTorch's \texttt{torch.compile} on a GPU, and (2) in a strictly CPU-only environment. All evaluations were conducted under the intra-instance, single-prompt configuration, with results detailed in Table~\ref{tab:compile_cpu}.

While the initial static graph compilation incurs a one-time overhead of approximately 6 to 7 minutes, it yields a $1.3\times$ to $1.5\times$ acceleration in subsequent inference runtimes across all datasets. Conversely, executing FullTilt purely on a CPU increases inference time by a factor of 35 to 40 relative to the uncompiled GPU baseline. Notably, despite this expected CPU bottleneck, FullTilt still successfully completes inference within a highly practical window of 10 to 30 seconds per tilt-series. This demonstrates the framework's high accessibility for resource-constrained laboratory settings lacking dedicated high-performance accelerators.

\begin{table*}[h]
\centering
\caption{\textbf{Extended computational efficiency benchmarks (compiled GPU and CPU-only execution).} Inference runtime (T.) and peak VRAM consumption (M.) are evaluated under the intra-instance, single-prompt configuration. For the compiled GPU setting, the one-time static graph compilation overhead is also reported in seconds.}
\label{tab:compile_cpu}
\resizebox{\textwidth}{!}{%
\begin{threeparttable}
\begin{tabular}{@{}l|rrr|rrr|rrr|rrr|rrr@{}}
\toprule
 & \multicolumn{3}{c}{\textbf{DS-10440}} & \multicolumn{3}{c}{\textbf{EMPIAR-10304}} & \multicolumn{3}{c}{\textbf{EMPIAR-10499}} & \multicolumn{3}{c}{\textbf{Simulation (Seen)}} & \multicolumn{3}{c}{\textbf{Simulation (Out-Distribution)}}\\ 
\cmidrule(lr){2-4}\cmidrule(lr){5-7}\cmidrule(lr){8-10}\cmidrule(lr){11-13}\cmidrule(lr){14-16}
\textbf{Hardware} & 
\textbf{Compile $\downarrow$} &
\textbf{T. $\downarrow$} &
\textbf{M. $\downarrow$} &
\textbf{Compile $\downarrow$} &
\textbf{T. $\downarrow$} &
\textbf{M. $\downarrow$} &
\textbf{Compile $\downarrow$} &
\textbf{T. $\downarrow$} &
\textbf{M. $\downarrow$} &
\textbf{Compile $\downarrow$} &
\textbf{T. $\downarrow$} &
\textbf{M. $\downarrow$} &
\textbf{Compile $\downarrow$} &
\textbf{T. $\downarrow$} &
\textbf{M. $\downarrow$} 
\\
\midrule
GPU &
n.a. & 0.499 & 3430 &
n.a. & 0.229 & 2410 &
n.a. & 0.389 & 3410 &
n.a. & 0.682 & 5160 &
n.a. & 0.694 & 5160 \\

GPU (Compiled) &
362 & 0.385 & 3830 &
372 & 0.149 & 2110 &
361 & 0.274 & 3080 &
413 & 0.509 & 5090 &
413 & 0.508 & 5090 \\

CPU-Only &
n.a. & 20.1 & n.a. &
n.a. & 9.33 & n.a. &
n.a. & 13.4 & n.a. &
n.a. & 24.6 & n.a. &
n.a. & 24.6 & n.a. \\
\bottomrule
\end{tabular}%

\begin{tablenotes}[flushleft]\footnotesize
\item Compile: Static graph compilation time (s); T.: Inference runtime (s); M.: VRAM (MB); $\downarrow$: lower is better; n.a.: not applicable
\end{tablenotes}
\end{threeparttable}
}
\end{table*}

\subsection{Statistical Significance}
We report statistical comparisons in Tables~\ref{tab:stats_intra} and~\ref{tab:stats_cross} for intra-instance and cross-instance detection, respectively. We use the Wilcoxon signed-rank test for paired comparisons between methods evaluated over 10 trials and the Mann-Whitney U test for comparisons with WSL baselines evaluated over 3 trials versus FullTilt's 10 trials. These sample sizes limit statistical power and, for exact tests, $p$-value resolution. A non-significant result indicates insufficient evidence of a difference at the current sample size, rather than equivalence.

For intra-instance detection, FullTilt significantly outperforms the competing methods across all real-world datasets, except for comparisons with TomoTwin on EMPIAR-10304: mAP@0.5r shows no significant difference with 1 prompt and is significantly lower with 2 and 4 prompts, while F1 shows no significant difference with 2 prompts. For cross-instance detection, FullTilt significantly outperforms all evaluated methods on EMPIAR-10499. On DS-10440, it significantly outperforms TomoTwin in mAP@0.5r and F1, with no significant difference in mAP@1r. On EMPIAR-10304, no significant differences from TomoTwin are observed across the evaluated metrics.

We use 10 trials because of the computational cost of the baselines. Running 20 additional trials across all datasets would require approximately 30 minutes for FullTilt but more than two months for the baselines, highlighting FullTilt's efficiency. The cross-instance comparisons with TomoTwin on DS-10440 and EMPIAR-10304 should therefore be interpreted in light of the limited statistical power at this sample size.

\begin{table*}[h]
\centering
\caption{\textbf{Statistical comparisons for intra-instance detection.}}
\label{tab:stats_intra}
\resizebox{\textwidth}{!}{%
\begin{threeparttable}
\begin{tabular}{@{}l|l|rrr|rrr|rrr@{}}
\toprule
& & \multicolumn{3}{c}{\textbf{DS-10440}}
  & \multicolumn{3}{c}{\textbf{EMPIAR-10304}}
  & \multicolumn{3}{c}{\textbf{EMPIAR-10499}} \\
\cmidrule(lr){3-5}\cmidrule(lr){6-8}\cmidrule(lr){9-11}
\textbf{\# P.} & \textbf{vs.\ Method}
& \textbf{mAP@0.5r} & \textbf{mAP@1r} & \textbf{F1}
& \textbf{mAP@0.5r} & \textbf{mAP@1r} & \textbf{F1}
& \textbf{mAP@0.5r} & \textbf{mAP@1r} & \textbf{F1} \\
\midrule
\multirow{6}{*}{1} & DETR3D
& $\blacktriangle^{**}$ & $\blacktriangle^{**}$ & $\blacktriangle^{**}$
& $\blacktriangle^{**}$ & $\blacktriangle^{**}$ & $\blacktriangle^{**}$
& $\blacktriangle^{**}$ & $\blacktriangle^{**}$ & $\blacktriangle^{**}$ \\
 & PETR
& $\blacktriangle^{**}$ & $\blacktriangle^{**}$ & $\blacktriangle^{**}$
& $\blacktriangle^{**}$ & $\blacktriangle^{**}$ & $\blacktriangle^{**}$
& $\blacktriangle^{**}$ & $\blacktriangle^{**}$ & $\blacktriangle^{**}$ \\
 & Zeng et al.
& -- & -- & $\blacktriangle^{**}$
& -- & -- & $\blacktriangle^{**}$
& -- & -- & $\blacktriangle^{**}$ \\
 & TomoTwin
& $\blacktriangle^{**}$ & $\blacktriangle^{*}$ & $\blacktriangle^{**}$
& ns & $\blacktriangle^{**}$ & $\blacktriangle^{*}$
& $\blacktriangle^{**}$ & $\blacktriangle^{**}$ & $\blacktriangle^{**}$ \\
 & CryoSAM
& -- & -- & $\blacktriangle^{**}$
& -- & -- & $\blacktriangle^{**}$
& -- & -- & $\blacktriangle^{**}$ \\
 & ProPicker
& -- & -- & $\blacktriangle^{**}$
& -- & -- & $\blacktriangle^{**}$
& -- & -- & $\blacktriangle^{**}$ \\
\midrule
\multirow{6}{*}{2} & DETR3D
& $\blacktriangle^{**}$ & $\blacktriangle^{**}$ & $\blacktriangle^{**}$
& $\blacktriangle^{**}$ & $\blacktriangle^{**}$ & $\blacktriangle^{**}$
& $\blacktriangle^{**}$ & $\blacktriangle^{**}$ & $\blacktriangle^{**}$ \\
 & PETR
& $\blacktriangle^{**}$ & $\blacktriangle^{**}$ & $\blacktriangle^{**}$
& $\blacktriangle^{**}$ & $\blacktriangle^{**}$ & $\blacktriangle^{**}$
& $\blacktriangle^{**}$ & $\blacktriangle^{**}$ & $\blacktriangle^{**}$ \\
 & Zeng et al.
& -- & -- & $\blacktriangle^{**}$
& -- & -- & $\blacktriangle^{**}$
& -- & -- & $\blacktriangle^{**}$ \\
 & TomoTwin
& $\blacktriangle^{**}$ & $\blacktriangle^{**}$ & $\blacktriangle^{**}$
& $\triangledown^{*}$ & $\blacktriangle^{**}$ & ns
& $\blacktriangle^{**}$ & $\blacktriangle^{**}$ & $\blacktriangle^{**}$ \\
 & CryoSAM
& -- & -- & $\blacktriangle^{**}$
& -- & -- & $\blacktriangle^{**}$
& -- & -- & $\blacktriangle^{**}$ \\
 & ProPicker
& -- & -- & $\blacktriangle^{**}$
& -- & -- & $\blacktriangle^{**}$
& -- & -- & $\blacktriangle^{**}$ \\
\midrule
\multirow{6}{*}{4} & DETR3D
& $\blacktriangle^{**}$ & $\blacktriangle^{**}$ & $\blacktriangle^{**}$
& $\blacktriangle^{**}$ & $\blacktriangle^{**}$ & $\blacktriangle^{**}$
& $\blacktriangle^{**}$ & $\blacktriangle^{**}$ & $\blacktriangle^{**}$ \\
 & PETR
& $\blacktriangle^{**}$ & $\blacktriangle^{**}$ & $\blacktriangle^{**}$
& $\blacktriangle^{**}$ & $\blacktriangle^{**}$ & $\blacktriangle^{**}$
& $\blacktriangle^{**}$ & $\blacktriangle^{**}$ & $\blacktriangle^{**}$ \\
 & Zeng et al.
& -- & -- & $\blacktriangle^{**}$
& -- & -- & $\blacktriangle^{**}$
& -- & -- & $\blacktriangle^{**}$ \\
 & TomoTwin
& $\blacktriangle^{**}$ & $\blacktriangle^{**}$ & $\blacktriangle^{**}$
& $\triangledown^{**}$ & $\blacktriangle^{**}$ & $\blacktriangle^{*}$
& $\blacktriangle^{**}$ & $\blacktriangle^{**}$ & $\blacktriangle^{**}$ \\
 & CryoSAM
& -- & -- & $\blacktriangle^{**}$
& -- & -- & $\blacktriangle^{**}$
& -- & -- & $\blacktriangle^{**}$ \\
 & ProPicker
& -- & -- & $\blacktriangle^{**}$
& -- & -- & $\blacktriangle^{**}$
& -- & -- & $\blacktriangle^{**}$ \\
\bottomrule
\end{tabular}

\begin{tablenotes}[flushleft]\footnotesize
\item Each entry compares FullTilt with the indicated method. \# P.: number of prompts; $\blacktriangle$/$\triangledown$: FullTilt performs significantly better/worse; ns: no significant difference; --: not applicable; ${}^{*}$: $p<0.05$; ${}^{**}$: $p<0.01$.
\end{tablenotes}
\end{threeparttable}%
}
\end{table*}

\begin{table*}[h]
\centering
\caption{\textbf{Statistical comparisons for cross-instance detection.}}
\label{tab:stats_cross}
\resizebox{\textwidth}{!}{%
\begin{threeparttable}
\begin{tabular}{@{}l|l|rrr|rrr|rrr@{}}
\toprule
& & \multicolumn{3}{c}{\textbf{DS-10440}}
  & \multicolumn{3}{c}{\textbf{EMPIAR-10304}}
  & \multicolumn{3}{c}{\textbf{EMPIAR-10499}} \\
\cmidrule(lr){3-5}\cmidrule(lr){6-8}\cmidrule(lr){9-11}
\textbf{\# P.} & \textbf{vs.\ Method}
& \textbf{mAP@0.5r} & \textbf{mAP@1r} & \textbf{F1}
& \textbf{mAP@0.5r} & \textbf{mAP@1r} & \textbf{F1}
& \textbf{mAP@0.5r} & \textbf{mAP@1r} & \textbf{F1} \\
\midrule
\multirow{3}{*}{1} & TomoPicker
& $\blacktriangle^{**}$ & $\blacktriangle^{**}$ & $\blacktriangle^{**}$
& $\blacktriangle^{**}$ & $\blacktriangle^{**}$ & $\blacktriangle^{**}$
& $\blacktriangle^{**}$ & $\blacktriangle^{**}$ & $\blacktriangle^{**}$ \\
 & TomoTwin
& $\blacktriangle^{*}$ & ns & $\blacktriangle^{**}$
& ns & ns & ns
& $\blacktriangle^{**}$ & $\blacktriangle^{*}$ & $\blacktriangle^{**}$ \\
 & ProPicker
& -- & -- & $\blacktriangle^{**}$
& -- & -- & $\blacktriangle^{**}$
& -- & -- & $\blacktriangle^{**}$ \\
\midrule
\multirow{4}{*}{2} & Huang et al.
& $\blacktriangle^{**}$ & $\blacktriangle^{**}$ & $\blacktriangle^{**}$
& $\blacktriangle^{**}$ & $\blacktriangle^{**}$ & $\blacktriangle^{**}$
& $\blacktriangle^{**}$ & $\blacktriangle^{**}$ & $\blacktriangle^{**}$ \\
 & TomoPicker
& $\blacktriangle^{**}$ & $\blacktriangle^{**}$ & $\blacktriangle^{**}$
& $\blacktriangle^{**}$ & $\blacktriangle^{**}$ & $\blacktriangle^{**}$
& $\blacktriangle^{**}$ & $\blacktriangle^{**}$ & $\blacktriangle^{**}$ \\
 & TomoTwin
& $\blacktriangle^{*}$ & ns & $\blacktriangle^{**}$
& ns & ns & ns
& $\blacktriangle^{**}$ & $\blacktriangle^{**}$ & $\blacktriangle^{*}$ \\
 & ProPicker
& -- & -- & $\blacktriangle^{**}$
& -- & -- & $\blacktriangle^{**}$
& -- & -- & $\blacktriangle^{**}$ \\
\midrule
\multirow{4}{*}{4} & Huang et al.
& $\blacktriangle^{**}$ & $\blacktriangle^{**}$ & $\blacktriangle^{**}$
& $\blacktriangle^{**}$ & $\blacktriangle^{**}$ & $\blacktriangle^{**}$
& $\blacktriangle^{**}$ & $\blacktriangle^{**}$ & $\blacktriangle^{**}$ \\
 & TomoPicker
& $\blacktriangle^{**}$ & $\blacktriangle^{**}$ & $\blacktriangle^{**}$
& $\blacktriangle^{**}$ & $\blacktriangle^{**}$ & $\blacktriangle^{**}$
& $\blacktriangle^{**}$ & $\blacktriangle^{**}$ & $\blacktriangle^{**}$ \\
 & TomoTwin
& $\blacktriangle^{*}$ & ns & $\blacktriangle^{**}$
& ns & ns & ns
& $\blacktriangle^{**}$ & $\blacktriangle^{**}$ & $\blacktriangle^{*}$ \\
 & ProPicker
& -- & -- & $\blacktriangle^{**}$
& -- & -- & $\blacktriangle^{**}$
& -- & -- & $\blacktriangle^{**}$ \\
\bottomrule
\end{tabular}

\begin{tablenotes}[flushleft]\footnotesize
\item Each entry compares FullTilt with the indicated method. \# P.: number of prompts; $\blacktriangle$/$\triangledown$: FullTilt performs significantly better/worse; ns: no significant difference; --: not applicable; ${}^{*}$: $p<0.05$; ${}^{**}$: $p<0.01$.
\end{tablenotes}
\end{threeparttable}%
}
\end{table*}

\subsection{Shape Generalization and Failure Analysis}
\label{sec:g_analysis}
We systematically analyze 20 protein types from simulated data, with axis ratios ranging from approximately 1.1 to 2.9 and molecular weights from approximately 43 to 2,366~kDa, to assess (i) performance on non-globular particles with and without $\mathcal{G}$, (ii) whether $\mathcal{G}$ introduces a globular bias, and (iii) failure cases. The axis ratio is defined as $\sqrt{\lambda_{\max}/\lambda_{\min}}$, where $\lambda_{\max}$ and $\lambda_{\min}$ are the largest and smallest PCA eigenvalues of the particle density. Higher axis ratios indicate less globular shapes. For each protein type, we compare performance with and without $\mathcal{G}$ using a Wilcoxon signed-rank test (Tables~\ref{tab:shape_seen} and~\ref{tab:shape_unseen}; rows sorted by axis ratio).

\begin{table*}[h]
\centering
\caption{\textbf{Effect of $\mathcal{G}$ on seen proteins in simulated data.}}
\label{tab:shape_seen}
\resizebox{\textwidth}{!}{%
\begin{threeparttable}
\begin{tabular}{@{}lrr|rr|rr|rr@{}}
\toprule
& & &
\multicolumn{2}{c}{\textbf{$\mathcal{G}$ vs.\ no $\mathcal{G}$}} &
\multicolumn{2}{c}{\textbf{FullTilt}} &
\multicolumn{2}{c}{\textbf{Without $\mathcal{G}$}} \\
\cmidrule(lr){4-5}\cmidrule(lr){6-7}\cmidrule(lr){8-9}
\textbf{PDB} & \textbf{Weight (kDa)} & \textbf{Axis ratio}
& \textbf{AP@0.5r} & \textbf{AP@1r}
& \textbf{AP@0.5r} & \textbf{AP@1r}
& \textbf{AP@0.5r} & \textbf{AP@1r} \\
\midrule
1BXN & 279.81 & 1.264
& $\blacktriangle^{*}$ & ns
& $0.802\pm0.053$ & $0.833\pm0.054$
& $0.730\pm0.029$ & $0.809\pm0.030$ \\
7ETM & 943.62 & 1.289
& $\blacktriangle^{**}$ & $\blacktriangle^{*}$
& $0.602\pm0.027$ & $0.799\pm0.020$
& $0.509\pm0.060$ & $0.728\pm0.075$ \\
5OOL & 1,797.38 & 1.388
& $\blacktriangle^{**}$ & $\blacktriangle^{**}$
& $0.806\pm0.035$ & $0.839\pm0.035$
& $0.744\pm0.011$ & $0.786\pm0.012$ \\
3GL1 & 84.61 & 1.504
& ns & ns
& $0.000\pm0.000$ & $0.000\pm0.000$
& $0.000\pm0.000$ & $0.000\pm0.000$ \\
1S3X & 42.75 & 1.521
& ns & ns
& $0.000\pm0.000$ & $0.000\pm0.000$
& $0.000\pm0.000$ & $0.001\pm0.001$ \\
7DD9 & 715.19 & 1.695
& $\blacktriangle^{**}$ & ns
& $0.809\pm0.065$ & $0.827\pm0.068$
& $0.671\pm0.089$ & $0.758\pm0.100$ \\
6IGC & 2,365.77 & 1.715
& ns & $\triangledown^{**}$
& $0.985\pm0.004$ & $0.991\pm0.002$
& $0.985\pm0.002$ & $0.995\pm0.001$ \\
3CF3 & 270.87 & 1.849
& $\blacktriangle^{**}$ & $\blacktriangle^{**}$
& $0.583\pm0.015$ & $0.647\pm0.015$
& $0.327\pm0.034$ & $0.455\pm0.037$ \\
6LXV & 751.32 & 1.902
& ns & ns
& $0.949\pm0.018$ & $0.951\pm0.017$
& $0.939\pm0.008$ & $0.950\pm0.007$ \\
5JH9 & 230.78 & 2.056
& $\blacktriangle^{**}$ & $\blacktriangle^{*}$
& $0.786\pm0.056$ & $0.808\pm0.055$
& $0.674\pm0.040$ & $0.759\pm0.042$ \\
\bottomrule
\end{tabular}

\begin{tablenotes}[flushleft]\footnotesize
\item $\blacktriangle$/$\triangledown$:
adding $\mathcal{G}$ significantly improves/degrades performance;
ns: no significant difference;
${}^{*}$: $p<0.05$; ${}^{**}$: $p<0.01$.
\end{tablenotes}
\end{threeparttable}%
}
\end{table*}

\begin{table*}[h]
\centering
\caption{\textbf{Effect of $\mathcal{G}$ on unseen proteins in simulated data.}}
\label{tab:shape_unseen}
\resizebox{\textwidth}{!}{%
\begin{threeparttable}
\begin{tabular}{@{}lrr|rr|rr|rr@{}}
\toprule
& & &
\multicolumn{2}{c}{\textbf{$\mathcal{G}$ vs.\ no $\mathcal{G}$}} &
\multicolumn{2}{c}{\textbf{FullTilt}} &
\multicolumn{2}{c}{\textbf{Without $\mathcal{G}$}} \\
\cmidrule(lr){4-5}\cmidrule(lr){6-7}\cmidrule(lr){8-9}
\textbf{PDB} & \textbf{Weight (kDa)} & \textbf{Axis ratio}
& \textbf{AP@0.5r} & \textbf{AP@1r}
& \textbf{AP@0.5r} & \textbf{AP@1r}
& \textbf{AP@0.5r} & \textbf{AP@1r} \\
\midrule
2DF7 & 987.25 & 1.116
& $\triangledown^{*}$ & ns
& $0.927\pm0.042$ & $0.968\pm0.042$
& $0.968\pm0.009$ & $0.994\pm0.009$ \\
1E9R & 293.47 & 1.151
& $\blacktriangle^{**}$ & $\blacktriangle^{*}$
& $0.654\pm0.055$ & $0.711\pm0.051$
& $0.519\pm0.031$ & $0.669\pm0.027$ \\
3HHB & 64.74 & 1.269
& ns & ns
& $0.000\pm0.000$ & $0.003\pm0.001$
& $0.000\pm0.000$ & $0.001\pm0.001$ \\
1FPY & 629.55 & 1.271
& ns & ns
& $0.860\pm0.039$ & $0.864\pm0.038$
& $0.866\pm0.056$ & $0.885\pm0.055$ \\
2H12 & 298.56 & 1.467
& $\blacktriangle^{**}$ & $\blacktriangle^{*}$
& $0.832\pm0.015$ & $0.863\pm0.013$
& $0.671\pm0.057$ & $0.790\pm0.058$ \\
4D4R & 63.66 & 1.786
& ns & ns
& $0.000\pm0.000$ & $0.000\pm0.000$
& $0.000\pm0.000$ & $0.000\pm0.000$ \\
1JZ8 & 478.40 & 2.061
& ns & ns
& $0.850\pm0.064$ & $0.857\pm0.065$
& $0.813\pm0.028$ & $0.835\pm0.024$ \\
6T3E & 78.11 & 2.076
& ns & ns
& $0.000\pm0.001$ & $0.005\pm0.003$
& $0.001\pm0.001$ & $0.002\pm0.002$ \\
1F1B & 103.55 & 2.220
& $\blacktriangle^{**}$ & ns
& $0.281\pm0.044$ & $0.486\pm0.060$
& $0.191\pm0.020$ & $0.456\pm0.031$ \\
1OAO & 316.52 & 2.936
& $\blacktriangle^{**}$ & ns
& $0.612\pm0.044$ & $0.660\pm0.046$
& $0.525\pm0.074$ & $0.602\pm0.075$ \\
\bottomrule
\end{tabular}

\begin{tablenotes}[flushleft]\footnotesize
\item $\blacktriangle$/$\triangledown$:
adding $\mathcal{G}$ significantly improves/degrades performance;
ns: no significant difference;
${}^{*}$: $p<0.05$; ${}^{**}$: $p<0.01$.
\end{tablenotes}
\end{threeparttable}%
}
\end{table*}

FullTilt detects both globular particles (1E9R, axis ratio 1.15, AP@0.5r $=0.65$) and non-globular particles (1OAO, axis ratio 2.94, AP@0.5r $=0.61$). Significant gains from $\mathcal{G}$ occur across the axis-ratio range, including for the most elongated particle (1OAO, axis ratio 2.94, AP@0.5r: $0.53\rightarrow0.61$), whereas performance declines for the most globular particle (2DF7, axis ratio 1.12, AP@0.5r: $0.97\rightarrow0.93$).

Spearman correlations between axis ratio and the gain from $\mathcal{G}$, defined as $\Delta\mathrm{AP} =\mathrm{AP}_{\text{FullTilt}} -\mathrm{AP}_{\text{without }\mathcal{G}}$, show no significant association (Table~\ref{tab:shape_gain_corr}). Finally, Spearman correlations between FullTilt's AP and particle properties help characterize its failure cases (Table~\ref{tab:particle_property_corr}). AP shows no significant association with axis ratio but a strong positive association with molecular weight.

Overall, FullTilt detects non-globular particles, and these results provide no evidence that $\mathcal{G}$ introduces a globular bias among the evaluated protein types. Instead, all evaluated particle types below 100~kDa have near-zero AP, consistent with the small-particle limitation.

\begin{table}[h]
\centering
\caption{\textbf{Spearman correlations between axis ratio
and the AP gain from $\mathcal{G}$.}}
\label{tab:shape_gain_corr}
\resizebox{0.5\textwidth}{!}{%
\begin{threeparttable}
\begin{tabular}{@{}lrr@{}}
\toprule
\textbf{Property}
& \textbf{$\Delta$AP@0.5r}
& \textbf{$\Delta$AP@1r} \\
\midrule
Axis ratio
& $0.155\;(p=0.515)$
& $0.203\;(p=0.390)$ \\
\bottomrule
\end{tabular}

\begin{tablenotes}[flushleft]\footnotesize
\item Each entry reports Spearman's $\rho$ and its $p$-value across the 20 evaluated protein types.
\end{tablenotes}
\end{threeparttable}
}
\end{table}

\begin{table}[h]
\centering
\caption{\textbf{Spearman correlations between FullTilt's AP
and particle properties.}}
\label{tab:particle_property_corr}
\resizebox{0.5\textwidth}{!}{%
\begin{threeparttable}
\begin{tabular}{@{}lrr@{}}
\toprule
\textbf{Property} & \textbf{AP@0.5r} & \textbf{AP@1r} \\
\midrule
Axis ratio
& $-0.161\;(p=0.499)$
& $-0.215\;(p=0.362)$ \\
Weight
& $0.844\;(p<10^{-5})$
& $0.858\;(p<10^{-5})$ \\
\bottomrule
\end{tabular}

\begin{tablenotes}[flushleft]\footnotesize
\item Each entry reports Spearman's $\rho$ and its $p$-value across the 20 evaluated protein types.
\end{tablenotes}
\end{threeparttable}
}
\end{table}

\subsection{Effect of Broader Training-Distribution Coverage}
\label{sec:broader}
The original FullTilt does not consistently outperform TomoTwin on simulated data. To investigate whether performance can be improved without architectural changes, we broaden the training distribution using an extended $\mathcal{G}$ that incorporates additional noise patterns (white, Poisson, and speckle noise) and geometric primitives (cube and shell). The architecture and optimization hyperparameters remain unchanged. Table~\ref{tab:extended_g} reports the results, with statistical comparisons based on Wilcoxon signed-rank tests over 10 trials.

With the same architecture, FullTilt with extended $\mathcal{G}$ significantly outperforms TomoTwin in 6 of the 10 dataset--metric comparisons, shows no significant difference in 3 (with higher point estimates in all three), and performs significantly worse in only 1. TomoTwin retains a higher point estimate only for mAP@1r on unseen simulated proteins, whereas FullTilt performs significantly better under the stricter mAP@0.5r criterion ($0.545$ vs.\ $0.531$). These results suggest that training-distribution coverage contributes to the original simulation performance gap and that broadening this distribution can improve performance on simulated data while retaining competitive real-world performance, without modifying the architecture.

\begin{table*}[h]
\centering
\caption{\textbf{FullTilt with extended $\mathcal{G}$.}
The architecture and optimization hyperparameters remain unchanged.}
\label{tab:extended_g}
\resizebox{\textwidth}{!}{%
\begin{threeparttable}
\begin{tabular}{@{}l|rr|rr|rr|rr|rr@{}}
\toprule
& \multicolumn{2}{c}{\textbf{DS-10440}}
& \multicolumn{2}{c}{\textbf{EMPIAR-10304}}
& \multicolumn{2}{c}{\textbf{EMPIAR-10499}}
& \multicolumn{2}{c}{\textbf{Simulation (Seen)}}
& \multicolumn{2}{c}{\textbf{Simulation (Unseen)}} \\
\cmidrule(lr){2-3}\cmidrule(lr){4-5}\cmidrule(lr){6-7}
\cmidrule(lr){8-9}\cmidrule(lr){10-11}
\textbf{Method}
& \textbf{mAP@0.5r} & \textbf{mAP@1r}
& \textbf{mAP@0.5r} & \textbf{mAP@1r}
& \textbf{mAP@0.5r} & \textbf{mAP@1r}
& \textbf{mAP@0.5r} & \textbf{mAP@1r}
& \textbf{mAP@0.5r} & \textbf{mAP@1r} \\
\midrule
TomoTwin
& $0.089\pm0.021$ & $0.132\pm0.024$
& $0.189\pm0.057$ & $0.339\pm0.063$
& $0.035\pm0.008$ & $0.113\pm0.020$
& $0.578\pm0.018$ & $0.708\pm0.021$
& $0.531\pm0.010$ & $0.678\pm0.021$ \\
FullTilt (extended $\mathcal{G}$)
& $0.163\pm0.013$ & $0.202\pm0.014$
& $0.199\pm0.007$ & $0.400\pm0.005$
& $0.251\pm0.004$ & $0.372\pm0.004$
& $0.694\pm0.011$ & $0.722\pm0.010$
& $0.545\pm0.011$ & $0.577\pm0.011$ \\
\midrule
FullTilt vs.\ TomoTwin
& $\blacktriangle^{**}$ & $\blacktriangle^{**}$
& ns & ns
& $\blacktriangle^{**}$ & $\blacktriangle^{**}$
& $\blacktriangle^{**}$ & ns
& $\blacktriangle^{**}$ & $\triangledown^{**}$ \\
\bottomrule
\end{tabular}

\begin{tablenotes}[flushleft]\footnotesize
\item Statistical comparisons use the Wilcoxon signed-rank test over 10 trials. $\blacktriangle$/$\triangledown$: FullTilt with extended $\mathcal{G}$ performs significantly better/worse than TomoTwin; ns: no significant difference; ${}^{*}$: $p<0.05$; ${}^{**}$: $p<0.01$.
\end{tablenotes}
\end{threeparttable}%
}
\end{table*}

\clearpage
\section{Qualitative Detection Results}
\label{sec:supp_result_visual}
We provide visual assessments of FullTilt alongside corresponding ground truth annotations and estimations from the state-of-the-art open-set baselines, TomoTwin~\cite{rice2023tomotwin} and ProPicker~\cite{wiedemann2026propicker}. To comprehensively demonstrate the capabilities of our framework, FullTilt's 3D detection results are projected back onto the raw 2D tilt-series and additionally visualized across reconstructed tomogram Z-slices. Because TomoTwin and ProPicker operate exclusively on 3D reconstructed volumes, their corresponding detection results are visualized solely on the tomogram Z-slices. All results are derived from the intra-instance, single-prompt configuration. For visual clarity, bounding boxes are uniformly color-coded: FullTilt estimations are displayed in \textcolor{red}{red} (with the provided visual prompt in \textcolor{blue}{blue}), ground truth annotations in \textcolor{green}{lime}, TomoTwin estimations in \textcolor{orange}{orange}, and ProPicker estimations in \textcolor{magenta}{magenta}.

\begin{figure}[h]
\begin{center}
\includegraphics[width=1.0\linewidth]{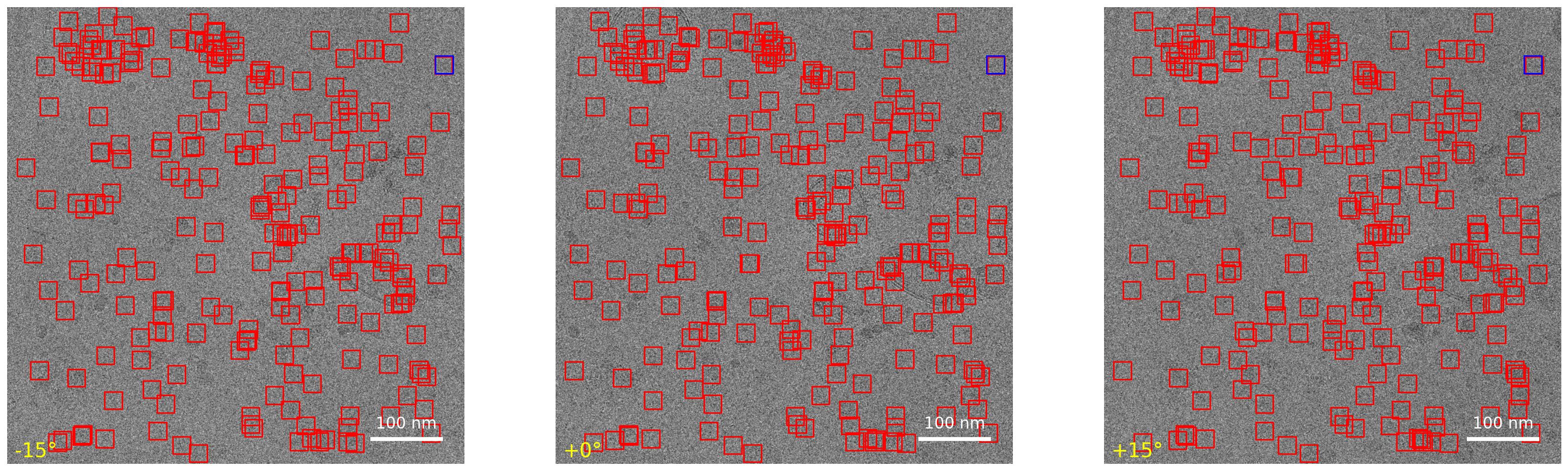}
\end{center}
\caption{\textbf{Projected FullTilt 3D detection results on the 2D tilt-series (DS-10440, \texttt{TS\_6\_6}).} We visualize the zero-shot detection of thyroglobulin particles evaluated under a single-prompt configuration. The reference visual prompt is highlighted by the blue bounding box. To prevent visual clutter, only a representative subset of the projection angles from the full tilt-series is displayed.}
\label{fig:detection_result_tiltseries_czii_TS_6_6}
\end{figure}

\begin{figure}[h]
\begin{center}
\includegraphics[width=1.0\linewidth]{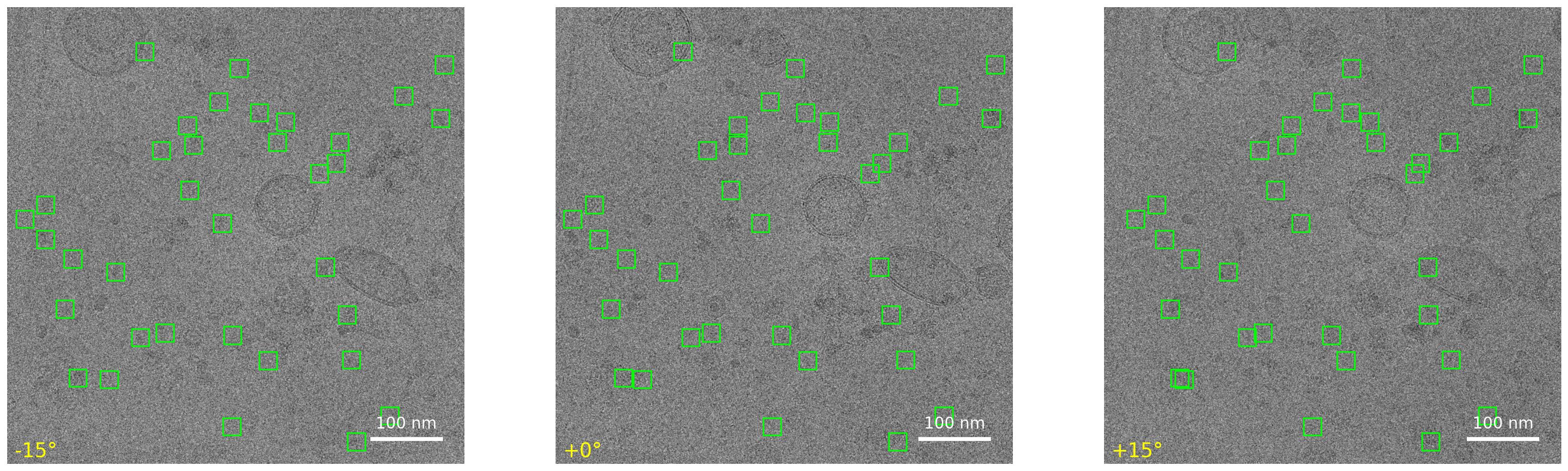}
\end{center}
\caption{\textbf{Projected 3D ground truth annotations on the 2D tilt-series (DS-10440, \texttt{TS\_6\_6}).} Ground truth localizations of thyroglobulin particles are shown. Only a representative subset of the projection angles from the full tilt-series is displayed.}
\label{fig:detection_result_tiltseries_czii_TS_6_6_gt}
\end{figure}

\begin{figure}[h]
\begin{center}
\includegraphics[width=1.0\linewidth]{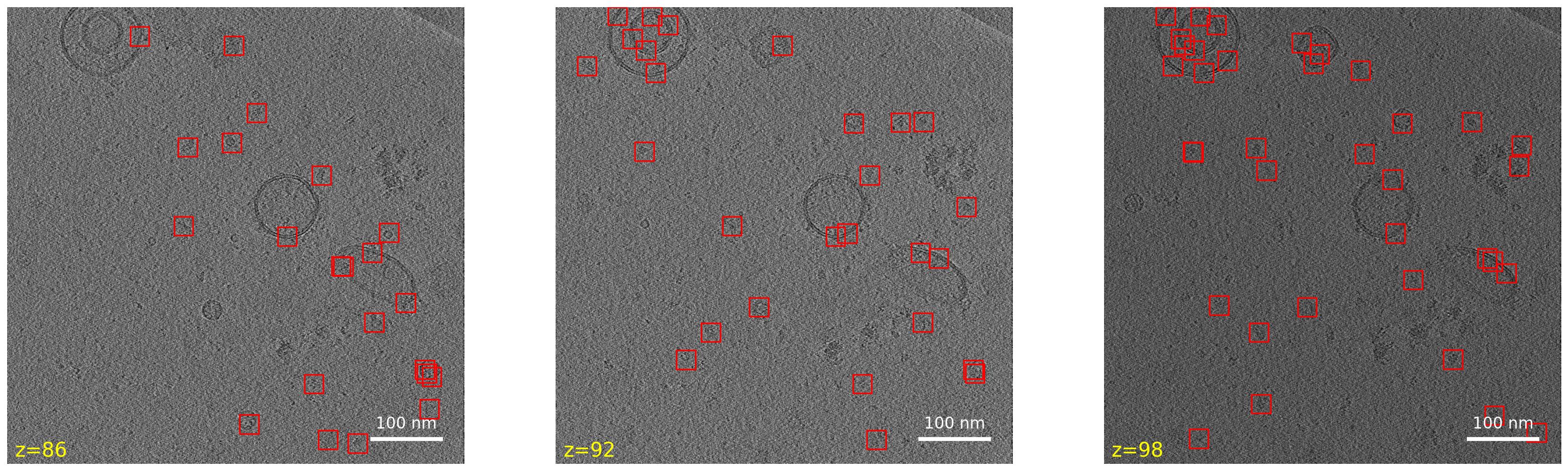}
\end{center}
\caption{\textbf{Tomogram Z-slice visualization of FullTilt detection results (DS-10440, \texttt{TS\_6\_6}).} We visualize the zero-shot detection of thyroglobulin particles evaluated under a single-prompt configuration. For each displayed $xy$-plane at depth $z$, we render only the estimated bounding boxes whose center coordinates fall within a local depth window defined by the particle's radius $r$ (i.e., $z \pm r$). A representative subset of the tomogram Z-slices is shown for visual clarity.}
\label{fig:detection_result_tomo_czii_TS_6_6}
\end{figure}

\begin{figure}[h]
\begin{center}
\includegraphics[width=1.0\linewidth]{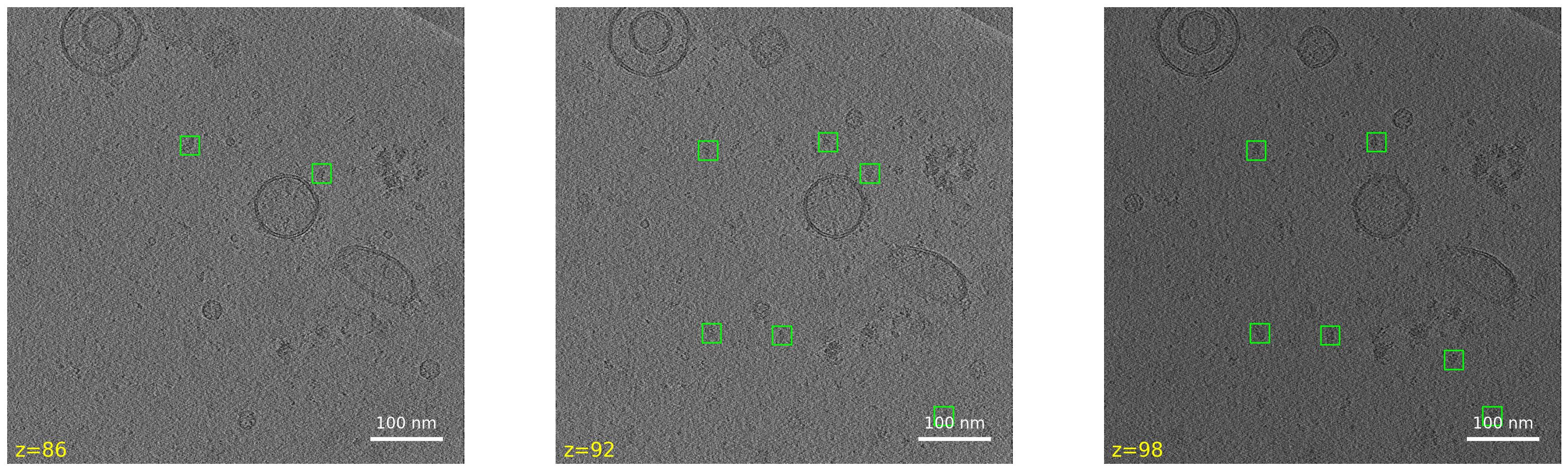}
\end{center}
\caption{\textbf{Tomogram Z-slice visualization of 3D ground truth (DS-10440, \texttt{TS\_6\_6}).} Ground truth localizations of thyroglobulin particles are shown. For each displayed $xy$-plane at depth $z$, we render only the bounding boxes whose center coordinates fall within a local depth window defined by the particle's radius $r$ (i.e., $z \pm r$). A representative subset of the tomogram Z-slices is shown for visual clarity.}
\label{fig:detection_result_tomo_czii_TS_6_6_gt}
\end{figure}

\begin{figure}[h]
\begin{center}
\includegraphics[width=1.0\linewidth]{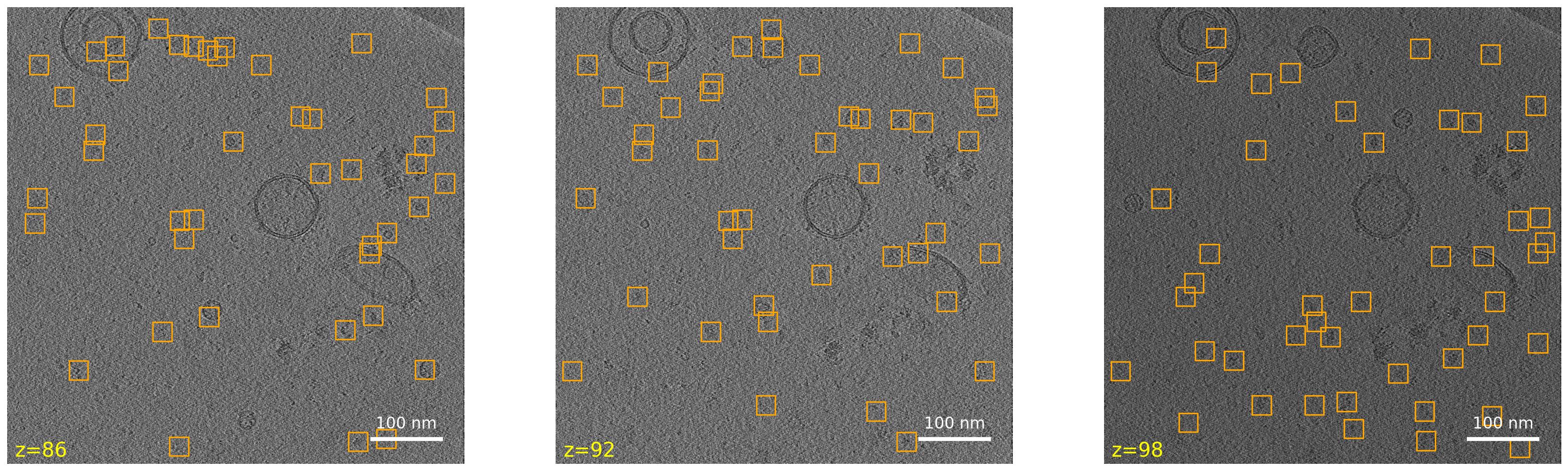}
\end{center}
\caption{\textbf{Tomogram Z-slice visualization of TomoTwin detection results (DS-10440, \texttt{TS\_6\_6}).} Estimated localizations of thyroglobulin particles are shown. For each displayed $xy$-plane at depth $z$, we render only the bounding boxes whose center coordinates fall within a local depth window defined by the particle's radius $r$ (i.e., $z \pm r$). A representative subset of the tomogram Z-slices is shown for visual clarity.}
\label{fig:tomotwin_detection_result_tomo_czii_TS_6_6}
\end{figure}

\begin{figure}[h]
\begin{center}
\includegraphics[width=1.0\linewidth]{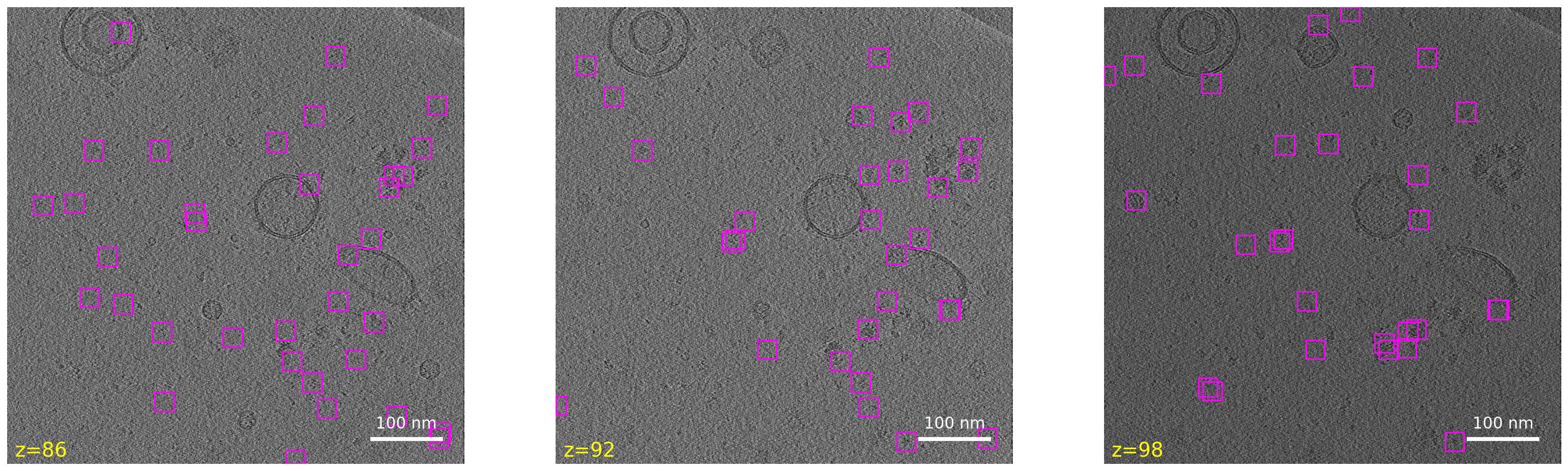}
\end{center}
\caption{\textbf{Tomogram Z-slice visualization of ProPicker detection results (DS-10440, \texttt{TS\_6\_6}).} Estimated localizations of thyroglobulin particles are shown. For each displayed $xy$-plane at depth $z$, we render only the bounding boxes whose center coordinates fall within a local depth window defined by the particle's radius $r$ (i.e., $z \pm r$). A representative subset of the tomogram Z-slices is shown for visual clarity.}
\label{fig:propicker_detection_result_tomo_czii_TS_6_6}
\end{figure}

\begin{figure}[h]
\begin{center}
\includegraphics[width=1.0\linewidth]{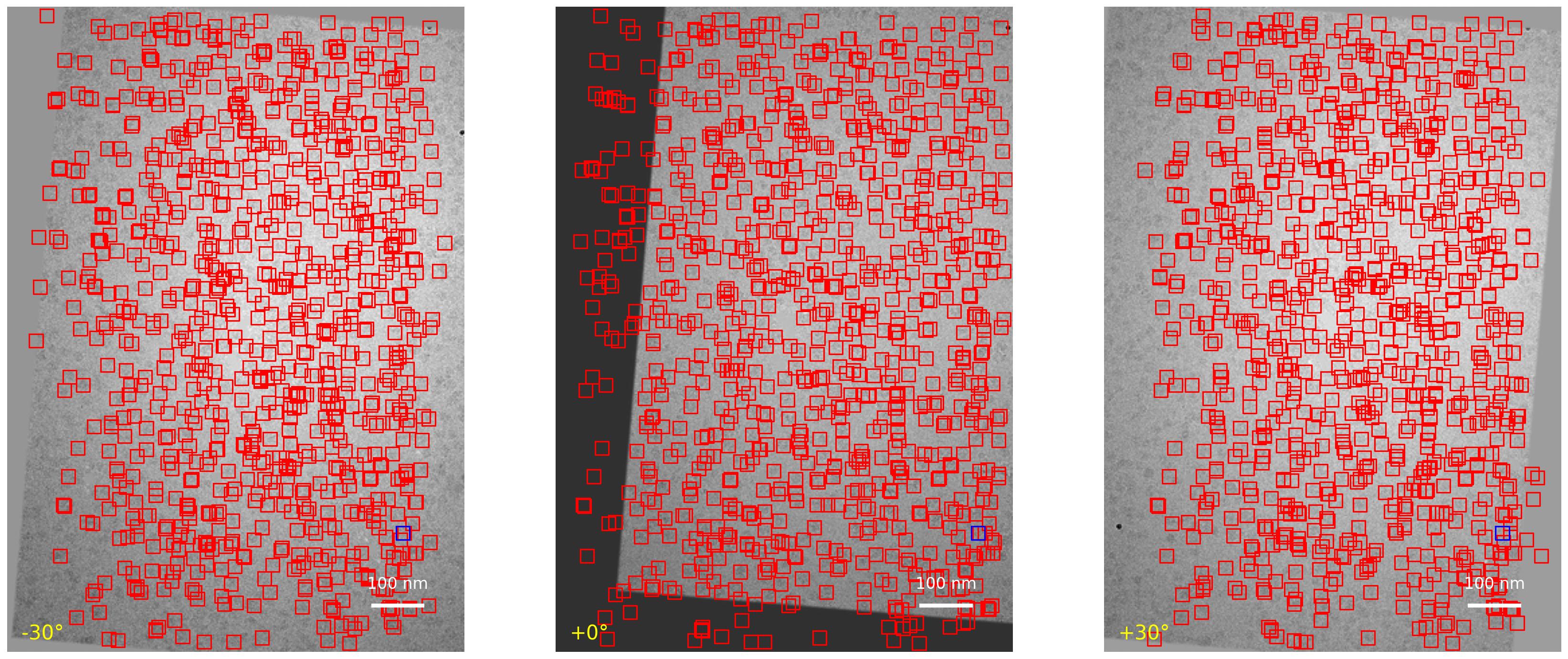}
\end{center}
\caption{\textbf{Projected FullTilt 3D detection results on the 2D tilt-series (EMPIAR-10304, \texttt{tilt1}).} We visualize the zero-shot detection of \textit{Escherichia coli} 70S ribosomes evaluated under a single-prompt configuration. The reference visual prompt is highlighted by the blue bounding box. To prevent visual clutter, only a representative subset of the projection angles from the full tilt-series is displayed.}
\label{fig:detection_result_tiltseries_10304_tilt1}
\end{figure}

\begin{figure}[h]
\begin{center}
\includegraphics[width=1.0\linewidth]{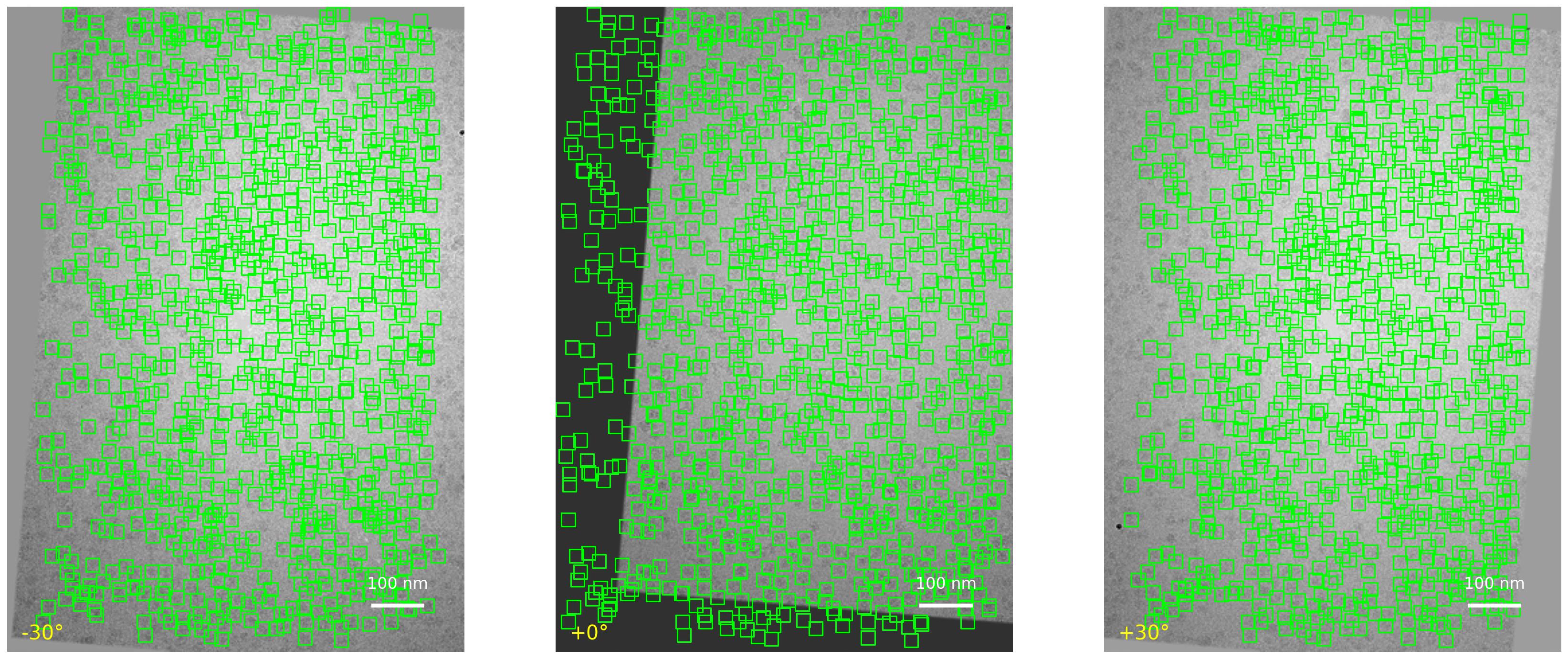}
\end{center}
\caption{\textbf{Projected 3D ground truth annotations on the 2D tilt-series (EMPIAR-10304, \texttt{tilt1}).} Ground truth localizations of \textit{Escherichia coli} 70S ribosomes are shown. Only a representative subset of the projection angles from the full tilt-series is displayed. Note that while a small fraction of particles may remain unlabeled due to the prohibitive difficulty of exhaustive 3D annotation, these empirical ground truth labels remain qualitatively robust.}
\label{fig:detection_result_tiltseries_10304_tilt1_gt}
\end{figure}

\begin{figure}[h]
\begin{center}
\includegraphics[width=1.0\linewidth]{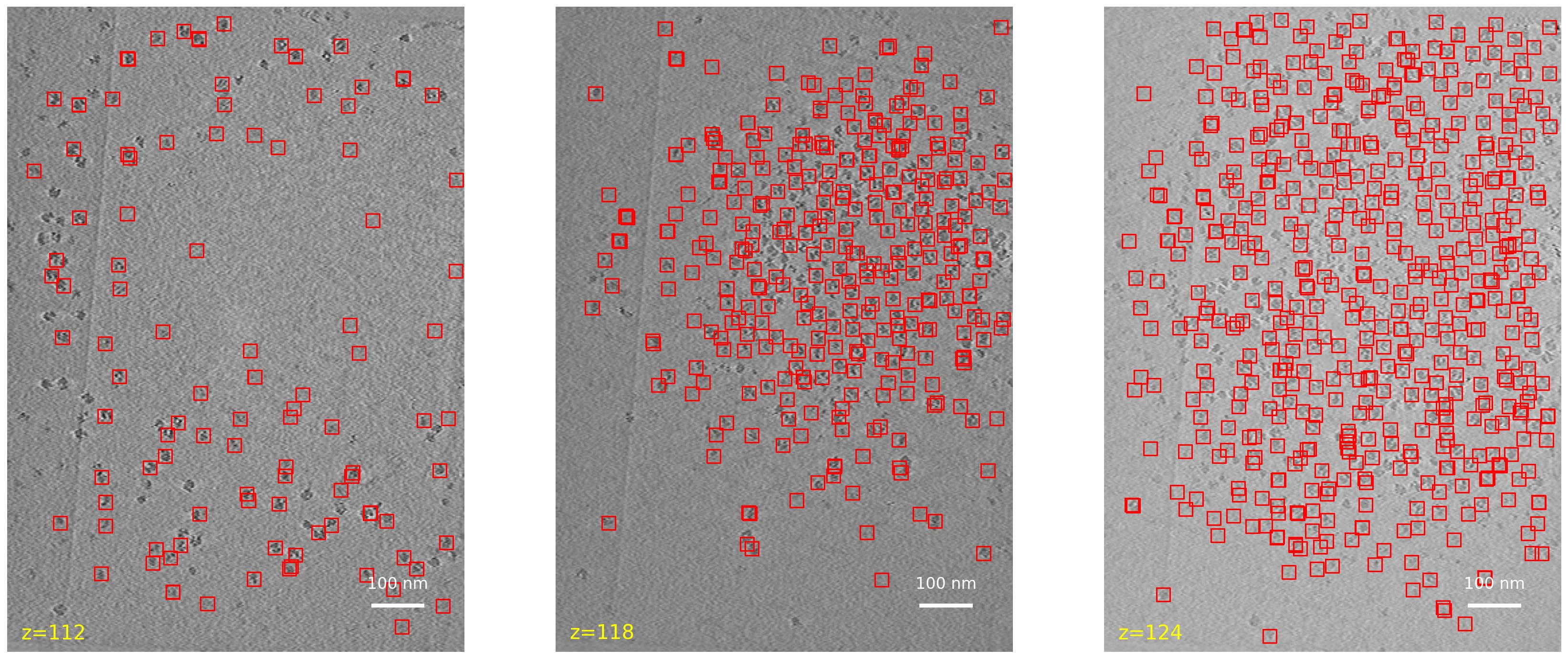}
\end{center}
\caption{\textbf{Tomogram Z-slice visualization of FullTilt detection results (EMPIAR-10304, \texttt{tilt1}).} We visualize the zero-shot detection of \textit{Escherichia coli} 70S ribosomes evaluated under a single-prompt configuration. For each displayed $xy$-plane at depth $z$, we render only the estimated bounding boxes whose center coordinates fall within a local depth window defined by the particle's radius $r$ (i.e., $z \pm r$). A representative subset of the tomogram Z-slices is shown for visual clarity.}
\label{fig:detection_result_tomo_10304_tilt1}
\end{figure}

\begin{figure}[h]
\begin{center}
\includegraphics[width=1.0\linewidth]{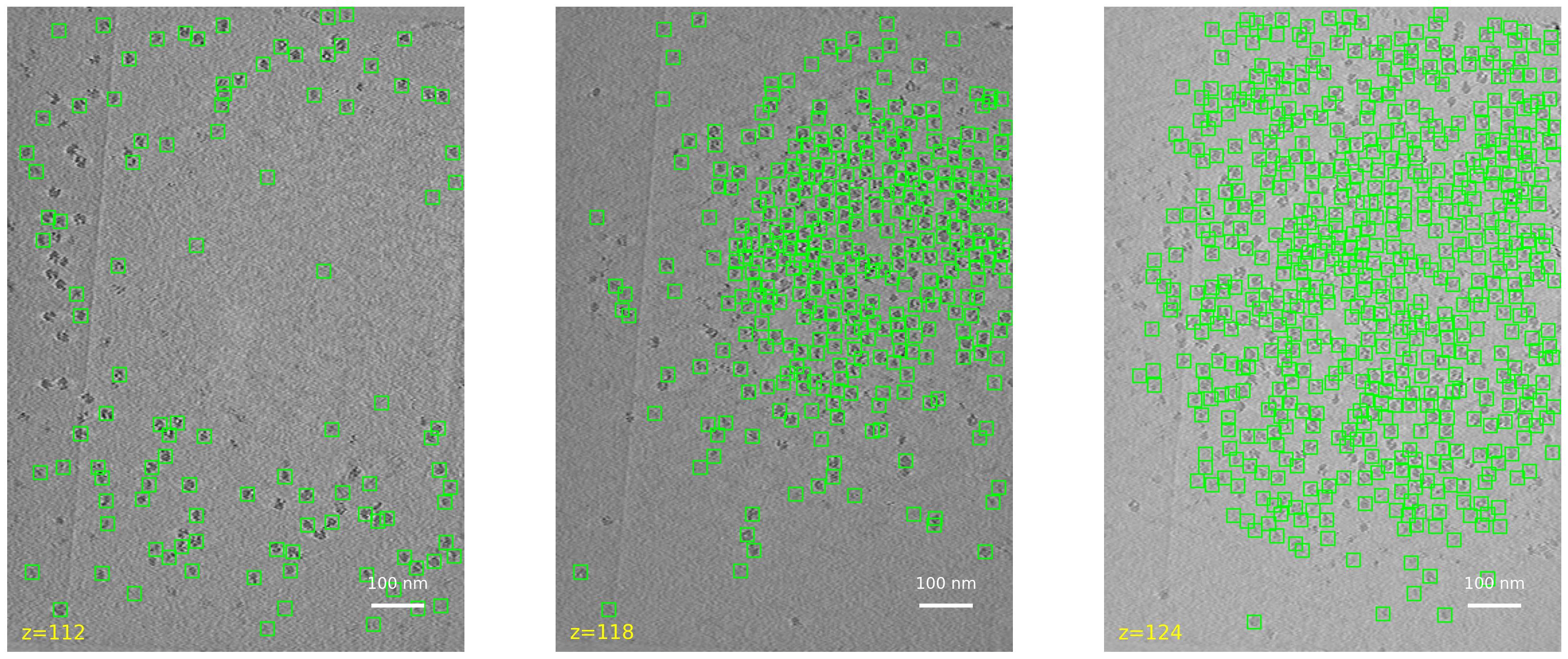}
\end{center}
\caption{\textbf{Tomogram Z-slice visualization of 3D ground truth (EMPIAR-10304, \texttt{tilt1}).} Ground truth localizations of \textit{Escherichia coli} 70S ribosomes are shown. For each displayed $xy$-plane at depth $z$, we render only the bounding boxes whose center coordinates fall within a local depth window defined by the particle's radius $r$ (i.e., $z \pm r$). A representative subset of the tomogram Z-slices is shown for visual clarity. Note that while a small fraction of particles may remain unlabeled due to the prohibitive difficulty of exhaustive 3D annotation, these empirical ground truth labels remain qualitatively robust.}
\label{fig:detection_result_tomo_10304_tilt1_gt}
\end{figure}

\begin{figure}[h]
\begin{center}
\includegraphics[width=1.0\linewidth]{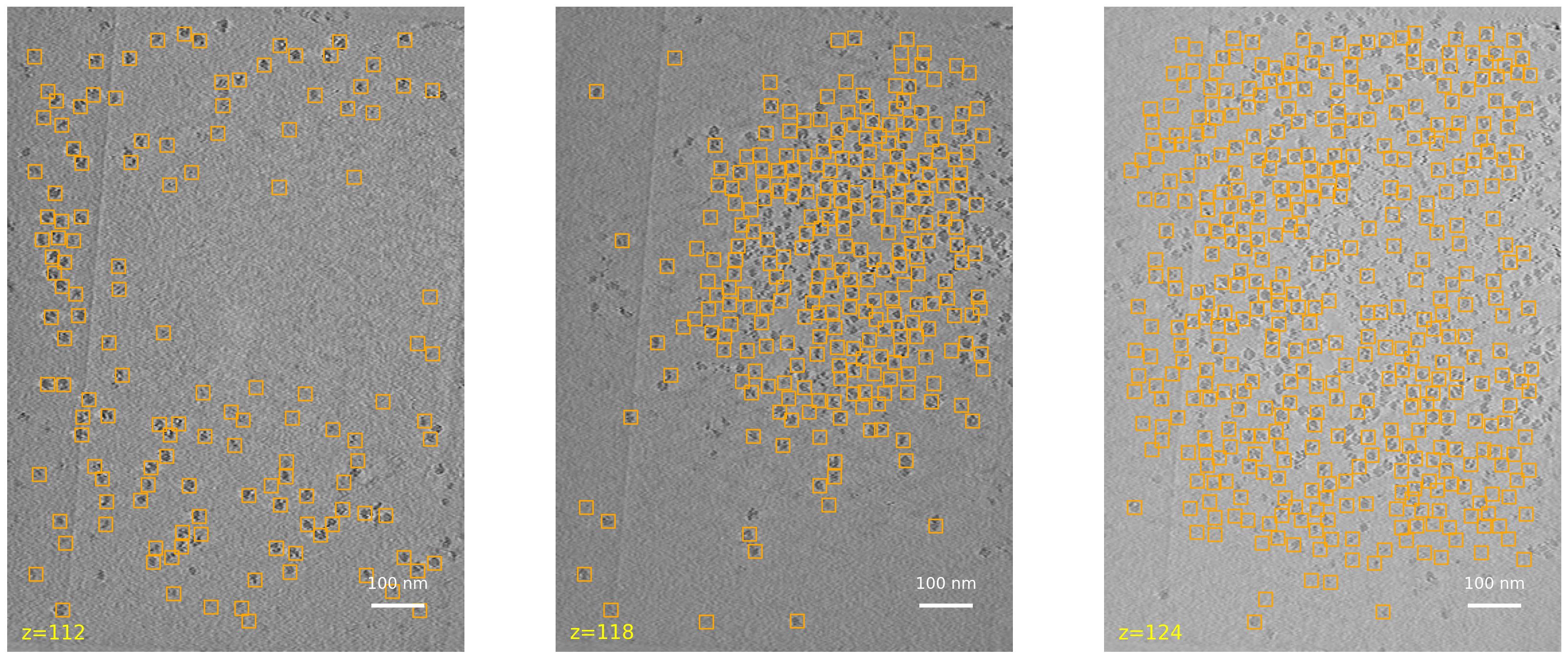}
\end{center}
\caption{\textbf{Tomogram Z-slice visualization of TomoTwin detection results (EMPIAR-10304, \texttt{tilt1}).} Estimated localizations of \textit{Escherichia coli} 70S ribosomes are shown. For each displayed $xy$-plane at depth $z$, we render only the bounding boxes whose center coordinates fall within a local depth window defined by the particle's radius $r$ (i.e., $z \pm r$). A representative subset of the tomogram Z-slices is shown for visual clarity.}
\label{fig:tomotwin_detection_result_tomo_10304_tilt1}
\end{figure}

\begin{figure}[h]
\begin{center}
\includegraphics[width=1.0\linewidth]{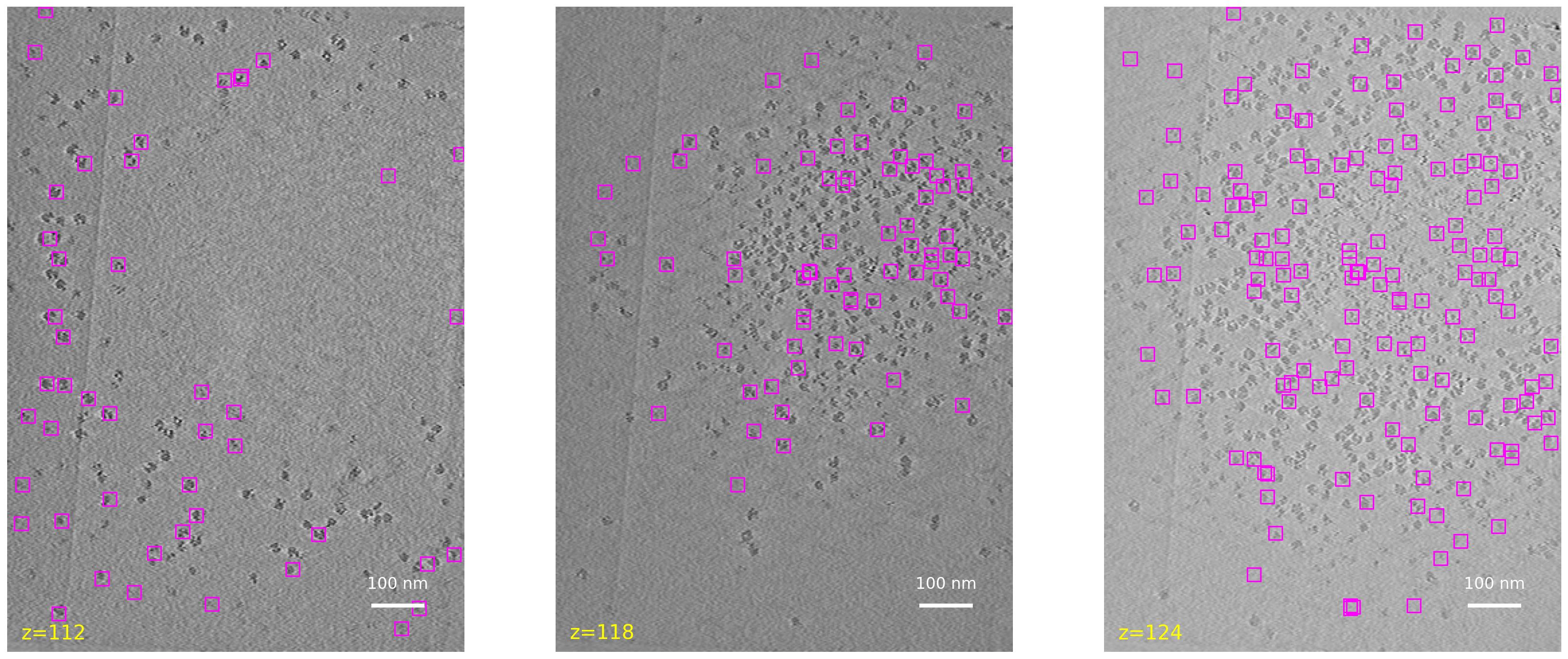}
\end{center}
\caption{\textbf{Tomogram Z-slice visualization of ProPicker detection results (EMPIAR-10304, \texttt{tilt1}).} Estimated localizations of \textit{Escherichia coli} 70S ribosomes are shown. For each displayed $xy$-plane at depth $z$, we render only the bounding boxes whose center coordinates fall within a local depth window defined by the particle's radius $r$ (i.e., $z \pm r$). A representative subset of the tomogram Z-slices is shown for visual clarity.}
\label{fig:propicker_detection_result_tomo_10304_tilt1}
\end{figure}

\begin{figure}[h]
\begin{center}
\includegraphics[width=1.0\linewidth]{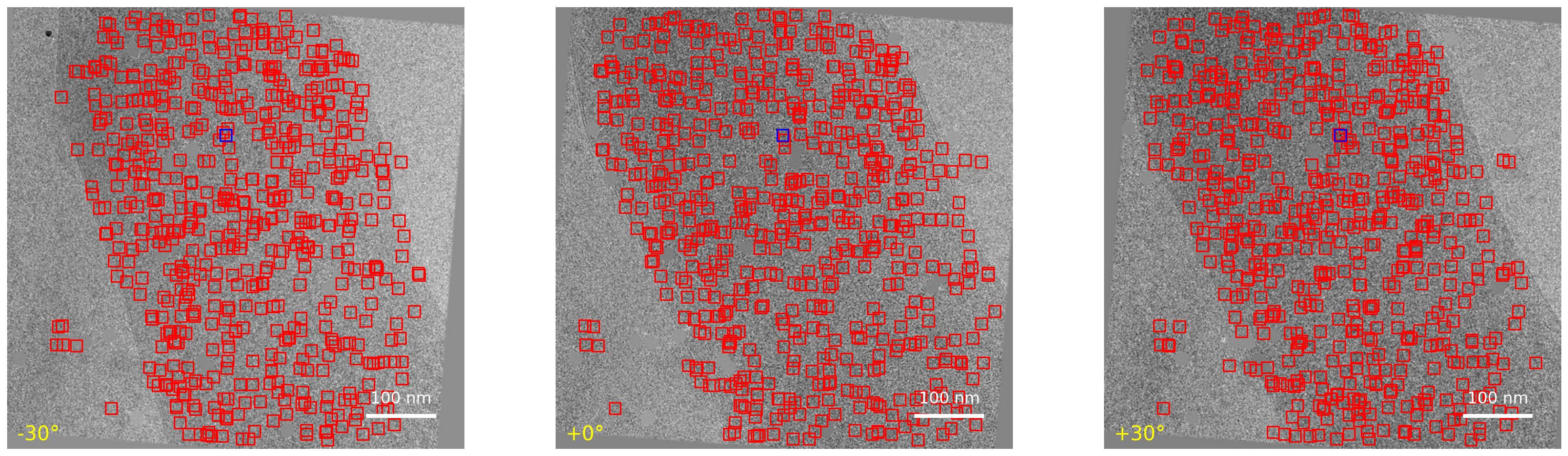}
\end{center}
\caption{\textbf{Projected FullTilt 3D detection results on the 2D tilt-series (EMPIAR-10499, \texttt{TS\_01}).} We visualize the zero-shot detection of \textit{Mycoplasma pneumoniae} ribosomes evaluated under a single-prompt configuration. The reference visual prompt is highlighted by the blue bounding box. To prevent visual clutter, only a representative subset of the projection angles from the full tilt-series is displayed.}
\label{fig:detection_result_tiltseries_10499_TS_01}
\end{figure}

\begin{figure}[h]
\begin{center}
\includegraphics[width=1.0\linewidth]{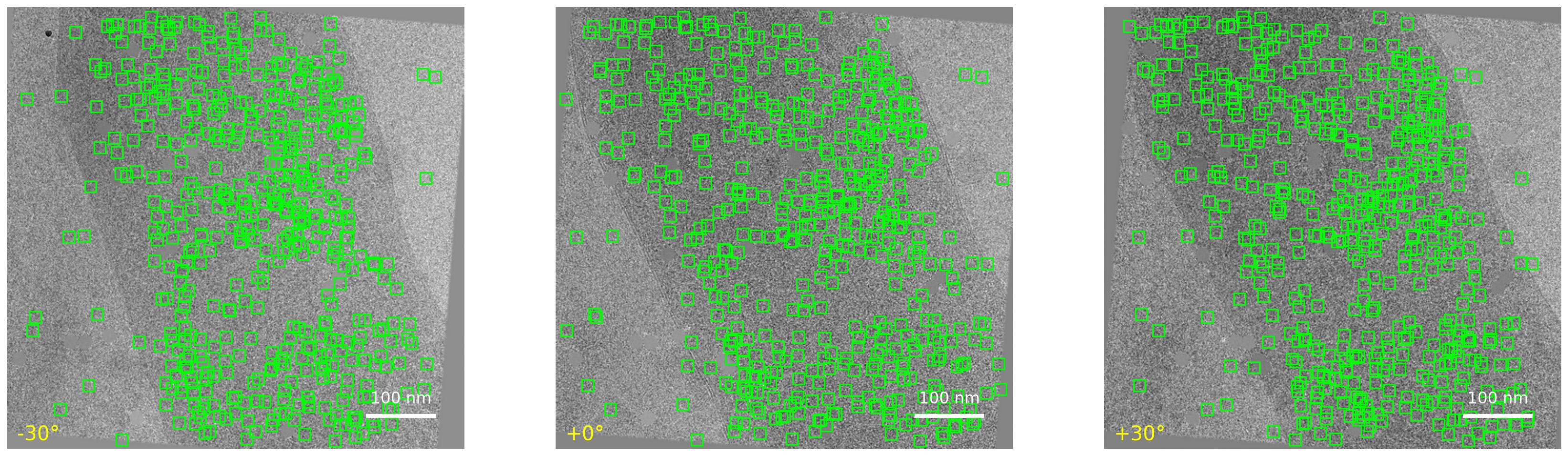}
\end{center}
\caption{\textbf{Projected 3D ground truth annotations on the 2D tilt-series (EMPIAR-10499, \texttt{TS\_01}).} Ground truth localizations of \textit{Mycoplasma pneumoniae} ribosomes are shown. Only a representative subset of the projection angles from the full tilt-series is displayed. Note that while a small fraction of particles may remain unlabeled due to the prohibitive difficulty of exhaustive 3D annotation, these empirical ground truth labels remain qualitatively robust.}
\label{fig:detection_result_tiltseries_10499_TS_01_gt}
\end{figure}

\begin{figure}[h]
\begin{center}
\includegraphics[width=1.0\linewidth]{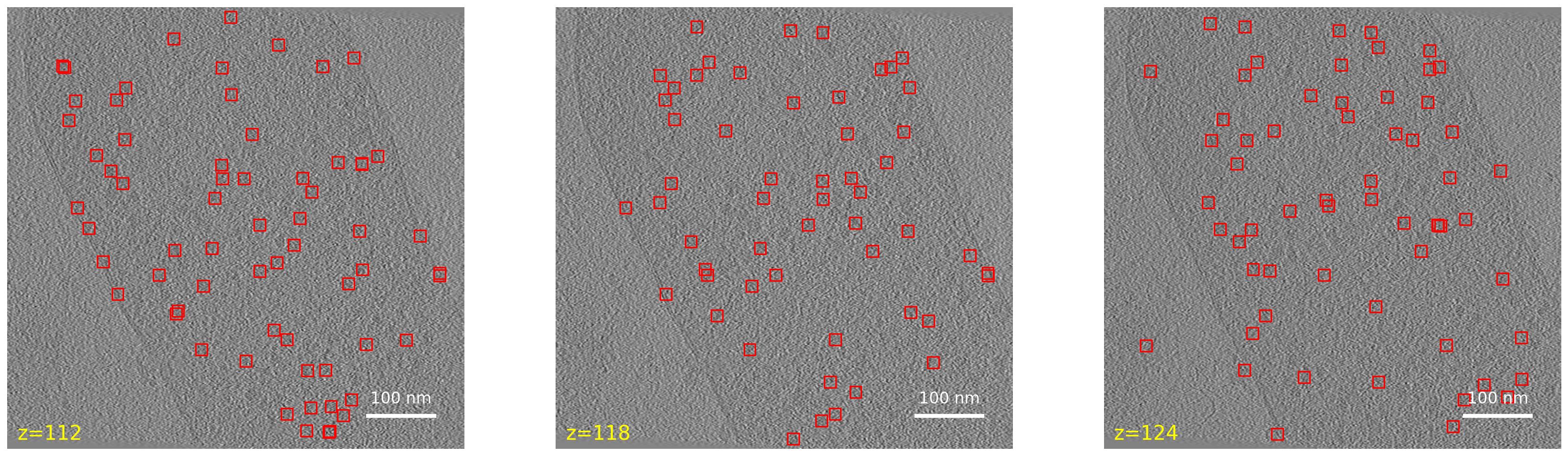}
\end{center}
\caption{\textbf{Tomogram Z-slice visualization of FullTilt detection results (EMPIAR-10499, \texttt{TS\_01}).} We visualize the zero-shot detection of \textit{Mycoplasma pneumoniae} ribosomes evaluated under a single-prompt configuration. For each displayed $xy$-plane at depth $z$, we render only the estimated bounding boxes whose center coordinates fall within a local depth window defined by the particle's radius $r$ (i.e., $z \pm r$). A representative subset of the tomogram Z-slices is shown for visual clarity.}
\label{fig:detection_result_tomo_10499_TS_01}
\end{figure}

\begin{figure}[h]
\begin{center}
\includegraphics[width=1.0\linewidth]{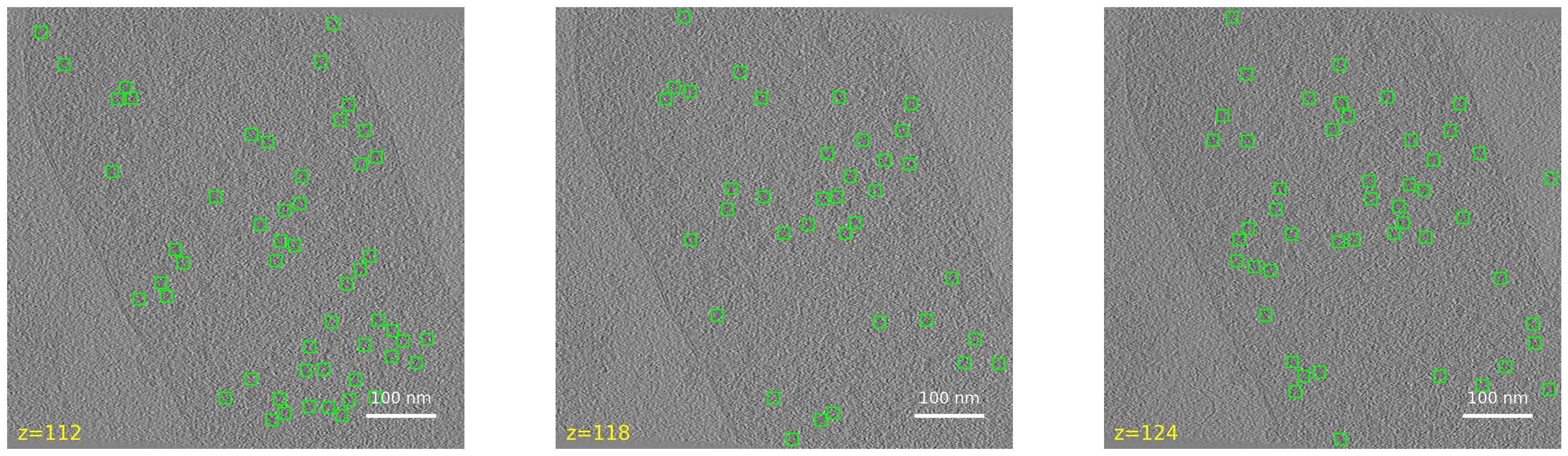}
\end{center}
\caption{\textbf{Tomogram Z-slice visualization of 3D ground truth (EMPIAR-10499, \texttt{TS\_01}).} Ground truth localizations of \textit{Mycoplasma pneumoniae} ribosomes are shown. For each displayed $xy$-plane at depth $z$, we render only the bounding boxes whose center coordinates fall within a local depth window defined by the particle's radius $r$ (i.e., $z \pm r$). A representative subset of the tomogram Z-slices is shown for visual clarity. Note that while a small fraction of particles may remain unlabeled due to the prohibitive difficulty of exhaustive 3D annotation, these empirical ground truth labels remain qualitatively robust.}
\label{fig:detection_result_tomo_10499_TS_01_gt}
\end{figure}

\begin{figure}[h]
\begin{center}
\includegraphics[width=1.0\linewidth]{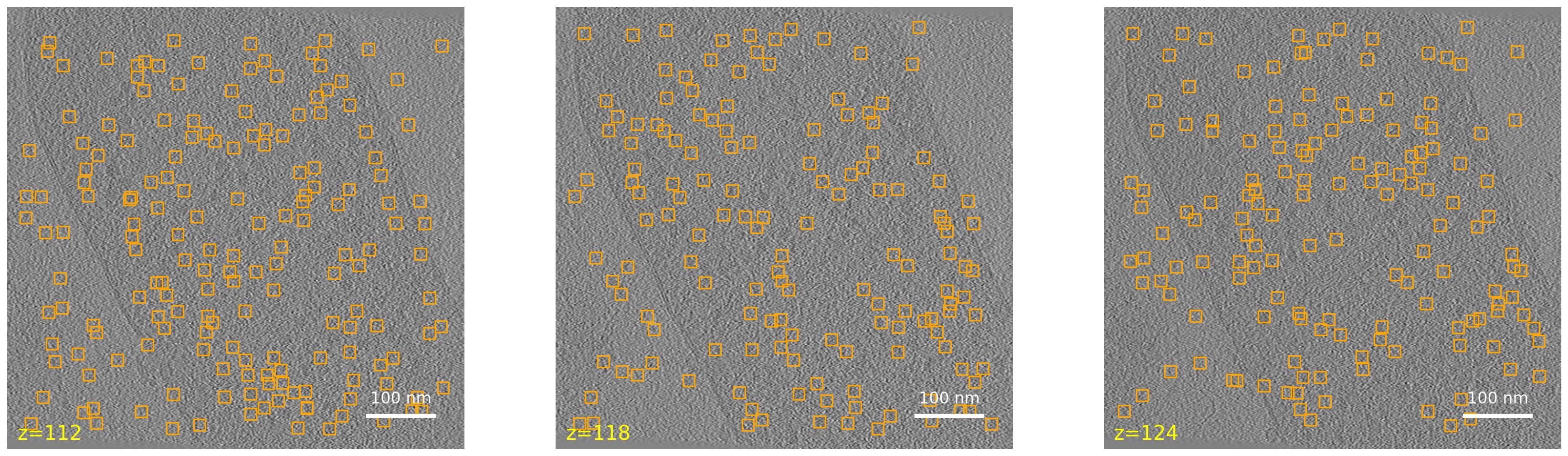}
\end{center}
\caption{\textbf{Tomogram Z-slice visualization of TomoTwin detection results (EMPIAR-10499, \texttt{TS\_01}).} Estimated localizations of \textit{Mycoplasma pneumoniae} ribosomes are shown. For each displayed $xy$-plane at depth $z$, we render only the bounding boxes whose center coordinates fall within a local depth window defined by the particle's radius $r$ (i.e., $z \pm r$). A representative subset of the tomogram Z-slices is shown for visual clarity.}
\label{fig:tomotwin_detection_result_tomo_10499_TS_01}
\end{figure}

\begin{figure}[h]
\begin{center}
\includegraphics[width=1.0\linewidth]{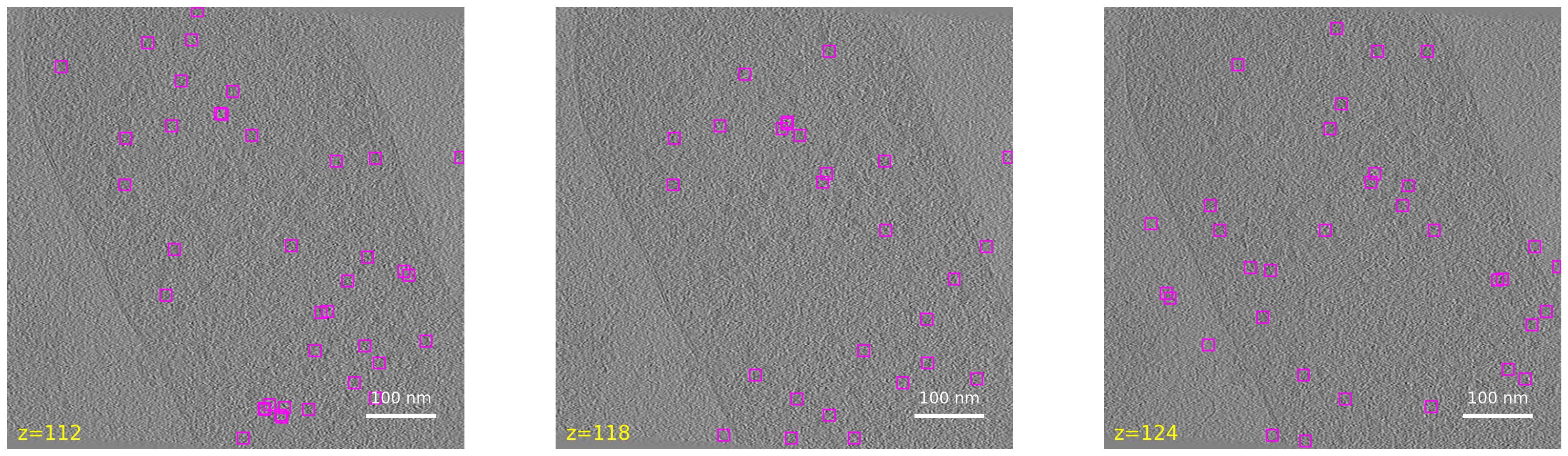}
\end{center}
\caption{\textbf{Tomogram Z-slice visualization of ProPicker detection results (EMPIAR-10499, \texttt{TS\_01}).} Estimated localizations of \textit{Mycoplasma pneumoniae} ribosomes are shown. For each displayed $xy$-plane at depth $z$, we render only the bounding boxes whose center coordinates fall within a local depth window defined by the particle's radius $r$ (i.e., $z \pm r$). A representative subset of the tomogram Z-slices is shown for visual clarity.}
\label{fig:propicker_detection_result_tomo_10499_TS_01}
\end{figure}

\begin{figure}[h]
\begin{center}
\includegraphics[width=1.0\linewidth]{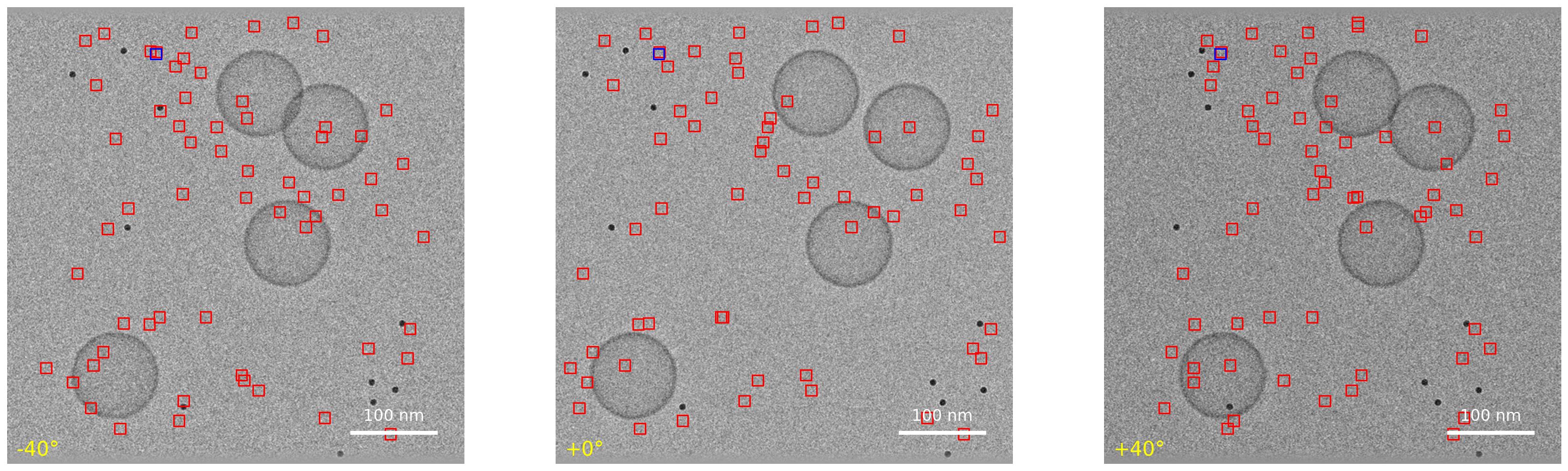}
\end{center}
\caption{\textbf{Projected FullTilt 3D detection results on the 2D tilt-series (seen simulation, \texttt{TS\_4\_0}).} We visualize the detection of \texttt{5JH9} particles evaluated under a single-prompt configuration. The reference visual prompt is highlighted by the blue bounding box. To prevent visual clutter, only a representative subset of the projection angles from the full tilt-series is displayed.}
\label{fig:detection_result_tiltseries_sim_same_4_0}
\end{figure}

\begin{figure}[h]
\begin{center}
\includegraphics[width=1.0\linewidth]{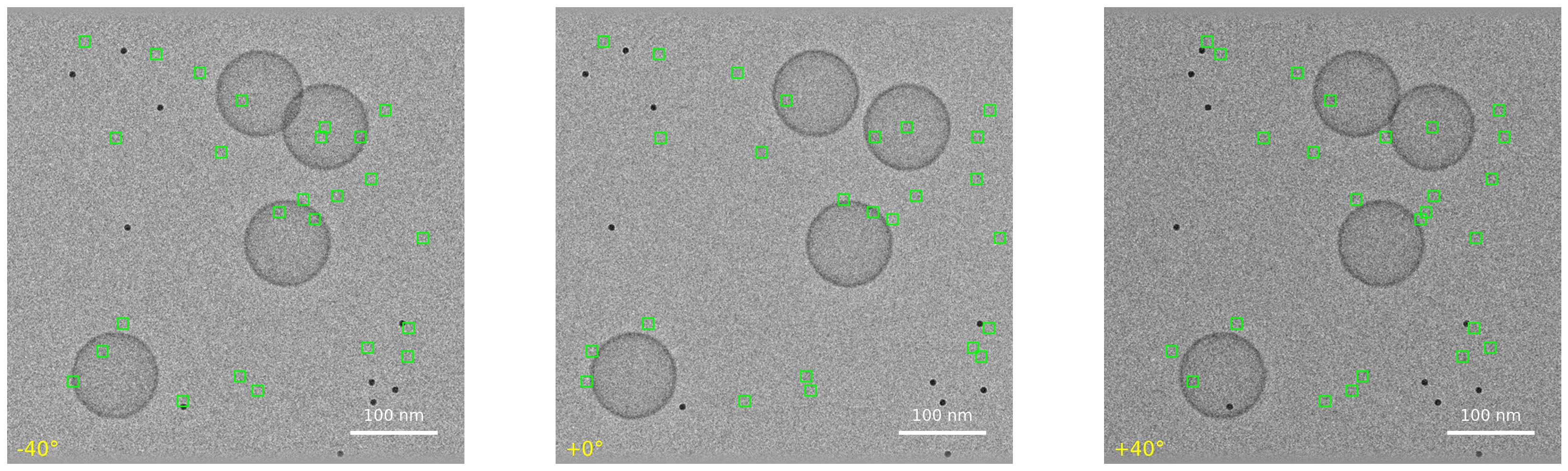}
\end{center}
\caption{\textbf{Projected 3D ground truth annotations on the 2D tilt-series (seen simulation, \texttt{TS\_4\_0}).} Ground truth localizations of \texttt{5JH9} particles are shown. Only a representative subset of the projection angles from the full tilt-series is displayed.}
\label{fig:detection_result_tiltseries_sim_same_4_0_gt}
\end{figure}

\begin{figure}[h]
\begin{center}
\includegraphics[width=1.0\linewidth]{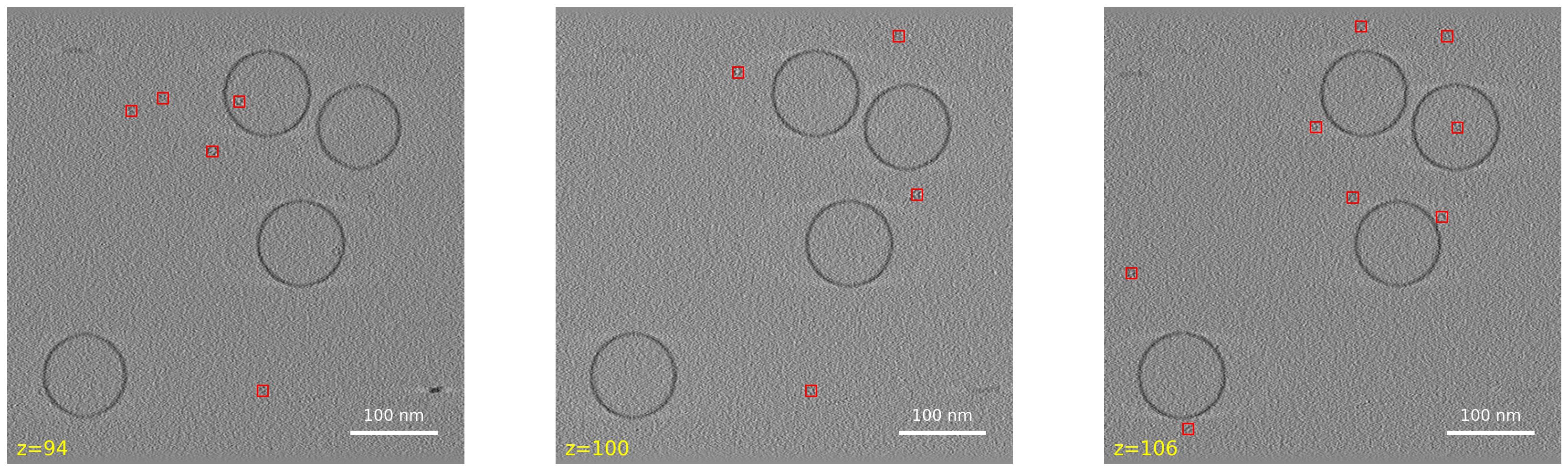}
\end{center}
\caption{\textbf{Tomogram Z-slice visualization of FullTilt detection results (``seen'' simulation, \texttt{TS\_4\_0}).} We visualize the detection of \texttt{5JH9} particles evaluated under a single-prompt configuration. For each displayed $xy$-plane at depth $z$, we render only the estimated bounding boxes whose center coordinates fall within a local depth window defined by the particle's radius $r$ (i.e., $z \pm r$). A representative subset of the tomogram Z-slices is shown for visual clarity.}
\label{fig:detection_result_tomo_sim_same_4_0}
\end{figure}

\begin{figure}[h]
\begin{center}
\includegraphics[width=1.0\linewidth]{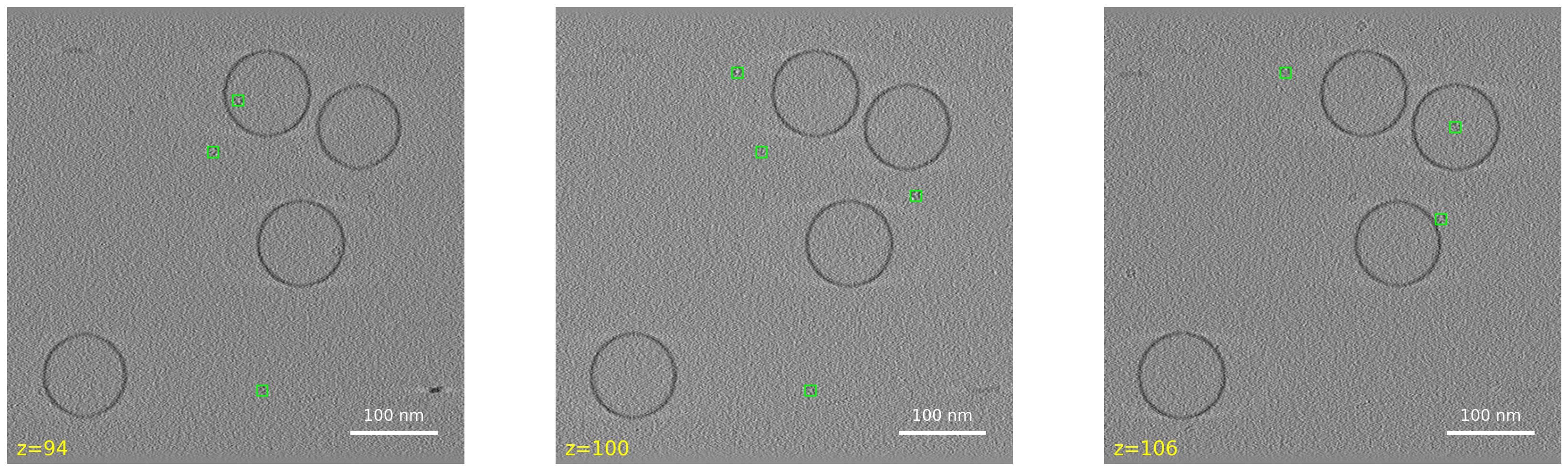}
\end{center}
\caption{\textbf{Tomogram Z-slice visualization of 3D ground truth (``seen'' simulation, \texttt{TS\_4\_0}).} Ground truth localizations of \texttt{5JH9} particles are shown. For each displayed $xy$-plane at depth $z$, we render only the bounding boxes whose center coordinates fall within a local depth window defined by the particle's radius $r$ (i.e., $z \pm r$). A representative subset of the tomogram Z-slices is shown for visual clarity.}
\label{fig:detection_result_tomo_sim_same_4_0_gt}
\end{figure}

\begin{figure}[h]
\begin{center}
\includegraphics[width=1.0\linewidth]{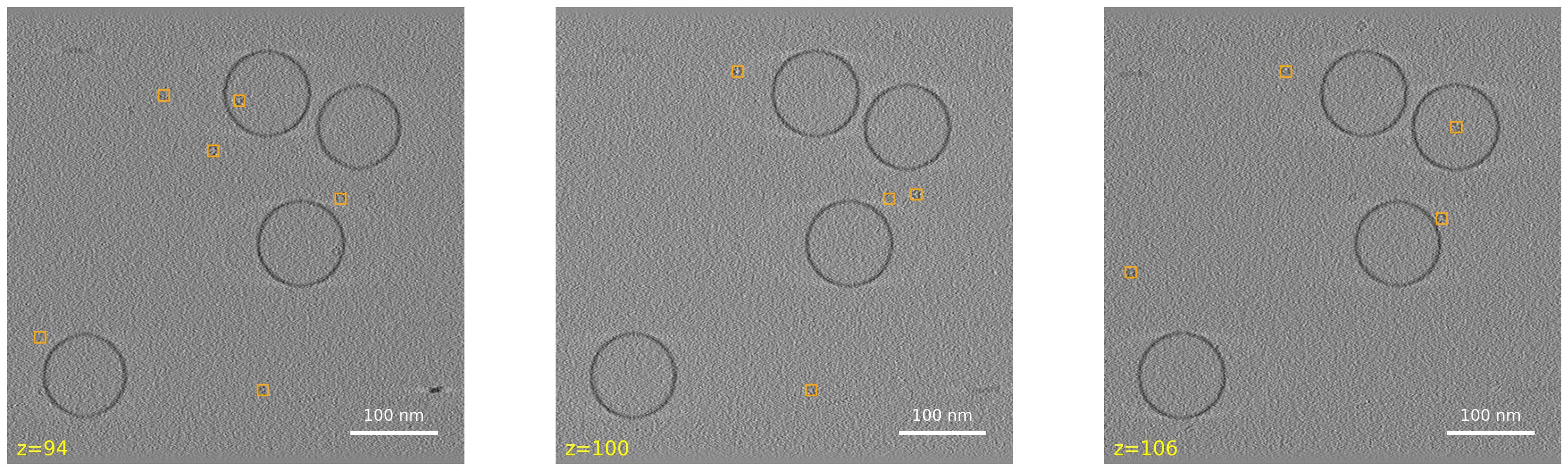}
\end{center}
\caption{\textbf{Tomogram Z-slice visualization of TomoTwin detection results (``seen'' simulation, \texttt{TS\_4\_0}).} Estimated localizations of \texttt{5JH9} particles are shown. For each displayed $xy$-plane at depth $z$, we render only the bounding boxes whose center coordinates fall within a local depth window defined by the particle's radius $r$ (i.e., $z \pm r$). A representative subset of the tomogram Z-slices is shown for visual clarity.}
\label{fig:tomotwin_detection_result_tomo_sim_same_4_0}
\end{figure}

\begin{figure}[h]
\begin{center}
\includegraphics[width=1.0\linewidth]{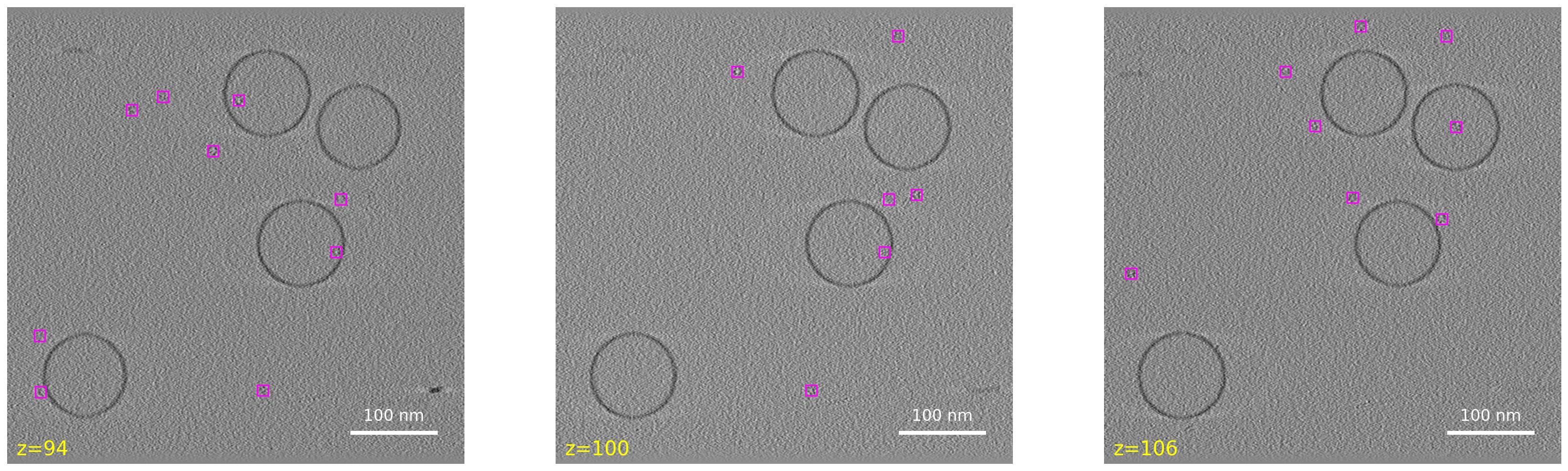}
\end{center}
\caption{\textbf{Tomogram Z-slice visualization of ProPicker detection results (``seen'' simulation, \texttt{TS\_4\_0}).} Estimated localizations of \texttt{5JH9} particles are shown. For each displayed $xy$-plane at depth $z$, we render only the bounding boxes whose center coordinates fall within a local depth window defined by the particle's radius $r$ (i.e., $z \pm r$). A representative subset of the tomogram Z-slices is shown for visual clarity.}
\label{fig:propicker_detection_result_tomo_sim_same_4_0}
\end{figure}

\begin{figure}[h]
\begin{center}
\includegraphics[width=1.0\linewidth]{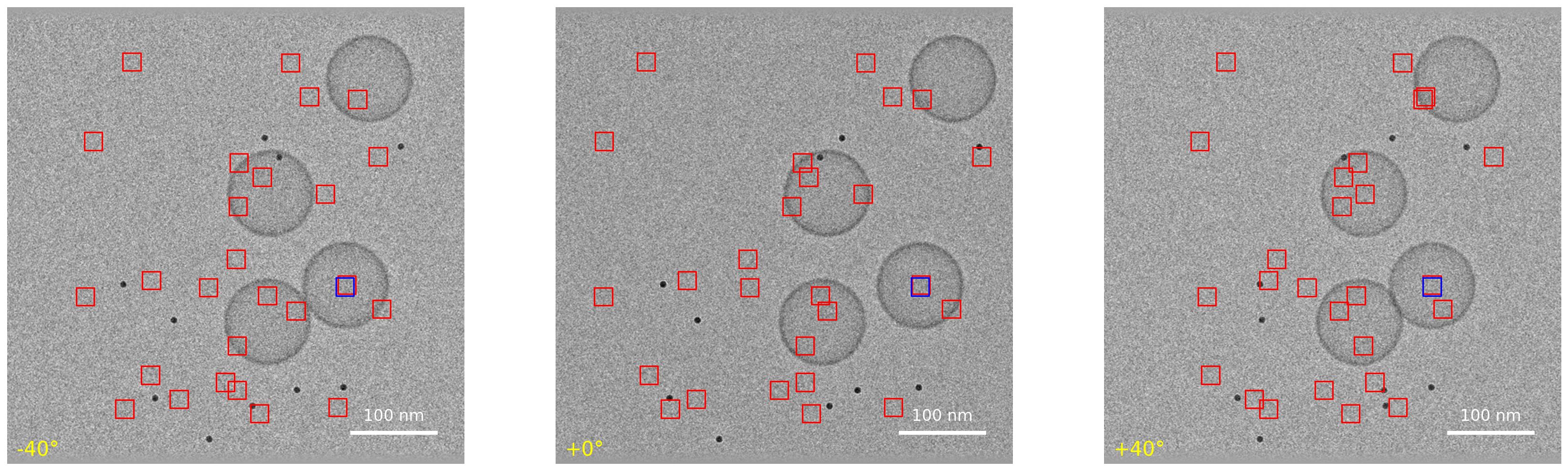}
\end{center}
\caption{\textbf{Projected FullTilt 3D detection results on the 2D tilt-series (``unseen'' simulation, \texttt{TS\_4\_1}).} We visualize the zero-shot detection of \texttt{2DF7} particles evaluated under a single-prompt configuration. The reference visual prompt is highlighted by the blue bounding box. To prevent visual clutter, only a representative subset of the projection angles from the full tilt-series is displayed.}
\label{fig:detection_result_tiltseries_sim_diff_4_1}
\end{figure}

\begin{figure}[h]
\begin{center}
\includegraphics[width=1.0\linewidth]{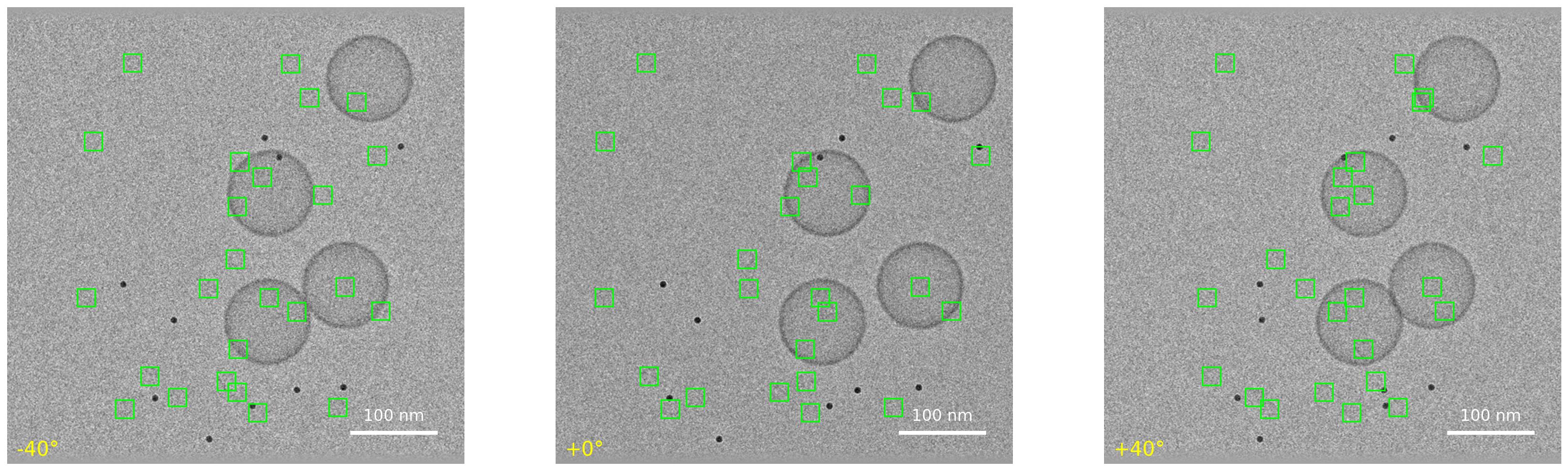}
\end{center}
\caption{\textbf{Projected 3D ground truth annotations on the 2D tilt-series (``unseen'' simulation, \texttt{TS\_4\_1}).} Ground truth localizations of \texttt{2DF7} particles are shown. Only a representative subset of the projection angles from the full tilt-series is displayed.}
\label{fig:detection_result_tiltseries_sim_diff_4_1_gt}
\end{figure}

\begin{figure}[h]
\begin{center}
\includegraphics[width=1.0\linewidth]{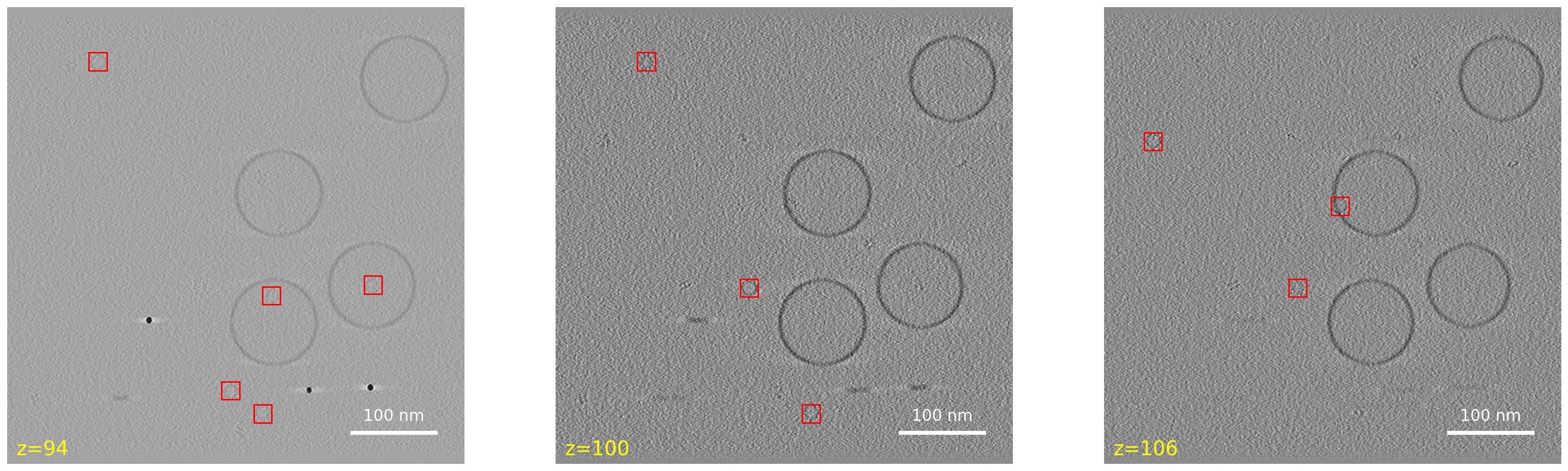}
\end{center}
\caption{\textbf{Tomogram Z-slice visualization of FullTilt detection results (``unseen'' simulation, \texttt{TS\_4\_1}).} We visualize the zero-shot detection of \texttt{2DF7} particles evaluated under a single-prompt configuration. For each displayed $xy$-plane at depth $z$, we render only the estimated bounding boxes whose center coordinates fall within a local depth window defined by the particle's radius $r$ (i.e., $z \pm r$). A representative subset of the tomogram Z-slices is shown for visual clarity.}
\label{fig:detection_result_tomo_sim_diff_4_1}
\end{figure}

\begin{figure}[h]
\begin{center}
\includegraphics[width=1.0\linewidth]{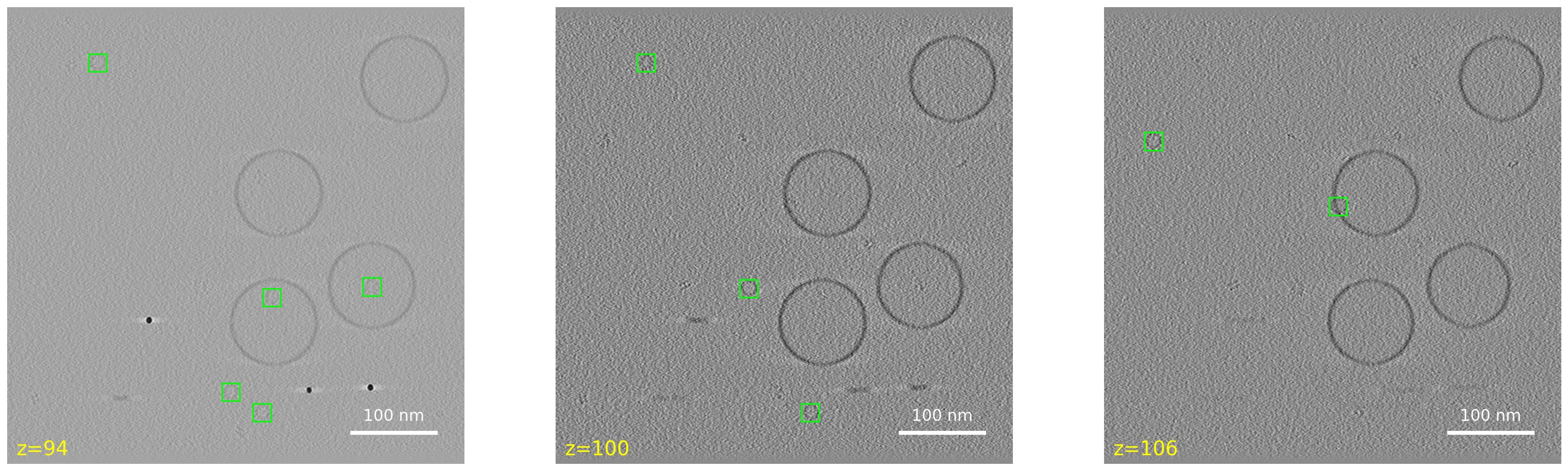}
\end{center}
\caption{\textbf{Tomogram Z-slice visualization of 3D ground truth (``unseen'' simulation, \texttt{TS\_4\_1}).} Ground truth localizations of \texttt{2DF7} particles are shown. For each displayed $xy$-plane at depth $z$, we render only the bounding boxes whose center coordinates fall within a local depth window defined by the particle's radius $r$ (i.e., $z \pm r$). A representative subset of the tomogram Z-slices is shown for visual clarity.}
\label{fig:detection_result_tomo_sim_diff_4_1_gt}
\end{figure}

\begin{figure}[h]
\begin{center}
\includegraphics[width=1.0\linewidth]{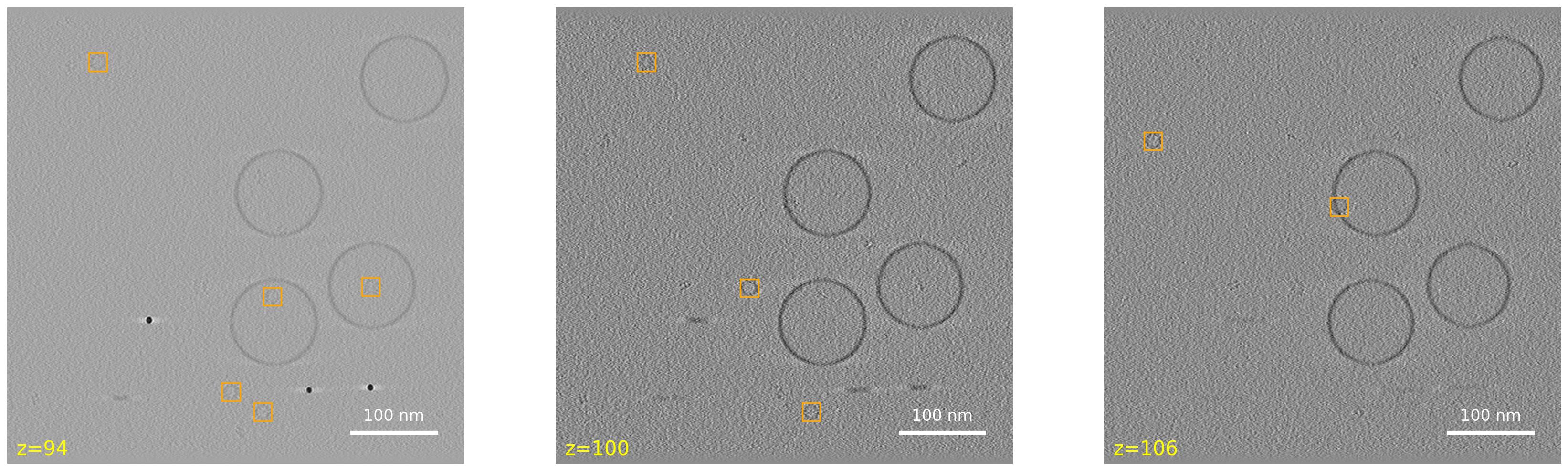}
\end{center}
\caption{\textbf{Tomogram Z-slice visualization of TomoTwin detection results (``seen'' simulation, \texttt{TS\_4\_1}).} Estimated localizations of \texttt{2DF7} particles are shown. For each displayed $xy$-plane at depth $z$, we render only the bounding boxes whose center coordinates fall within a local depth window defined by the particle's radius $r$ (i.e., $z \pm r$). A representative subset of the tomogram Z-slices is shown for visual clarity.}
\label{fig:tomotwin_detection_result_tomo_sim_diff_4_1}
\end{figure}

\begin{figure}[h]
\begin{center}
\includegraphics[width=1.0\linewidth]{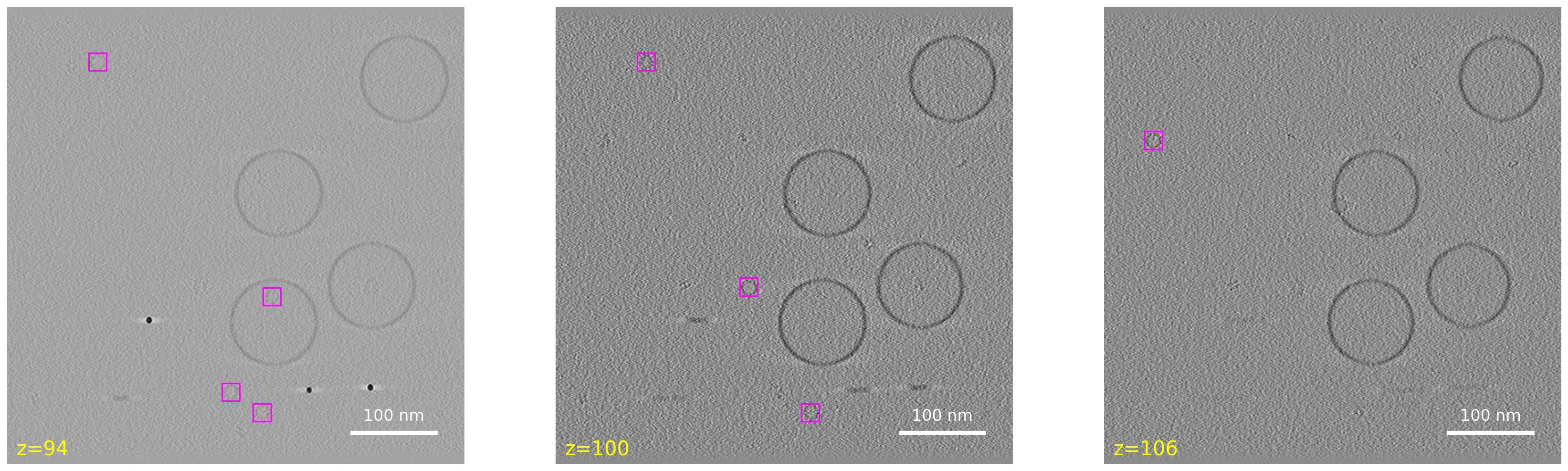}
\end{center}
\caption{\textbf{Tomogram Z-slice visualization of ProPicker detection results (``seen'' simulation, \texttt{TS\_4\_1}).} Estimated localizations of \texttt{2DF7} particles are shown. For each displayed $xy$-plane at depth $z$, we render only the bounding boxes whose center coordinates fall within a local depth window defined by the particle's radius $r$ (i.e., $z \pm r$). A representative subset of the tomogram Z-slices is shown for visual clarity.}
\label{fig:propicker_detection_result_tomo_sim_diff_4_1}
\end{figure}

Visualizations for representative instances are presented in the following order: the DS-10440 dataset (Figs.~\ref{fig:detection_result_tiltseries_czii_TS_6_6}--\ref{fig:propicker_detection_result_tomo_czii_TS_6_6}), EMPIAR-10304 (Figs.~\ref{fig:detection_result_tiltseries_10304_tilt1}--\ref{fig:propicker_detection_result_tomo_10304_tilt1}), EMPIAR-10499 (Figs.~\ref{fig:detection_result_tiltseries_10499_TS_01}--\ref{fig:propicker_detection_result_tomo_10499_TS_01}), ``seen'' simulations (Figs.~\ref{fig:detection_result_tiltseries_sim_same_4_0}--\ref{fig:propicker_detection_result_tomo_sim_same_4_0}), and ``unseen'' simulations (Figs.~\ref{fig:detection_result_tiltseries_sim_diff_4_1}--\ref{fig:propicker_detection_result_tomo_sim_diff_4_1}). As previously noted, while empirical EMPIAR datasets may contain a small fraction of unlabeled particles due to the inherent challenge of obtaining exhaustive ground truth, the annotations remain qualitatively robust for visual comparison.

\subsection{DS-10440 Qualitative Detection Results}

\clearpage

\subsection{EMPIAR-10304 Qualitative Detection Results}

\clearpage

\subsection{EMPIAR-10499 Qualitative Detection Results}

\clearpage

\subsection{``Seen'' Simulation Qualitative Detection Results}

\clearpage

\subsection{``Unseen'' Simulation Qualitative Detection Results}

\clearpage

\section{Comprehensive Ablation Study}
\label{sec:supp_abl}
\subsection{Additional Results on the Number of Tilt Images}
We evaluate the impact of utilizing fewer tilt images by systematically testing retention rates of 100\%, 50\%, 25\%, and 12.5\%. These sparse subsets are generated by progressively doubling the angular stride between sampled projections. The additional results are detailed in Tables~\ref{tab:reduce_tilt_images_real_1p} (1 prompt), ~\ref{tab:reduce_tilt_images_real_2p} (2 prompts), and \ref{tab:reduce_tilt_images_real_4p} (4 prompts) for the real-world datasets, and Tables~\ref{tab:reduce_tilt_images_sim_1p} (1 prompt), \ref{tab:reduce_tilt_images_sim_2p} (2 prompts), and \ref{tab:reduce_tilt_images_sim_4p} (4 prompts) for the simulation datasets. As anticipated, reducing the density of the tilt-series leads to a consistent degradation in detection performance, which is accompanied by proportional gains in inference speed and reduced VRAM consumption.

\begin{table*}[h]
\centering
\caption{\textbf{Ablation study on the number of tilt images (real-world datasets; 1 prompt).} Reducing the input sequence inherently decreases runtime and memory at the cost of overall detection performance.}
\label{tab:reduce_tilt_images_real_1p}
\resizebox{\textwidth}{!}{%
\begin{threeparttable}
\begin{tabular}{@{}l|rrrr|rrrr|rrrr@{}}
\toprule
 & \multicolumn{4}{c}{\textbf{DS-10440}} & \multicolumn{4}{c}{\textbf{EMPIAR-10304}} & \multicolumn{4}{c}{\textbf{EMPIAR-10499}}\\ 
\cmidrule(lr){2-5}\cmidrule(lr){6-9}\cmidrule(lr){10-13}
\textbf{\# Tilt Images} & 
\textbf{mAP@0.5r $\uparrow$} & \textbf{mAP@1r $\uparrow$} &
\textbf{T. $\downarrow$} & \textbf{M. $\downarrow$} &
\textbf{mAP@0.5r $\uparrow$} & \textbf{mAP@1r $\uparrow$} &
\textbf{T. $\downarrow$} & \textbf{M. $\downarrow$} &
\textbf{mAP@0.5r $\uparrow$} & \textbf{mAP@1r $\uparrow$} &
\textbf{T. $\downarrow$} & \textbf{M. $\downarrow$}
\\

\midrule
100\% & 
0.118 ± 0.013 & 0.148 ± 0.013 & 0.499 & 3430 &
0.199 ± 0.004 & 0.437 ± 0.002 & 0.229 & 2410 &
0.159 ± 0.004 & 0.288 ± 0.004 & 0.389 & 3410 \\

50\% & 
0.100 ± 0.013 & 0.133 ± 0.011 & 0.246 & 1860 & 
0.184 ± 0.002 & 0.427 ± 0.002 & 0.118 & 1380 &
0.127 ± 0.006 & 0.260 ± 0.007 & 0.202 & 1950 \\

25\% & 
0.085 ± 0.010 & 0.120 ± 0.011 & 0.153 & 1280 &
0.157 ± 0.003 & 0.391 ± 0.003 & 0.079 & 860 &
0.060 ± 0.005 & 0.165 ± 0.008 & 0.116 & 1220 \\

12.5\% & 
0.038 ± 0.007 & 0.073 ± 0.014 & 0.087 & 700 & 
0.084 ± 0.003 & 0.268 ± 0.003 & 0.069 & 600 &
0.010 ± 0.001 & 0.048 ± 0.003 & 0.083 & 790 \\

\bottomrule
\end{tabular}%

\begin{tablenotes}[flushleft]\footnotesize
\item T.: runtime (s); M.: VRAM (MB); $\downarrow$: lower is better; $\uparrow$: higher is better
\end{tablenotes}
\end{threeparttable}
}
\end{table*}

\begin{table*}[h]
\centering
\caption{\textbf{Ablation study on the number of tilt images (real-world datasets; 2 prompts).} Reducing the input sequence inherently decreases runtime and memory at the cost of overall detection performance.}
\label{tab:reduce_tilt_images_real_2p}
\resizebox{\textwidth}{!}{%
\begin{threeparttable}
\begin{tabular}{@{}l|rrrr|rrrr|rrrr@{}}
\toprule
 & \multicolumn{4}{c}{\textbf{DS-10440}} & \multicolumn{4}{c}{\textbf{EMPIAR-10304}} & \multicolumn{4}{c}{\textbf{EMPIAR-10499}}\\ 
\cmidrule(lr){2-5}\cmidrule(lr){6-9}\cmidrule(lr){10-13}
\textbf{\# Tilt Images} & 
\textbf{mAP@0.5r $\uparrow$} & \textbf{mAP@1r $\uparrow$} &
\textbf{T. $\downarrow$} & \textbf{M. $\downarrow$} &
\textbf{mAP@0.5r $\uparrow$} & \textbf{mAP@1r $\uparrow$} &
\textbf{T. $\downarrow$} & \textbf{M. $\downarrow$} &
\textbf{mAP@0.5r $\uparrow$} & \textbf{mAP@1r $\uparrow$} &
\textbf{T. $\downarrow$} & \textbf{M. $\downarrow$}
\\

\midrule
100\% & 
0.129 ± 0.010 & 0.161 ± 0.011 & 0.503 & 3430 &
0.200 ± 0.004 & 0.437 ± 0.003 & 0.230 & 2410 &
0.169 ± 0.006 & 0.302 ± 0.009 & 0.394 & 3410 \\

50\% & 
0.114 ± 0.007 & 0.150 ± 0.008 & 0.246 & 1860 &
0.183 ± 0.003 & 0.426 ± 0.002 & 0.119 & 1380 &
0.134 ± 0.006 & 0.272 ± 0.008 & 0.205 & 1950 \\

25\% & 
0.097 ± 0.006 & 0.134 ± 0.006 & 0.153 & 1280 &
0.156 ± 0.002 & 0.390 ± 0.002 & 0.082 & 860 &
0.065 ± 0.003 & 0.177 ± 0.007 & 0.116 & 1220 \\

12.5\% & 
0.044 ± 0.006 & 0.082 ± 0.009 & 0.088 & 700 &
0.085 ± 0.002 & 0.270 ± 0.004 & 0.072 & 600 &
0.011 ± 0.001 & 0.050 ± 0.002 & 0.083 & 792 \\

\bottomrule
\end{tabular}%

\begin{tablenotes}[flushleft]\footnotesize
\item T.: runtime (s); M.: VRAM (MB); $\downarrow$: lower is better; $\uparrow$: higher is better
\end{tablenotes}
\end{threeparttable}
}
\end{table*}

\begin{table*}[h]
\centering
\caption{\textbf{Ablation study on the number of tilt images (real-world datasets; 4 prompts).} Reducing the input sequence inherently decreases runtime and memory at the cost of overall detection performance.}
\label{tab:reduce_tilt_images_real_4p}
\resizebox{\textwidth}{!}{%
\begin{threeparttable}
\begin{tabular}{@{}l|rrrr|rrrr|rrrr@{}}
\toprule
 & \multicolumn{4}{c}{\textbf{DS-10440}} & \multicolumn{4}{c}{\textbf{EMPIAR-10304}} & \multicolumn{4}{c}{\textbf{EMPIAR-10499}}\\ 
\cmidrule(lr){2-5}\cmidrule(lr){6-9}\cmidrule(lr){10-13}
\textbf{\# Tilt Images} & 
\textbf{mAP@0.5r $\uparrow$} & \textbf{mAP@1r $\uparrow$} &
\textbf{T. $\downarrow$} & \textbf{M. $\downarrow$} &
\textbf{mAP@0.5r $\uparrow$} & \textbf{mAP@1r $\uparrow$} &
\textbf{T. $\downarrow$} & \textbf{M. $\downarrow$} &
\textbf{mAP@0.5r $\uparrow$} & \textbf{mAP@1r $\uparrow$} &
\textbf{T. $\downarrow$} & \textbf{M. $\downarrow$}
\\

\midrule
100\% &
0.133 ± 0.007 & 0.180 ± 0.007 & 0.510 & 3430 &
0.201 ± 0.004 & 0.439 ± 0.001 & 0.230 & 2410 &
0.173 ± 0.002 & 0.309 ± 0.003 & 0.394 & 3410 \\

50\% &
0.117 ± 0.007 & 0.157 ± 0.007 & 0.244 & 1860 &
0.186 ± 0.003 & 0.429 ± 0.002 & 0.122 & 1380 &
0.138 ± 0.002 & 0.281 ± 0.004 & 0.203 & 1950 \\

25\% &
0.099 ± 0.006 & 0.139 ± 0.005 & 0.150 & 1280 &
0.155 ± 0.002 & 0.389 ± 0.003 & 0.083 & 860 &
0.069 ± 0.004 & 0.185 ± 0.008 & 0.115 & 1220 \\

12.5\% &
0.046 ± 0.006 & 0.086 ± 0.010 & 0.081 & 700 &
0.083 ± 0.002 & 0.269 ± 0.001 & 0.077 & 600 &
0.011 ± 0.001 & 0.052 ± 0.003 & 0.082 & 792 \\

\bottomrule
\end{tabular}%

\begin{tablenotes}[flushleft]\footnotesize
\item T.: runtime (s); M.: VRAM (MB); $\downarrow$: lower is better; $\uparrow$: higher is better
\end{tablenotes}
\end{threeparttable}
}
\end{table*}

\begin{table*}[h]
\centering
\caption{\textbf{Ablation study on the number of tilt images (simulation datasets; 1 prompt).} Reducing the input sequence inherently decreases runtime and memory at the cost of overall detection performance.}
\label{tab:reduce_tilt_images_sim_1p}
\resizebox{\textwidth}{!}{%
\begin{threeparttable}
\begin{tabular}{@{}l|rrrr|rrrr@{}}
\toprule
 & \multicolumn{4}{c}{\textbf{Simulation (Seen)}} & \multicolumn{4}{c}{\textbf{Simulation (Unseen)}} \\ 
\cmidrule(lr){2-5}\cmidrule(lr){6-9}
\textbf{\# Tilt Images} & 
\textbf{mAP@0.5r $\uparrow$} & \textbf{mAP@1r $\uparrow$} &
\textbf{T. $\downarrow$} & \textbf{M. $\downarrow$} &
\textbf{mAP@0.5r $\uparrow$} & \textbf{mAP@1r $\uparrow$} &
\textbf{T. $\downarrow$} & \textbf{M. $\downarrow$}\\

\midrule
100\% &
0.632 ± 0.006 & 0.670 ± 0.007 & 0.682 & 5160 &
0.502 ± 0.012 & 0.542 ± 0.013 & 0.694 & 5160 \\

50\% &
0.567 ± 0.018 & 0.639 ± 0.019 & 0.329 & 2820 &
0.426 ± 0.023 & 0.487 ± 0.021 & 0.334 & 2820 \\

25\% &
0.377 ± 0.018 & 0.532 ± 0.022 & 0.163 & 1570 &
0.272 ± 0.025 & 0.380 ± 0.031 & 0.167 & 1570 \\

12.5\% &
0.151 ± 0.029 & 0.244 ± 0.037 & 0.093 & 949 &
0.069 ± 0.018 & 0.135 ± 0.025 & 0.092 & 949 \\

\bottomrule
\end{tabular}%

\begin{tablenotes}[flushleft]\footnotesize
\item T.: runtime (s); M.: VRAM (MB); $\downarrow$: lower is better; $\uparrow$: higher is better
\end{tablenotes}
\end{threeparttable}
}
\end{table*}

\begin{table*}[h]
\centering
\caption{\textbf{Ablation study on the number of tilt images (simulation datasets; 2 prompts).} Reducing the input sequence inherently decreases runtime and memory at the cost of overall detection performance.}
\label{tab:reduce_tilt_images_sim_2p}
\resizebox{\textwidth}{!}{%
\begin{threeparttable}
\begin{tabular}{@{}l|rrrr|rrrr@{}}
\toprule
 & \multicolumn{4}{c}{\textbf{Simulation (Seen)}} & \multicolumn{4}{c}{\textbf{Simulation (Unseen)}} \\ 
\cmidrule(lr){2-5}\cmidrule(lr){6-9}
\textbf{\# Tilt Images} & 
\textbf{mAP@0.5r $\uparrow$} & \textbf{mAP@1r $\uparrow$} &
\textbf{T. $\downarrow$} & \textbf{M. $\downarrow$} &
\textbf{mAP@0.5r $\uparrow$} & \textbf{mAP@1r $\uparrow$} &
\textbf{T. $\downarrow$} & \textbf{M. $\downarrow$}\\

\midrule
100\% &
0.631 ± 0.008 & 0.670 ± 0.007 & 0.685 & 5160 &
0.513 ± 0.010 & 0.552 ± 0.011 & 0.698 & 5160 \\

50\% &
0.577 ± 0.011 & 0.648 ± 0.014 & 0.329 & 2820 &
0.452 ± 0.013 & 0.511 ± 0.012 & 0.334 & 2820 \\

25\% &
0.400 ± 0.010 & 0.559 ± 0.014 & 0.163 & 1570 &
0.295 ± 0.015 & 0.408 ± 0.017 & 0.162 & 1570 \\

12.5\% &
0.192 ± 0.021 & 0.311 ± 0.033 & 0.090 & 949 &
0.101 ± 0.018 & 0.192 ± 0.025 & 0.094 & 949 \\

\bottomrule
\end{tabular}%

\begin{tablenotes}[flushleft]\footnotesize
\item T.: runtime (s); M.: VRAM (MB); $\downarrow$: lower is better; $\uparrow$: higher is better
\end{tablenotes}
\end{threeparttable}
}
\end{table*}

\begin{table*}[h]
\centering
\caption{\textbf{Ablation study on the number of tilt images (simulation datasets; 4 prompts).} Reducing the input sequence inherently decreases runtime and memory at the cost of overall detection performance.}
\label{tab:reduce_tilt_images_sim_4p}
\resizebox{\textwidth}{!}{%
\begin{threeparttable}
\begin{tabular}{@{}l|rrrr|rrrr@{}}
\toprule
 & \multicolumn{4}{c}{\textbf{Simulation (Seen)}} & \multicolumn{4}{c}{\textbf{Simulation (Unseen)}} \\ 
\cmidrule(lr){2-5}\cmidrule(lr){6-9}
\textbf{\# Tilt Images} & 
\textbf{mAP@0.5r $\uparrow$} & \textbf{mAP@1r $\uparrow$} &
\textbf{T. $\downarrow$} & \textbf{M. $\downarrow$} &
\textbf{mAP@0.5r $\uparrow$} & \textbf{mAP@1r $\uparrow$} &
\textbf{T. $\downarrow$} & \textbf{M. $\downarrow$}\\

\midrule
100\% &
0.627 ± 0.008 & 0.662 ± 0.007 & 0.687 & 5160 &
0.513 ± 0.004 & 0.553 ± 0.005 & 0.699 & 5160 \\

50\% &
0.574 ± 0.007 & 0.644 ± 0.008 & 0.331 & 2820 &
0.463 ± 0.003 & 0.521 ± 0.004 & 0.334 & 2820 \\

25\% &
0.408 ± 0.007 & 0.562 ± 0.009 & 0.166 & 1570 &
0.324 ± 0.009 & 0.436 ± 0.007 & 0.165 & 1570 \\

12.5\% &
0.222 ± 0.010 & 0.354 ± 0.018 & 0.094 & 949 &
0.129 ± 0.016 & 0.239 ± 0.025 & 0.089 & 949 \\

\bottomrule
\end{tabular}%

\begin{tablenotes}[flushleft]\footnotesize
\item T.: runtime (s); M.: VRAM (MB); $\downarrow$: lower is better; $\uparrow$: higher is better
\end{tablenotes}
\end{threeparttable}
}
\end{table*}

\clearpage

\subsection{Number of Projections per Prompt}
To demonstrate the flexibility of our visual prompt encoder ($\mathcal{E}_{\text{Prompt}}$), we evaluate scenarios where visual prompts are not provided across the entire tilt-series. Specifically, we test retention settings where prompts are present in 100\%, 50\%, 25\%, and 12.5\% of the tilt images, alongside an extreme ``single-image prompt'' setting. The comprehensive results across the 1-, 2-, and 4-prompt configurations are detailed in Tables~\ref{tab:reduce_prompt_in_tilt_images_real_1p},~\ref{tab:reduce_prompt_in_tilt_images_real_2p}, and~\ref{tab:reduce_prompt_in_tilt_images_real_4p} for the real-world datasets, and Tables~\ref{tab:reduce_prompt_in_tilt_images_sim_1p},~\ref{tab:reduce_prompt_in_tilt_images_sim_2p}, and~\ref{tab:reduce_prompt_in_tilt_images_sim_4p} for the simulation datasets. Remarkably, FullTilt remains highly robust even in the most extreme scenario—providing just a single visual prompt on a single tilt image (specifically, the 0° projection)—incurring only a marginal degradation in detection performance.

\begin{table*}[h]
\centering
\caption{\textbf{Ablation study on the number of projections per prompt (real-world datasets; 1 prompt).} Evaluation of FullTilt's robustness when prompts are provided on only a subset of the tilt-series images.}
\label{tab:reduce_prompt_in_tilt_images_real_1p}
\resizebox{\textwidth}{!}{%
\begin{threeparttable}
\begin{tabular}{@{}l|rrrr|rrrr|rrrr@{}}
\toprule
 & \multicolumn{4}{c}{\textbf{DS-10440}} & \multicolumn{4}{c}{\textbf{EMPIAR-10304}} & \multicolumn{4}{c}{\textbf{EMPIAR-10499}}\\ 
\cmidrule(lr){2-5}\cmidrule(lr){6-9}\cmidrule(lr){10-13}
\textbf{\# Projections per Prompt} & 
\textbf{mAP@0.5r $\uparrow$} & \textbf{mAP@1r $\uparrow$} &
\textbf{T. $\downarrow$} & \textbf{M. $\downarrow$} &
\textbf{mAP@0.5r $\uparrow$} & \textbf{mAP@1r $\uparrow$} &
\textbf{T. $\downarrow$} & \textbf{M. $\downarrow$} &
\textbf{mAP@0.5r $\uparrow$} & \textbf{mAP@1r $\uparrow$} &
\textbf{T. $\downarrow$} & \textbf{M. $\downarrow$}
\\

\midrule
100\% & 
0.118 ± 0.013 & 0.148 ± 0.013 & 0.499 & 3430 &
0.199 ± 0.004 & 0.437 ± 0.002 & 0.229 & 2410 &
0.159 ± 0.004 & 0.288 ± 0.004 & 0.389 & 3410 \\

50\% & 
0.111 ± 0.008 & 0.142 ± 0.009 & 0.514 & 3430 &
0.199 ± 0.003 & 0.437 ± 0.002 & 0.223 & 2420 &
0.162 ± 0.005 & 0.290 ± 0.005 & 0.392 & 3410 \\

25\% & 
0.107 ± 0.013 & 0.136 ± 0.014 & 0.513 & 3430 & 
0.199 ± 0.004 & 0.437 ± 0.003 & 0.223 & 2420 &
0.158 ± 0.006 & 0.285 ± 0.007 & 0.391 & 3410 \\

12.5\% & 
0.108 ± 0.009 & 0.137 ± 0.010 & 0.512 & 3430 &
0.197 ± 0.005 & 0.436 ± 0.003 & 0.224 & 2420 &
0.158 ± 0.005 & 0.285 ± 0.006 & 0.391 & 3410 \\

Only One & 
0.106 ± 0.014 & 0.134 ± 0.016 & 0.513 & 3430 &
0.190 ± 0.004 & 0.430 ± 0.004 & 0.224 & 2420 &
0.155 ± 0.009 & 0.278 ± 0.011 & 0.390 & 3410 \\

\bottomrule
\end{tabular}%

\begin{tablenotes}[flushleft]\footnotesize
\item T.: runtime (s); M.: VRAM (MB); $\downarrow$: lower is better; $\uparrow$: higher is better
\end{tablenotes}
\end{threeparttable}
}
\end{table*}

\begin{table*}[h]
\centering
\caption{\textbf{Ablation study on the number of projections per prompt (real-world datasets; 2 prompts).} Evaluation of FullTilt's robustness when prompts are provided on only a subset of the tilt-series images.}
\label{tab:reduce_prompt_in_tilt_images_real_2p}
\resizebox{\textwidth}{!}{%
\begin{threeparttable}
\begin{tabular}{@{}l|rrrr|rrrr|rrrr@{}}
\toprule
 & \multicolumn{4}{c}{\textbf{DS-10440}} & \multicolumn{4}{c}{\textbf{EMPIAR-10304}} & \multicolumn{4}{c}{\textbf{EMPIAR-10499}}\\ 
\cmidrule(lr){2-5}\cmidrule(lr){6-9}\cmidrule(lr){10-13}
\textbf{\# Projections per Prompt} & 
\textbf{mAP@0.5r $\uparrow$} & \textbf{mAP@1r $\uparrow$} &
\textbf{T. $\downarrow$} & \textbf{M. $\downarrow$} &
\textbf{mAP@0.5r $\uparrow$} & \textbf{mAP@1r $\uparrow$} &
\textbf{T. $\downarrow$} & \textbf{M. $\downarrow$} &
\textbf{mAP@0.5r $\uparrow$} & \textbf{mAP@1r $\uparrow$} &
\textbf{T. $\downarrow$} & \textbf{M. $\downarrow$}
\\

\midrule
100\% & 
0.129 ± 0.010 & 0.161 ± 0.011 & 0.503 & 3430 &
0.200 ± 0.004 & 0.437 ± 0.003 & 0.230 & 2410 &
0.169 ± 0.006 & 0.302 ± 0.009 & 0.394 & 3410 \\

50\% & 
0.124 ± 0.009 & 0.157 ± 0.009 & 0.513 & 3430 &
0.199 ± 0.003 & 0.437 ± 0.003 & 0.226 & 2420 &
0.168 ± 0.003 & 0.301 ± 0.004 & 0.392 & 3410 \\

25\% & 
0.123 ± 0.011 & 0.155 ± 0.012 & 0.514 & 3430 &
0.199 ± 0.003 & 0.438 ± 0.003 & 0.225 & 2420 &
0.166 ± 0.006 & 0.297 ± 0.009 & 0.391 & 3410 \\

12.5\% & 
0.126 ± 0.012 & 0.156 ± 0.013 & 0.514 & 3430 &
0.198 ± 0.003 & 0.436 ± 0.002 & 0.225 & 2420 &
0.167 ± 0.005 & 0.300 ± 0.007 & 0.390 & 3410 \\

Only One & 
0.108 ± 0.011 & 0.137 ± 0.009 & 0.513 & 3430 &
0.190 ± 0.005 & 0.429 ± 0.006 & 0.225 & 2420 &
0.164 ± 0.006 & 0.293 ± 0.008 & 0.390 & 3410 \\

\bottomrule
\end{tabular}%

\begin{tablenotes}[flushleft]\footnotesize
\item T.: runtime (s); M.: VRAM (MB); $\downarrow$: lower is better; $\uparrow$: higher is better
\end{tablenotes}
\end{threeparttable}
}
\end{table*}

\begin{table*}[h]
\centering
\caption{\textbf{Ablation study on the number of projections per prompt (real-world datasets; 4 prompts).} Evaluation of FullTilt's robustness when prompts are provided on only a subset of the tilt-series images.}
\label{tab:reduce_prompt_in_tilt_images_real_4p}
\resizebox{\textwidth}{!}{%
\begin{threeparttable}
\begin{tabular}{@{}l|rrrr|rrrr|rrrr@{}}
\toprule
 & \multicolumn{4}{c}{\textbf{DS-10440}} & \multicolumn{4}{c}{\textbf{EMPIAR-10304}} & \multicolumn{4}{c}{\textbf{EMPIAR-10499}}\\ 
\cmidrule(lr){2-5}\cmidrule(lr){6-9}\cmidrule(lr){10-13}
\textbf{\# Projections per Prompt} & 
\textbf{mAP@0.5r $\uparrow$} & \textbf{mAP@1r $\uparrow$} &
\textbf{T. $\downarrow$} & \textbf{M. $\downarrow$} &
\textbf{mAP@0.5r $\uparrow$} & \textbf{mAP@1r $\uparrow$} &
\textbf{T. $\downarrow$} & \textbf{M. $\downarrow$} &
\textbf{mAP@0.5r $\uparrow$} & \textbf{mAP@1r $\uparrow$} &
\textbf{T. $\downarrow$} & \textbf{M. $\downarrow$}
\\

\midrule
100\% & 
0.133 ± 0.007 & 0.180 ± 0.007 & 0.510 & 3430 &
0.201 ± 0.004 & 0.439 ± 0.001 & 0.230 & 2410 &
0.173 ± 0.002 & 0.309 ± 0.003 & 0.394 & 3410 \\

50\% & 
0.130 ± 0.007 & 0.165 ± 0.006 & 0.514 & 3430 &
0.199 ± 0.002 & 0.438 ± 0.002 & 0.227 & 2420 &
0.175 ± 0.003 & 0.311 ± 0.004 & 0.392 & 3410 \\

25\% & 
0.133 ± 0.005 & 0.167 ± 0.005 & 0.514 & 3430 &
0.199 ± 0.002 & 0.438 ± 0.002 & 0.226 & 2420 &
0.172 ± 0.004 & 0.307 ± 0.005 & 0.392 & 3410 \\

12.5\% & 
0.130 ± 0.006 & 0.160 ± 0.007 & 0.514 & 3430 &
0.197 ± 0.004 & 0.437 ± 0.002 & 0.226 & 2420 &
0.170 ± 0.005 & 0.306 ± 0.007 & 0.391 & 3410 \\

Only One & 
0.117 ± 0.009 & 0.146 ± 0.009 & 0.514 & 3430 &
0.191 ± 0.004 & 0.433 ± 0.004 & 0.225 & 2420 &
0.175 ± 0.006 & 0.306 ± 0.006 & 0.392 & 3410 \\

\bottomrule
\end{tabular}%

\begin{tablenotes}[flushleft]\footnotesize
\item T.: runtime (s); M.: VRAM (MB); $\downarrow$: lower is better; $\uparrow$: higher is better
\end{tablenotes}
\end{threeparttable}
}
\end{table*}

\begin{table*}[h]
\centering
\caption{\textbf{Ablation study on reducing the number of prompted tilt images (simulation datasets; 1 prompt).} Evaluation of FullTilt's robustness when prompts are provided on only a subset of the tilt-series images.}
\label{tab:reduce_prompt_in_tilt_images_sim_1p}
\resizebox{\textwidth}{!}{%
\begin{threeparttable}
\begin{tabular}{@{}l|rrrr|rrrr@{}}
\toprule
 & \multicolumn{4}{c}{\textbf{Simulation (Seen)}} & \multicolumn{4}{c}{\textbf{Simulation (Unseen)}} \\ 
\cmidrule(lr){2-5}\cmidrule(lr){6-9}
\textbf{\# Projections per Prompt} & 
\textbf{mAP@0.5r $\uparrow$} & \textbf{mAP@1r $\uparrow$} &
\textbf{T. $\downarrow$} & \textbf{M. $\downarrow$} &
\textbf{mAP@0.5r $\uparrow$} & \textbf{mAP@1r $\uparrow$} &
\textbf{T. $\downarrow$} & \textbf{M. $\downarrow$}\\

\midrule
100\% &
0.632 ± 0.006 & 0.670 ± 0.007 & 0.682 & 5160 &
0.502 ± 0.012 & 0.542 ± 0.013 & 0.694 & 5160 \\

50\% &
0.625 ± 0.012 & 0.665 ± 0.012 & 0.682 & 5160 &
0.500 ± 0.008 & 0.543 ± 0.008 & 0.694 & 5160 \\

25\% &
0.629 ± 0.013 & 0.666 ± 0.014 & 0.681 & 5160 &
0.483 ± 0.016 & 0.523 ± 0.018 & 0.694 & 5160 \\

12.5\% &
0.609 ± 0.019 & 0.645 ± 0.019 & 0.680 & 5160 &
0.486 ± 0.025 & 0.525 ± 0.025 & 0.693 & 5160 \\

Only One & 
0.572 ± 0.025 & 0.608 ± 0.026 & 0.661 & 5160 &
0.431 ± 0.017 & 0.467 ± 0.015 & 0.693 & 5160 \\

\bottomrule
\end{tabular}%

\begin{tablenotes}[flushleft]\footnotesize
\item T.: runtime (s); M.: VRAM (MB); $\downarrow$: lower is better; $\uparrow$: higher is better
\end{tablenotes}
\end{threeparttable}
}
\end{table*}

\begin{table*}[h]
\centering
\caption{\textbf{Ablation study on reducing the number of prompted tilt images (simulation datasets; 2 prompts).} Evaluation of FullTilt's robustness when prompts are provided on only a subset of the tilt-series images.}
\label{tab:reduce_prompt_in_tilt_images_sim_2p}
\resizebox{\textwidth}{!}{%
\begin{threeparttable}
\begin{tabular}{@{}l|rrrr|rrrr@{}}
\toprule
 & \multicolumn{4}{c}{\textbf{Simulation (Seen)}} & \multicolumn{4}{c}{\textbf{Simulation (Unseen)}} \\ 
\cmidrule(lr){2-5}\cmidrule(lr){6-9}
\textbf{\# Projections per Prompt} & 
\textbf{mAP@0.5r $\uparrow$} & \textbf{mAP@1r $\uparrow$} &
\textbf{T. $\downarrow$} & \textbf{M. $\downarrow$} &
\textbf{mAP@0.5r $\uparrow$} & \textbf{mAP@1r $\uparrow$} &
\textbf{T. $\downarrow$} & \textbf{M. $\downarrow$}\\

\midrule
100\% &
0.631 ± 0.008 & 0.670 ± 0.007 & 0.685 & 5160 &
0.513 ± 0.010 & 0.552 ± 0.011 & 0.698 & 5160 \\

50\% &
0.630 ± 0.010 & 0.668 ± 0.010 & 0.682 & 5160 &
0.510 ± 0.011 & 0.551 ± 0.011 & 0.695 & 5160 \\

25\% &
0.626 ± 0.006 & 0.662 ± 0.006 & 0.681 & 5160 &
0.510 ± 0.009 & 0.549 ± 0.008 & 0.695 & 5160 \\

12.5\% &
0.628 ± 0.009 & 0.663 ± 0.008 & 0.681 & 5160 &
0.505 ± 0.013 & 0.543 ± 0.013 & 0.694 & 5160 \\

Only One & 
0.594 ± 0.013 & 0.629 ± 0.015 & 0.681 & 5160 &
0.475 ± 0.027 & 0.513 ± 0.027 & 0.692 & 5160 \\

\bottomrule
\end{tabular}%

\begin{tablenotes}[flushleft]\footnotesize
\item T.: runtime (s); M.: VRAM (MB); $\downarrow$: lower is better; $\uparrow$: higher is better
\end{tablenotes}
\end{threeparttable}
}
\end{table*}

\begin{table*}[h]
\centering
\caption{\textbf{Ablation study on reducing the number of prompted tilt images (simulation datasets; 4 prompts).} Evaluation of FullTilt's robustness when prompts are provided on only a subset of the tilt-series images.}
\label{tab:reduce_prompt_in_tilt_images_sim_4p}
\resizebox{\textwidth}{!}{%
\begin{threeparttable}
\begin{tabular}{@{}l|rrrr|rrrr@{}}
\toprule
 & \multicolumn{4}{c}{\textbf{Simulation (Seen)}} & \multicolumn{4}{c}{\textbf{Simulation (Unseen)}} \\ 
\cmidrule(lr){2-5}\cmidrule(lr){6-9}
\textbf{\# Projections per Prompt} & 
\textbf{mAP@0.5r $\uparrow$} & \textbf{mAP@1r $\uparrow$} &
\textbf{T. $\downarrow$} & \textbf{M. $\downarrow$} &
\textbf{mAP@0.5r $\uparrow$} & \textbf{mAP@1r $\uparrow$} &
\textbf{T. $\downarrow$} & \textbf{M. $\downarrow$}\\

\midrule
100\% &
0.627 ± 0.008 & 0.662 ± 0.007 & 0.687 & 5160 &
0.513 ± 0.004 & 0.553 ± 0.005 & 0.699 & 5160 \\

50\% &
0.625 ± 0.010 & 0.660 ± 0.009 & 0.681 & 5160 &
0.514 ± 0.005 & 0.554 ± 0.005 & 0.692 & 5160 \\

25\% &
0.629 ± 0.012 & 0.663 ± 0.012 & 0.682 & 5160 &
0.511 ± 0.005 & 0.550 ± 0.007 & 0.697 & 5160 \\

12.5\% &
0.625 ± 0.006 & 0.659 ± 0.007 & 0.682 & 5160 &
0.510 ± 0.009 & 0.548 ± 0.010 & 0.698 & 5160 \\

Only One & 
0.579 ± 0.020 & 0.614 ± 0.020 & 0.683 & 5160 &
0.482 ± 0.022 & 0.520 ± 0.022 & 0.697 & 5160 \\

\bottomrule
\end{tabular}%

\begin{tablenotes}[flushleft]\footnotesize
\item T.: runtime (s); M.: VRAM (MB); $\downarrow$: lower is better; $\uparrow$: higher is better
\end{tablenotes}
\end{threeparttable}
}
\end{table*}

\clearpage

\subsection{Scaling the Number of Queries}
In dense macromolecular environments, such as specific tomograms within the EMPIAR-10304 dataset, the total number of target particles frequently exceeds the default query capacity ($M = 900$) typical of DETR-like architectures. To address this bottleneck without the computational burden of retraining, we investigate the zero-shot scaling of our query capacity by directly duplicating the learned content queries during inference. The quantitative results are detailed in Table~\ref{tab:increase_queries}. 

Remarkably, scaling the number of queries to 1350 (1.5$\times$) and 1800 (2.0$\times$) yields steady improvements in detection performance across all prompting configurations. However, this performance gain involves a distinct computational trade-off: peak VRAM consumption exhibits a severe super-linear growth. In practice, attempting to train FullTilt from scratch with 1800 queries strictly exceeds the 48GB VRAM capacity of an NVIDIA RTX 6000 Ada GPU, resulting in out-of-memory errors. Consequently, developing more memory-efficient strategies to handle ultra-dense particle distributions remains an important direction for future work. 

\begin{table*}[h]
\centering
\caption{\textbf{Ablation study on zero-shot query scaling (EMPIAR-10304).} We show the impact of heuristically duplicating the default learned content queries ($M=900$) during inference to accommodate high-density particle distributions. While scaling the query capacity to 1.5$\times$ and 2.0$\times$ strictly improves detection performance across all prompting settings, it incurs a super-linear penalty in peak VRAM consumption.}
\label{tab:increase_queries}
\resizebox{\textwidth}{!}{%
\begin{threeparttable}
\begin{tabular}{@{}l|rrrr|rrrr|rrrr@{}}
\toprule
 & \multicolumn{4}{c}{\textbf{1 Prompt}} & \multicolumn{4}{c}{\textbf{2 Prompts}} & \multicolumn{4}{c}{\textbf{4 Prompts}}\\ 
\cmidrule(lr){2-5}\cmidrule(lr){6-9}\cmidrule(lr){10-13}
\textbf{\# Queries} & 
\textbf{mAP@0.5r $\uparrow$} & \textbf{mAP@1r $\uparrow$} &
\textbf{T. $\downarrow$} & \textbf{M. $\downarrow$} &
\textbf{mAP@0.5r $\uparrow$} & \textbf{mAP@1r $\uparrow$} &
\textbf{T. $\downarrow$} & \textbf{M. $\downarrow$} &
\textbf{mAP@0.5r $\uparrow$} & \textbf{mAP@1r $\uparrow$} &
\textbf{T. $\downarrow$} & \textbf{M. $\downarrow$}
\\

\midrule
900 (1.0$\times$) & 
0.199 ± 0.004 & 0.437 ± 0.002 & 0.229 & 2410 &
0.200 ± 0.004 & 0.437 ± 0.003 & 0.230 & 2410 &
0.201 ± 0.004 & 0.439 ± 0.001 & 0.230 & 2410 \\

1350 (1.5$\times$)& 
0.218 ± 0.004 & 0.496 ± 0.002 & 0.296 & 3950 &
0.219 ± 0.004 & 0.497 ± 0.002 & 0.293 & 3950 &
0.219 ± 0.002 & 0.498 ± 0.001 & 0.295 & 3950 \\

1800 (2.0$\times$) & 
0.224 ± 0.002 & 0.515 ± 0.002 & 0.356 & 6060 &
0.225 ± 0.005 & 0.516 ± 0.002 & 0.356 & 6060 &
0.225 ± 0.002 & 0.516 ± 0.003 & 0.356 & 6060 \\

\bottomrule
\end{tabular}%

\begin{tablenotes}[flushleft]\footnotesize
\item T.: runtime (s); M.: VRAM (MB); $\downarrow$: lower is better; $\uparrow$: higher is better
\end{tablenotes}
\end{threeparttable}
}
\end{table*}

\subsection{Additional Component Ablation Results}
We conduct a comprehensive ablation study to isolate the impact of our proposed modules: the tilt-series encoder ($\mathcal{E}_{\text{Tilt}}$), the tilt-aware query initializer ($\mathcal{I}_{\text{Tilt}}$), and the auxiliary geometric primitives module ($\mathcal{G}$). Quantitative results on the simulation datasets are detailed in Table~\ref{tab:ablation_module_sim}. As expected, the naive baseline—omitting all proposed modules—exhibits severely degraded performance. When evaluating components in isolation, the integration of the tilt-aware query initializer ($\mathcal{I}_{\text{Tilt}}$) yields the most substantial independent performance gains. In dual-module configurations, the combination of $\mathcal{E}_{\text{Tilt}}$ and $\mathcal{I}_{\text{Tilt}}$ demonstrates a strong synergistic effect, significantly elevating detection capabilities. Finally, incorporating the auxiliary geometric primitives module ($\mathcal{G}$) completes the FullTilt framework, achieving optimal peak performance across all evaluated metrics.

\begin{table*}[h]
\centering
\caption{\textbf{Ablation study on FullTilt components (simulation datasets).} We show the impact of the tilt-series encoder ($\mathcal{E}_{\text{Tilt}}$), tilt-aware query initializer ($\mathcal{I}_{\text{Tilt}}$), and auxiliary geometric primitives ($\mathcal{G}$) on detection performance across two simulation datasets. The integration of all three modules yields the optimal performance.}
\label{tab:ablation_module_sim}
\resizebox{\textwidth}{!}{%
\begin{threeparttable}
\begin{tabular}{@{}ccc|rrrr|rrrr@{}}
\toprule
 & & & \multicolumn{4}{c}{\textbf{Simulation (Seen)}} & \multicolumn{4}{c}{\textbf{Simulation (Unseen)}} \\ 
\cmidrule(lr){4-7}\cmidrule(lr){8-11}
$\mathcal{E}_{\text{Tilt}}$ & $\mathcal{I}_{\text{Tilt}}$ & $\mathcal{G}$ & 
\textbf{mAP@0.5r $\uparrow$} & \textbf{mAP@1r $\uparrow$} &
\textbf{T. $\downarrow$} & \textbf{M. $\downarrow$} &
\textbf{mAP@0.5r $\uparrow$} & \textbf{mAP@1r $\uparrow$} &
\textbf{T. $\downarrow$} & \textbf{M. $\downarrow$}\\

\midrule 
& & & 
0.001 ± 0.000 & 0.010 ± 0.001 & 0.319 & 4790 &
0.000 ± 0.000 & 0.003 ± 0.001 & 0.320 & 4790 \\
\midrule

\checkmark & & & 
0.004 ± 0.001 & 0.011 ± 0.002 & 0.677 & 4810 &
0.004 ± 0.001 & 0.008 ± 0.002 & 0.670 & 4810 \\

& \checkmark & & 
0.332 ± 0.006 & 0.545 ± 0.007 & 0.331 & 5140 &
0.237 ± 0.008 & 0.439 ± 0.018 & 0.331 & 5140 \\

& & \checkmark & 
0.027 ± 0.002 & 0.103 ± 0.005 & 0.323 & 4790 &
0.012 ± 0.002 & 0.068 ± 0.006 & 0.322 & 4790 \\

\midrule

\checkmark & \checkmark & & 
0.558 ± 0.013 & 0.624 ± 0.015 & 0.678 & 5160 &
0.455 ± 0.014 & 0.523 ± 0.015 & 0.692 & 5160 \\

& \checkmark & \checkmark & 
0.382 ± 0.010 & 0.569 ± 0.013 & 0.331 & 5140 &
0.290 ± 0.014 & 0.460 ± 0.012 & 0.330 & 5140 \\

\checkmark & & \checkmark & 
0.014 ± 0.003 & 0.032 ± 0.002 & 0.677 & 4810 & 
0.008 ± 0.002 & 0.018 ± 0.001 & 0.660 & 4810 \\

\midrule

\checkmark & \checkmark & \checkmark & 
0.632 ± 0.006 & 0.670 ± 0.007 & 0.682 & 5160 &
0.502 ± 0.012 & 0.542 ± 0.013 & 0.694 & 5160 \\

\bottomrule
\end{tabular}%

\begin{tablenotes}[flushleft]\footnotesize
\item \checkmark: has this feature; T.: runtime (s); M.: VRAM (MB); $\downarrow$: lower is better; $\uparrow$: higher is better
\end{tablenotes}
\end{threeparttable}
}
\end{table*}

\clearpage
\subsection{Detailed Ablation Study of the Tilt-Series Encoder ($\mathcal{E}_{\text{Tilt}}$)}
\label{sec:abl_tilt_encoder}
We conducted a granular ablation study to evaluate the specific components of the tilt-series encoder: the integration of tilt embeddings, the application of alternating-attention, and the adoption of global row attention. Comprehensive results for the real-world and simulation datasets are presented in Tables~\ref{tab:ablation_tse_real} and \ref{tab:ablation_tse_sim}, respectively. 

As anticipated from our PETR benchmark, relying solely on static tilt embeddings without cross-tilt information fusion yields minimal performance improvement. However, cross-tilt fusion intrinsically depends on these embeddings; omitting them deprives the network of the explicit angle information necessary to geometrically correlate projections. While introducing dense alternating-attention successfully enables this fusion and drives a dramatic increase in detection performance, it incurs severe computational penalties. This dense formulation increases runtime by $10\times$ to $20\times$ and demands excessive VRAM, strictly necessitating high-capacity NVIDIA H200 GPUs for training. Although integrating hardware-aware optimizations like FlashAttention~\cite{dao2022flashattention} reduces the memory footprint to $\mathcal{O}(N)$, the temporal complexity fundamentally remains $\mathcal{O}(N^2)$. By transitioning to our proposed global row attention mechanism, we achieve an approximate $10\times$ reduction in inference runtime, while fully preserving the robust performance gains established by the cross-tilt information fusion.

\begin{table*}[h]
\centering
\caption{\textbf{Ablation study of the tilt-series encoder ($\mathcal{E}_{\text{Tilt}}$) on real-world datasets.} Quantitative evaluation of the components within the tilt-series encoder, including the tilt embedding, alternating-attention (AA), and global row attention is shown. Performance is measured across the DS-10440, EMPIAR-10304, and EMPIAR-10499 datasets. Incorporating global row attention achieves an order-of-magnitude reduction in inference runtime while preserving the significant performance gains established by cross-tilt information fusion.}
\label{tab:ablation_tse_real}
\resizebox{\textwidth}{!}{%
\begin{threeparttable}
\begin{tabular}{@{}ccc|rrrr|rrrr|rrrr@{}}
\toprule
 & & & \multicolumn{4}{c}{\textbf{DS-10440}} & \multicolumn{4}{c}{\textbf{EMPIAR-10304}} & \multicolumn{4}{c}{\textbf{EMPIAR-10499}}\\ 
\cmidrule(lr){4-7}\cmidrule(lr){8-11}\cmidrule(lr){12-15}
Tilt Embedding & AA & Global Row Attn. &
\textbf{mAP@0.5r $\uparrow$} & \textbf{mAP@1r $\uparrow$} &
\textbf{T. $\downarrow$} & \textbf{M. $\downarrow$} &
\textbf{mAP@0.5r $\uparrow$} & \textbf{mAP@1r $\uparrow$} &
\textbf{T. $\downarrow$} & \textbf{M. $\downarrow$} &
\textbf{mAP@0.5r $\uparrow$} & \textbf{mAP@1r $\uparrow$} &
\textbf{T. $\downarrow$} & \textbf{M. $\downarrow$}
\\

\midrule

& & & 
0.094 ± 0.004 & 0.153 ± 0.005 & 0.222 & 3430 &
0.030 ± 0.001 & 0.181 ± 0.003 & 0.142 & 2400 &
0.033 ± 0.001 & 0.127 ± 0.003 & 0.207 & 3380 \\

\checkmark & & & 
0.099 ± 0.006 & 0.149 ± 0.008 & 0.224 & 3430 &
0.031 ± 0.002 & 0.198 ± 0.004 & 0.140 & 2400 &
0.048 ± 0.001 & 0.167 ± 0.004 & 0.207 & 3380 \\

\checkmark & \checkmark & & 
0.125 ± 0.007 & 0.158 ± 0.009 & 3.73 & 3430 &
0.176 ± 0.005 & 0.384 ± 0.005 & 0.723 & 2420 &
0.152 ± 0.003 & 0.280 ± 0.004 & 2.05 & 3390 \\

& \checkmark & \checkmark & 
0.084 ± 0.004 & 0.134 ± 0.005 & 0.512 & 3430 &
0.062 ± 0.002 & 0.239 ± 0.002 & 0.231 & 2420 &
0.049 ± 0.001 & 0.164 ± 0.004 & 0.384 & 3390 \\

\checkmark & \checkmark & \checkmark & 
0.118 ± 0.013 & 0.148 ± 0.013 & 0.499 & 3430 &
0.199 ± 0.004 & 0.437 ± 0.002 & 0.229 & 2410 &
0.159 ± 0.004 & 0.288 ± 0.004 & 0.389 & 3410 \\

\bottomrule
\end{tabular}%

\begin{tablenotes}[flushleft]\footnotesize
\item \checkmark: has this feature; AA: alternating-attention; Global Row Attn.: global row attention; T.: runtime (s); M.: VRAM (MB); $\downarrow$: lower is better; $\uparrow$: higher is better
\end{tablenotes}
\end{threeparttable}
}
\end{table*}

\begin{table*}[h]
\centering
\caption{\textbf{Ablation study of the tilt-series encoder ($\mathcal{E}_{\text{Tilt}}$) on simulation datasets.} Quantitative evaluation of the tilt-series encoder's components under both ``seen'' and ``unseen'' simulated settings is shown. Consistent with real-world evaluations, the transition to global row attention reduces computational overhead, dramatically accelerating inference with near-zero degradation in detection performance.}
\label{tab:ablation_tse_sim}
\resizebox{\textwidth}{!}{%
\begin{threeparttable}
\begin{tabular}{@{}ccc|rrrr|rrrr@{}}
\toprule
 & & & \multicolumn{4}{c}{\textbf{Simulation (Seen)}} & \multicolumn{4}{c}{\textbf{Simulation (Unseen)}} \\ 
\cmidrule(lr){4-7}\cmidrule(lr){8-11}
Tilt Embedding & AA & Global Row Attn. & 
\textbf{mAP@0.5r $\uparrow$} & \textbf{mAP@1r $\uparrow$} &
\textbf{T. $\downarrow$} & \textbf{M. $\downarrow$} &
\textbf{mAP@0.5r $\uparrow$} & \textbf{mAP@1r $\uparrow$} &
\textbf{T. $\downarrow$} & \textbf{M. $\downarrow$}\\

\midrule 
& & & 
0.382 ± 0.010 & 0.569 ± 0.013 & 0.331 & 5140 &
0.290 ± 0.014 & 0.460 ± 0.012 & 0.330 & 5140 \\

\checkmark & & & 
0.377 ± 0.005 & 0.571 ± 0.009 & 0.333 & 5140 &
0.283 ± 0.014 & 0.450 ± 0.017 & 0.331 & 5140 \\

\checkmark & \checkmark & & 
0.639 ± 0.006 & 0.669 ± 0.006 & 6.07 & 5160 &
0.497 ± 0.015 & 0.535 ± 0.013 & 5.89 & 5160 \\

& \checkmark & \checkmark & 
0.511 ± 0.014 & 0.617 ± 0.009 & 0.697 & 5160 &
0.405 ± 0.008 & 0.514 ± 0.009 & 0.688 & 5160 \\

\checkmark & \checkmark& \checkmark & 
0.632 ± 0.006 & 0.670 ± 0.007 & 0.682 & 5160 &
0.502 ± 0.012 & 0.542 ± 0.013 & 0.694 & 5160 \\

\bottomrule
\end{tabular}%

\begin{tablenotes}[flushleft]\footnotesize
\item \checkmark: has this feature; AA: alternating-attention; Global Row Attn.: global row attention; T.: runtime (s); M.: VRAM (MB); $\downarrow$: lower is better; $\uparrow$: higher is better
\end{tablenotes}
\end{threeparttable}
}
\end{table*}

\clearpage
\subsection{Detailed Ablation Study of the Tilt-Aware Query Initializer ($\mathcal{I}_{\text{Tilt}}$)}
\label{sec:abl_tilt_initializer}
We conducted a detailed ablation study to evaluate the specific initialization strategies of the tilt-aware query initializer. Specifically, we ablate the design choice of initializing the 3D anchors' $x$- and $y$-coordinates from the closest 0° projection, as well as explicitly setting the initial $z$-coordinate to 0.5 (representing the geometric midplane of the $z$-axis). Quantitative results for the real-world and simulation datasets are provided in Tables~\ref{tab:ablation_initialization_real} and \ref{tab:ablation_initialization_sim}, respectively. 

Without explicit initialization (i.e., relying entirely on randomly initialized, fully learnable 3D anchors), the model's detection performance is severely degraded. Utilizing the closest 0° projection to initialize the $x$- and $y$-coordinates of each 3D anchor provides a critical spatial prior, dramatically improving overall performance. Finally, incorporating the depth-wise $z$-coordinate initialization stabilizes the spatial learning process, allowing the model to achieve optimal performance. This full initialization strategy yields particularly significant gains on the EMPIAR-10304 and EMPIAR-10499 datasets.

\begin{table*}[h]
\centering
\caption{\textbf{Ablation study of the Tilt-Aware Query Initializer ($\mathcal{I}_{\text{Tilt}}$) on real-world datasets.} We show the impact of explicitly initializing the $x$-, $y$-, and $z$-coordinates of the 3D anchors prior to the decoder stages. Relying on purely learned anchors fails to establish a strong spatial prior, whereas our proposed 0° projection initialization and midplane $z$-initialization drastically improve detection performance.}
\label{tab:ablation_initialization_real}
\resizebox{\textwidth}{!}{%
\begin{threeparttable}
\begin{tabular}{@{}cc|rrrr|rrrr|rrrr@{}}
\toprule
 & & \multicolumn{4}{c}{\textbf{DS-10440}} & \multicolumn{4}{c}{\textbf{EMPIAR-10304}} & \multicolumn{4}{c}{\textbf{EMPIAR-10499}}\\ 
\cmidrule(lr){3-6}\cmidrule(lr){7-10}\cmidrule(lr){11-14}
\textbf{0$^\circ$ XY Init.} & \textbf{Midplane Z Init.} & 
\textbf{mAP@0.5r $\uparrow$} & \textbf{mAP@1r $\uparrow$} &
\textbf{T. $\downarrow$} & \textbf{M. $\downarrow$} &
\textbf{mAP@0.5r $\uparrow$} & \textbf{mAP@1r $\uparrow$} &
\textbf{T. $\downarrow$} & \textbf{M. $\downarrow$} &
\textbf{mAP@0.5r $\uparrow$} & \textbf{mAP@1r $\uparrow$} &
\textbf{T. $\downarrow$} & \textbf{M. $\downarrow$}
\\

\midrule

& & 
0.002 ± 0.001 & 0.008 ± 0.002 & 0.520 & 3430 &
0.001 ± 0.000 & 0.017 ± 0.001 & 0.229 & 2300 &
0.000 ± 0.000 & 0.004 ± 0.000 & 0.379 & 3190 \\

\checkmark & & 
0.122 ± 0.008 & 0.152 ± 0.008 & 0.510 & 3430 &
0.139 ± 0.003 & 0.348 ± 0.004 & 0.229 & 2420 &
0.155 ± 0.007 & 0.284 ± 0.011 & 0.389 & 3390 \\

\checkmark & \checkmark & 
0.118 ± 0.013 & 0.148 ± 0.013 & 0.499 & 3430 &
0.199 ± 0.004 & 0.437 ± 0.002 & 0.229 & 2410 &
0.159 ± 0.004 & 0.288 ± 0.004 & 0.389 & 3410 \\

\bottomrule
\end{tabular}%

\begin{tablenotes}[flushleft]\footnotesize
\item \checkmark: has this feature; Init.: initialization; T.: runtime (s); M.: VRAM (MB); $\downarrow$: lower is better; $\uparrow$: higher is better
\end{tablenotes}
\end{threeparttable}
}
\end{table*}

\begin{table*}[h]
\centering
\caption{\textbf{Ablation study of the Tilt-Aware Query Initializer ($\mathcal{I}_{\text{Tilt}}$) on simulation datasets.} We show the impact of explicitly initializing the $x$-, $y$-, and $z$-coordinates of the 3D anchors prior to the decoder stages. Relying on purely learned anchors fails to establish a strong spatial prior, whereas our proposed 0° projection initialization and midplane $z$-initialization drastically improve detection performance.}
\label{tab:ablation_initialization_sim}
\resizebox{\textwidth}{!}{%
\begin{threeparttable}
\begin{tabular}{@{}cc|rrrr|rrrr@{}}
\toprule
 & & \multicolumn{4}{c}{\textbf{Simulation (Seen)}} & \multicolumn{4}{c}{\textbf{Simulation (Unseen)}} \\ 
\cmidrule(lr){3-6}\cmidrule(lr){7-10}
\textbf{0$^\circ$ XY Init.} & \textbf{Midplane Z Init.}  & 
\textbf{mAP@0.5r $\uparrow$} & \textbf{mAP@1r $\uparrow$} &
\textbf{T. $\downarrow$} & \textbf{M. $\downarrow$} &
\textbf{mAP@0.5r $\uparrow$} & \textbf{mAP@1r $\uparrow$} &
\textbf{T. $\downarrow$} & \textbf{M. $\downarrow$}\\

\midrule 
& & 
0.014 ± 0.003 & 0.032 ± 0.002 & 0.677 & 4810 & 
0.008 ± 0.002 & 0.018 ± 0.001 & 0.660 & 4810 \\

\checkmark & & 
0.614 ± 0.013 & 0.650 ± 0.012 & 0.681 & 5160 &
0.493 ± 0.011 & 0.534 ± 0.010 & 0.691 & 5160 \\

\checkmark & \checkmark & 
0.632 ± 0.006 & 0.670 ± 0.007 & 0.682 & 5160 &
0.502 ± 0.012 & 0.542 ± 0.013 & 0.694 & 5160 \\

\bottomrule
\end{tabular}%

\begin{tablenotes}[flushleft]\footnotesize
\item \checkmark: has this feature; Init.: initialization; T.: runtime (s); M.: VRAM (MB); $\downarrow$: lower is better; $\uparrow$: higher is better
\end{tablenotes}
\end{threeparttable}
}
\end{table*}

\clearpage

\begin{algorithm}[H]
\caption{PyTorch style pseudocode for Local and Global Row Attention}
\label{alg:attention_pseudo}
\vspace{1mm} 
\begin{lstlisting}
def |local_attention|(x):
    """
    x:      features shaped (N, H, W, C)
    """
    N, H, W, C = x.shape
    
    # [N, H, W, C] -> [N, H$\times$W, C]
    x = x.flatten(1, 2)
    x = SelfAttention(x)
    
    # [N, H$\times$W, C] -> [N, H, W, C]
    x = x.view(N, H, W, C)
    
    return x


def |global_row_attention|(x):
    """
    x:      features shaped (N, H, W, C)
    """
    N, H, W, C = x.shape
    
    # [N, H, W, C] -> [H, N, W, C]
    x = x.permute(1, 0, 2, 3)
    
    # [H, N, W, C] -> [H, N$\times$W, C]
    x = x.flatten(1, 2)
    x = SelfAttention(x)
    
    # [H, N$\times$W, C] -> [H, N, W, C]
    x = x.view(H, N, W, C)
    
    # [H, N, W, C] -> [N, H, W, C]
    x = x.permute(1, 0, 2, 3)
    
    return x
\end{lstlisting}
\end{algorithm}


\end{document}